\documentclass[oneside,11pt]{article}

\usepackage{times,url,mathrsfs}
\usepackage{amsmath,wrapfig,color}
\usepackage{amsfonts}
\usepackage{graphicx}
\usepackage{subfigure}
\newtheorem{theorem}{Theorem}

\begin{document}
\title{\huge Coding for Random Projections and Approximate Near Neighbor Search\vspace{0.2in}}

\author{ \bf{Ping Li} \\
         Department of Statistics \& Biostatistics\\\hspace{0.1in}
         Department of Computer Science\\
       Rutgers University\\
          Piscataway, NJ 08854, USA\\
       \texttt{pingli@stat.rutgers.edu}\\\\
       \and
         \bf{Michael Mitzenmacher}\\
         School of Engineering and Applied Sciences\\
         Harvard University\\
         Cambridge, MA 02138, USA\\
         \texttt{michaelm@eecs.harvard.edu}
       \and
         \bf{Anshumali Shrivastava}\\
         Department of Computer Science\\
         Cornell University\\
         {Ithaca, NY 14853, USA}\\
        \texttt{anshu@cs.cornell.edu}}

\date{}

\maketitle

\begin{abstract}\vspace{-0.1in}
\noindent This technical note compares two coding (quantization) schemes for random projections in the context of sub-linear time approximate near neighbor search. The first scheme is based on uniform quantization~\cite{Report:RPCode2013} while the second scheme utilizes a uniform quantization plus a uniformly random offset~\cite{Proc:Datar_SCG04} (which has been  popular in practice). The  prior work~\cite{Report:RPCode2013} compared the two schemes in the context of similarity estimation and training linear classifiers, with the conclusion that the step of random offset is not necessary and may hurt the performance (depending on the similarity level).   The task of near neighbor search is  related to similarity estimation with importance distinctions and requires own study. In this paper, we demonstrate that in the context of near neighbor search, the step of random offset is not needed either and may hurt the performance (sometimes significantly so, depending on the similarity  and other parameters). \\

\noindent  For approximate near neighbor search, when the target similarity level is high (e.g.,  correlation $>0.85$), our analysis suggest to use a uniform quantization to build hash tables, with a  bin width $w=1\sim 1.5$. On the other hand, when the target similarity level is not that high, it is preferable to use larger $w$ values (e.g., $w \ge 2 \sim 3$). This is equivalent to say that it suffices to use only a small number of bits (or even just 1 bit) to code each hashed value in the context of sublinear time near neighbor search. An extensive experimental study on two reasonably large datasets confirms the theoretical finding. \\

\noindent Coding for building hash tables is a  different task from coding for similarity estimation. For near neighbor search, we need coding of the projected data to determine which buckets the data points should be placed in (and the coded values are not stored). For similarity estimation, the purpose of coding is for accurately estimating the similarities using small storage space. Therefore, if necessary, we can actually code the projected data twice (with different bin widths).  \\

\noindent In this paper, we do not study the important issue of ``re-ranking'' of retrieved data points by using estimated similarities. That step is needed when exact (all pairwise) similarities can not be practically stored or computed on the fly. In a concurrent work~\cite{Report:RPCodeNonlinear2014}, we demonstrate that the retrieval accuracy can be further improved by using nonlinear estimators of the similarities based on a 2-bit coding scheme.

\end{abstract}

\newpage\clearpage

\section{Introduction}

This paper focuses on the comparison of two quantization schemes for random projections in the context of sublinear time near neighbor search. The  task of {near neighbor search} is to identify a set of data points which are ``most similar'' (in some measure of similarity) to a query data point. Efficient algorithms for near neighbor search  have  numerous applications in search, databases, machine learning,  recommending systems, computer vision, etc. Developing efficient algorithms for finding near neighbors has been an active  research topic since the early days of modern computing~\cite{Article:Friedman_75}. Near neighbor search with extremely high-dimensional data (e.g., texts or images) is still a challenging task and an active research problem.

Among many types of similarity measures, the (squared) Euclidian distance (denoted by $d$) and the correlation (denoted by $\rho$) are most commonly used. Without loss of generality, consider two high-dimensional data vectors $u, v\in\mathbb{R}^D$. The squared Euclidean distance and  correlation are defined as follows:
\begin{align}
d = \sum_{i=1}^D |u_i - v_i|^2, \hspace{0.4in} \rho =  \frac{\sum_{i=1}^Du_iv_i}{\sqrt{\sum_{i=1}^D u_i^2} \sqrt{\sum_{i=1}^D v_i^2} }
\end{align}

In practice, it appears that the correlation is more often used than the distance, partly because $|\rho|$ is nicely normalized within 0 and 1. In fact, in this study, we will assume that the marginal $l_2$ norms $\sum_{i=1}^D |u_i|^2$ and $\sum_{i=1}^D |v_i|^2$ are known. This is a  reasonable assumption. Computing the marginal $l_2$ norms only requires scanning the data once, which is anyway needed during the data collection process. In machine learning practice, it is  common  to first normalize the data (to have unit $l_2$ norm) before feeding the data to classification (e.g., SVM) or clustering (e.g., K-means) algorithms.

For convenience, throughout this paper, we  assume  unit $l_2$ norms, i.e.,
\begin{align}
\rho =  \frac{\sum_{i=1}^Du_iv_i}{\sqrt{\sum_{i=1}^D u_i^2} \sqrt{\sum_{i=1}^D v_i^2} } = \sum_{i=1}^D u_iv_i,\hspace{0.5in} \text{where } \ \ \sum_{i=1}^D u_i^2 = \sum_{i=1}^D v_i^2 = 1
\end{align}

\subsection{Random Projections}
As an effective tool for dimensionality reduction, the idea  of random projections is to multiply the data, e.g.,   $u, v\in\mathbb{R}^D$, with a random normal projection matrix $\mathbf{R}\in\mathbb{R}^{D\times k}$ (where $k\ll D$), to generate:
\begin{align}
x = u\times \mathbf{R} \in\mathbb{R}^k,\hspace{0.2in} y = v\times \mathbf{R} \in\mathbb{R}^k, \hspace{0.2in} \mathbf{R} = \{r_{ij}\}{_{i=1}^D}{_{j=1}^k}, \hspace{0.2in} r_{ij} \sim N(0,1) \text{ i.i.d. }
\end{align}
The method of random projections has become popular for large-scale machine learning applications such as classification, regression, matrix factorization, singular value decomposition, near neighbor search, etc.%~\cite{Proc:Papadimitriou_PODS98,Proc:Dasgupta_FOCS99,Proc:Bingham_KDD01,Article:Buher_Tompa,Proc:Fradkin_KDD03,Proc:Li_Hastie_Church_COLT06,Proc:Frund_NIPS08,Book:Vempala,Proc:Dasgupta_UAI00,Article:JL84,Proc:Wang_Li_SDM10}.

The potential benefits of coding with a small number of bits arise because the (uncoded) projected data, $x_j = \sum_{i=1}^D u_i r_{ij}$ and $y_j = \sum_{i=1}^D v_i r_{ij}$, being real-valued numbers, are neither convenient/economical for storage and transmission, nor well-suited for indexing. The focus of this paper is on approximate (sublinear time) near neighbor search in the framework of {\em locality sensitive hashing}~\cite{Proc:Indyk_STOC98}. In particular, we will compare two \textbf{coding} (quantization) schemes of random projections~\cite{Proc:Datar_SCG04,Report:RPCode2013} in the context of near neighbor search.

%\newpage

\subsection{Uniform Quantization}

The recent work~\cite{Report:RPCode2013} proposed an intuitive coding scheme, based on a simple uniform quantization:
\begin{align}\label{eqn_hw}
h_{w}^{(j)}(u) = \left\lfloor x_j/w\right\rfloor,\hspace{0.5in} h_{w}^{(j)}(v) = \left\lfloor y_j/w\right\rfloor
\end{align}
where $w>0$ is the bin width and $\left\lfloor . \right\rfloor$ is the standard floor operation.

The following theorem is proved in~\cite{Report:RPCode2013} about the collision probability $P_{w} = \mathbf{Pr}\left(h_{w}^{(j)}(u) = h_{w}^{(j)}(v) \right)$.
\begin{theorem}\label{eqn_Pw}
\begin{align}\label{eqn_Pw}
P_{w} =\mathbf{Pr}\left(h_{w}^{(j)}(u) = h_{w}^{(j)}(v) \right) = 2\sum_{i=0}^\infty\int_{iw}^{(i+1)w}\phi(z)\left[\Phi\left(\frac{(i+1)w-\rho z}{\sqrt{1-\rho^2}}\right)- \Phi\left(\frac{iw-\rho z}{\sqrt{1-\rho^2}}\right)\right]dz
\end{align}
In addition, $P_w$ is a monotonically increasing function of $\rho$.\\
\end{theorem}
The fact that $P_w$ is a monotonically increasing function of $\rho$ makes   (\ref{eqn_hw}) a suitable coding scheme for approximate near neighbor search in the general framework of locality sensitive hashing (LSH).

\subsection{Uniform Quantization with Random Offset}

 \cite{Proc:Datar_SCG04} proposed the following well-known coding scheme, which uses
windows and a random offset:
\begin{align}\label{eqn_hwq}
h_{w,q}^{(j)}(u) = \left\lfloor\frac{x_j + q_j}{w}\right\rfloor,\hspace{0.3in} h_{w,q}^{(j)}(v) = \left\lfloor\frac{y_j + q_j}{w}\right\rfloor
\end{align}
where $q_j\sim uniform(0,w)$. \cite{Proc:Datar_SCG04} showed that the collision probability can be written as
\begin{align}\label{eqn_Pwq}
P_{w,q} = &\mathbf{Pr}\left(h_{w,q}^{(j)}(u) = h_{w,q}^{(j)}(v)\right)
= \int_0^w\frac{1}{\sqrt{d}}2\phi\left(\frac{t}{\sqrt{d}}\right)\left(1-\frac{t}{w}\right)dt
\end{align}
where $d = ||u-v||^2= 2(1-\rho)$ is the Euclidean distance between $u$ and $v$.  Compared with (\ref{eqn_hwq}), the scheme  (\ref{eqn_hw}) does  not use the additional randomization with $q\sim uniform(0,w)$ (i.e., the offset). \cite{Report:RPCode2013} elaborated the following advantages of (\ref{eqn_hw}) in the context of similarity estimation:
\begin{enumerate}
\item Operationally,  $h_{w}$ is  simpler than $h_{w,q}$.
\item With a fixed $w$,   $h_{w}$  is always more accurate than $h_{w,q}$, often significantly so.
\item For each coding scheme, one can separately find the optimum bin width $w$. The optimized  $h_w$ is also more accurate than optimized $h_{w,q}$, often significantly so.
\item $h_w$ requires a smaller number of bits than $h_{w,q}$.
\end{enumerate}

\noindent In this paper, we will compare $h_{w,q}$ with $h_w$ in the context of sublinear time near neighbor search.

\subsection{Sublinear Time $c$-Approximate Near Neighbor Search}

Consider a data vector  $u$. Suppose there exists another vector whose Euclidian distance ($\sqrt{d}$) from $u$ is at most $\sqrt{d_0}$ (the target distance). The goal of {\em$c$-approximate $\sqrt{d_0}$-near neighbor} algorithms is to return data vectors (with high probability) whose Euclidian distances from $u$ are at most $c\times \sqrt{d_0}$ with $c>1$.

Recall that, in our definition, $d = 2(1-\rho)$ is the squared Euclidian distance. To be consistent with~\cite{Proc:Datar_SCG04}, we present the results in terms of $\sqrt{d}$. Corresponding to the target distance $\sqrt{d_0}$,  the target similarity $\rho_0$ can be computed from $d_0 = 2(1-\rho_0)$ i.e., $\rho_0 = 1-d_0/2$. To simplify the presentation, we focus on $\rho\geq 0$ (as is common in practice), i.e., $0\leq d\leq 2$. Once we fix a target similarity $\rho_0$, $c$ can not exceed a certain value:
\begin{align}
c\sqrt{2(1-\rho_0)}\leq \sqrt{2} \Longrightarrow c \leq \sqrt{\frac{1}{1-\rho_0}}
\end{align}
For example, when $\rho_0 =0.5$, we must have $1\leq c \leq \sqrt{2}$. \\

Under the general framework, the performance of an LSH algorithm largely depends on the difference (gap) between the two collision probabilities $P^{(1)}$ and $P^{(2)}$ (respectively corresponding to $\sqrt{d_0}$ and $c\sqrt{d_0}$):
\begin{align}
&P^{(1)}_w =  \mathbf{Pr}\left(h_w(u) = h_w(v) \right) \hspace{0.2in} \text{when } d = ||u - v||^2_2 = d_0\\
&P^{(2)}_w =  \mathbf{Pr}\left(h_w(u) = h_w(v) \right) \hspace{0.2in} \text{when } d = ||u - v||^2_2 = c^2d_0
\end{align}
Corresponding to $h_{w,q}$, the collision probabilities $P^{(1)}_{w,q}$ and $P^{(2)}_{w,q}$ are analogously defined.  \\

A larger difference between $P^{(1)}$ and $P^{(2)}$ implies a more efficient LSH algorithm. The following ``$G$'' values ($G_w$ for $h_w$ and $G_{w,q}$ for $h_{w,q}$) characterize the gaps:
\begin{align}\label{eqn_rho_M}
&G_w =\frac{\log 1/P_w^{(1)} }{\log 1/P_w^{(2)} },\hspace{0.5in} G_{w,q} =\frac{\log 1/P_{w,q}^{(1)} }{\log 1/P_{w,q}^{(2)} }
\end{align}

\noindent A smaller $G$ (i.e., larger difference between $P^{(1)}$ and $P^{(2)}$) leads to a potentially more efficient LSH algorithm and $\rho <\frac{1}{c}$ is particularly desirable~\cite{Proc:Indyk_STOC98}. The general theory  says  the query time for $c$-approximate $d_0$-near neighbor is dominated by $O(N^G)$ distance evaluations, where $N$ is the total number of data vectors in the collection. This is better than $O(N)$, the cost of a linear scan.

\section{Comparison of the Collision Probabilities}\label{sec_hw}

To help understand the intuition why $h_w$ may lead to better performance than $h_{w,q}$, in this section we examine their collision probabilities $P_{w}$ and $P_{w,q}$, which can be expressed in terms of the standard normal pdf and cdf functions: $\phi(x) = \frac{1}{\sqrt{2\pi}} e^{-\frac{x^2}{2}}$ and $\Phi(x) = \int_{-\infty}^x \phi(x) dx$,
\begin{align}\label{eqn_Pwq2}
&P_{w,q} = \mathbf{Pr}\left(h_{w,q}^{(j)}(u) = h_{w,q}^{(j)}(v)\right)
= 2\Phi\left(\frac{w}{\sqrt{d}}\right)-1-\frac{2}{\sqrt{2\pi}w/\sqrt{d}}+\frac{2}{w/\sqrt{d}}\phi\left(\frac{w}{\sqrt{d}}\right)\\
&P_{w} = \mathbf{Pr}\left(h_{w}^{(j)}(u) = h_{w}^{(j)}(v)\right)=2\sum_{i=0}^\infty\int_{iw}^{(i+1)w}\phi(z)\left[\Phi\left(\frac{(i+1)w-\rho z}{\sqrt{1-\rho^2}}\right)- \Phi\left(\frac{iw-\rho z}{\sqrt{1-\rho^2}}\right)\right]dz
\end{align}
 It is clear that $P_{w,q}\rightarrow 1$ as $w\rightarrow\infty$.\\

\begin{figure}[h!]
\begin{center}
\mbox{
\includegraphics[width = 2.2in]{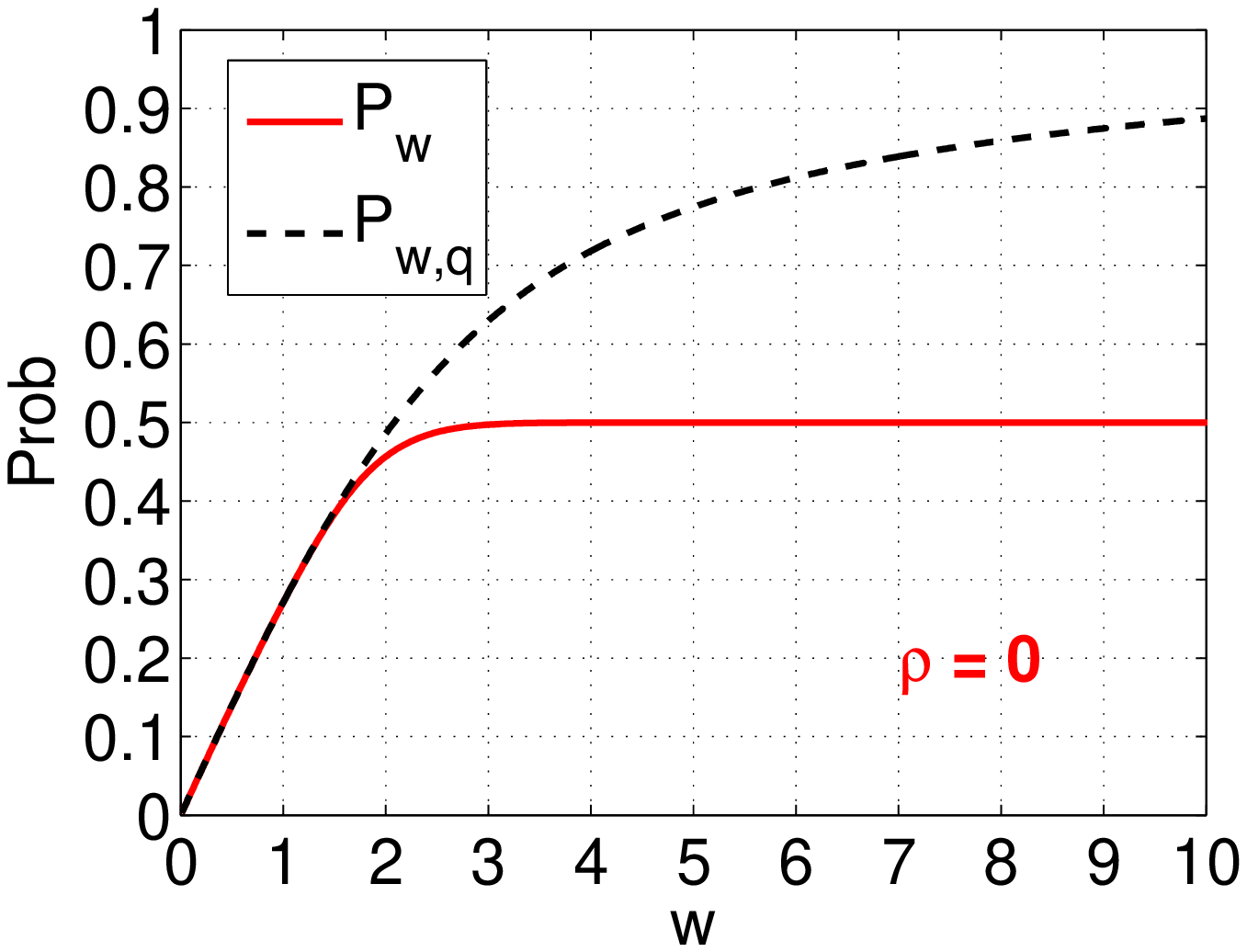}
\includegraphics[width = 2.2in]{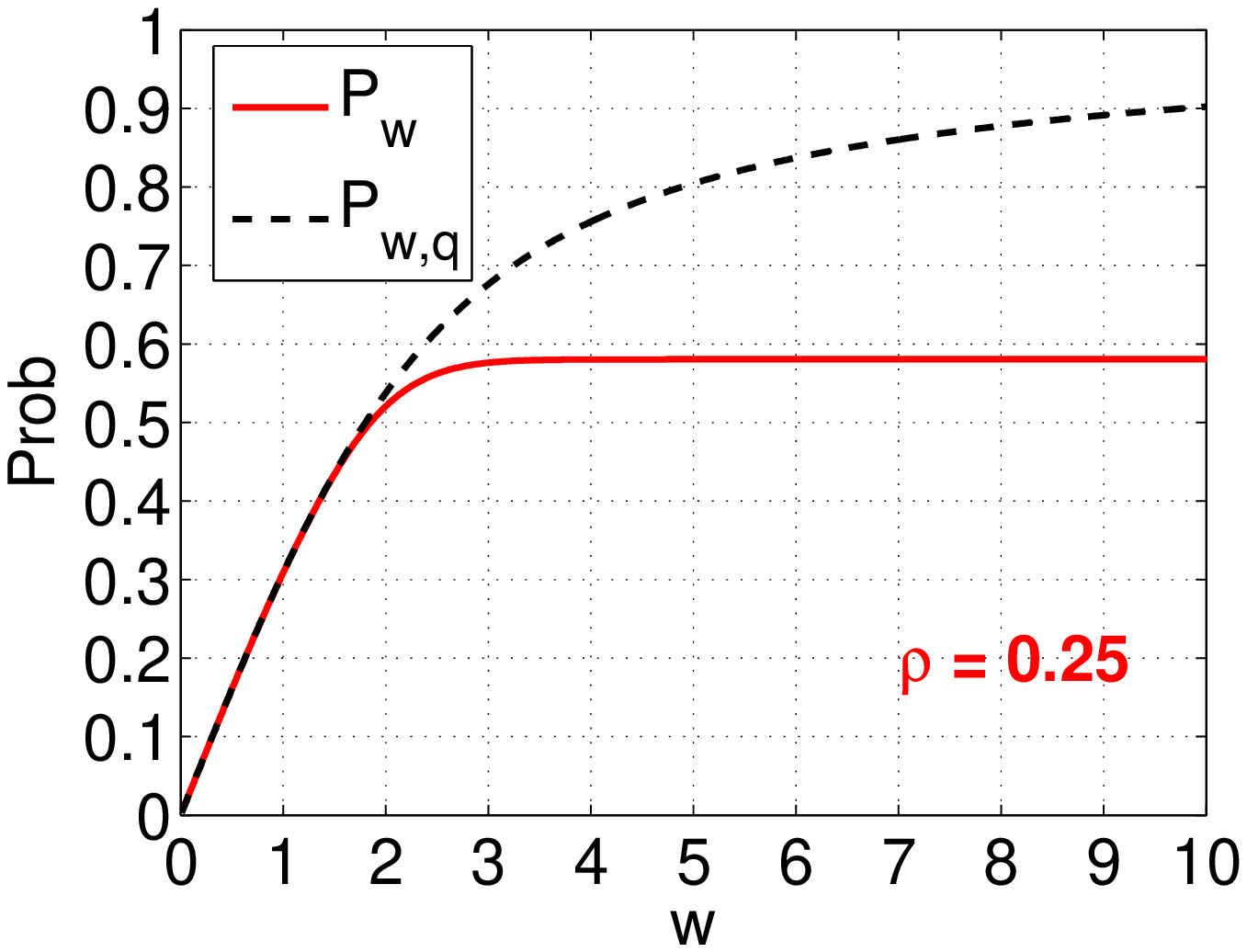}
\includegraphics[width = 2.2in]{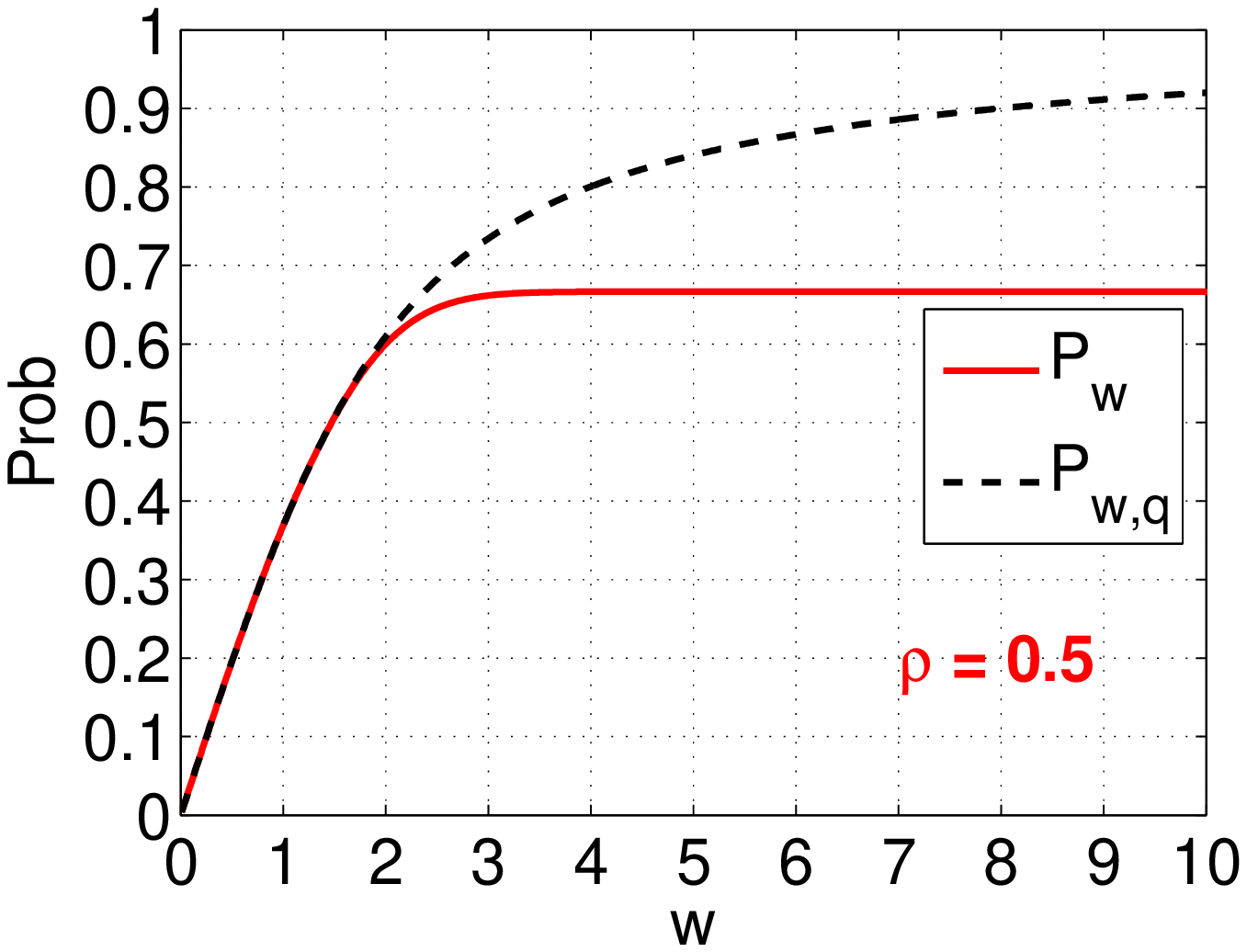}
}
\mbox{
\includegraphics[width = 2.2in]{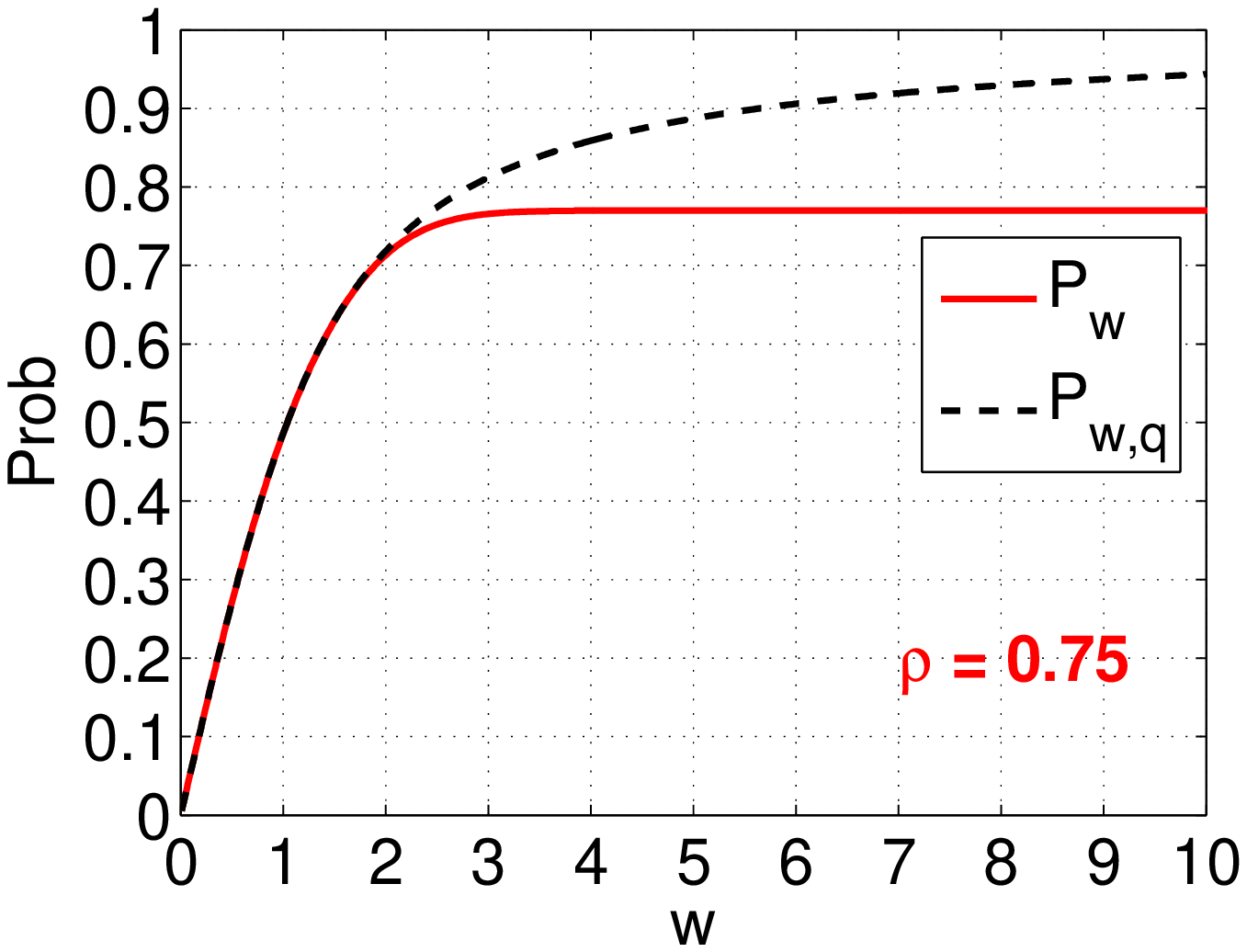}
\includegraphics[width = 2.2in]{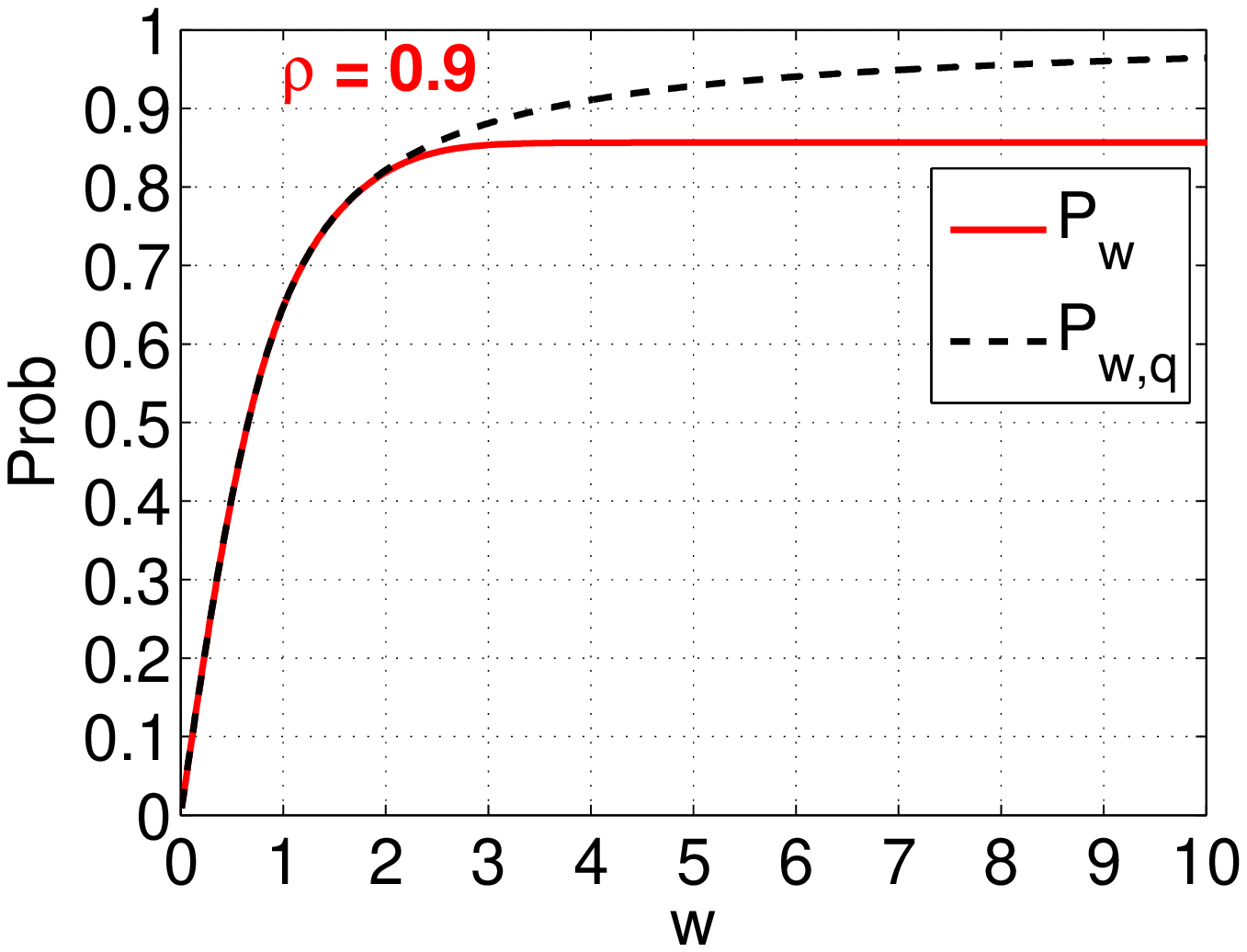}
\includegraphics[width = 2.2in]{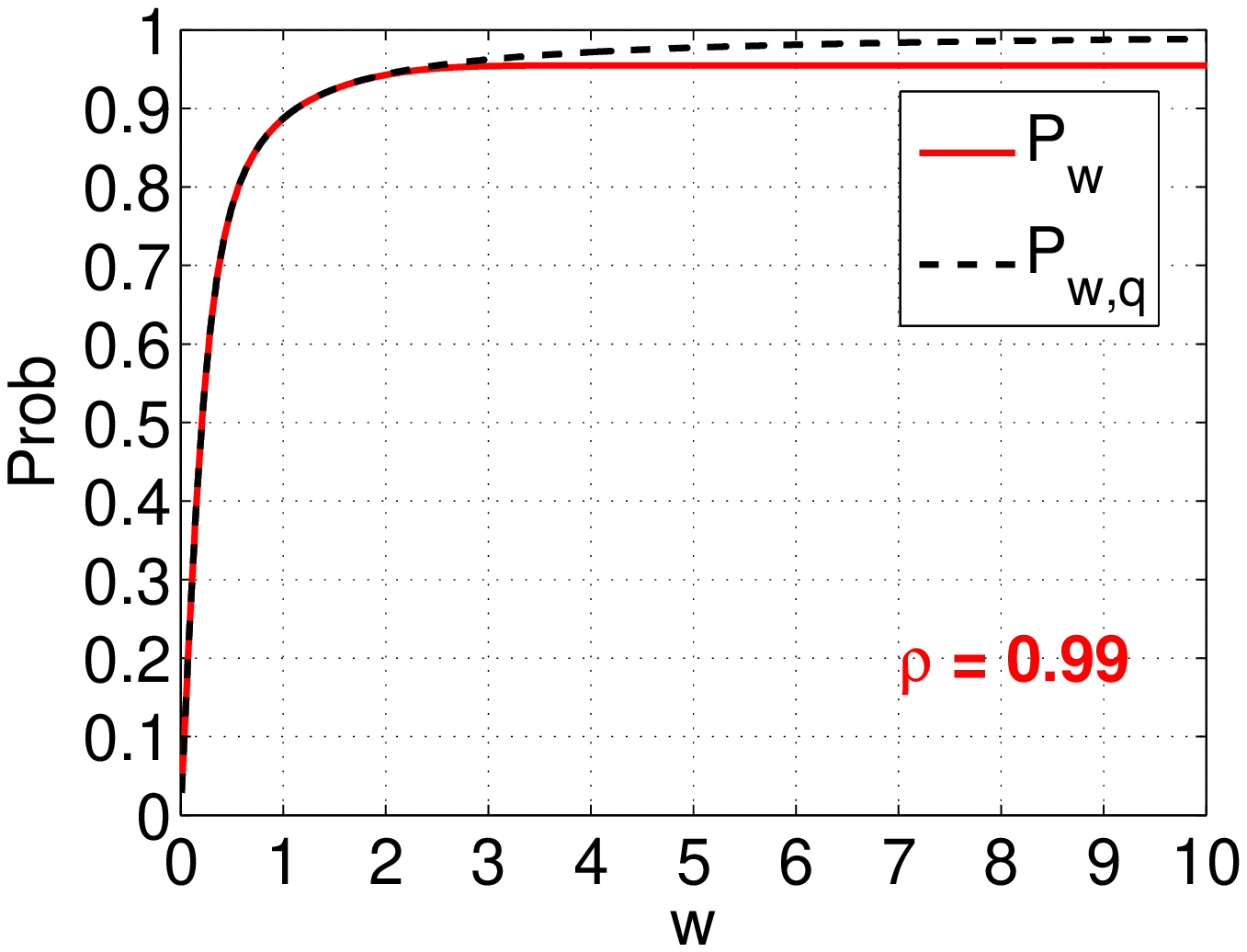}
}
\end{center}
\vspace{-.2in}
\caption{Collision probabilities, $P_w$ and $P_{w,q}$, for  $\rho = 0, 0.25, 0.5, 0.75, 0.9$, and $0.99$.  The scheme  $h_w$ has smaller collision probabilities than the scheme~\cite{Proc:Datar_SCG04} $h_{w,q}$, especially when $w>2$.}\label{fig_Pwq}
\end{figure}

Figure~\ref{fig_Pwq} plots both $P_w$ and $P_{w,q}$ for selected $\rho$ values.  The difference between $P_w$ and $P_{w,q}$ becomes apparent when $w$ is not small. For example, when $\rho=0$, $P_{w}$ quickly approaches the limit 0.5 while $P_{w,q}$ keeps increasing (to 1) as $w$ increases. Intuitively, the fact that $P_{w,q}\rightarrow1$ when $\rho=0$, is undesirable because it means two orthogonal vectors will have the same coded value. Thus, it is  not surprising that  $h_w$ will have better performance than $h_{w,q}$, for both similarity estimation and sublinear time near neighbor search.

%\newpage\clearpage
\newpage

\section{Theoretical Comparison of the Gaps}

\begin{figure}[h!]
\begin{center}
\mbox{
\includegraphics[width = 2.2in]{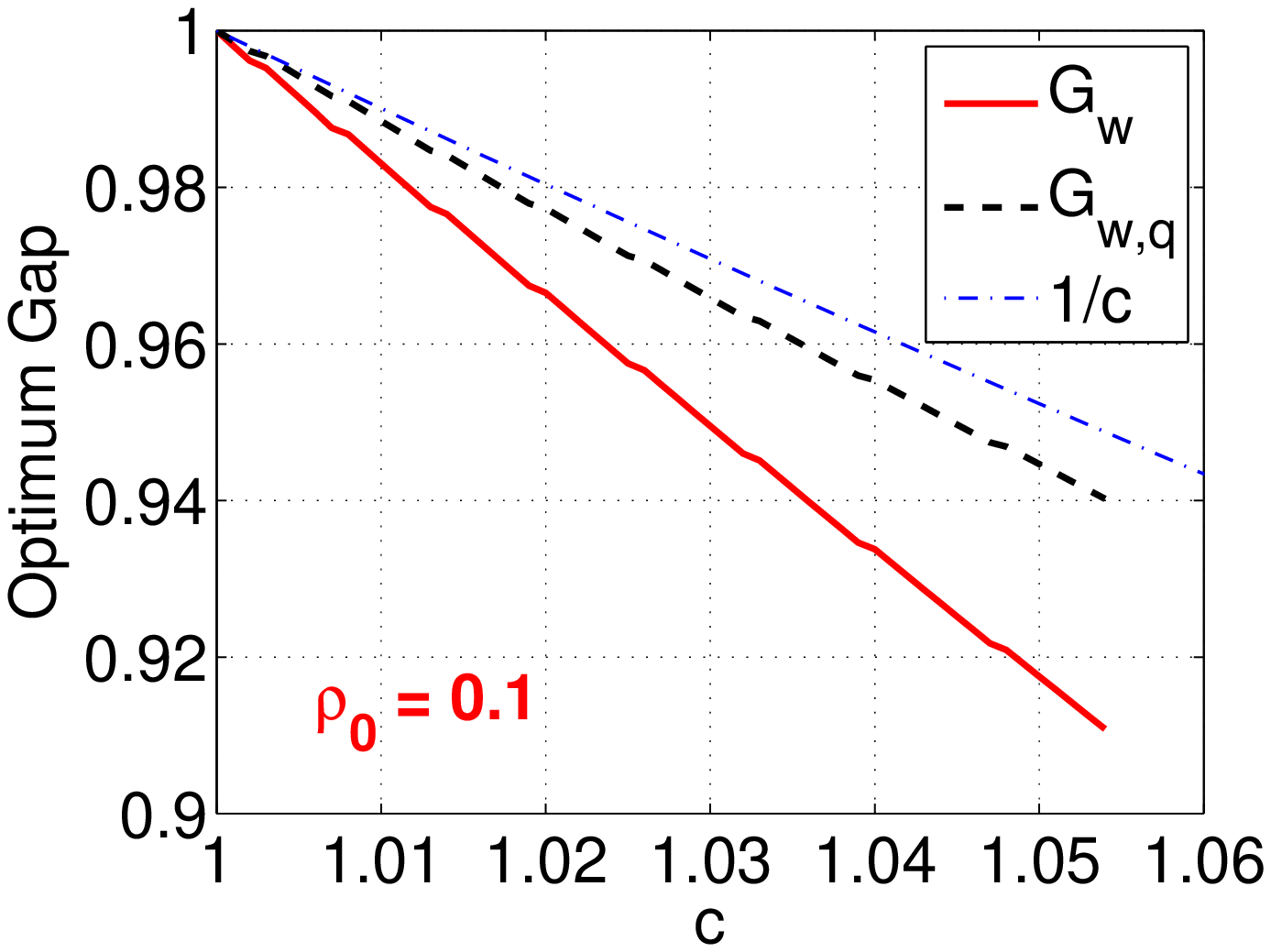}
\includegraphics[width = 2.2in]{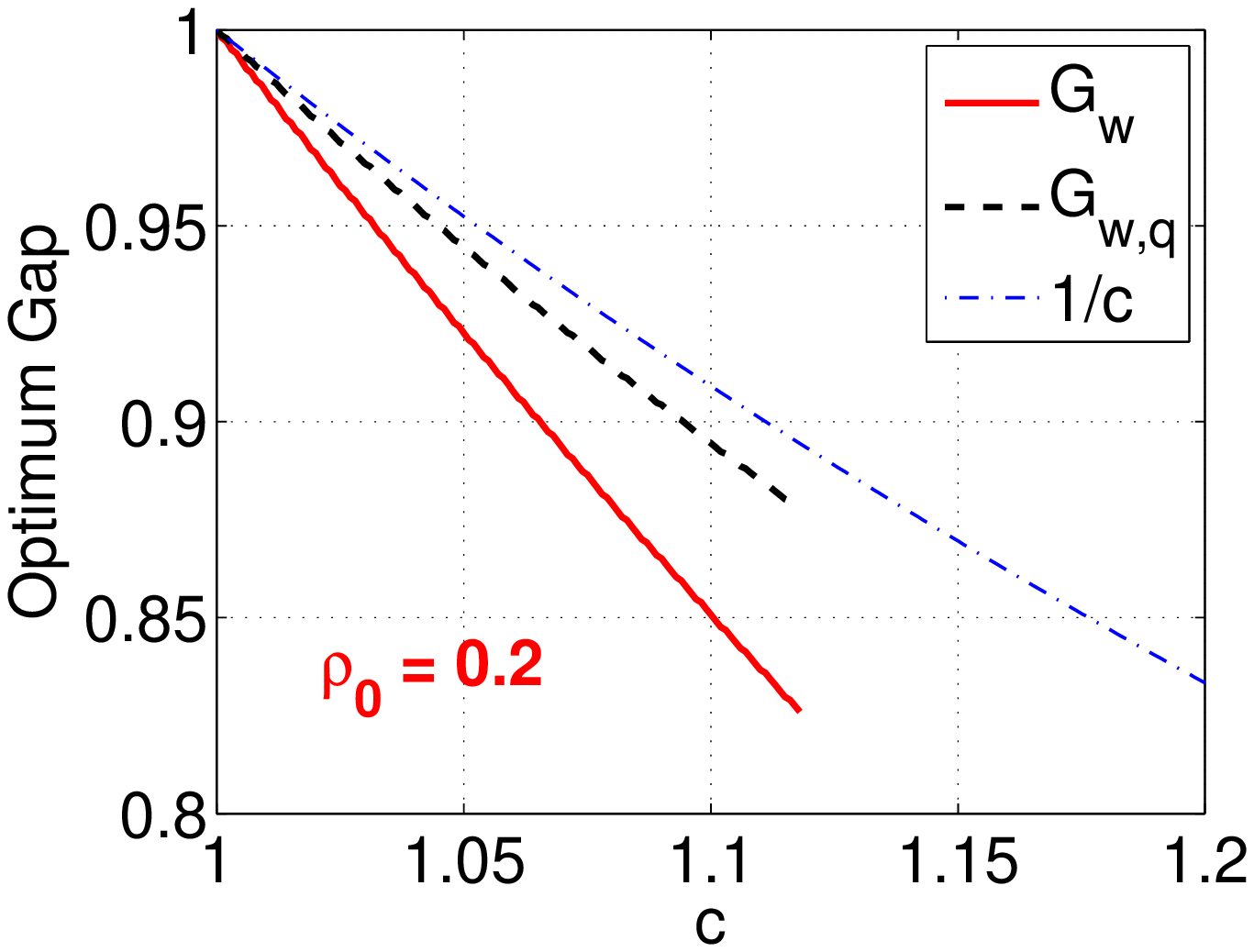}
\includegraphics[width = 2.2in]{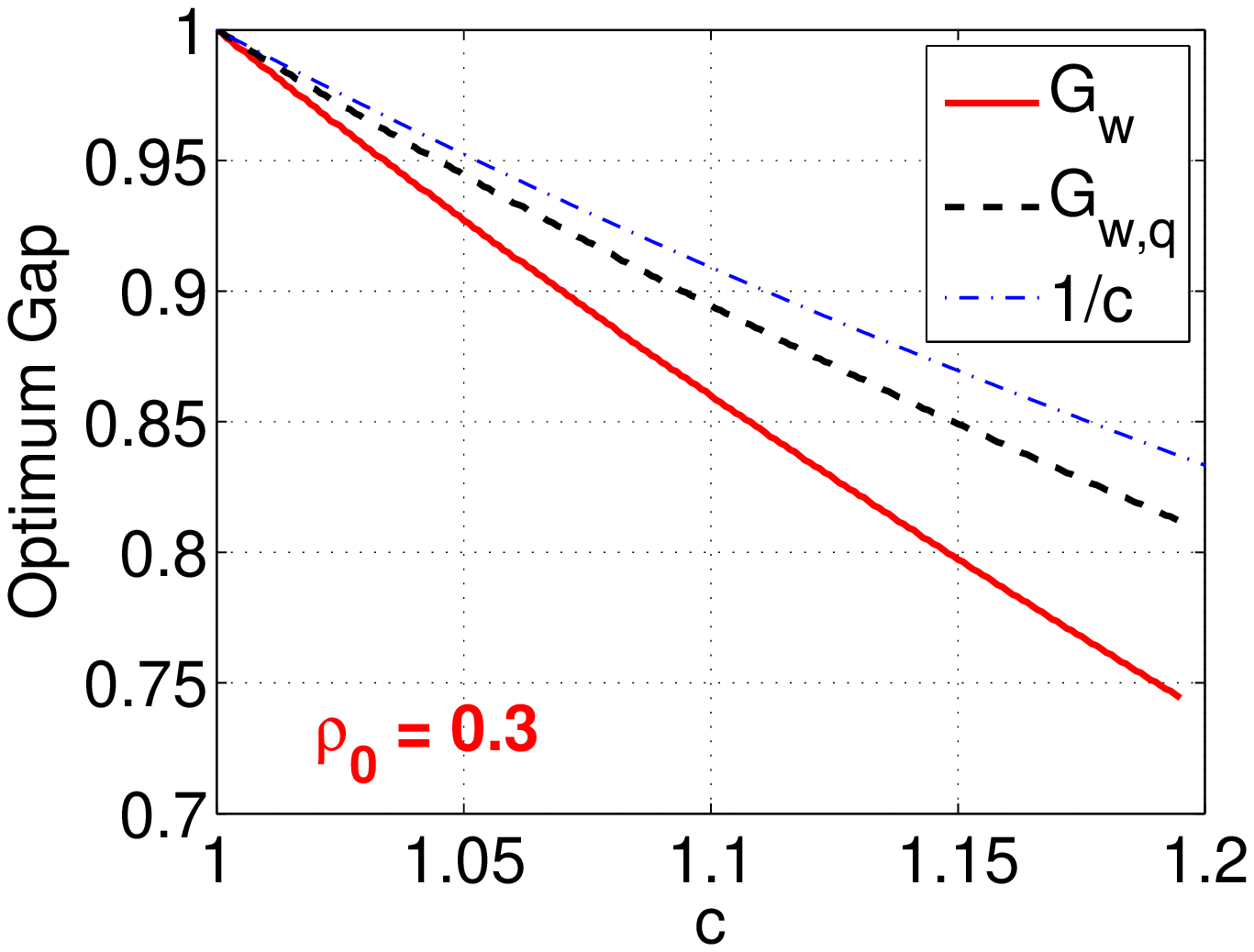}
}
\mbox{
\includegraphics[width = 2.2in]{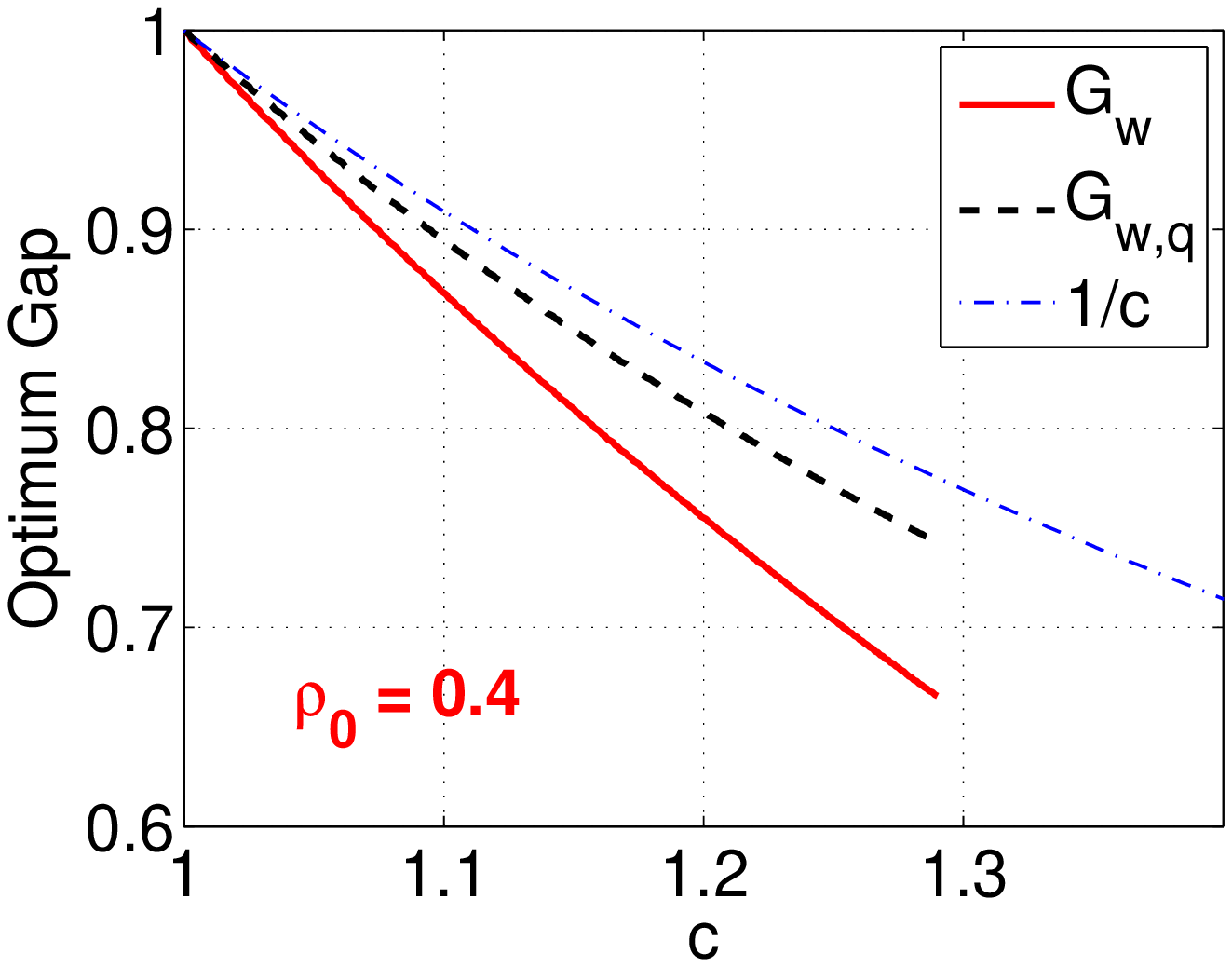}
\includegraphics[width = 2.2in]{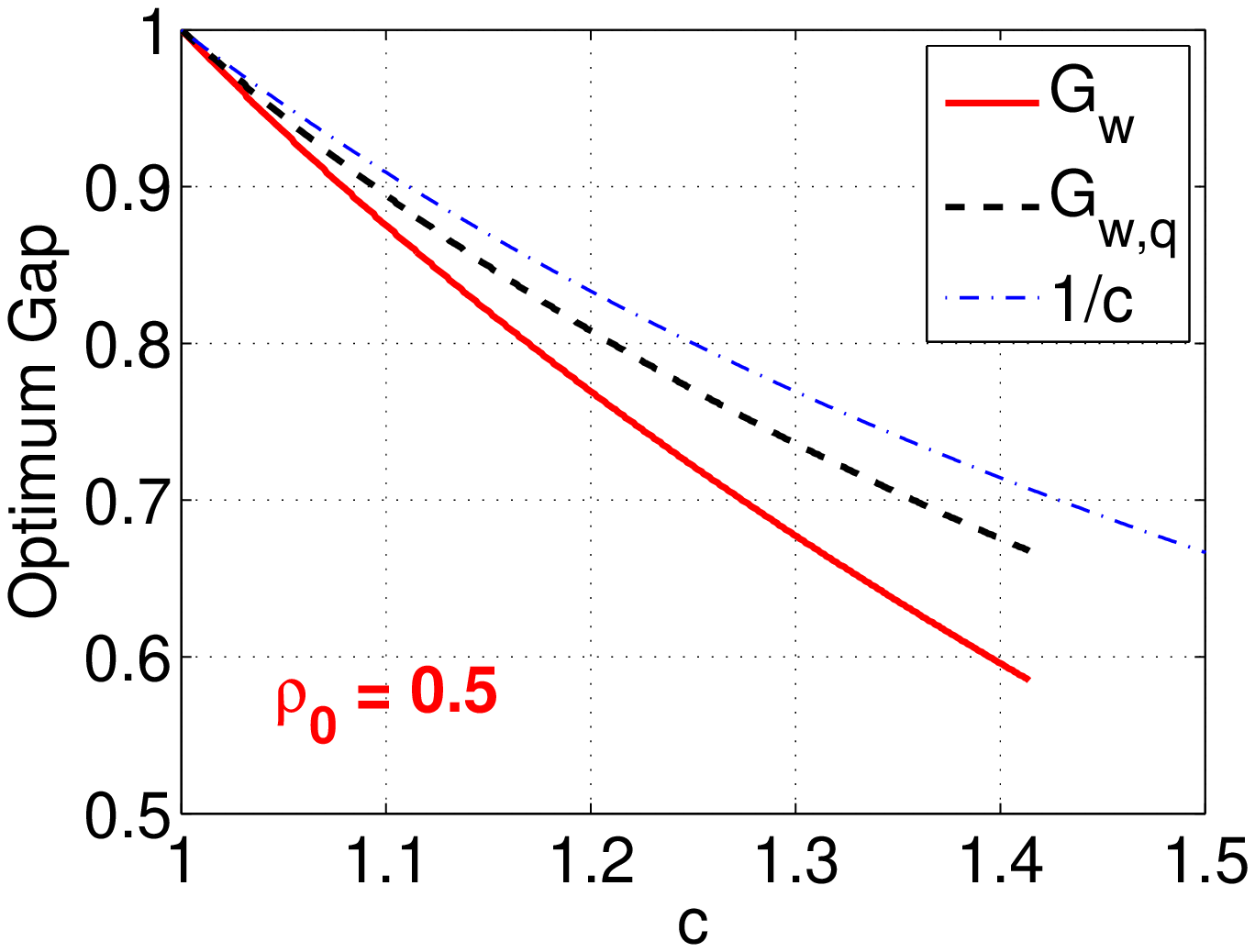}
\includegraphics[width = 2.2in]{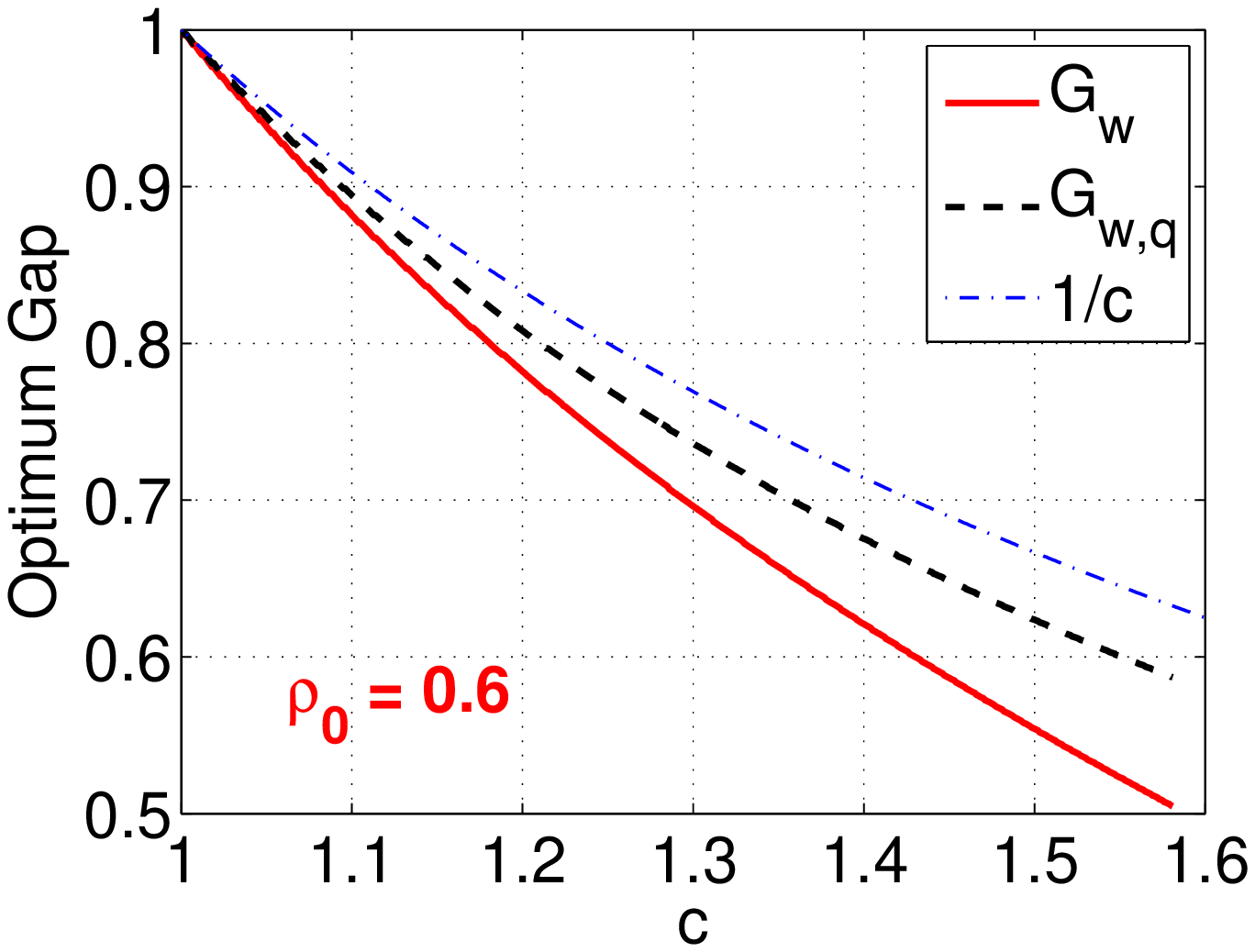}
}

\mbox{
\includegraphics[width = 2.2in]{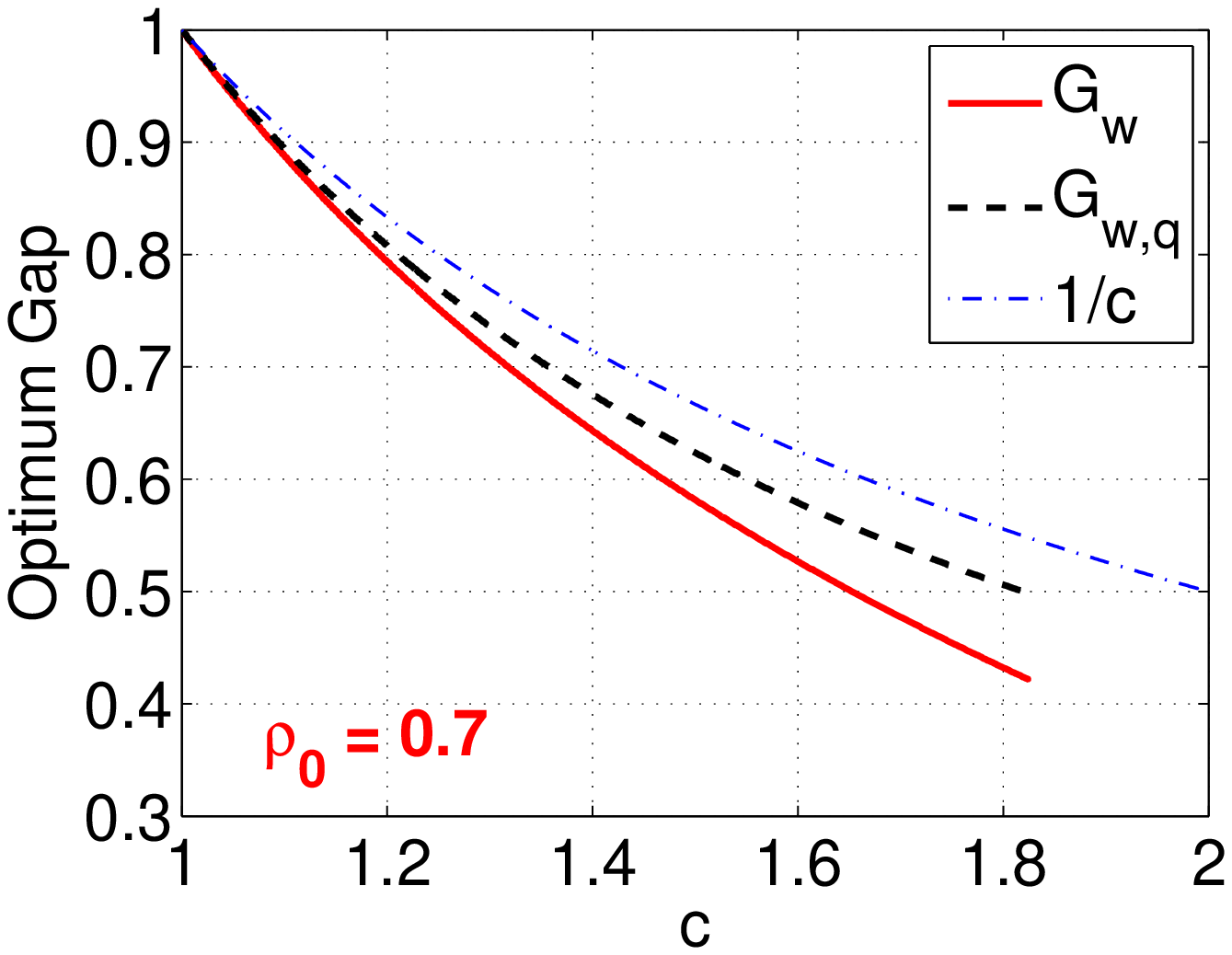}
\includegraphics[width = 2.2in]{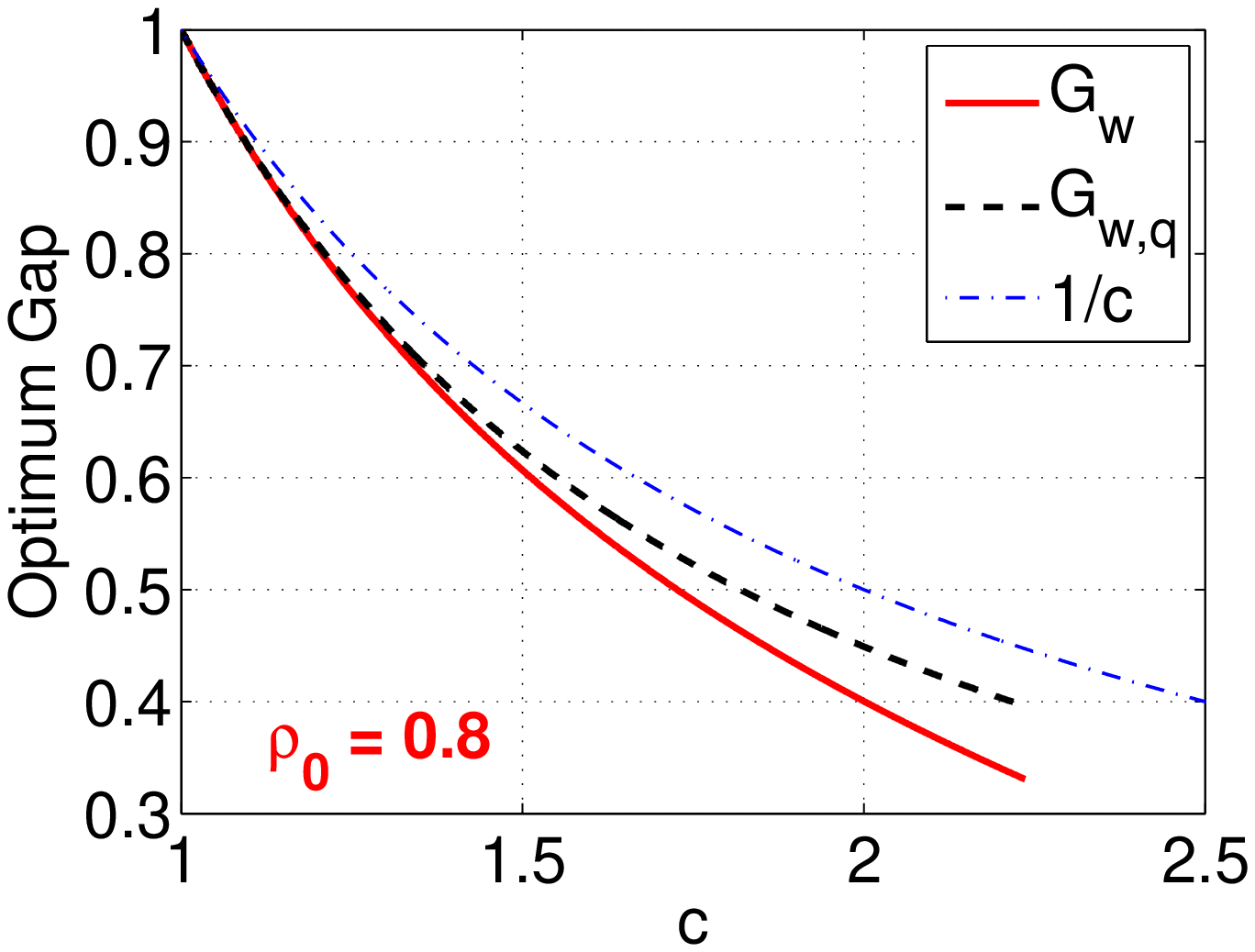}
\includegraphics[width = 2.2in]{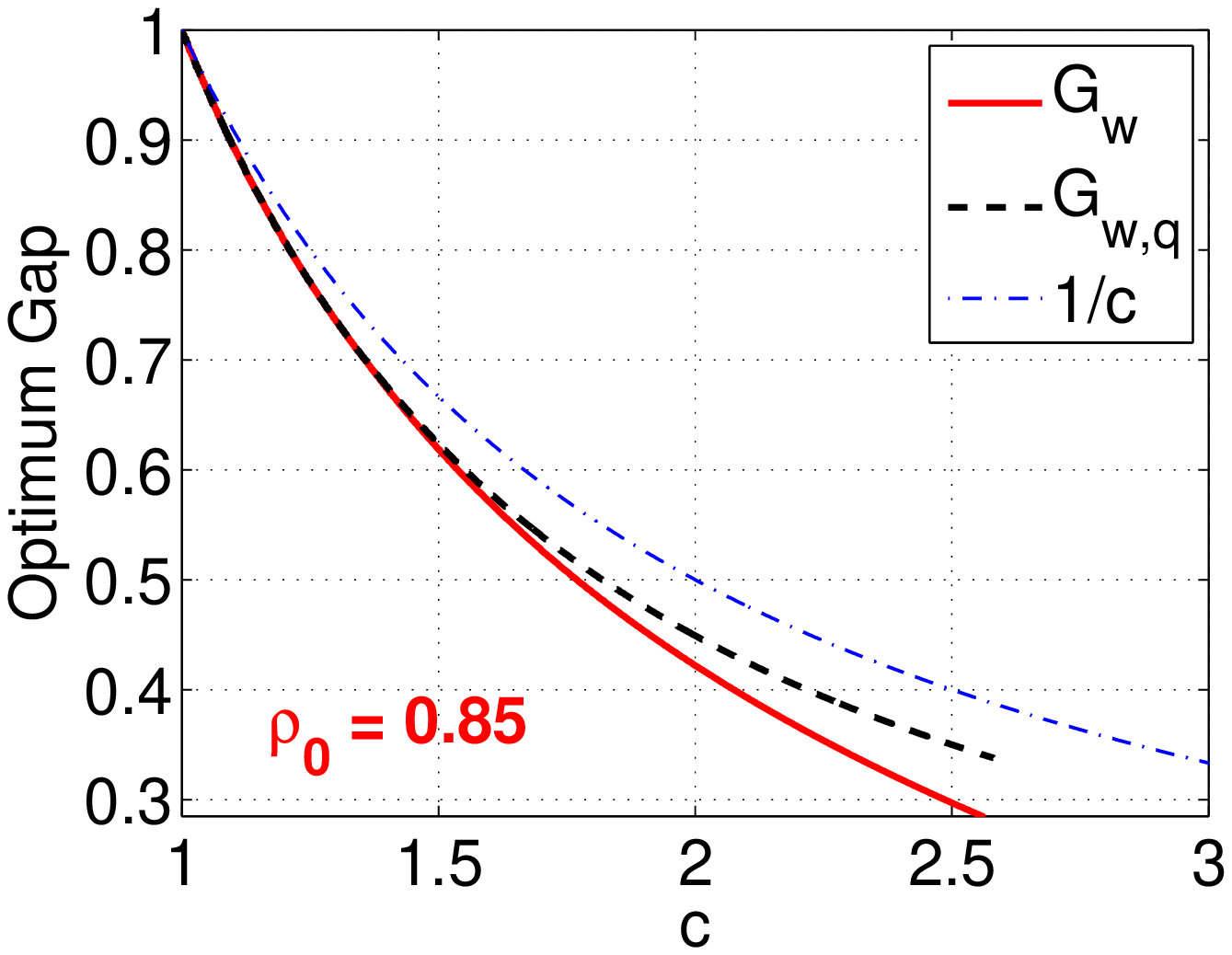}
}

\mbox{
\includegraphics[width = 2.2in]{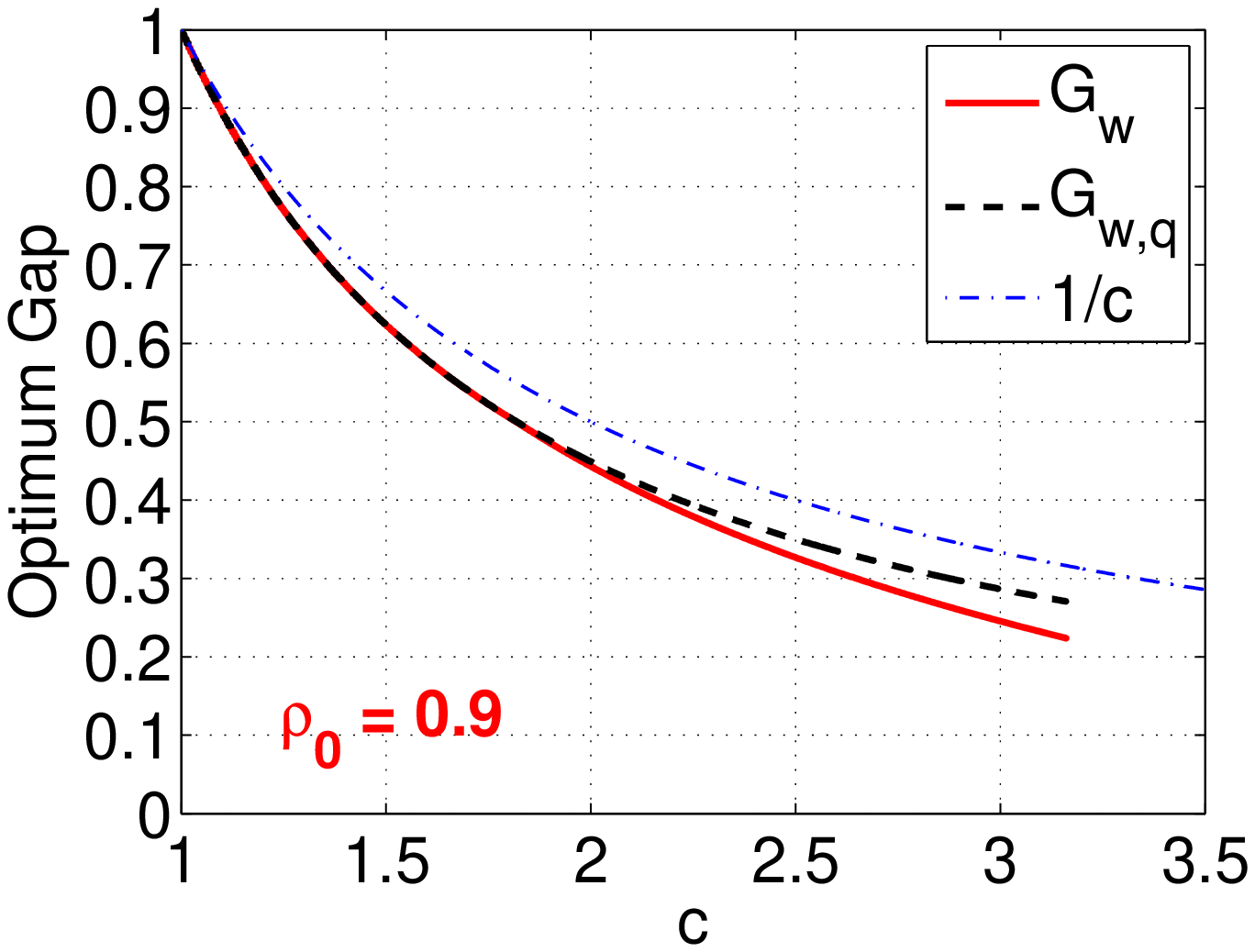}
\includegraphics[width = 2.2in]{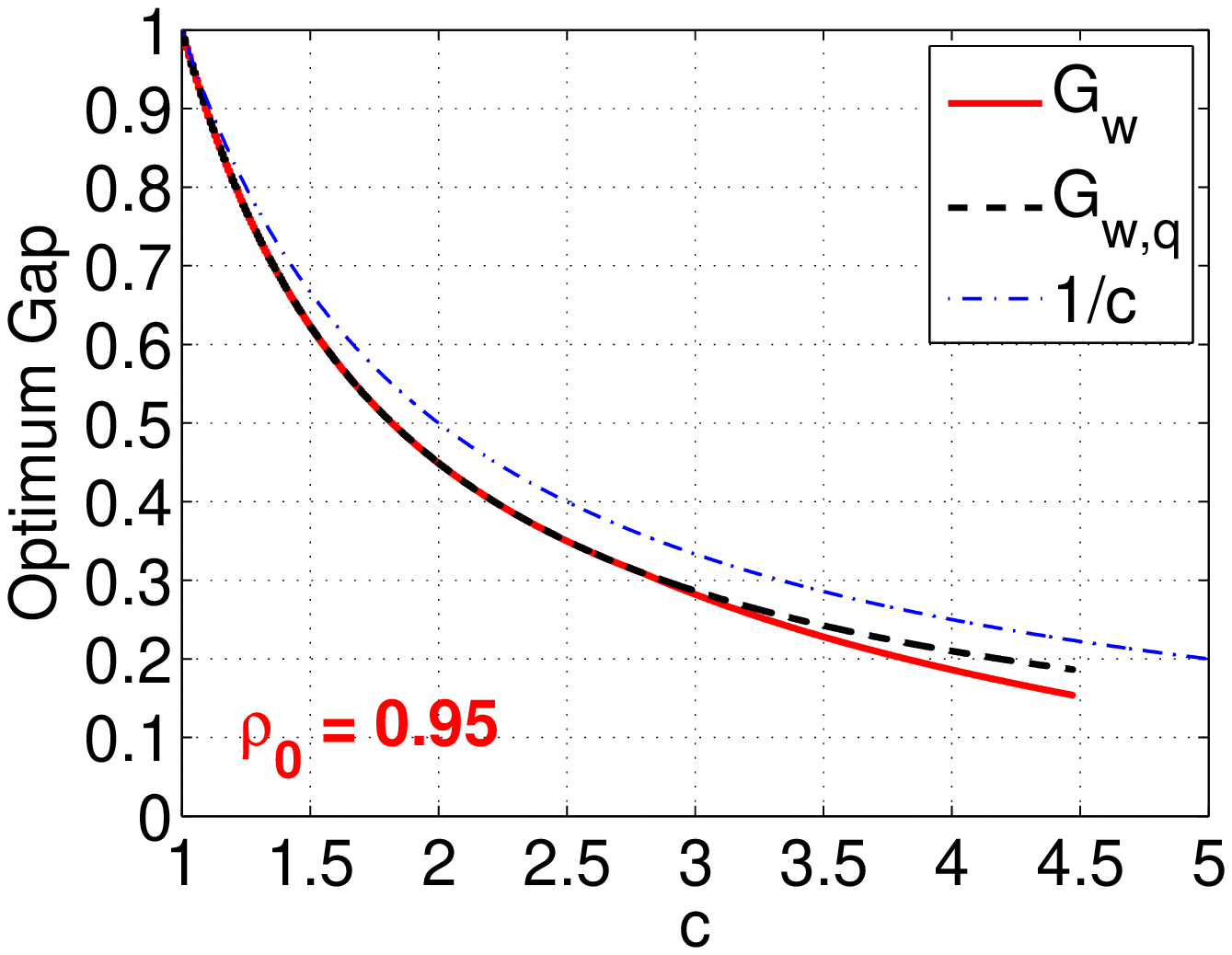}
\includegraphics[width = 2.2in]{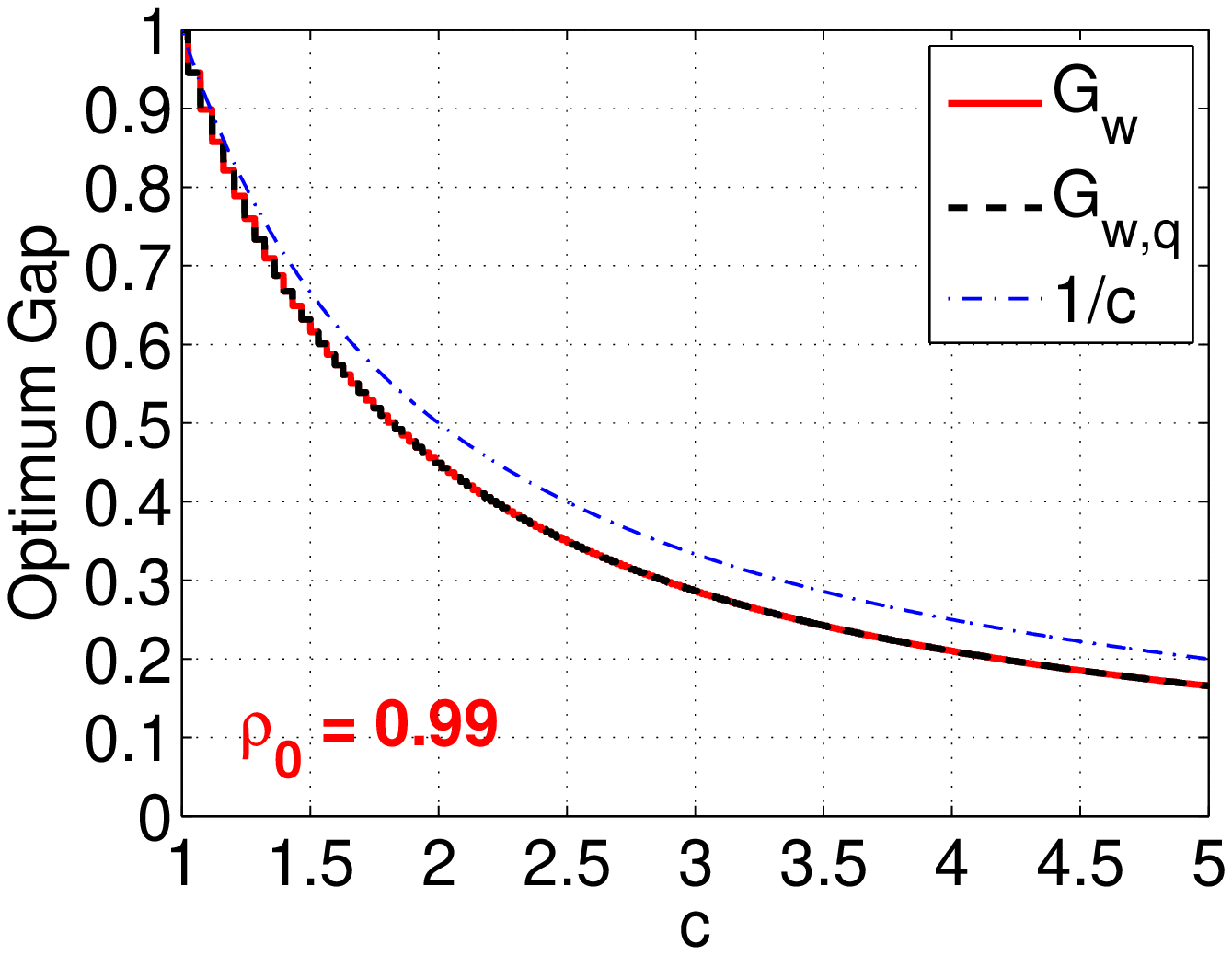}
}
\end{center}
\vspace{-.2in}
\caption{Comparison of the optimum gaps (smaller the better) for $h_w$ and $h_{w,q}$. For each $\rho_0$ and $c$, we can find the smallest gaps individually for $h_w$ and $h_{w,q}$, over the entire range of $w$. We can see that for all target similarity levels $\rho_0$, both $h_{w,q}$ and $h_w$ exhibit better performance than $1/c$. $h_w$ always has smaller gap than $h_{w,q}$, although in high similarity region both schemes perform similarly. }\label{fig_GwqOpt}
\end{figure}

Figure~\ref{fig_GwqOpt} compares $G_w$ with $G_{w,q}$ at their ``optimum'' $w$ values, as functions of $c$, for a wide range of target similarity $\rho_0$ levels. Basically, at each $c$ and $\rho_0$, we choose the $w$ to minimize $G_w$ and the  $w$ to minimize $G_{w,q}$. This figure illustrates that $G_w$ is smaller than $G_{w,q}$, noticeably so in the low similarity region.\\

\begin{figure}[h!]
\begin{center}
\mbox{
\includegraphics[width = 2.2in]{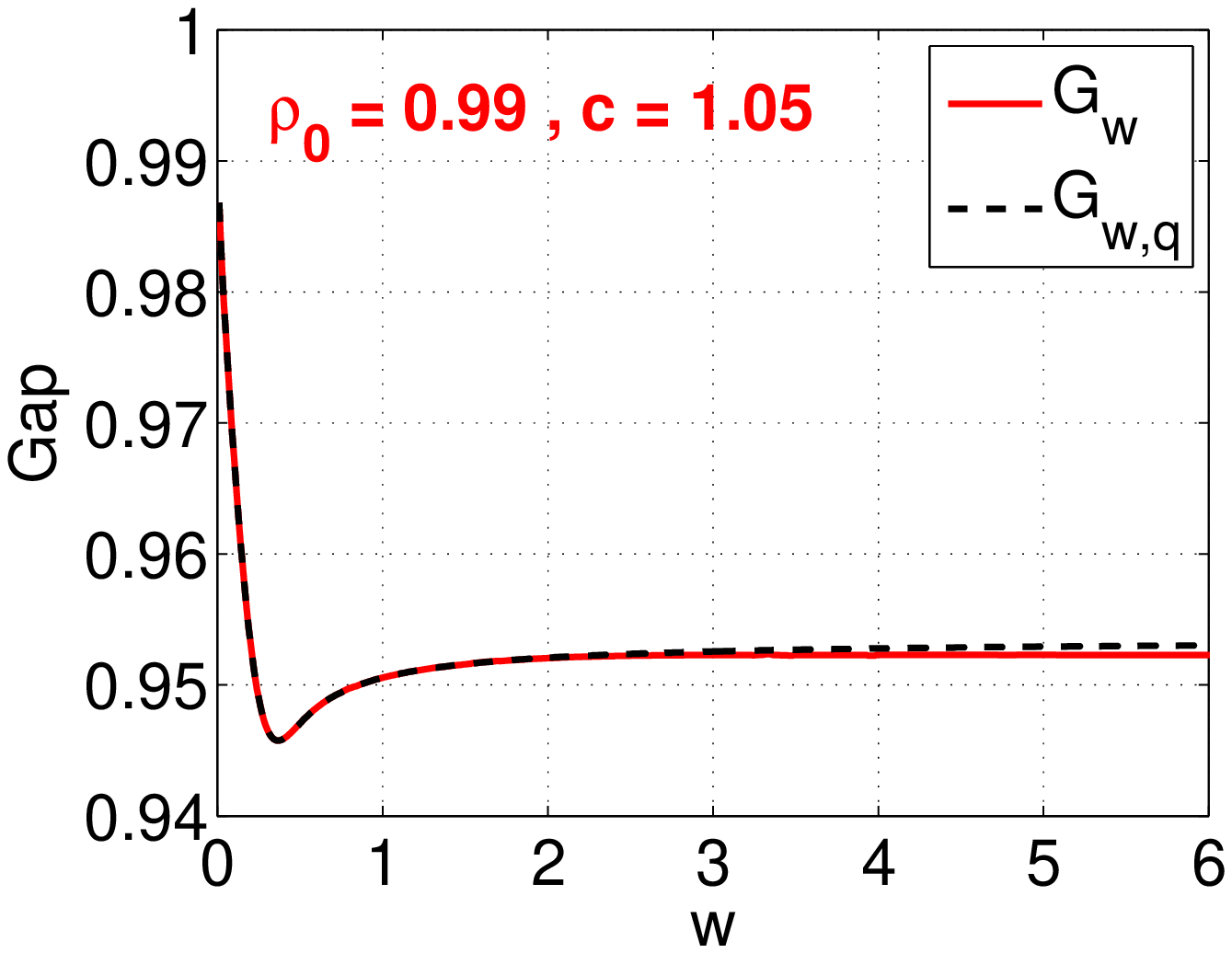}
\includegraphics[width = 2.2in]{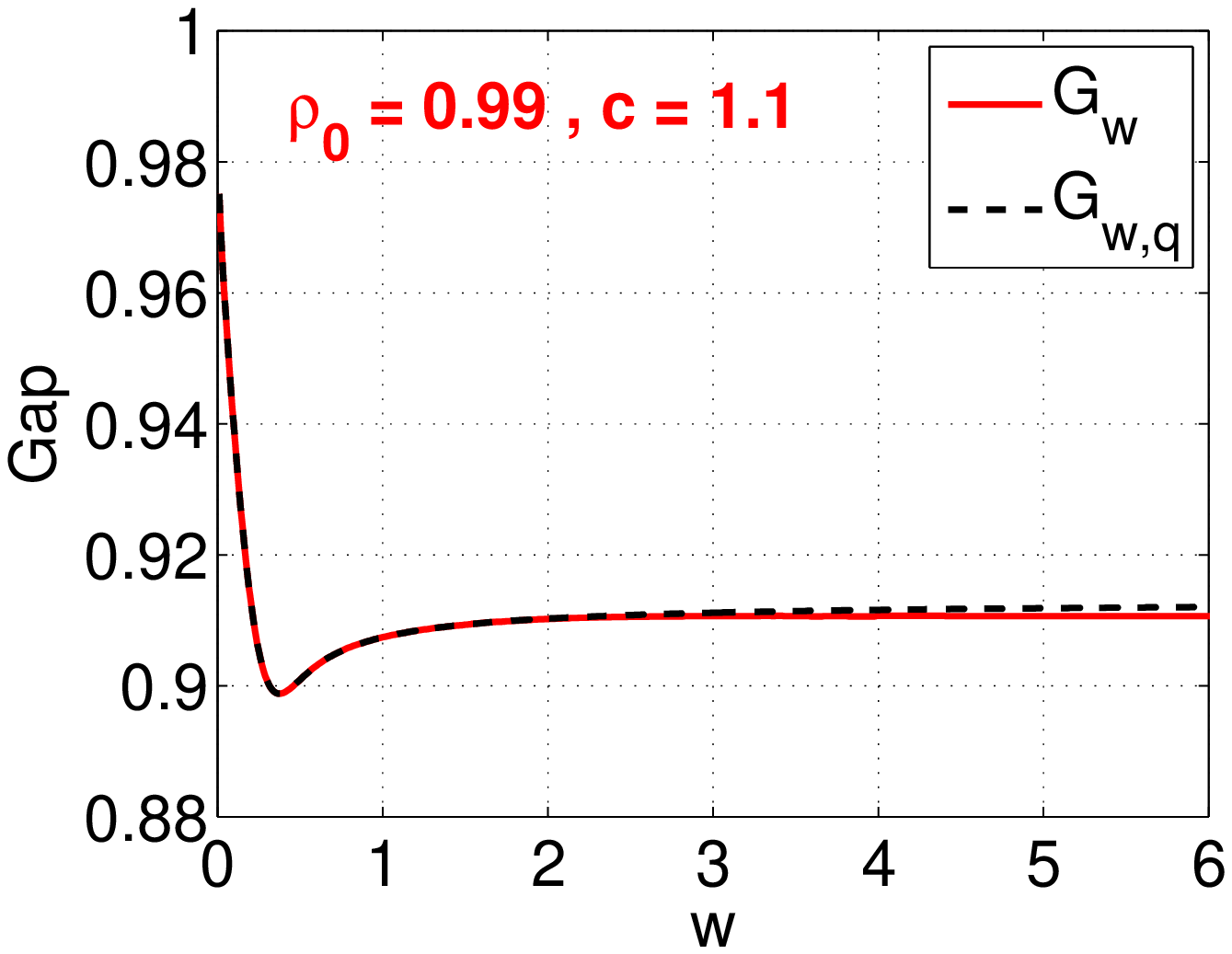}
\includegraphics[width = 2.2in]{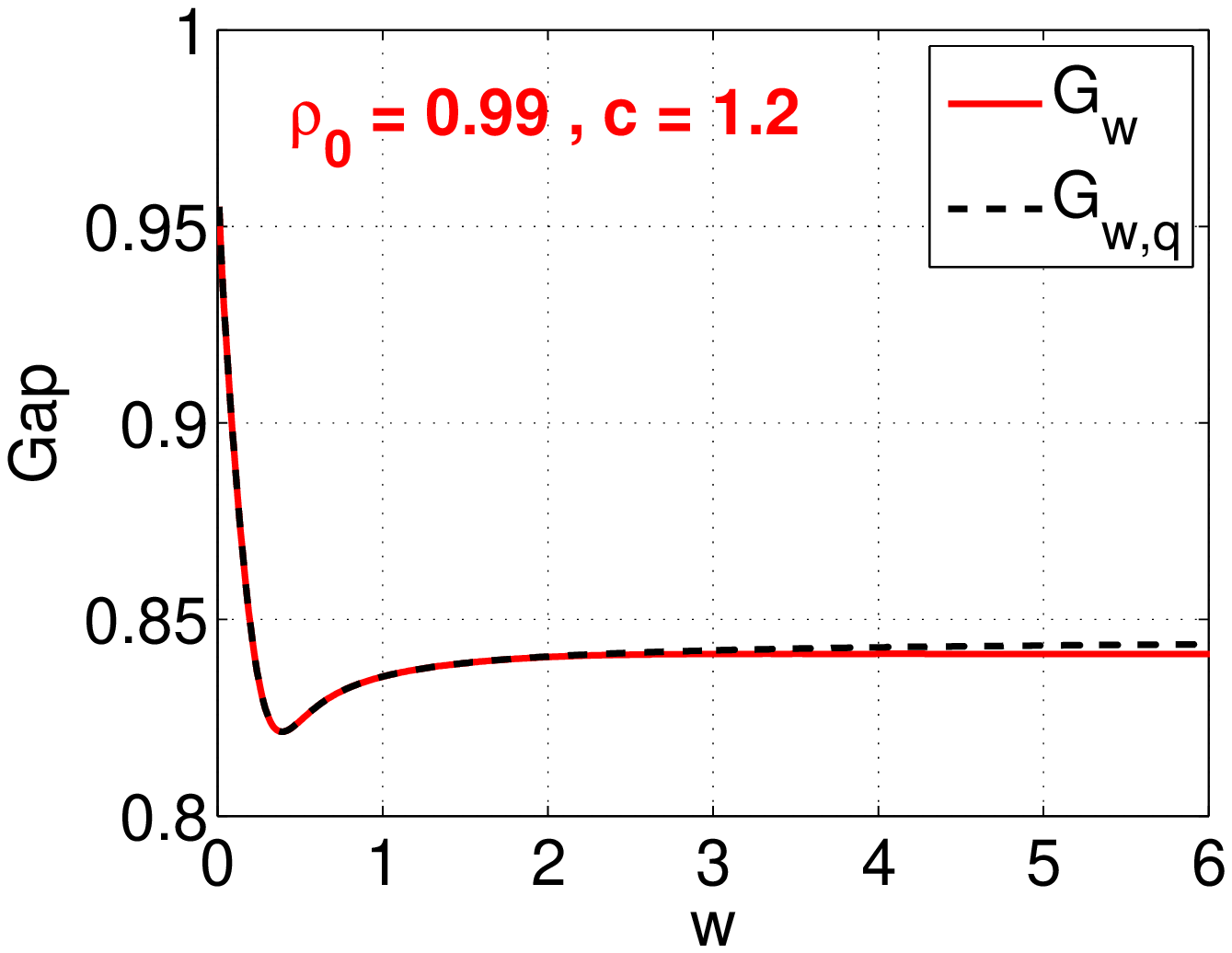}
}
\mbox{
\includegraphics[width = 2.2in]{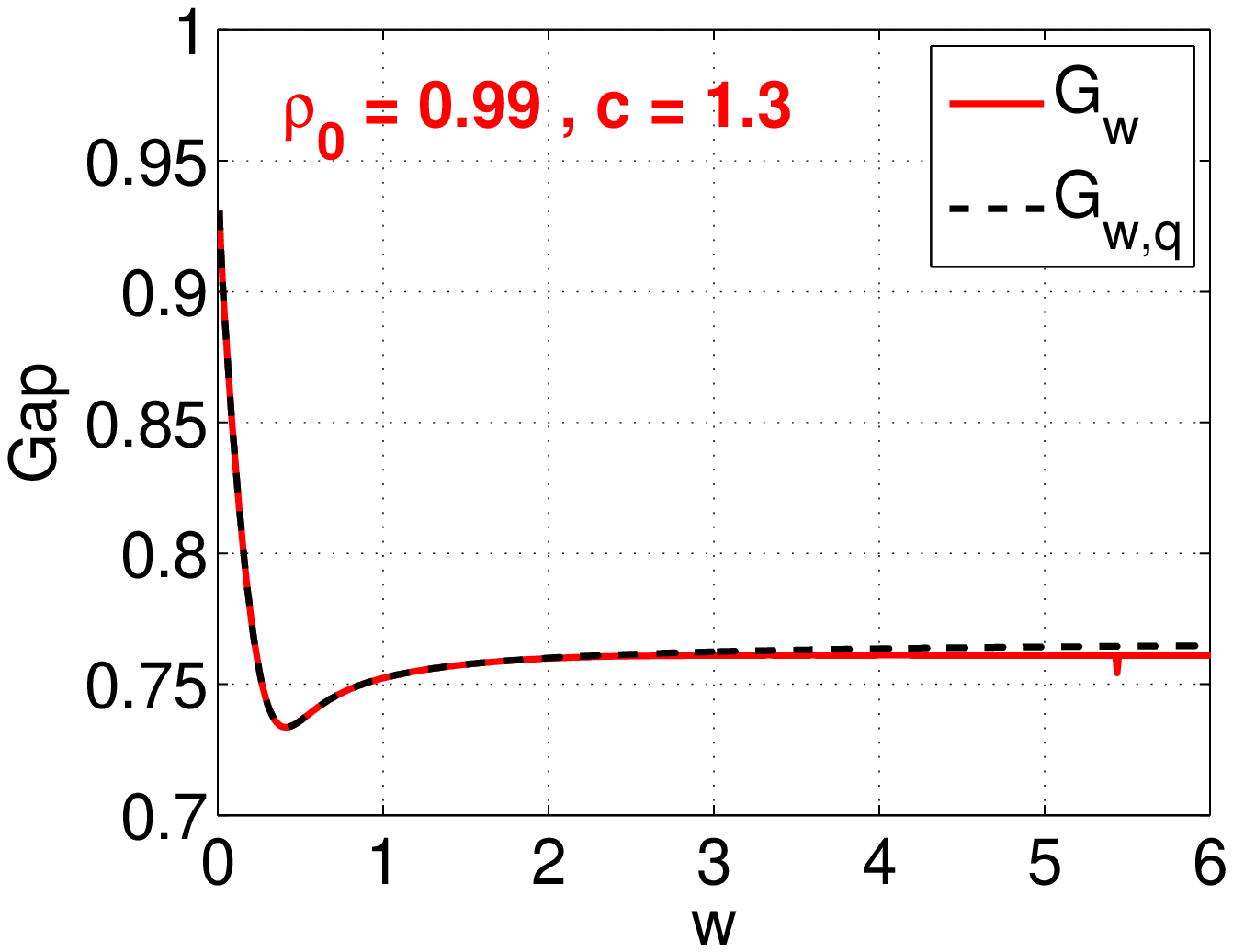}
\includegraphics[width = 2.2in]{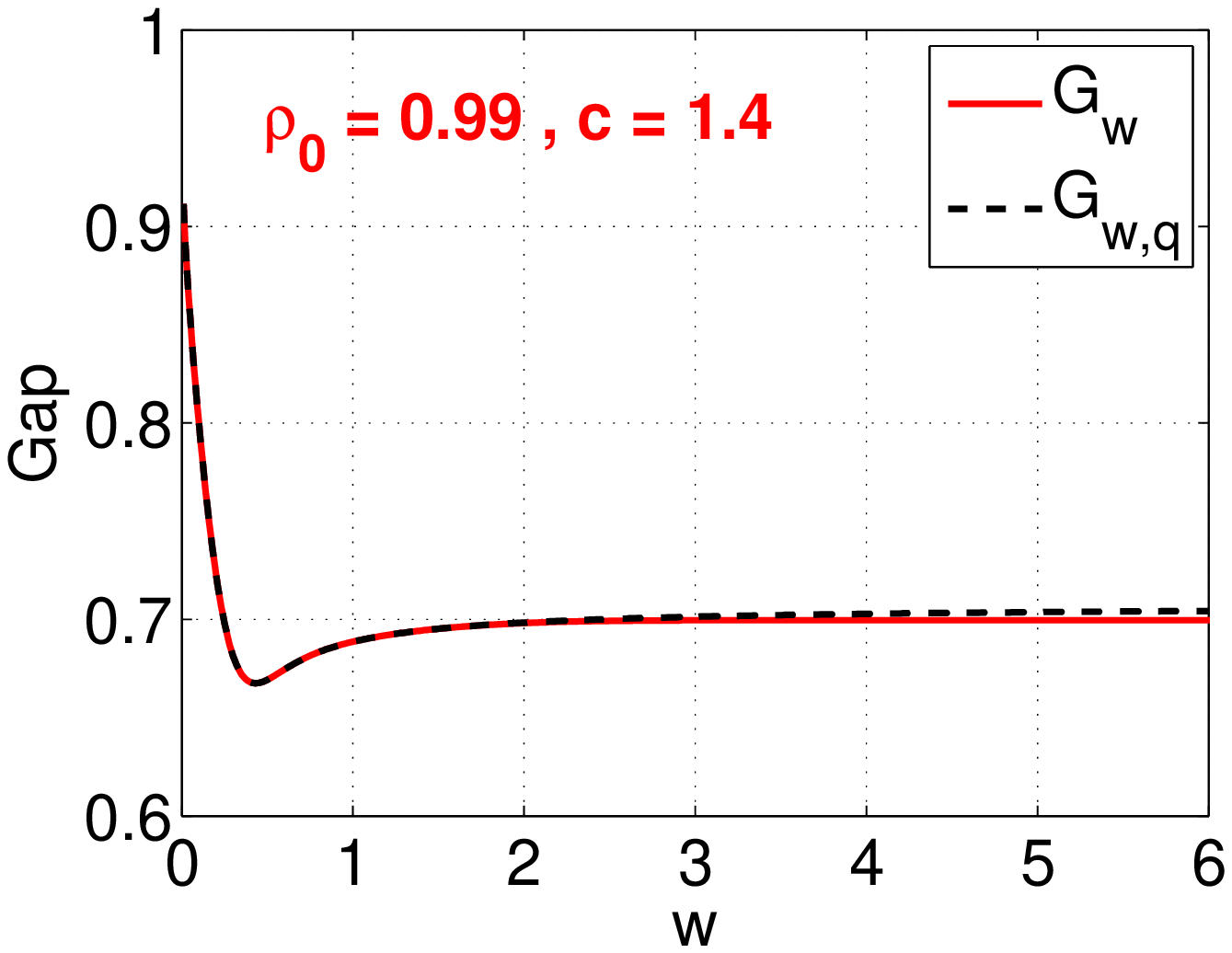}
\includegraphics[width = 2.2in]{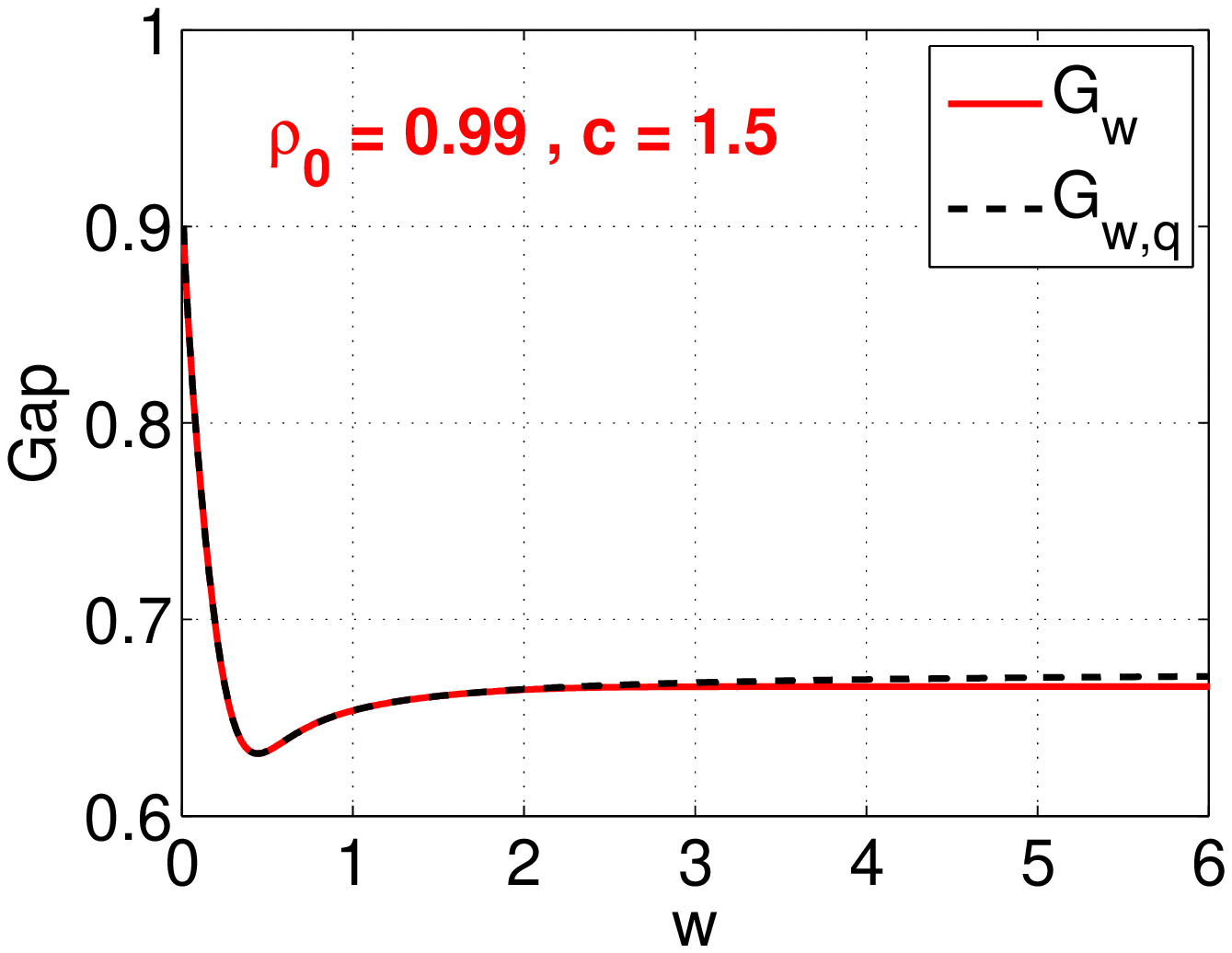}
}

\mbox{
\includegraphics[width = 2.2in]{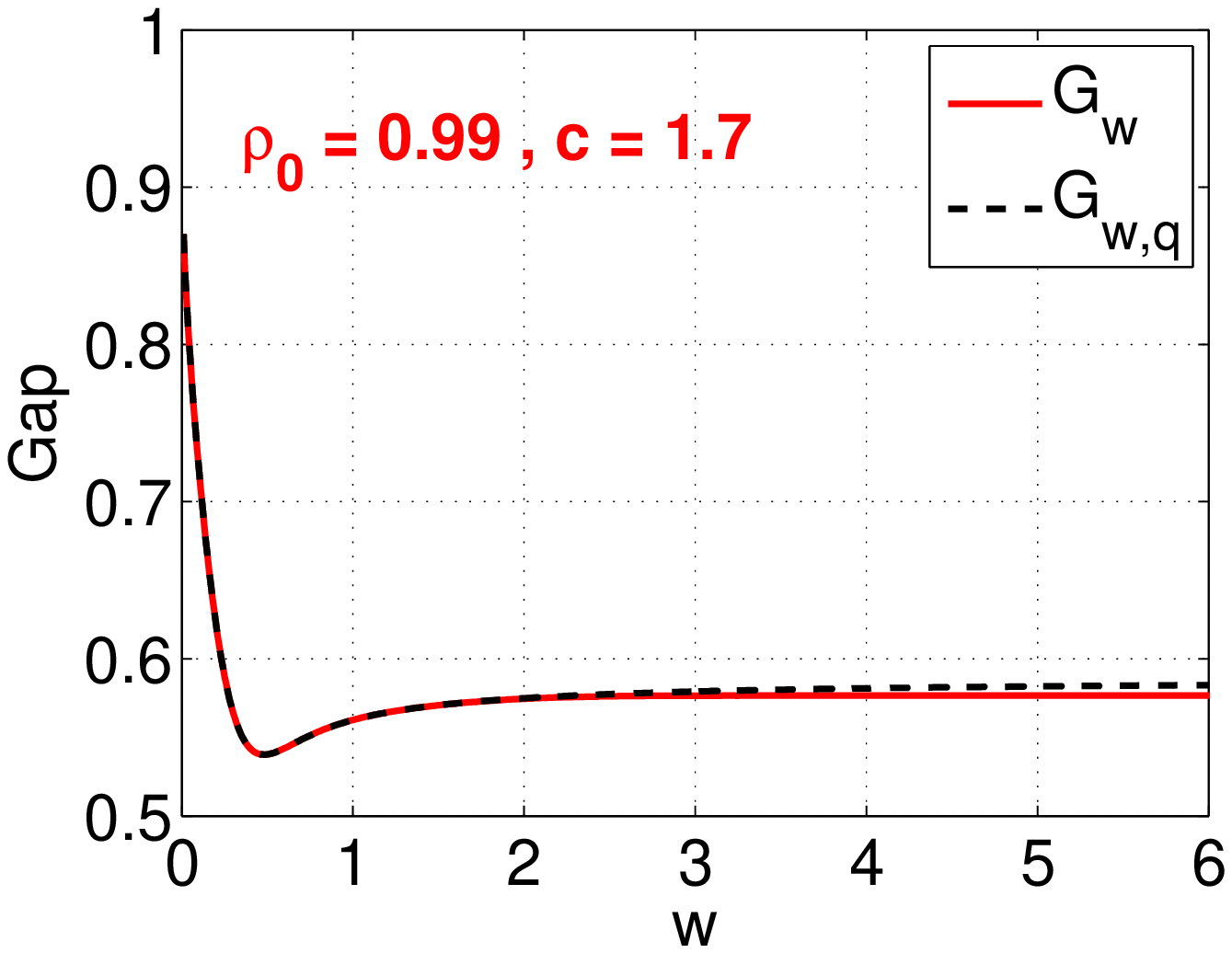}
\includegraphics[width = 2.2in]{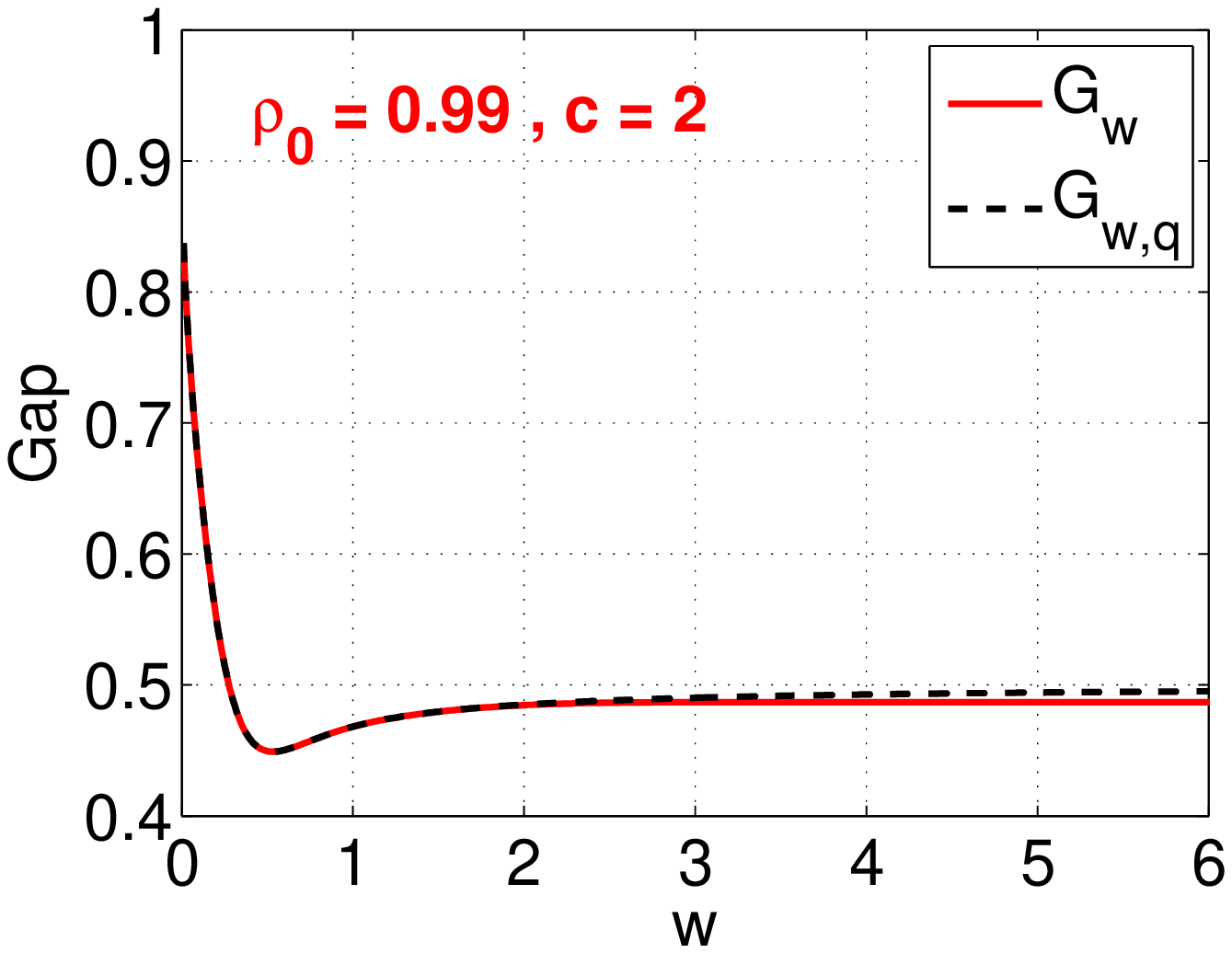}
\includegraphics[width = 2.2in]{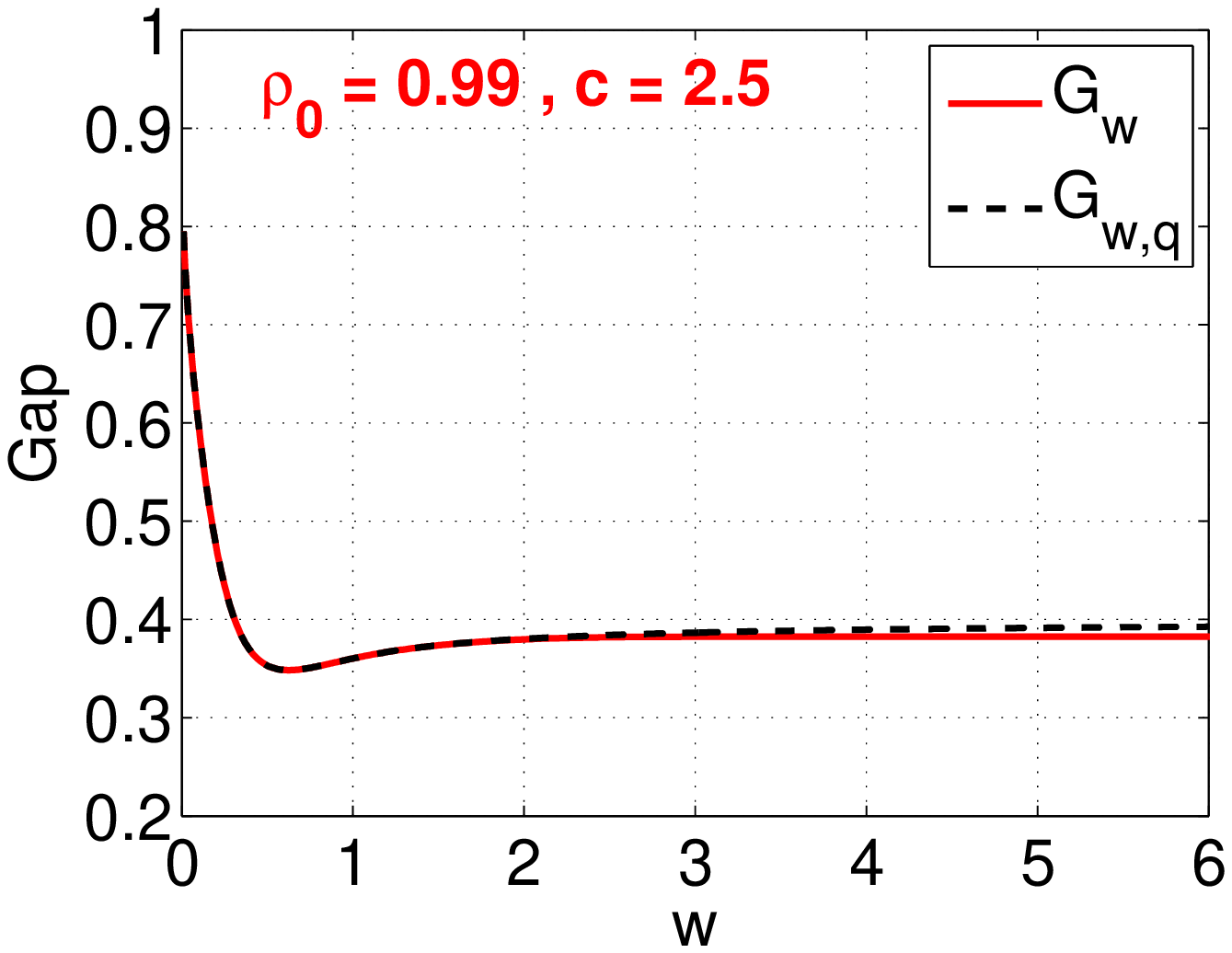}
}

\mbox{
\includegraphics[width = 2.2in]{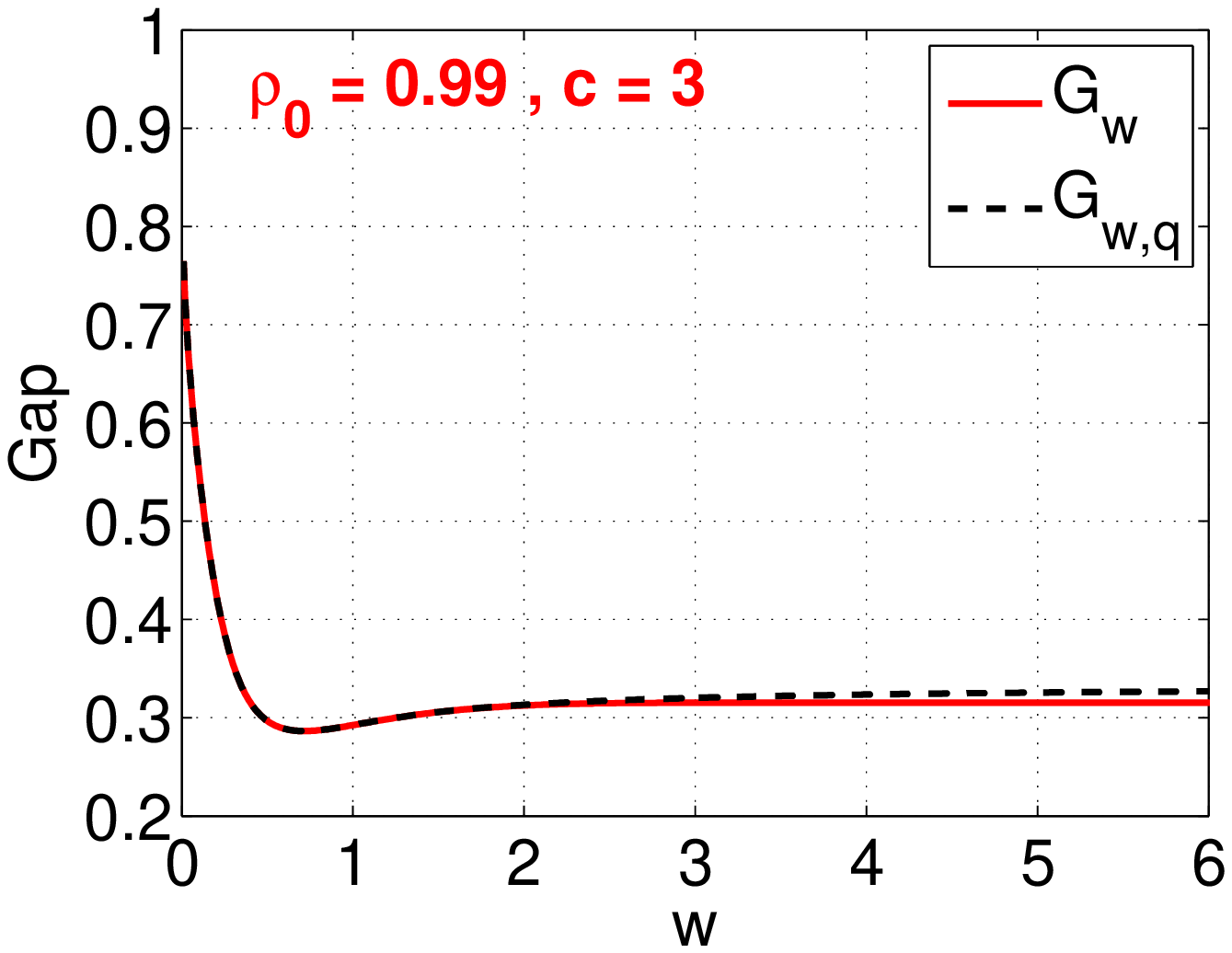}
\includegraphics[width = 2.2in]{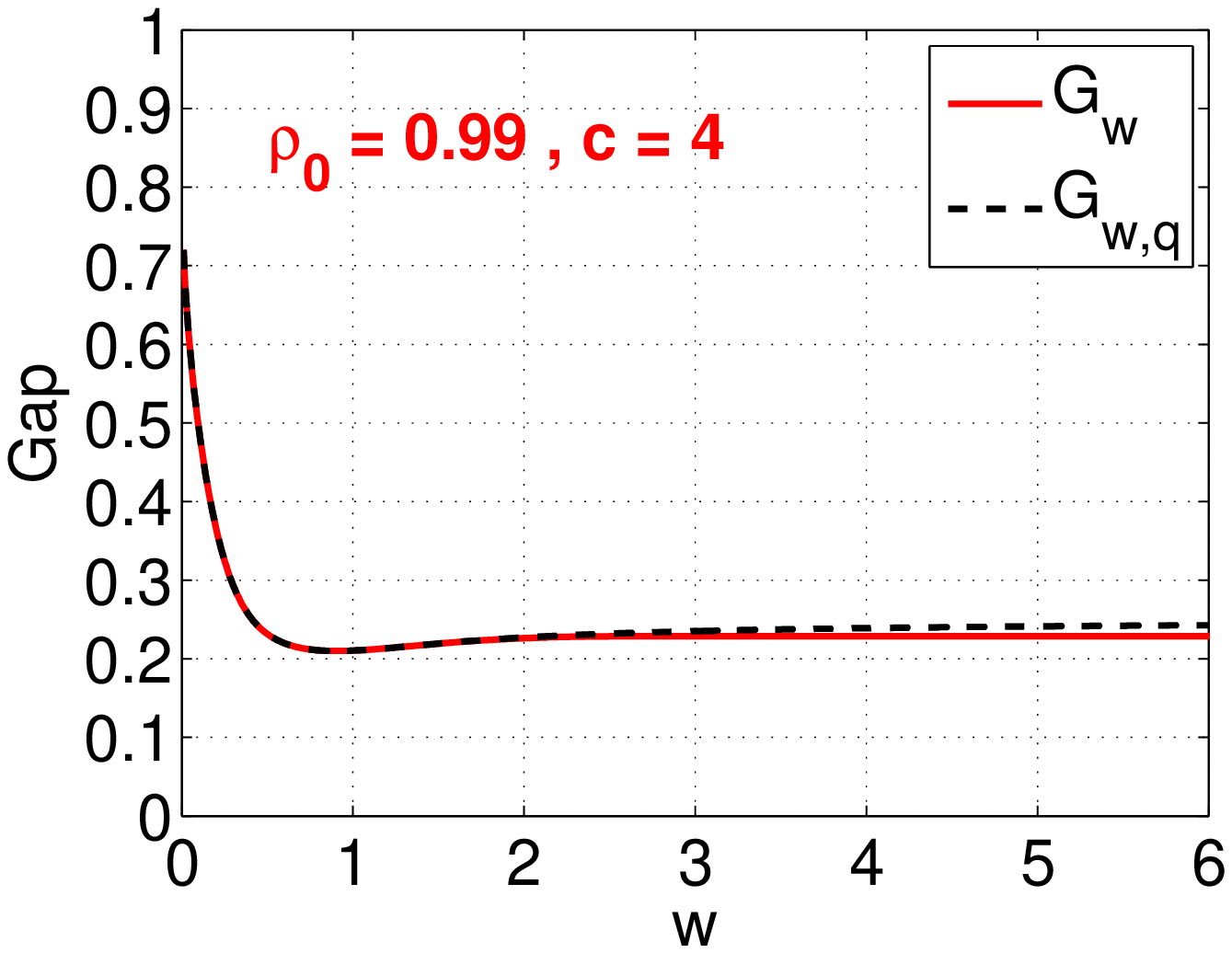}
\includegraphics[width = 2.2in]{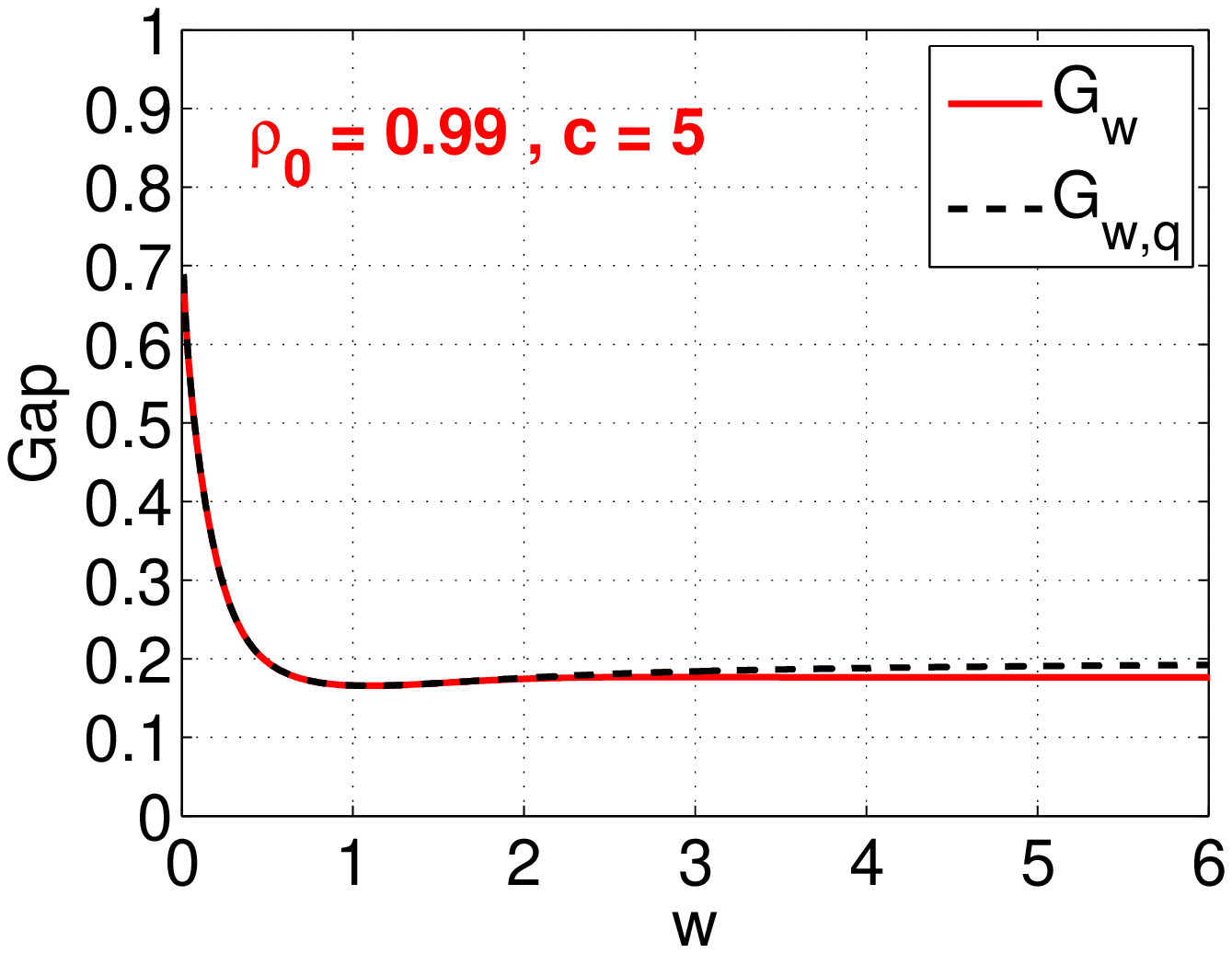}
}

\end{center}
\vspace{-.2in}
\caption{The gaps $G_w$ and $G_{w,q}$ as functions of $w$, for $\rho_0 = 0.99$. In each panel, we plot both $G_w$ and $G_{w,q}$ for a particular $c$ value.  The plots illustrate where the optimum $w$ values are obtained. }\label{fig_GwqR099C}
\end{figure}

\begin{figure}[h!]
\begin{center}
\mbox{
\includegraphics[width = 2.2in]{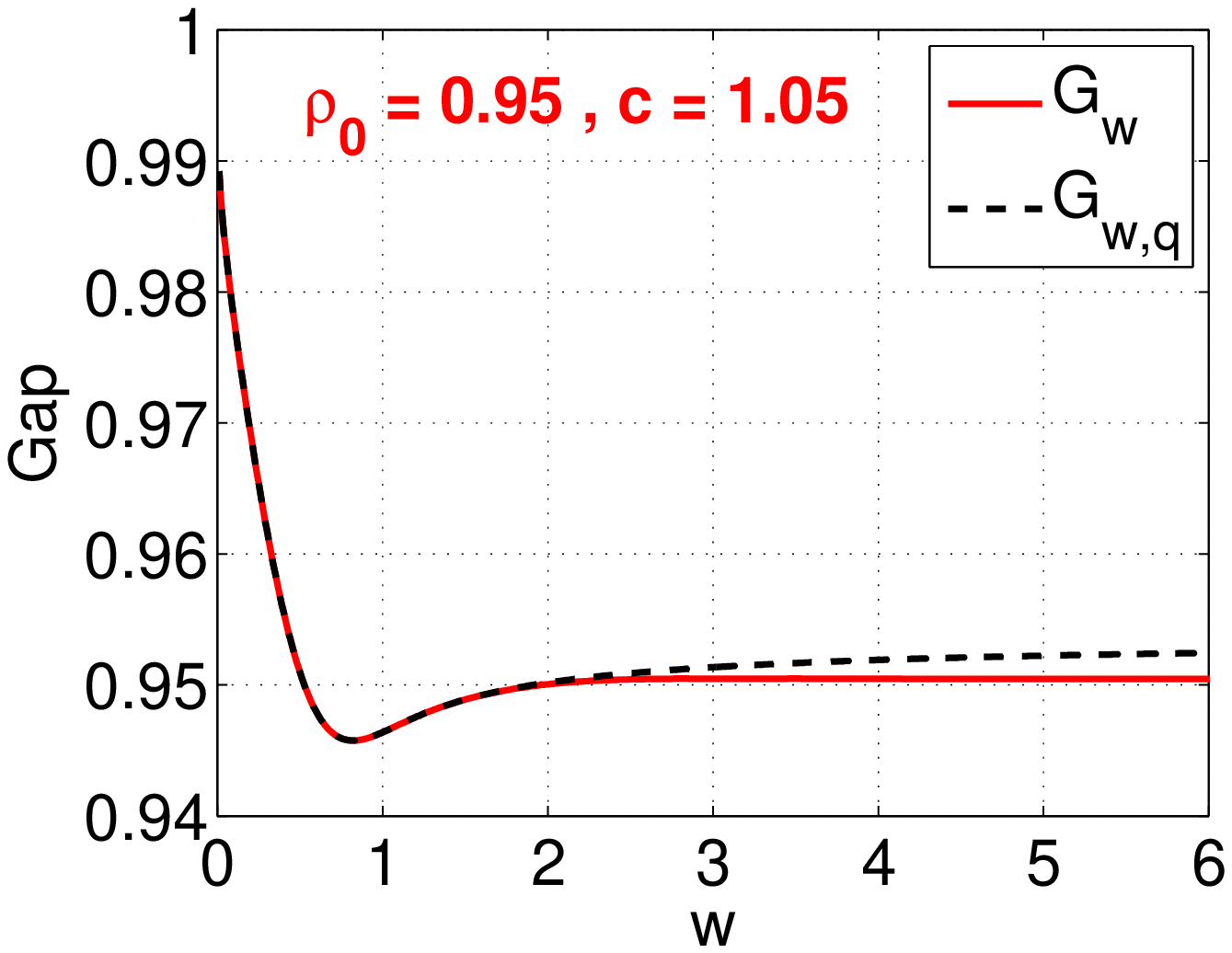}
\includegraphics[width = 2.2in]{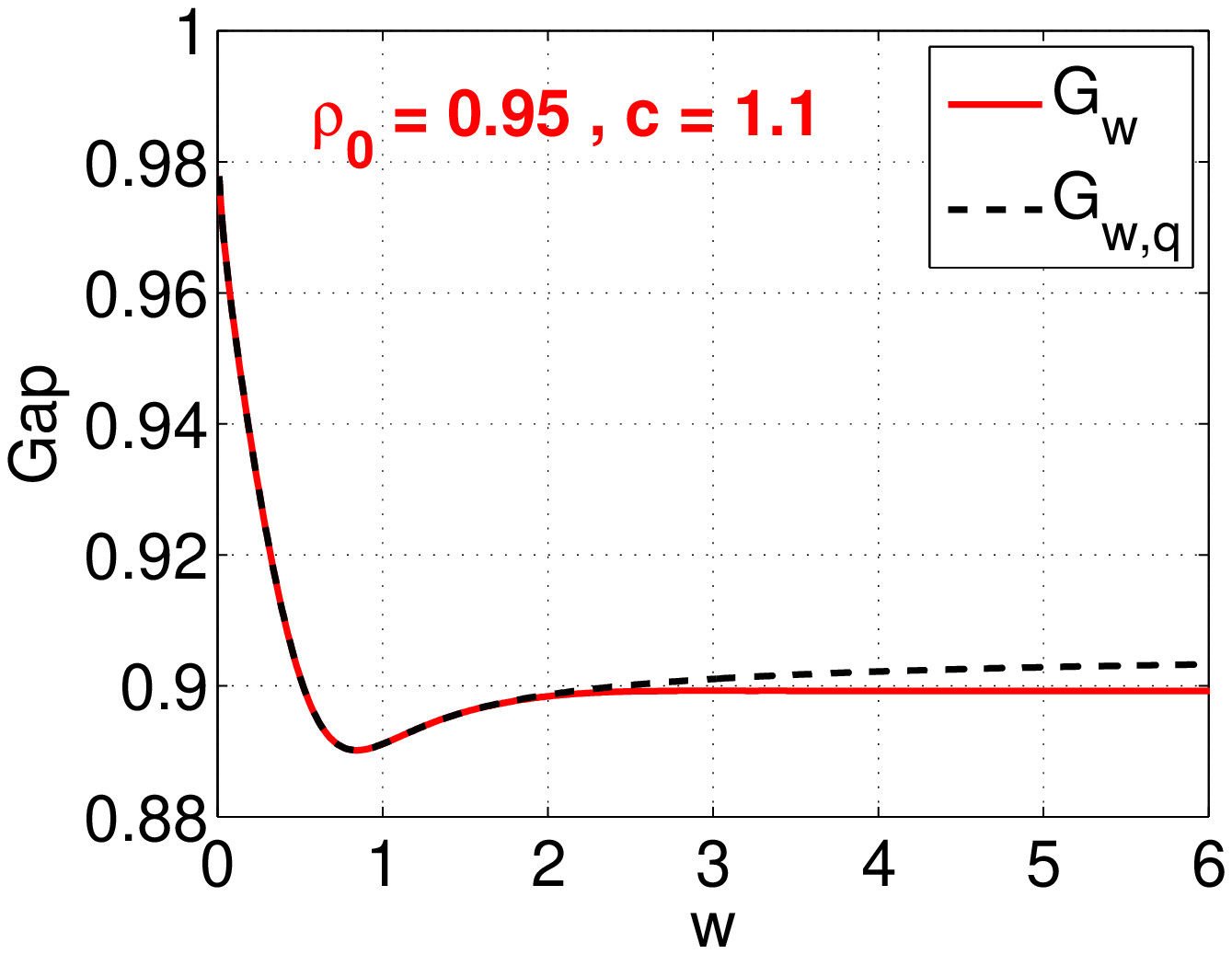}
\includegraphics[width = 2.2in]{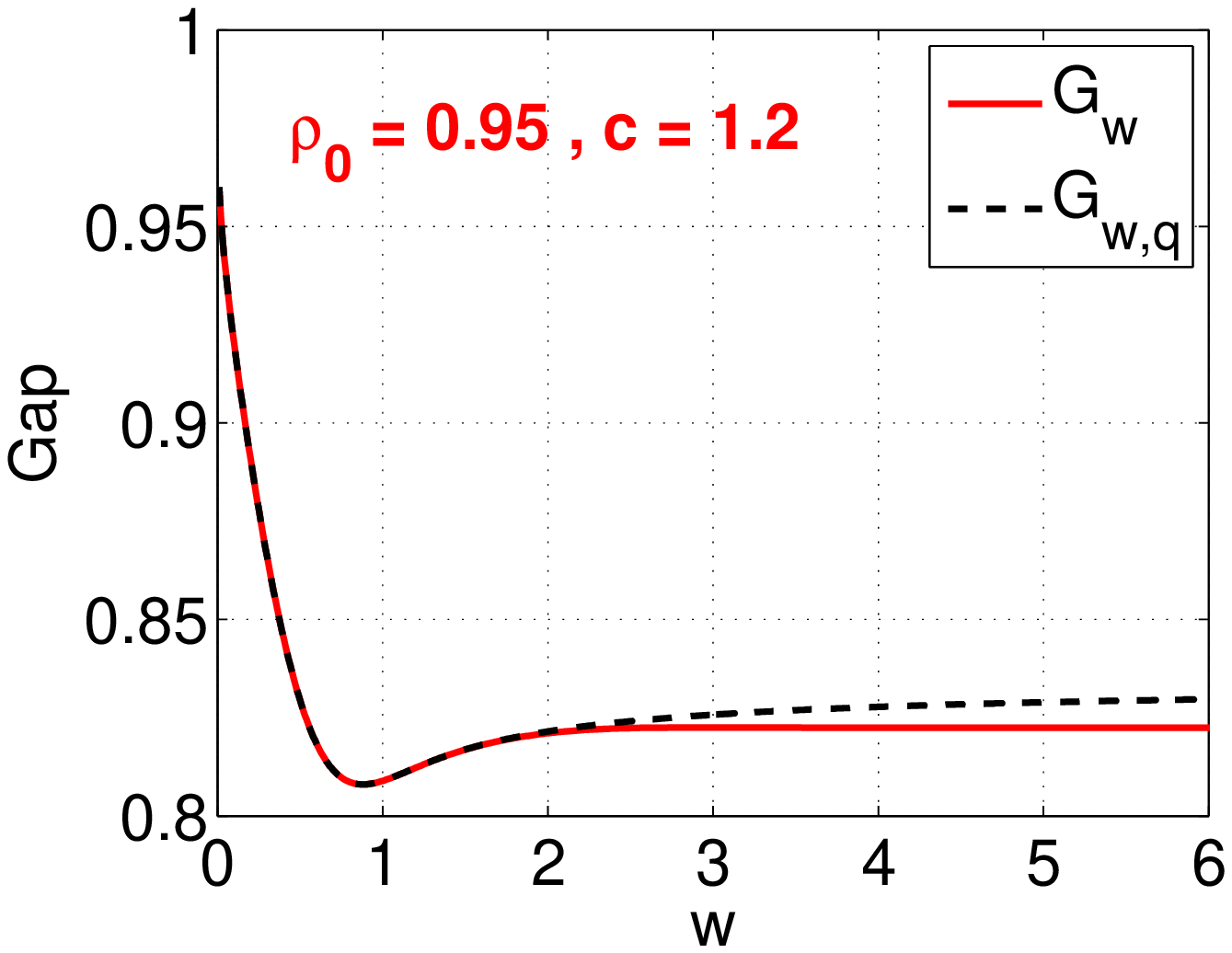}
}
\mbox{
\includegraphics[width = 2.2in]{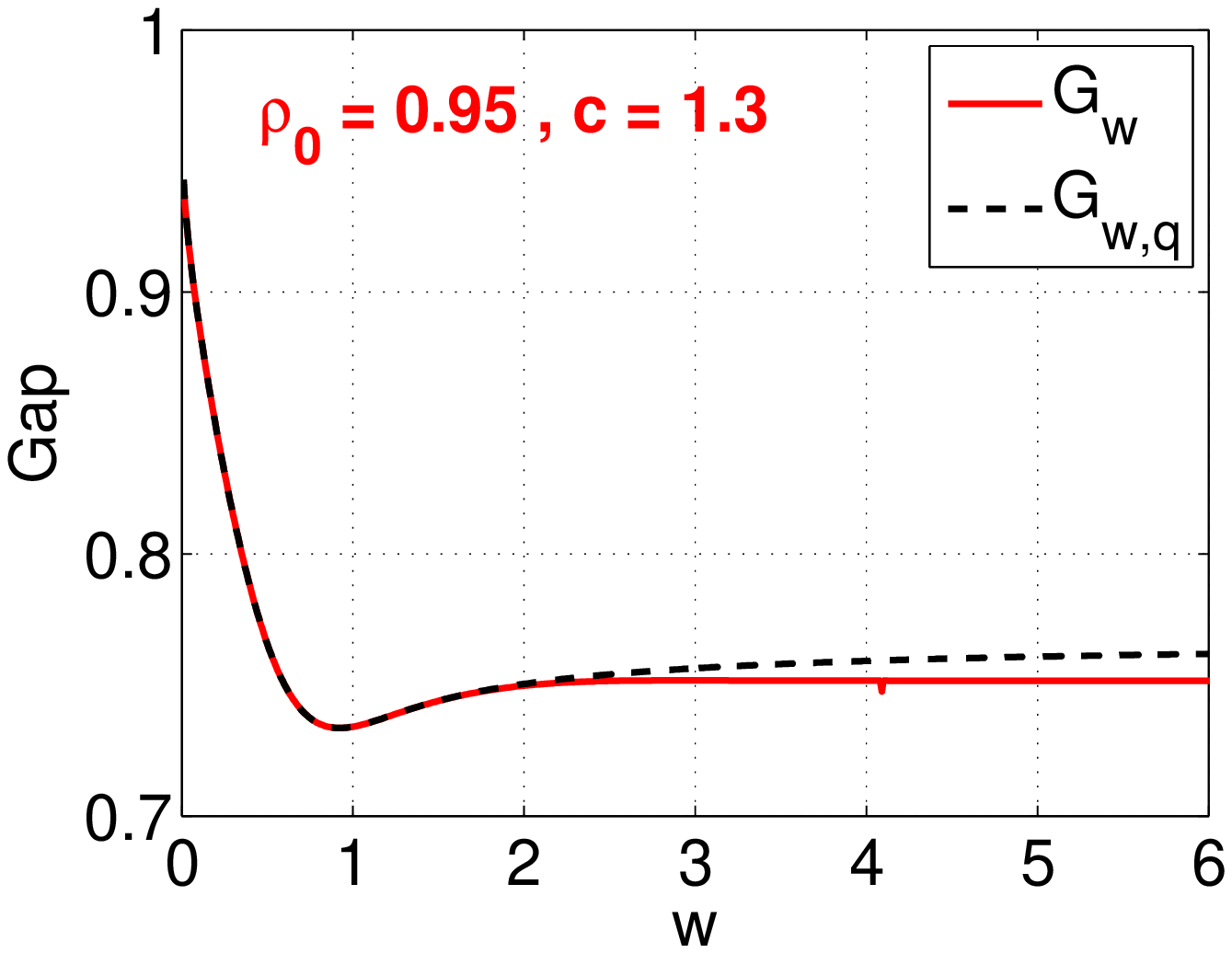}
\includegraphics[width = 2.2in]{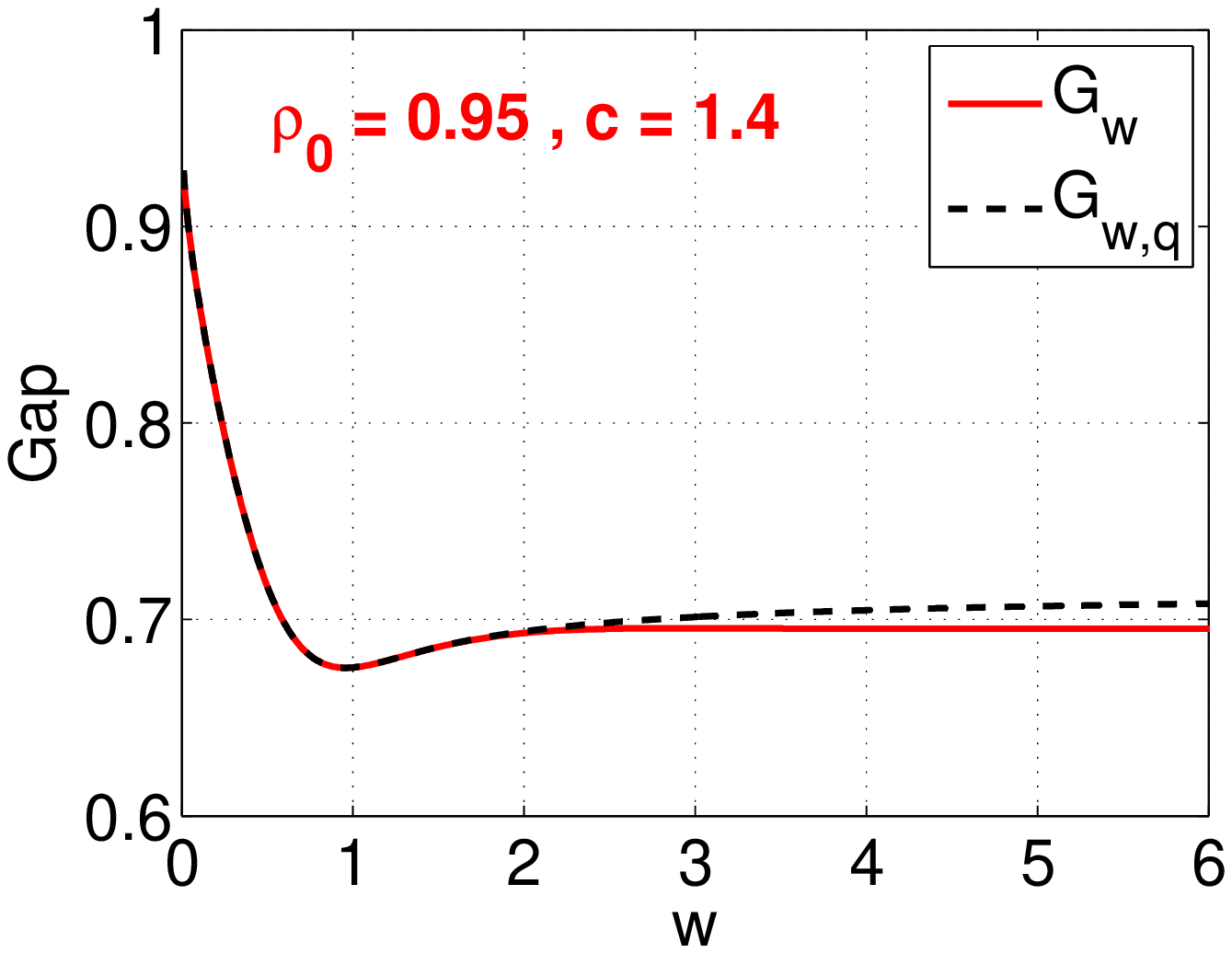}
\includegraphics[width = 2.2in]{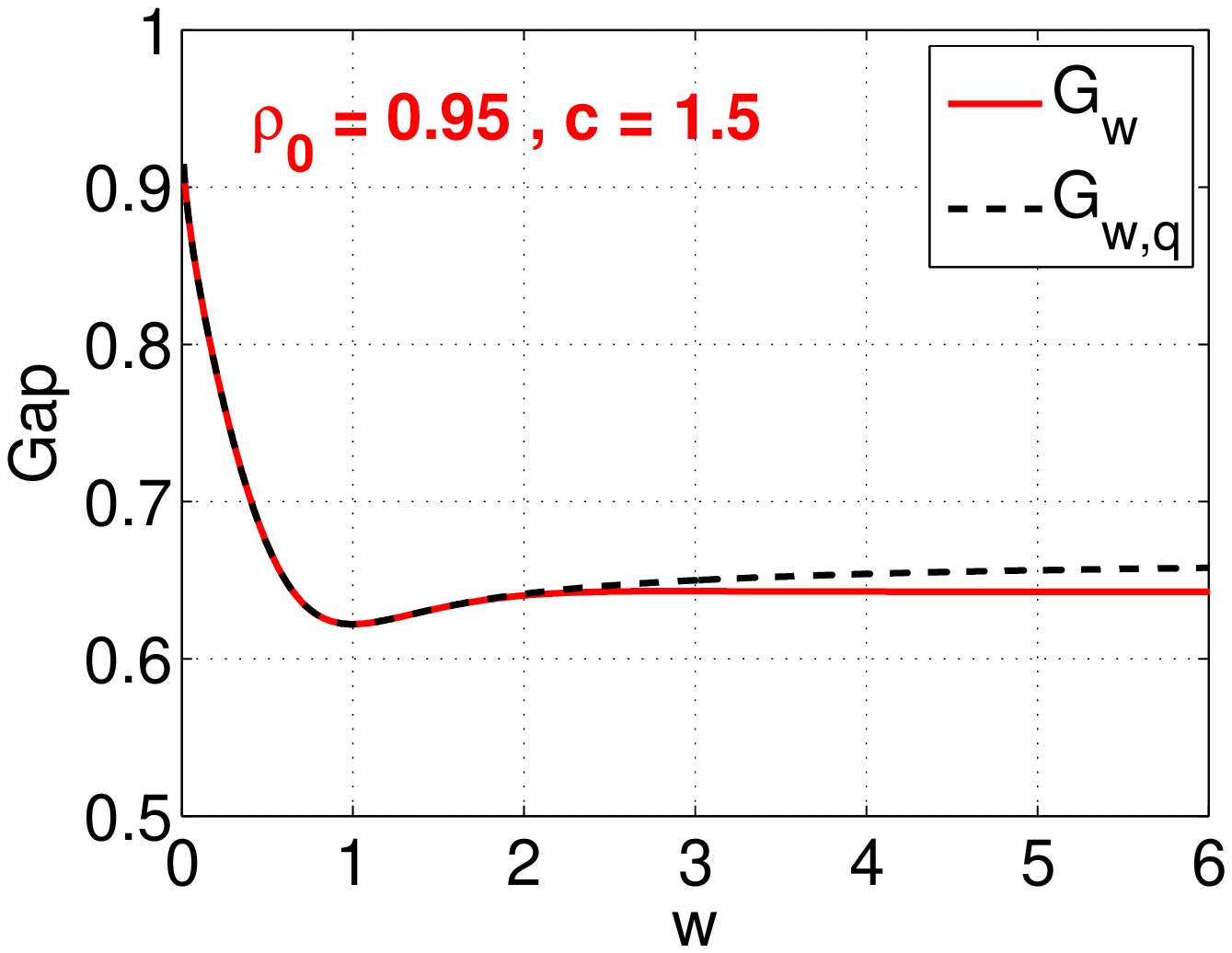}
}

\mbox{
\includegraphics[width = 2.2in]{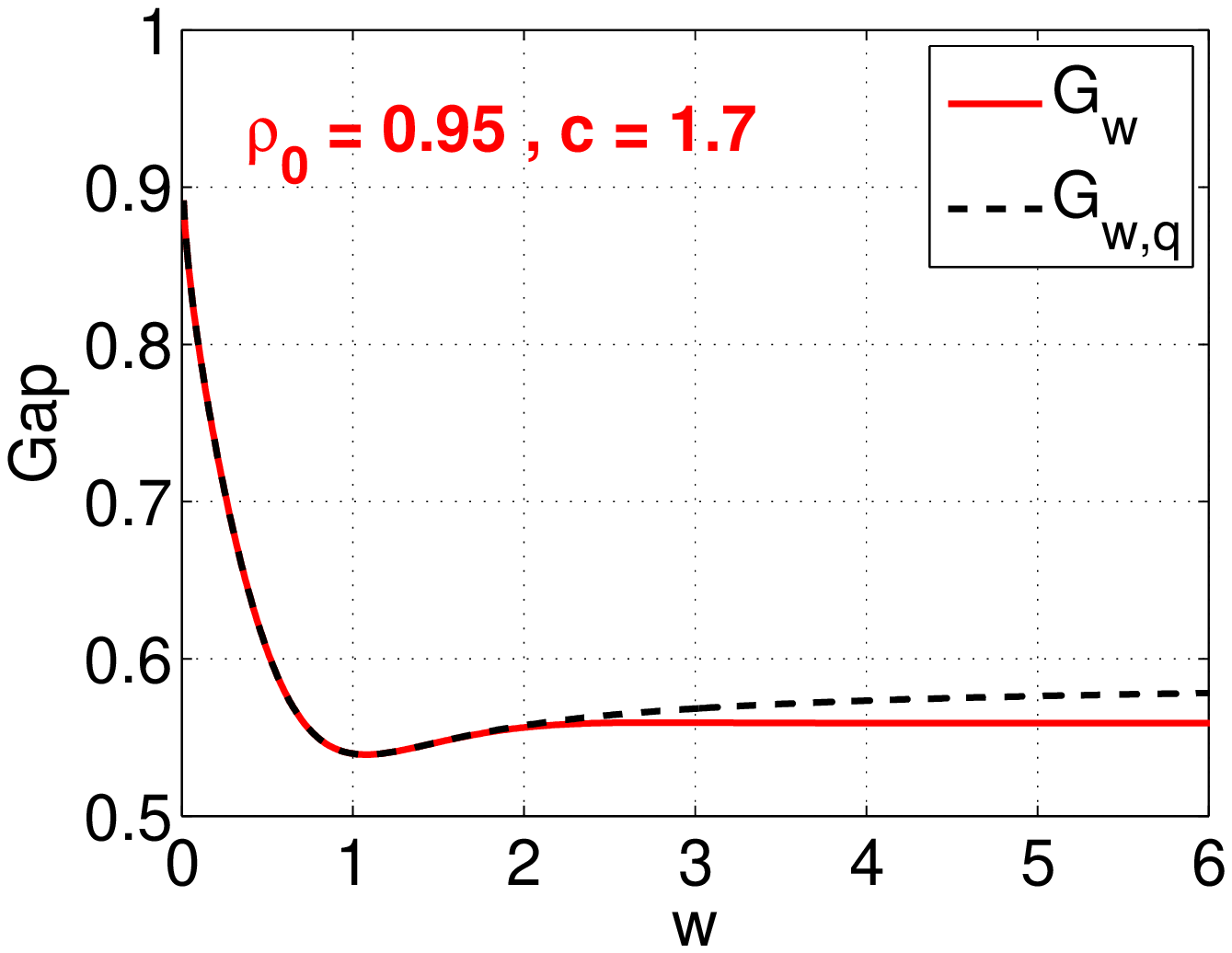}
\includegraphics[width = 2.2in]{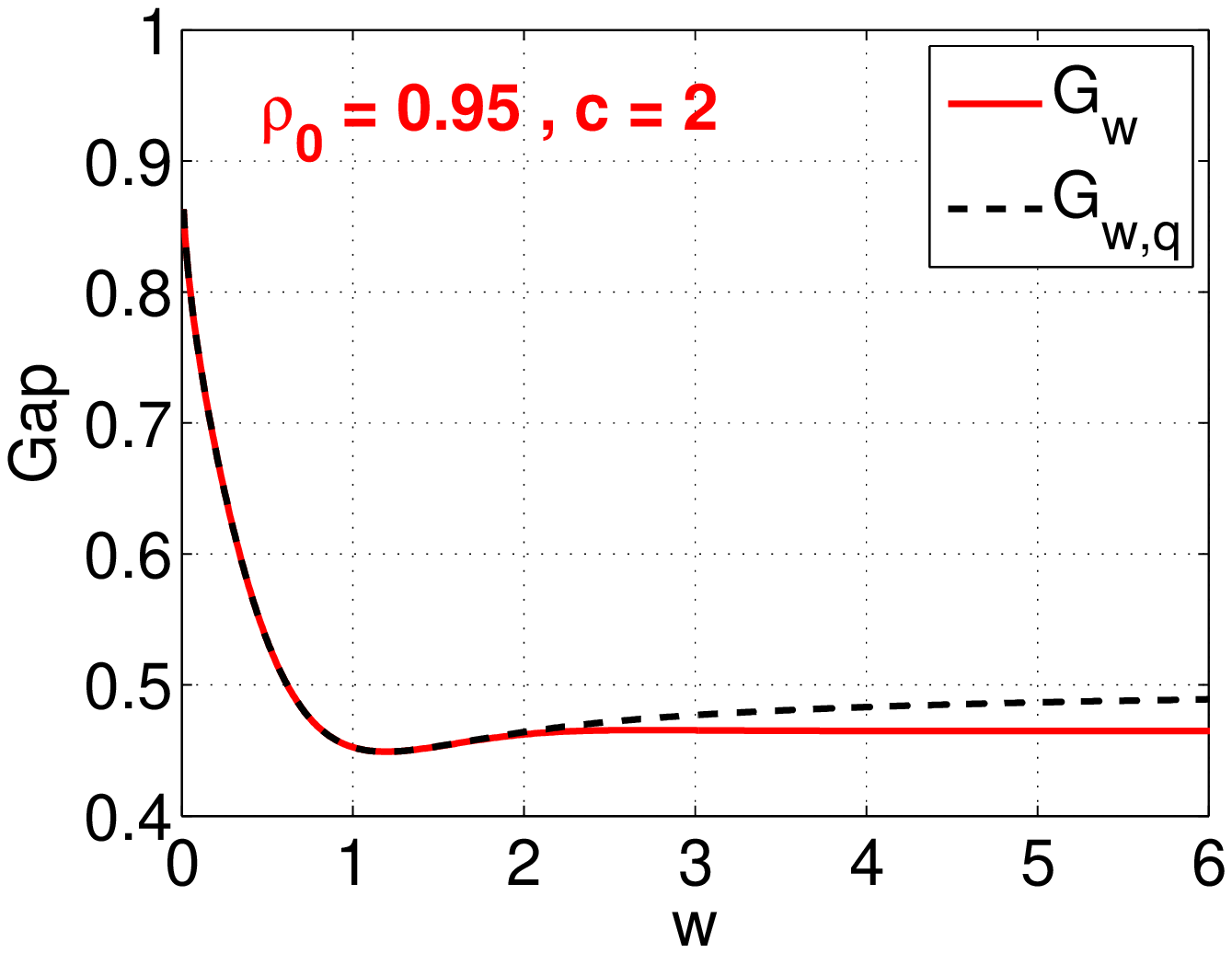}
\includegraphics[width = 2.2in]{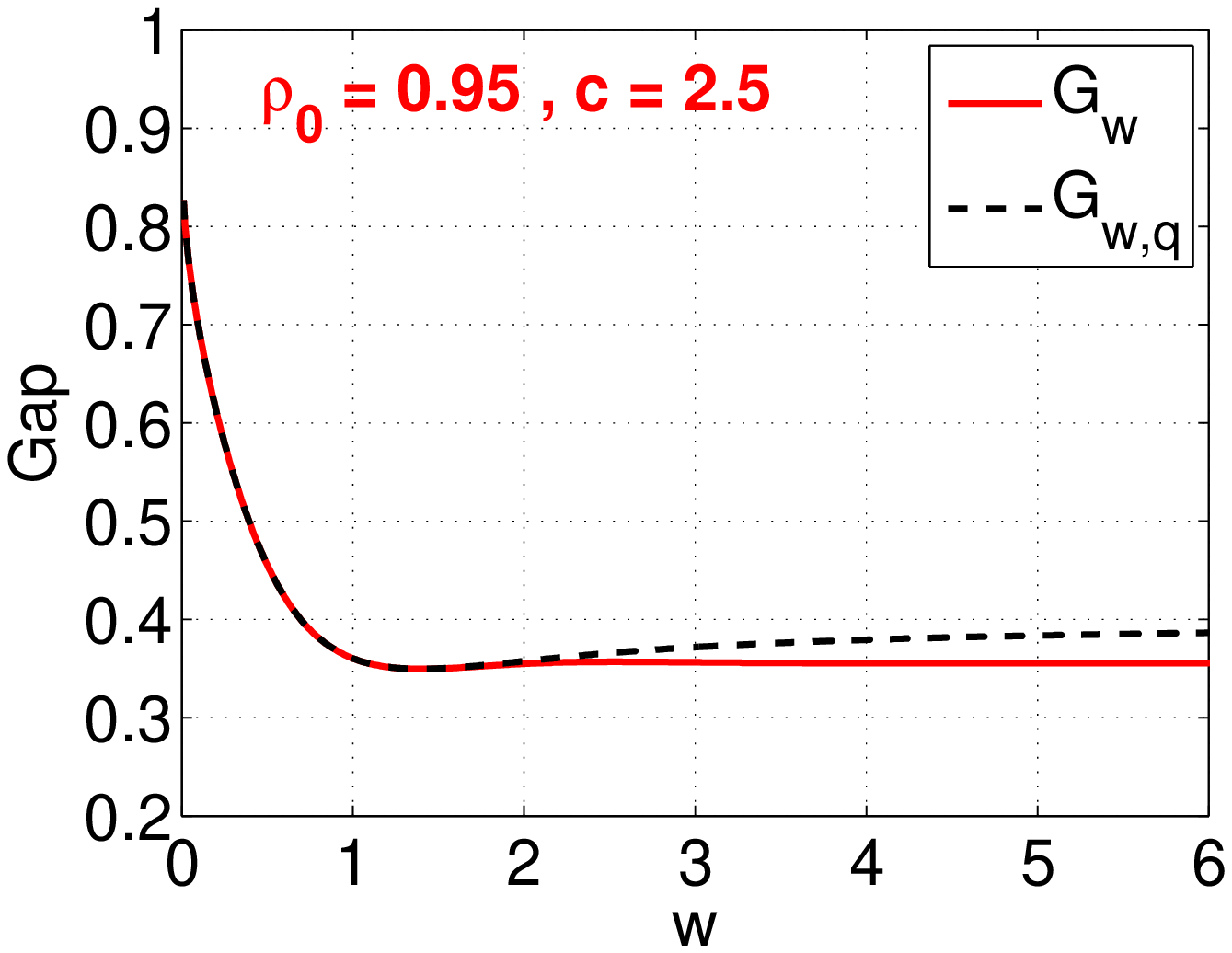}
}

\mbox{
\includegraphics[width = 2.2in]{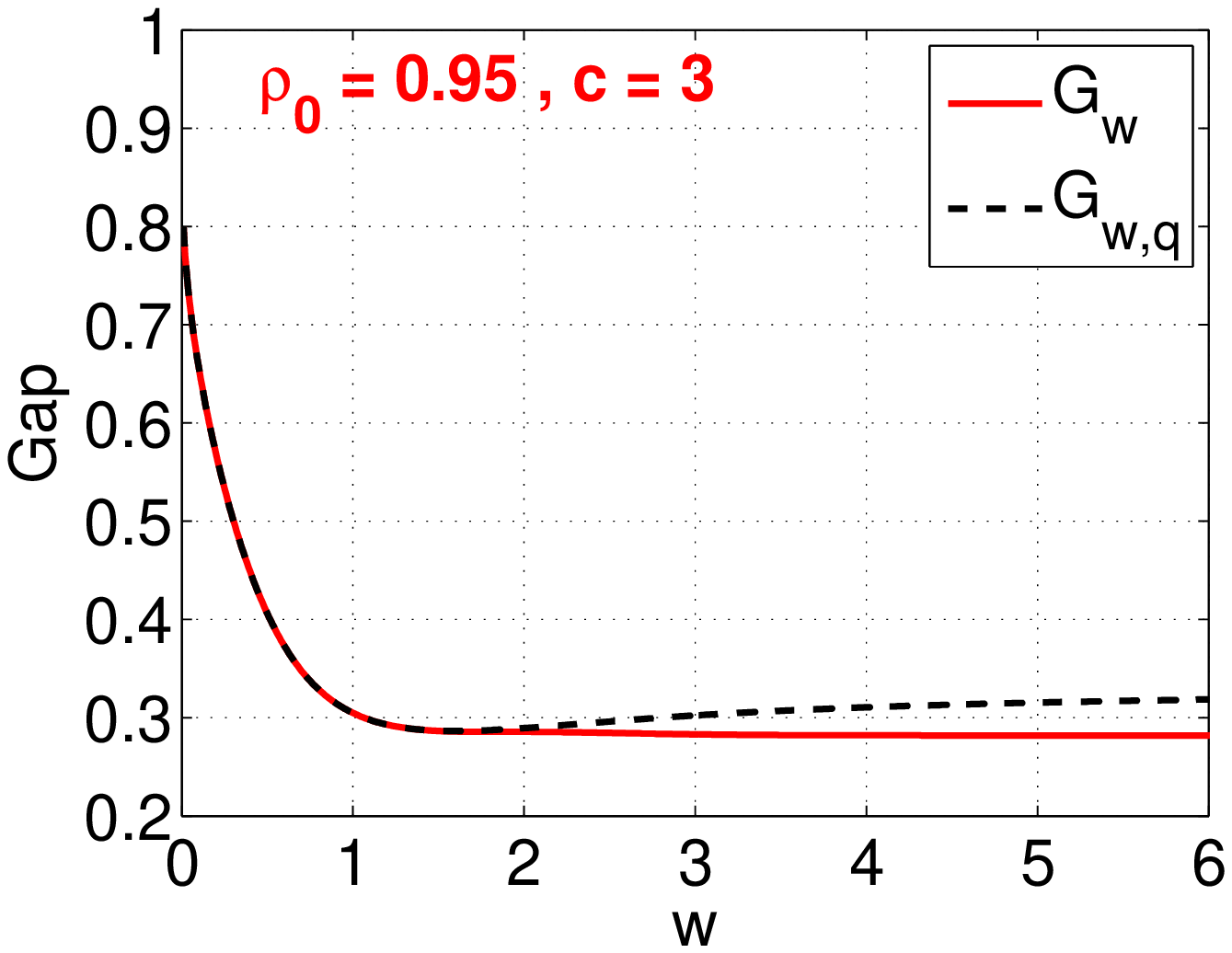}
\includegraphics[width = 2.2in]{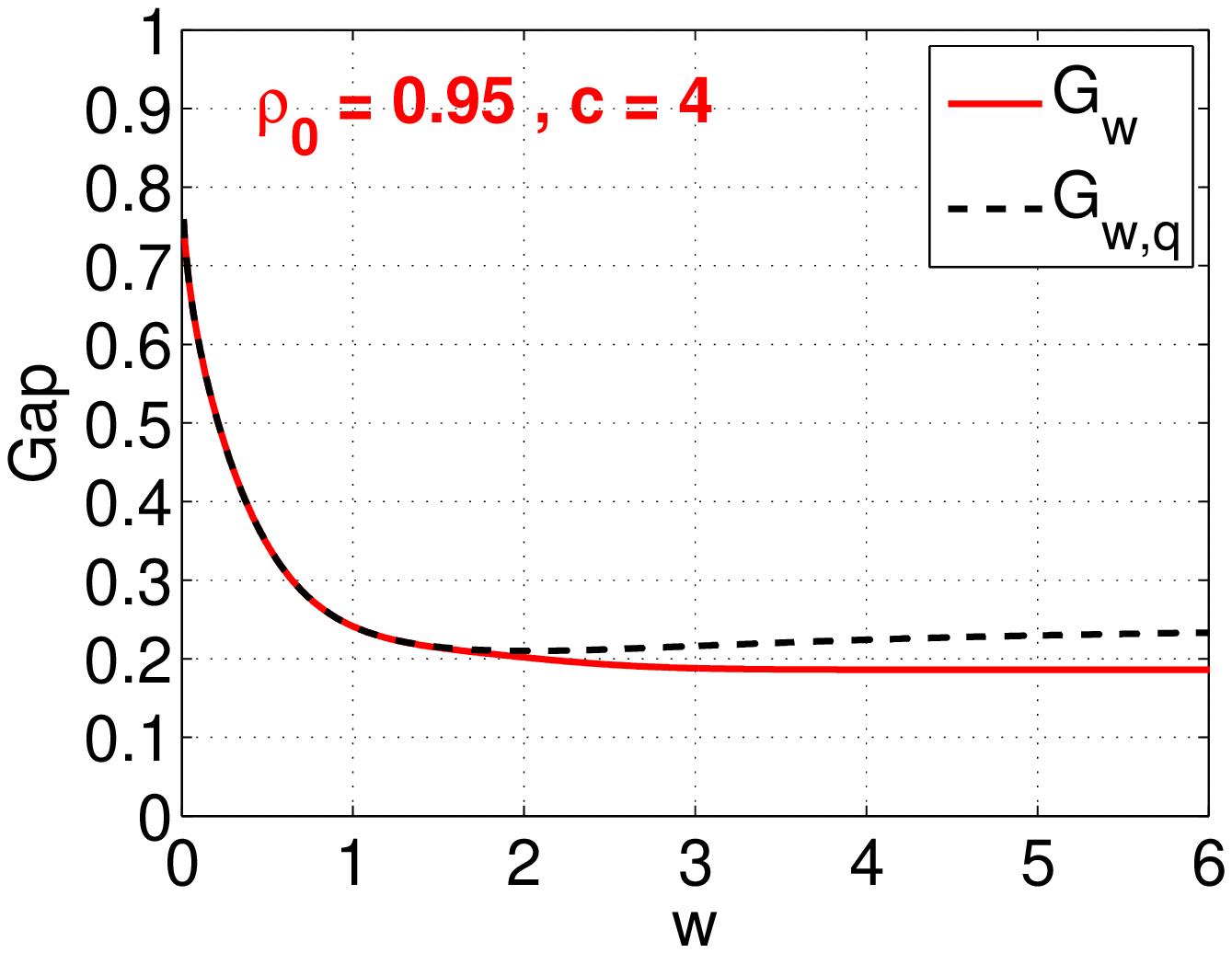}
\includegraphics[width = 2.2in]{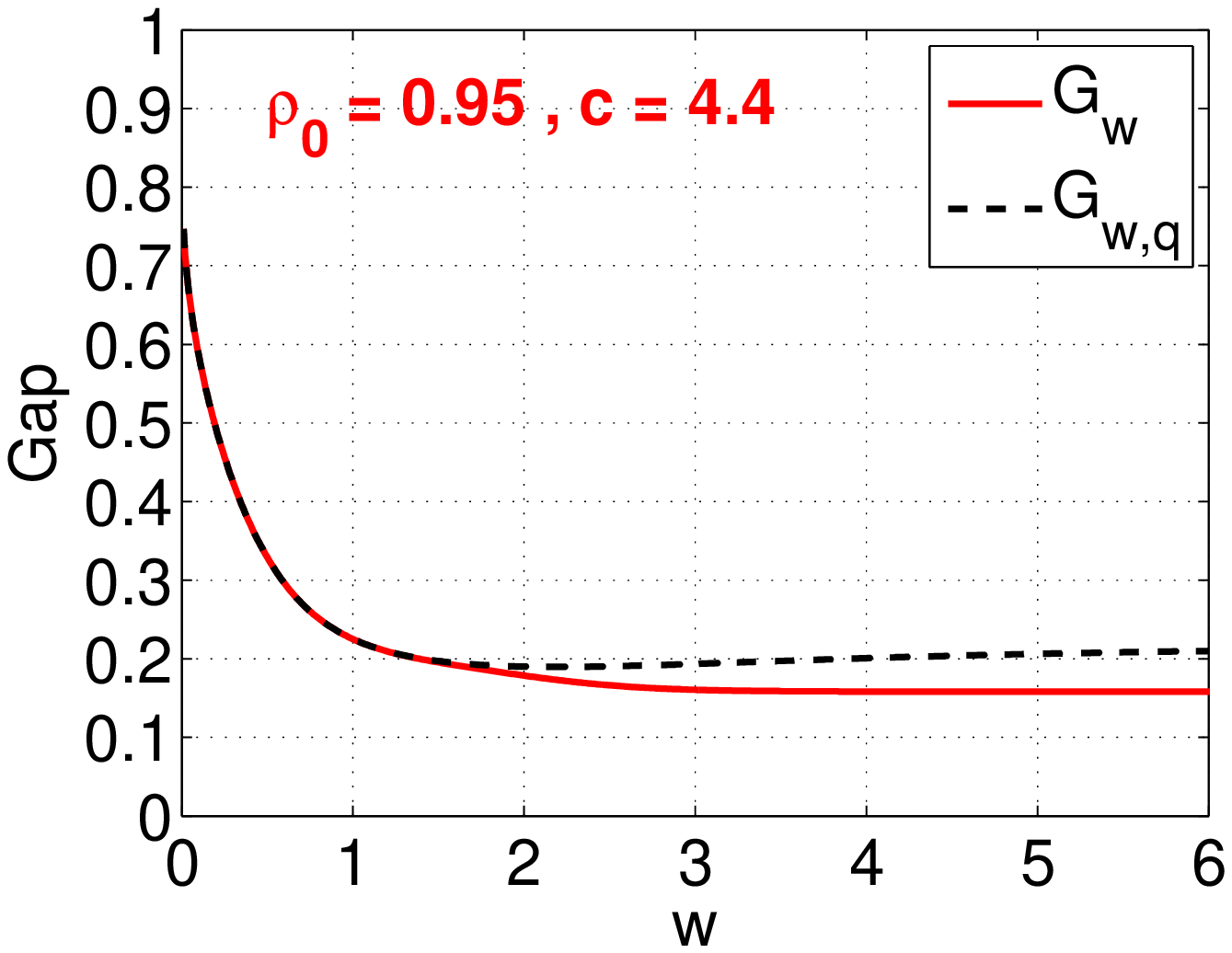}
}

\end{center}
\vspace{-.2in}
\caption{The gaps $G_w$ and $G_{w,q}$ as functions of $w$, for $\rho_0 = 0.95$ and a range of $c$ values. }\label{fig_GwqR095C}
\end{figure}

\begin{figure}[h!]
\begin{center}
\mbox{
\includegraphics[width = 2.2in]{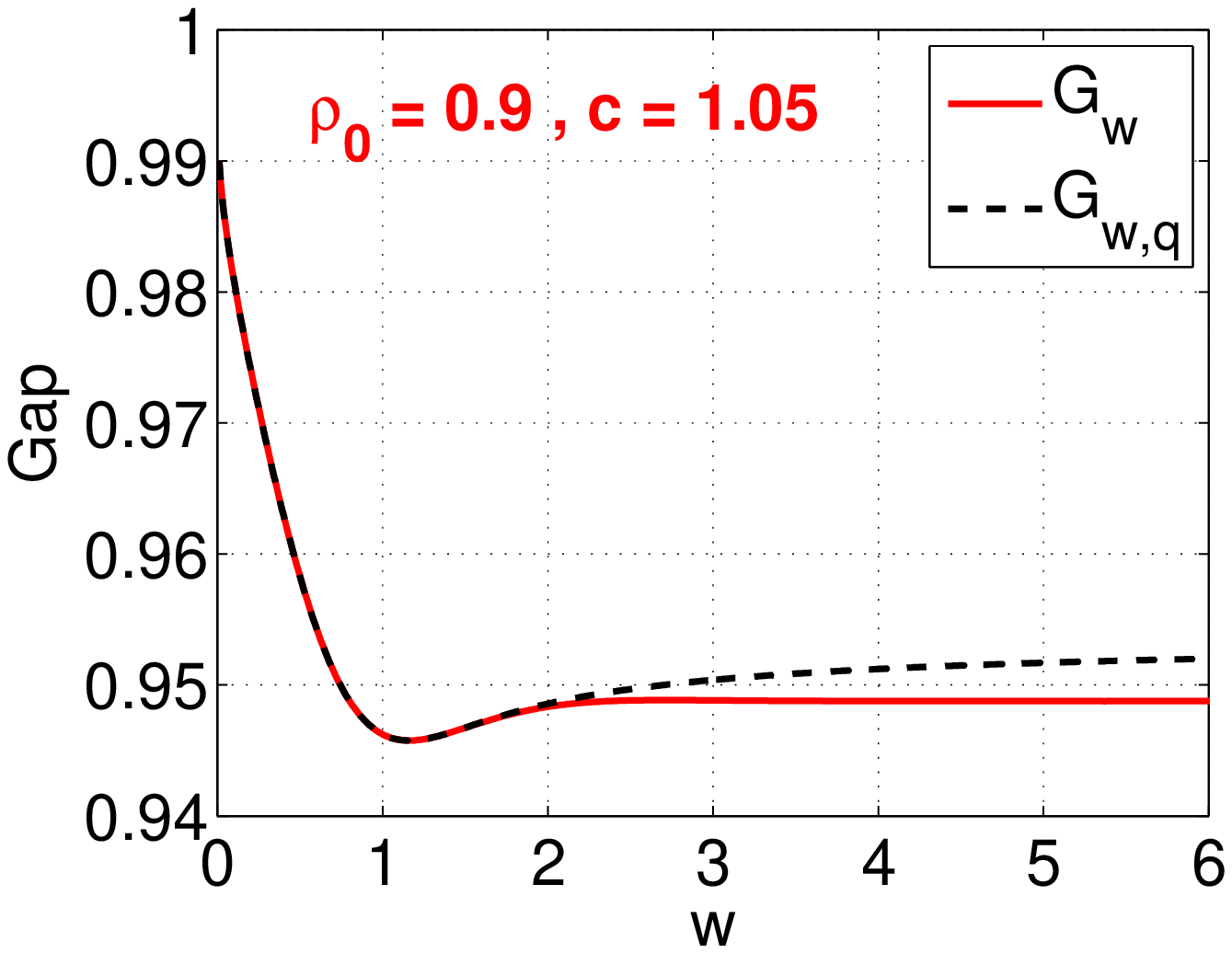}
\includegraphics[width = 2.2in]{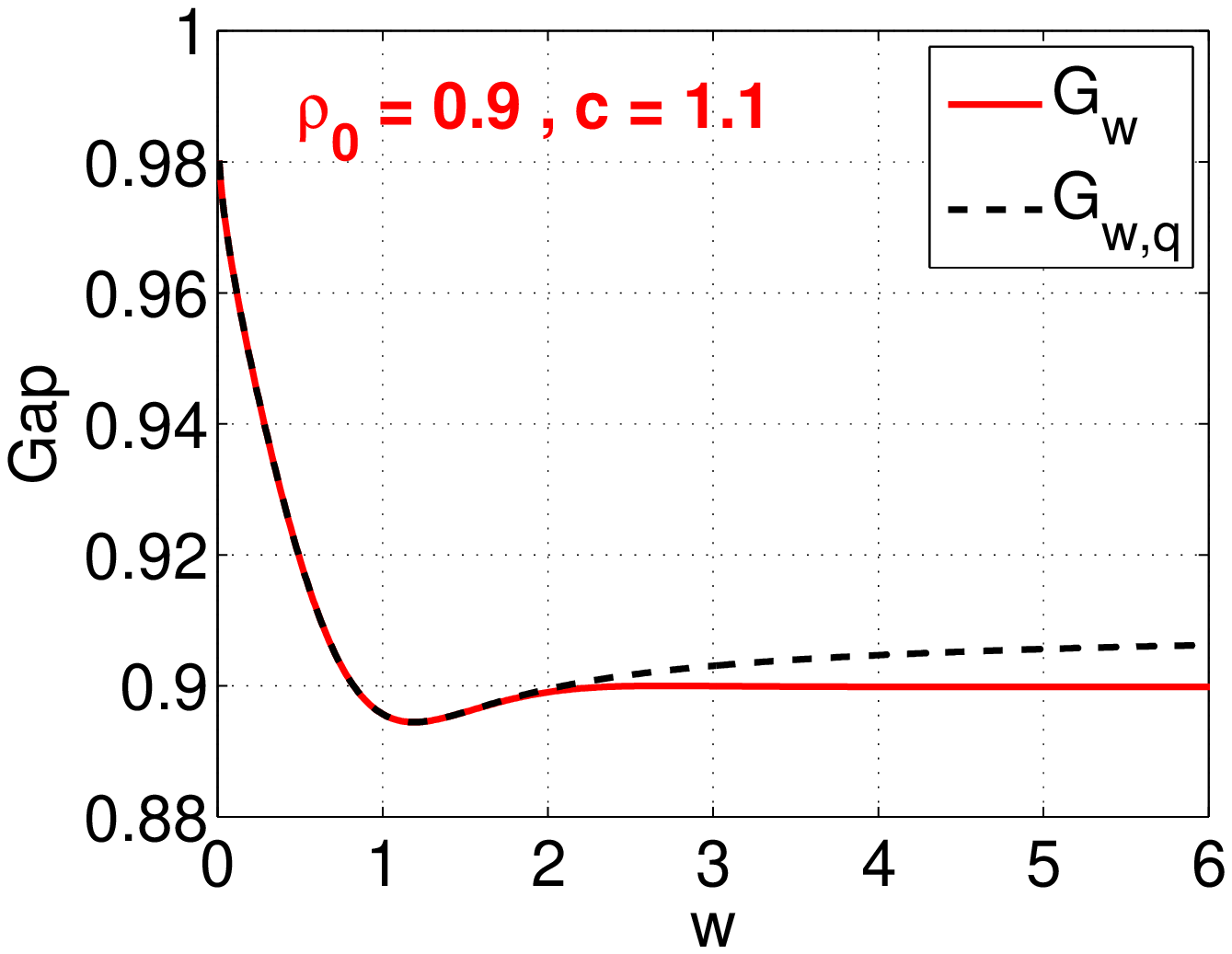}
\includegraphics[width = 2.2in]{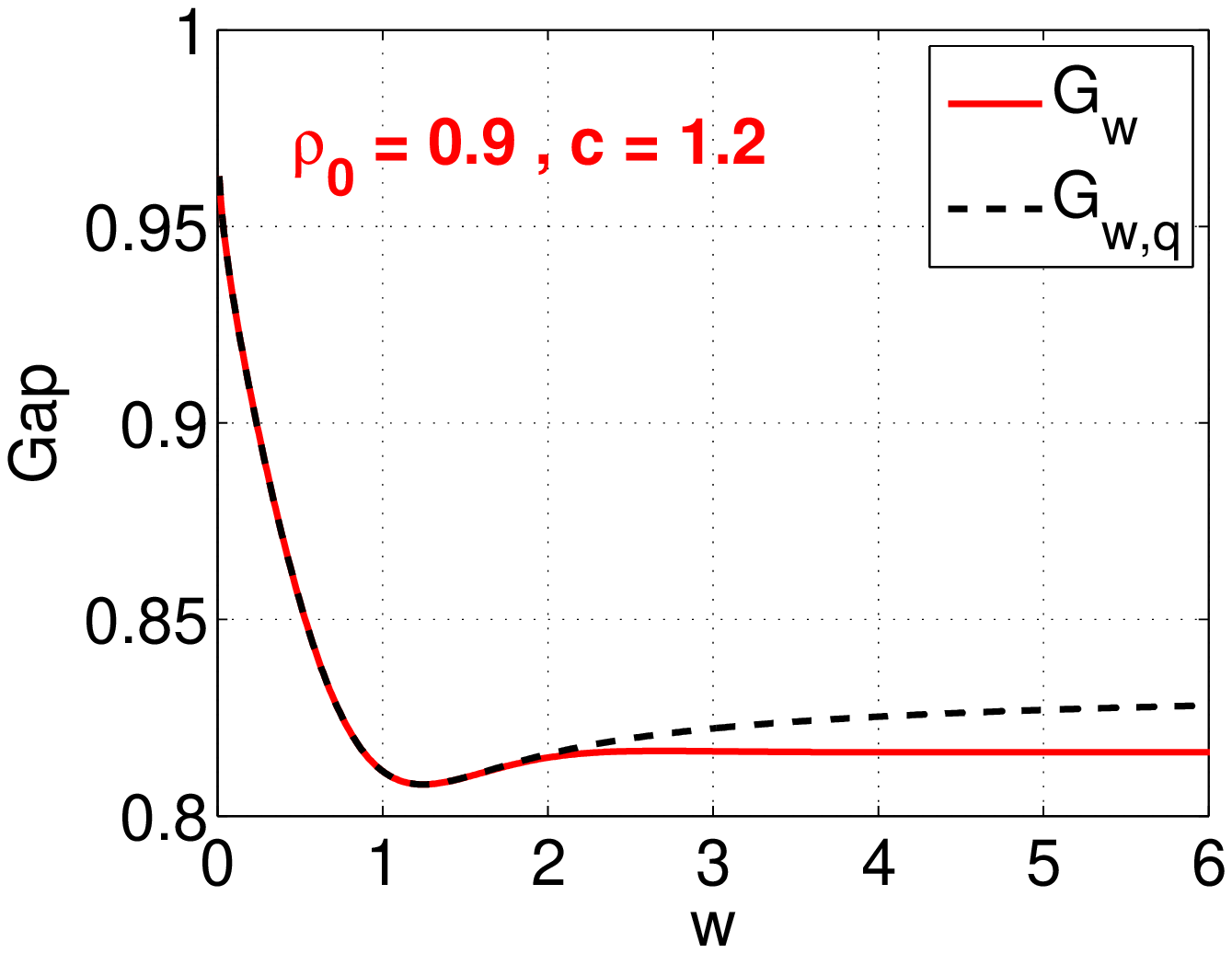}
}
\mbox{
\includegraphics[width = 2.2in]{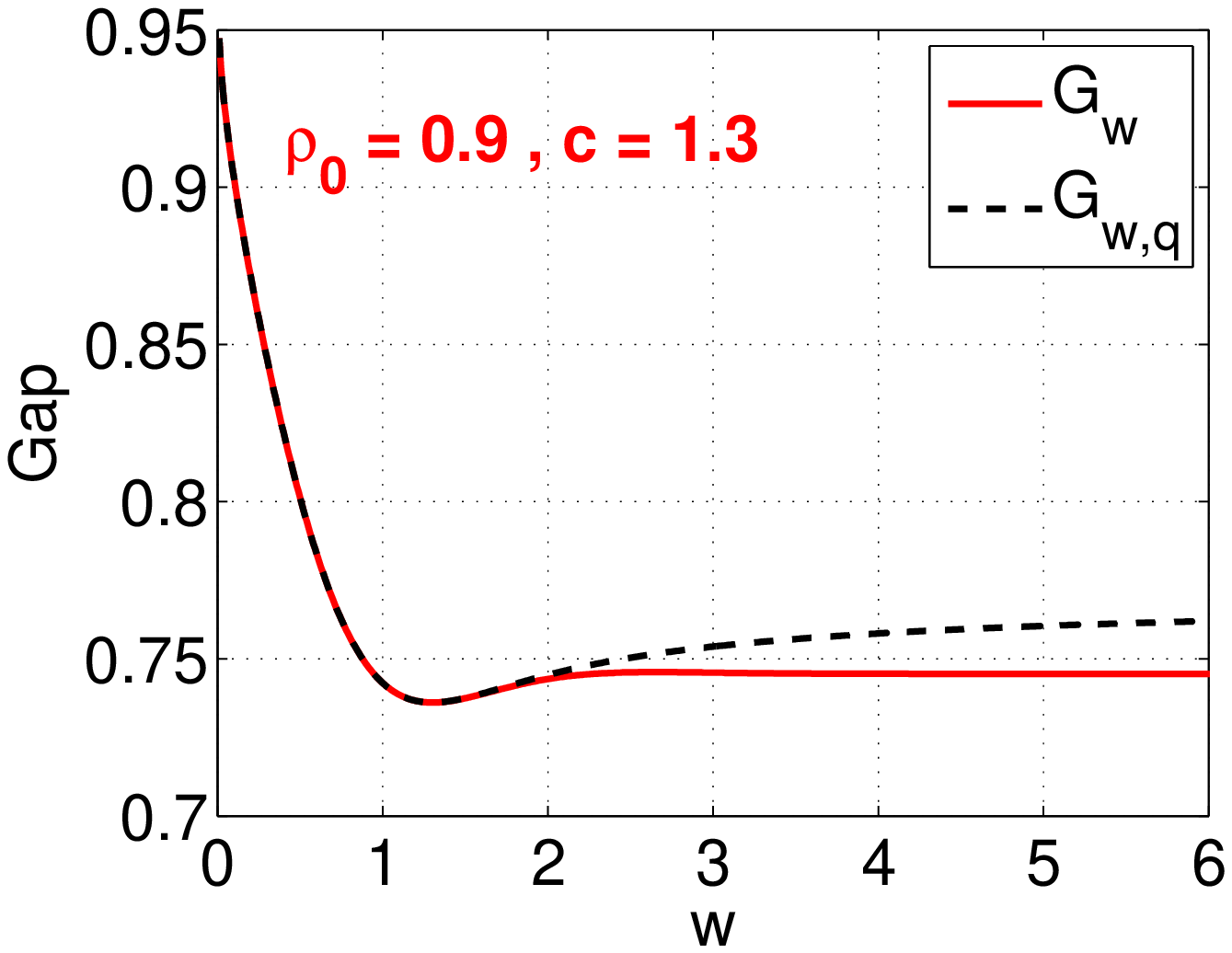}
\includegraphics[width = 2.2in]{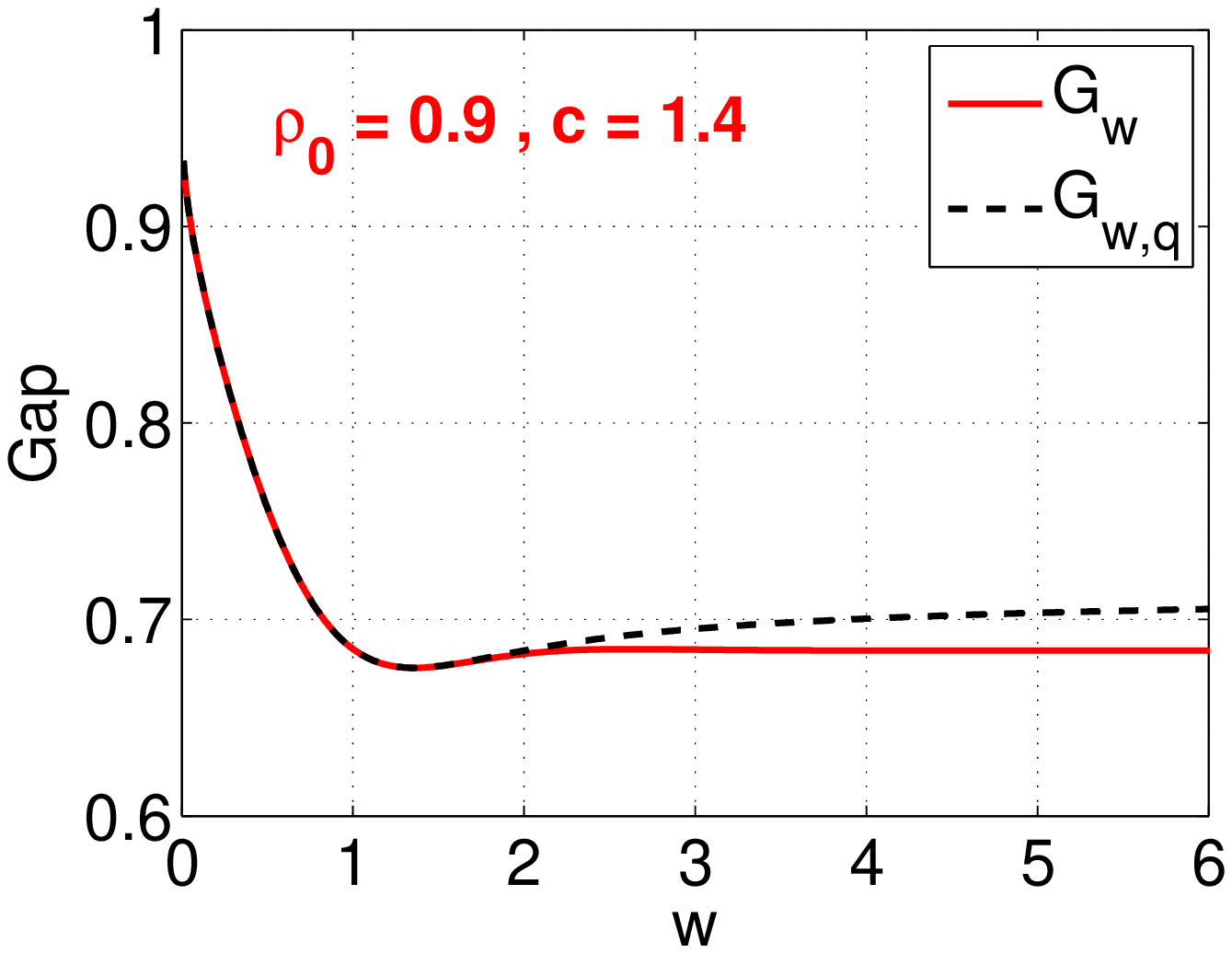}
\includegraphics[width = 2.2in]{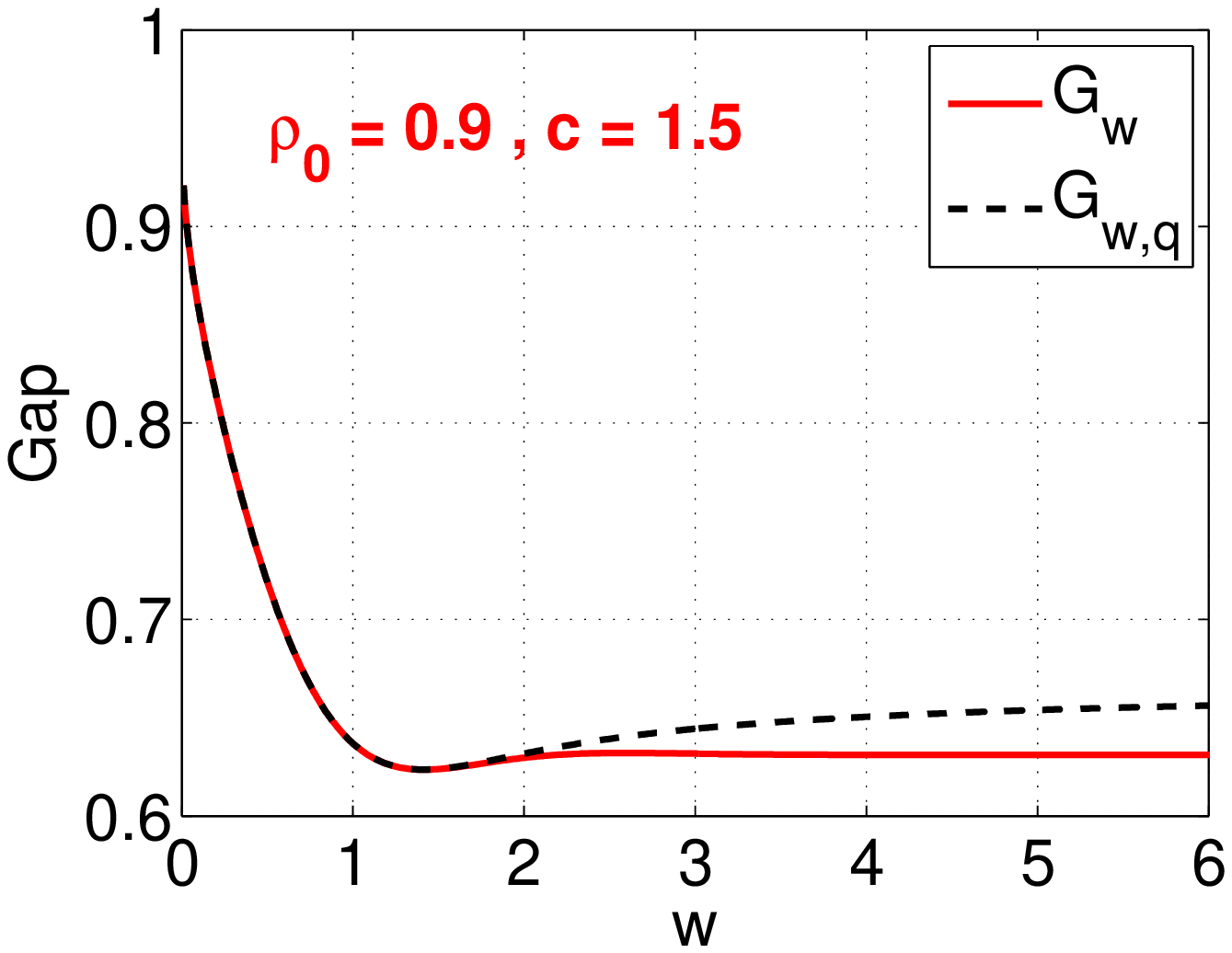}
}

\mbox{
\includegraphics[width = 2.2in]{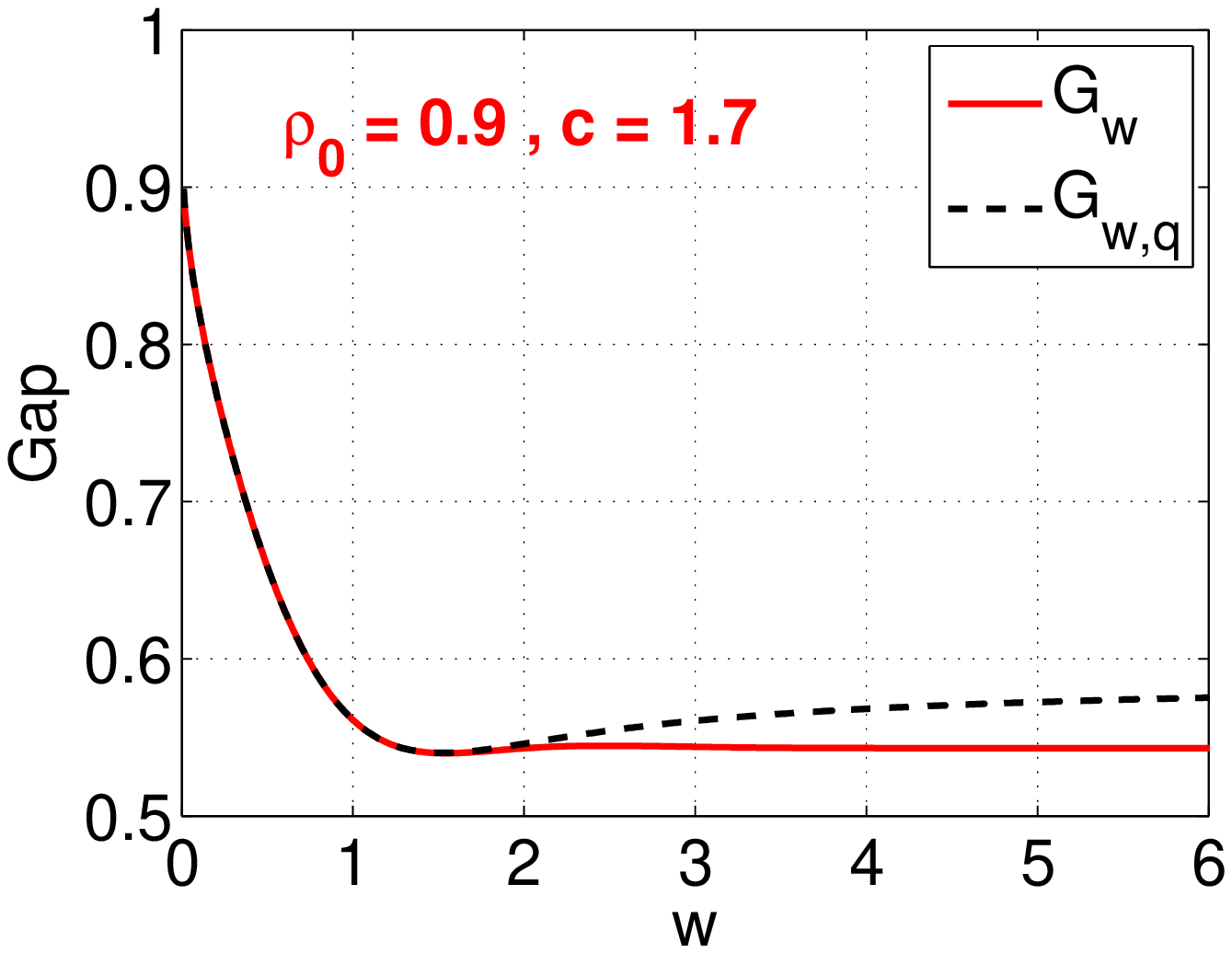}
\includegraphics[width = 2.2in]{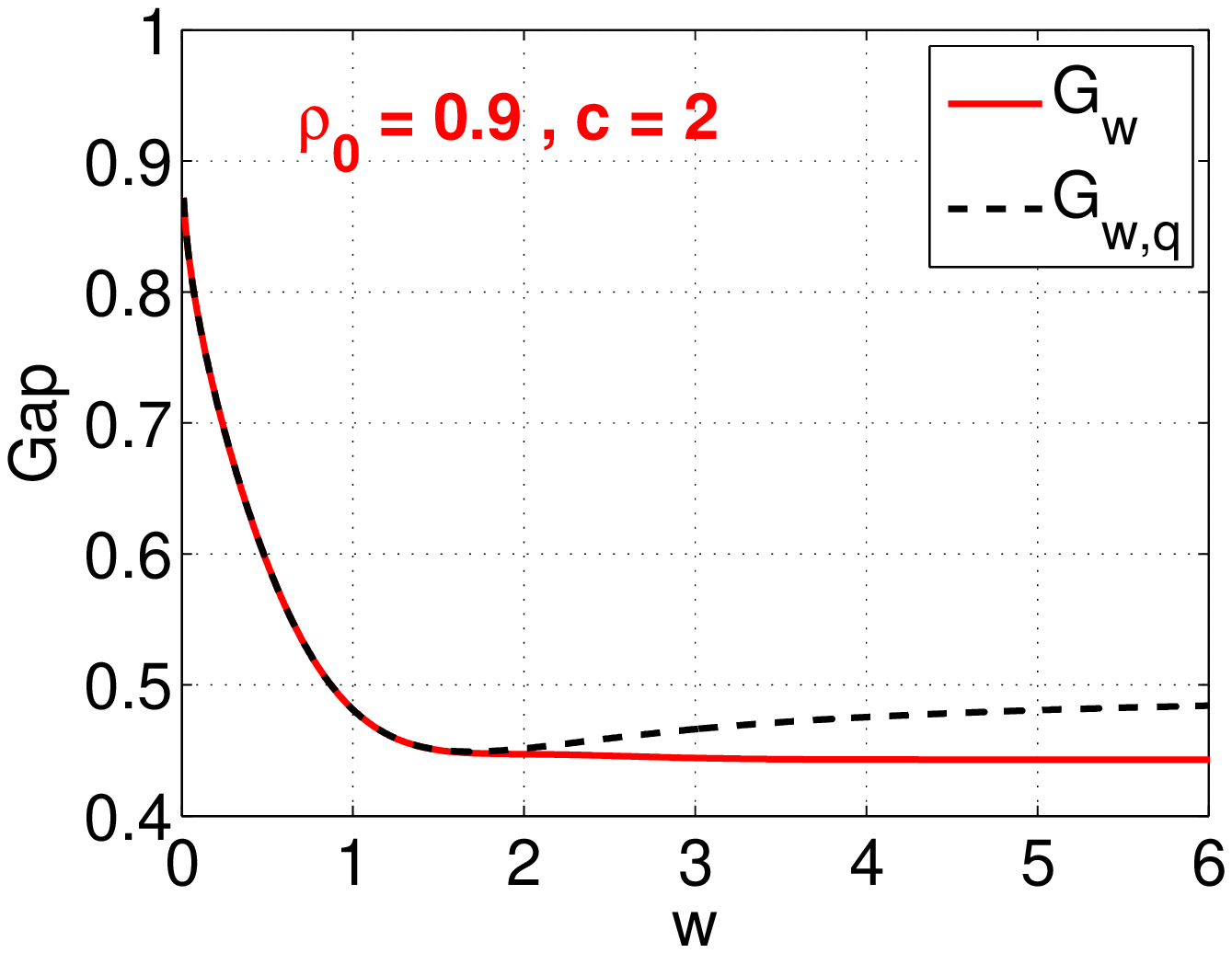}
\includegraphics[width = 2.2in]{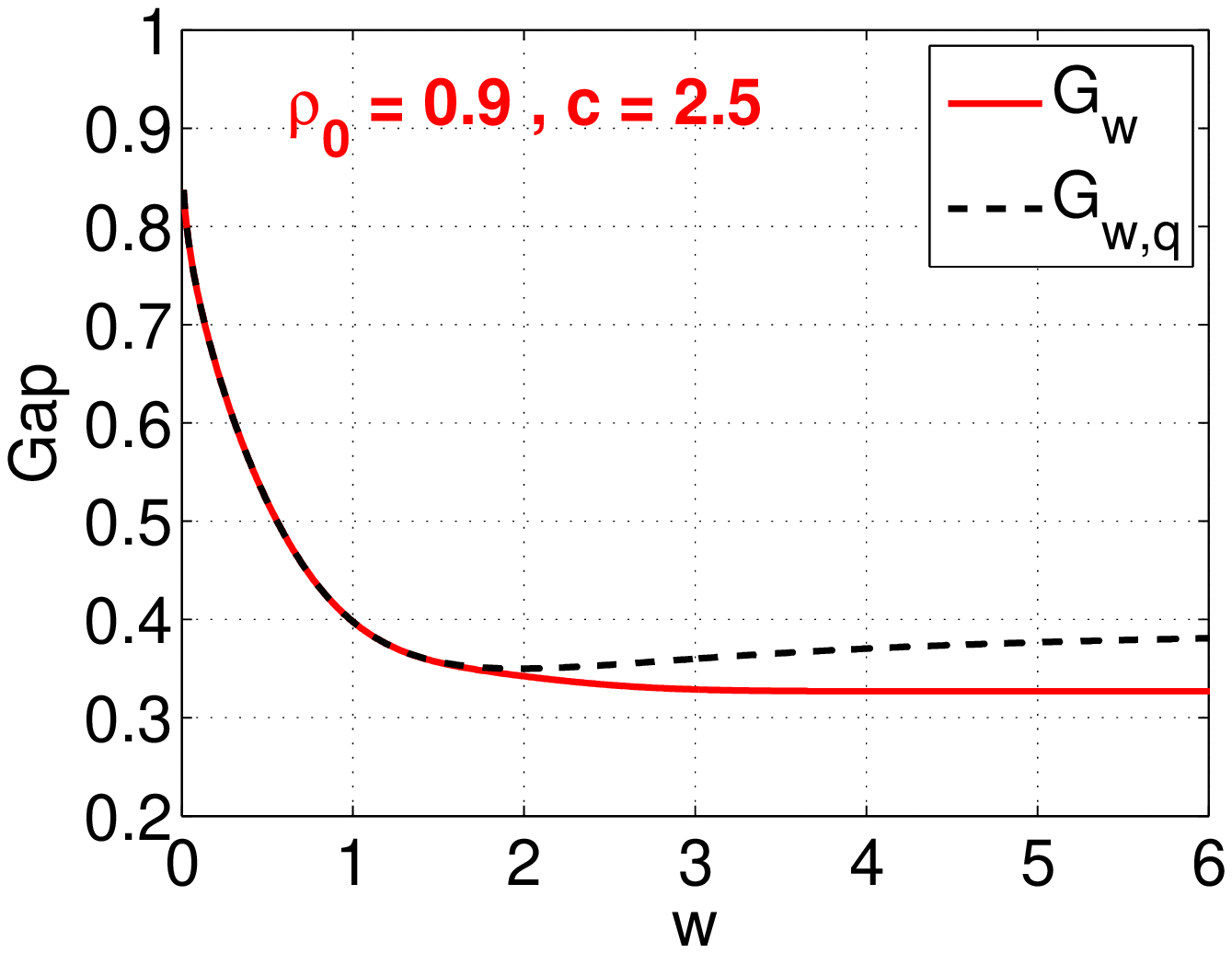}
}

\end{center}
\vspace{-.2in}
\caption{The gaps $G_w$ and $G_{w,q}$ as functions of $w$, for $\rho_0 = 0.9$ and a range of $c$ values.}\label{fig_GwqR09C}
\end{figure}

\begin{figure}[h!]
\begin{center}
\mbox{
\includegraphics[width = 2.2in]{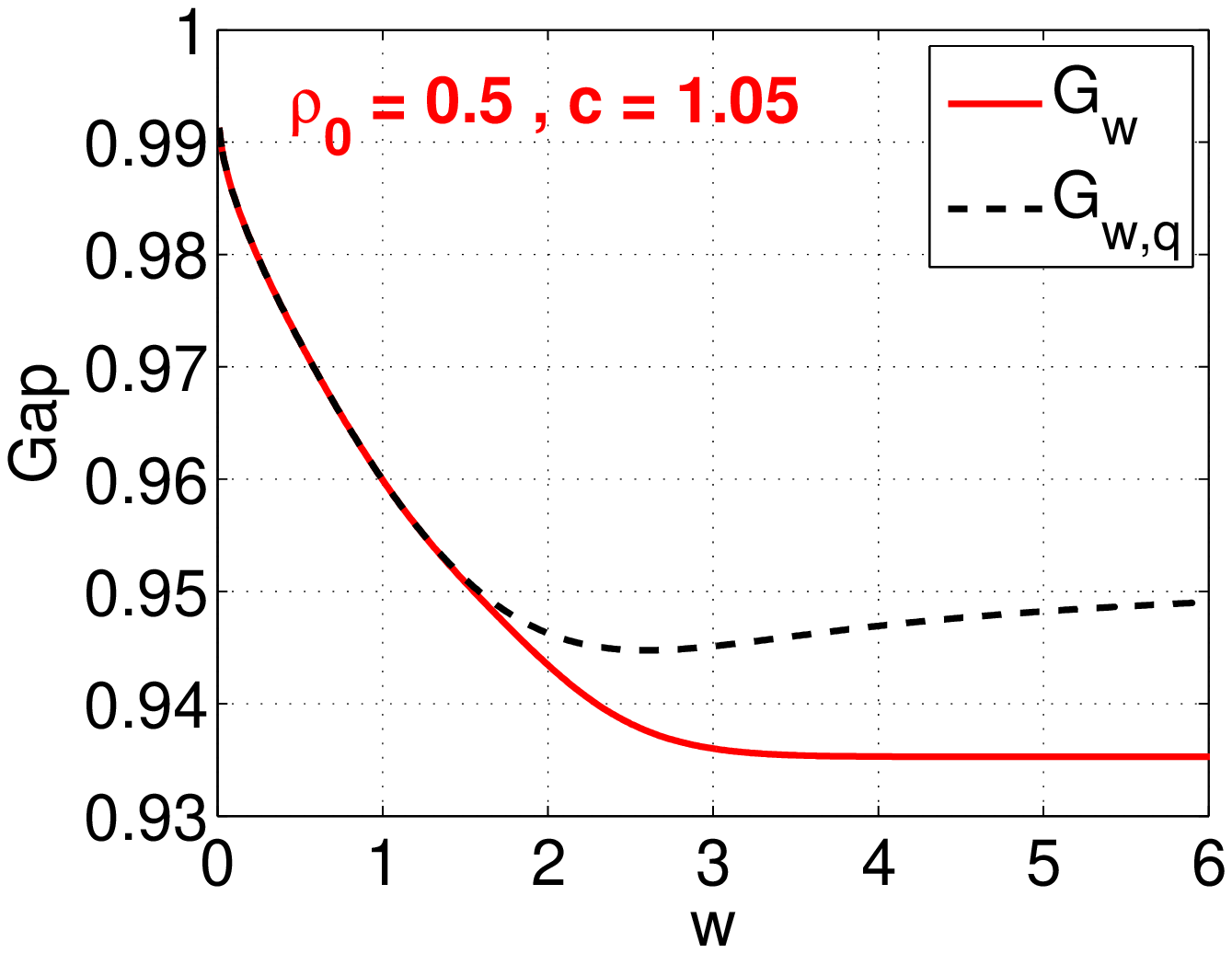}
\includegraphics[width = 2.2in]{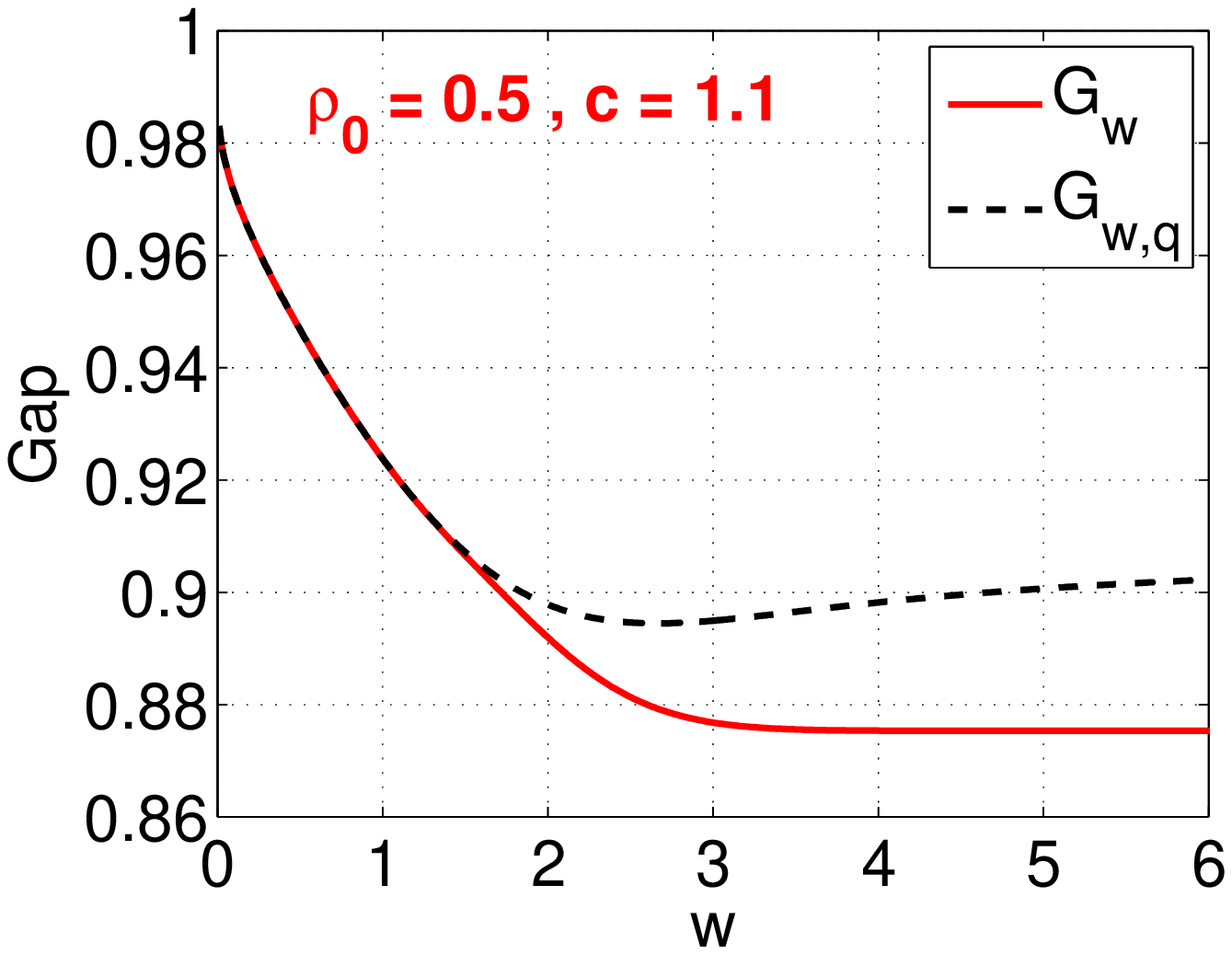}
\includegraphics[width = 2.2in]{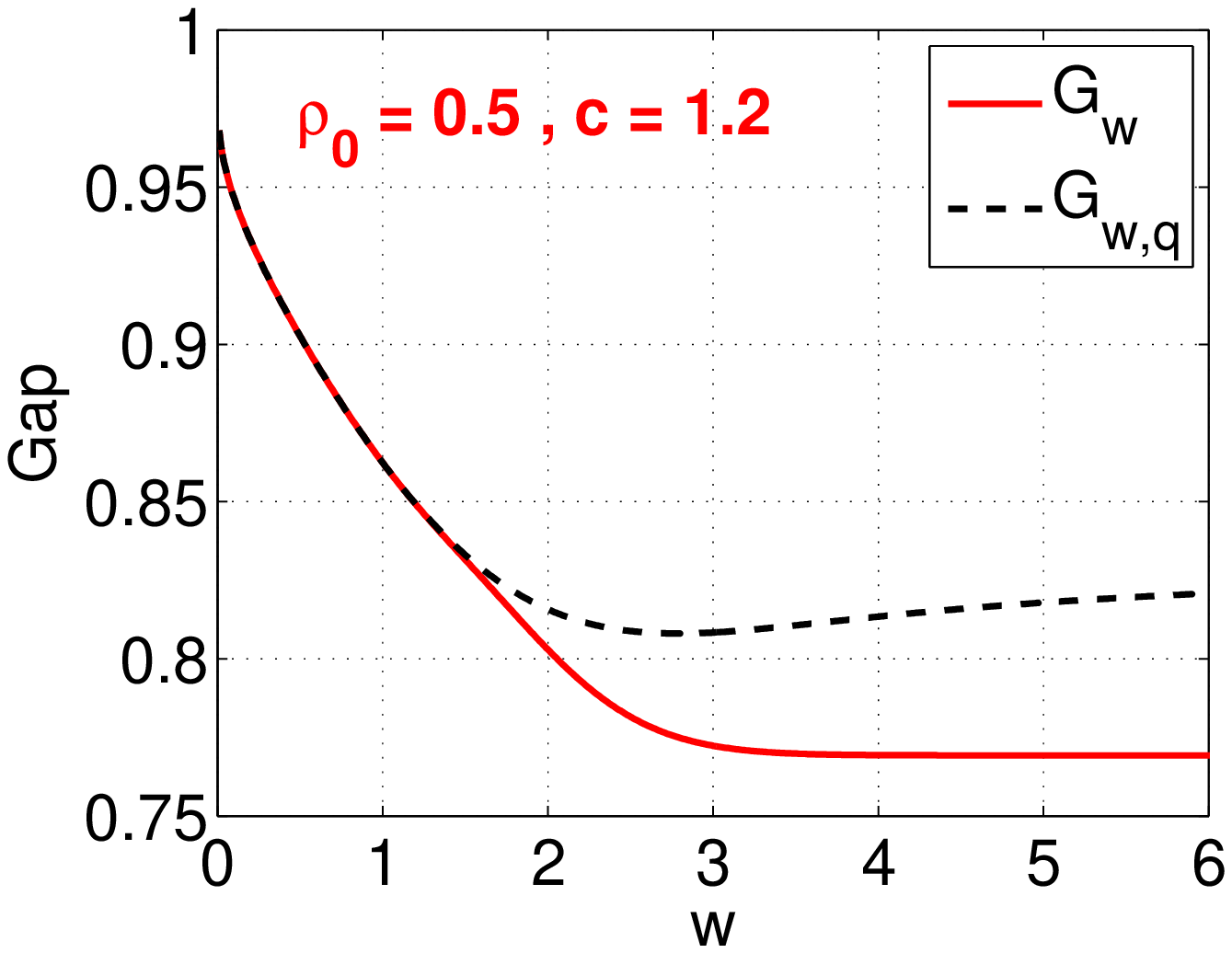}
}
\mbox{
\includegraphics[width = 2.2in]{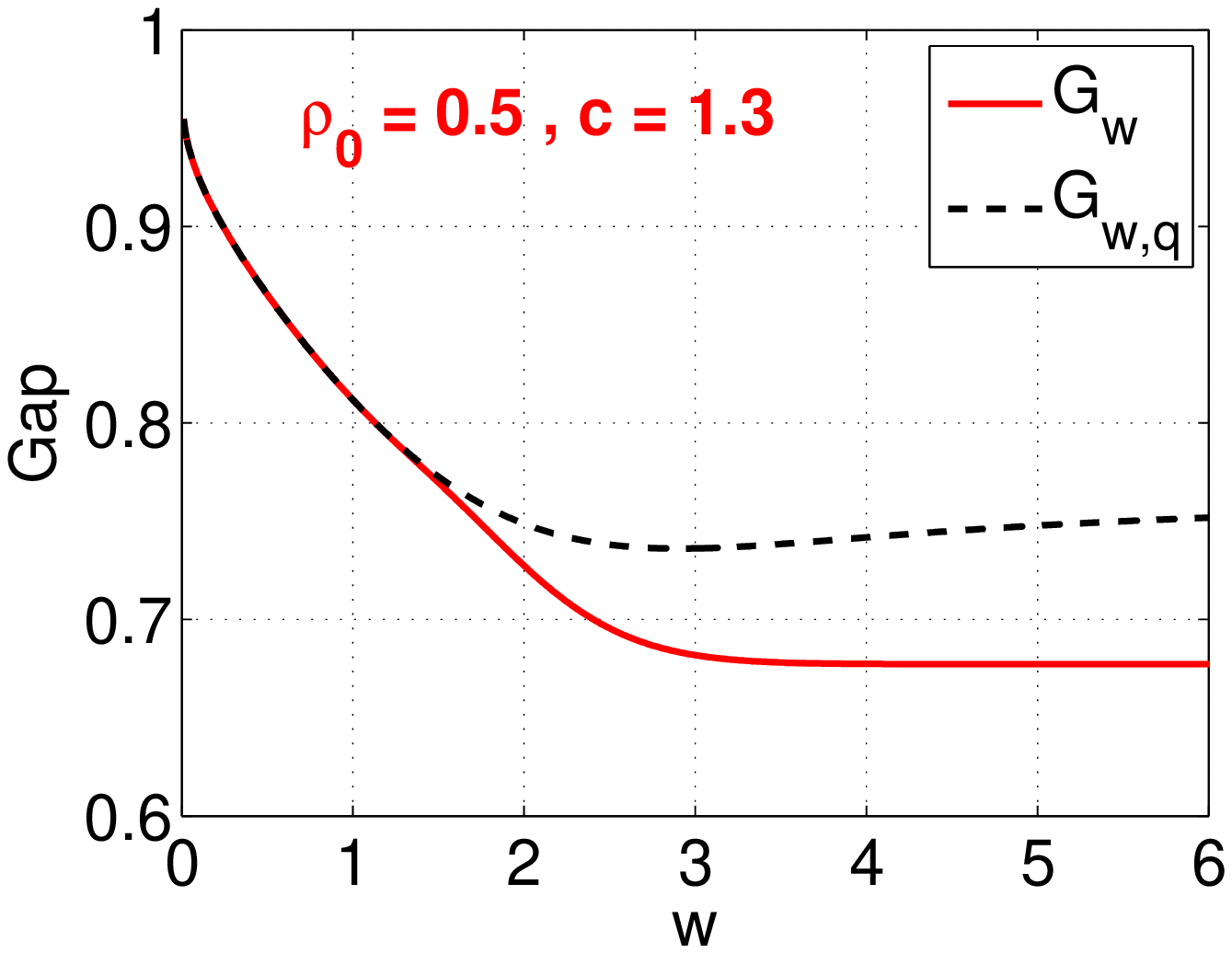}
\includegraphics[width = 2.2in]{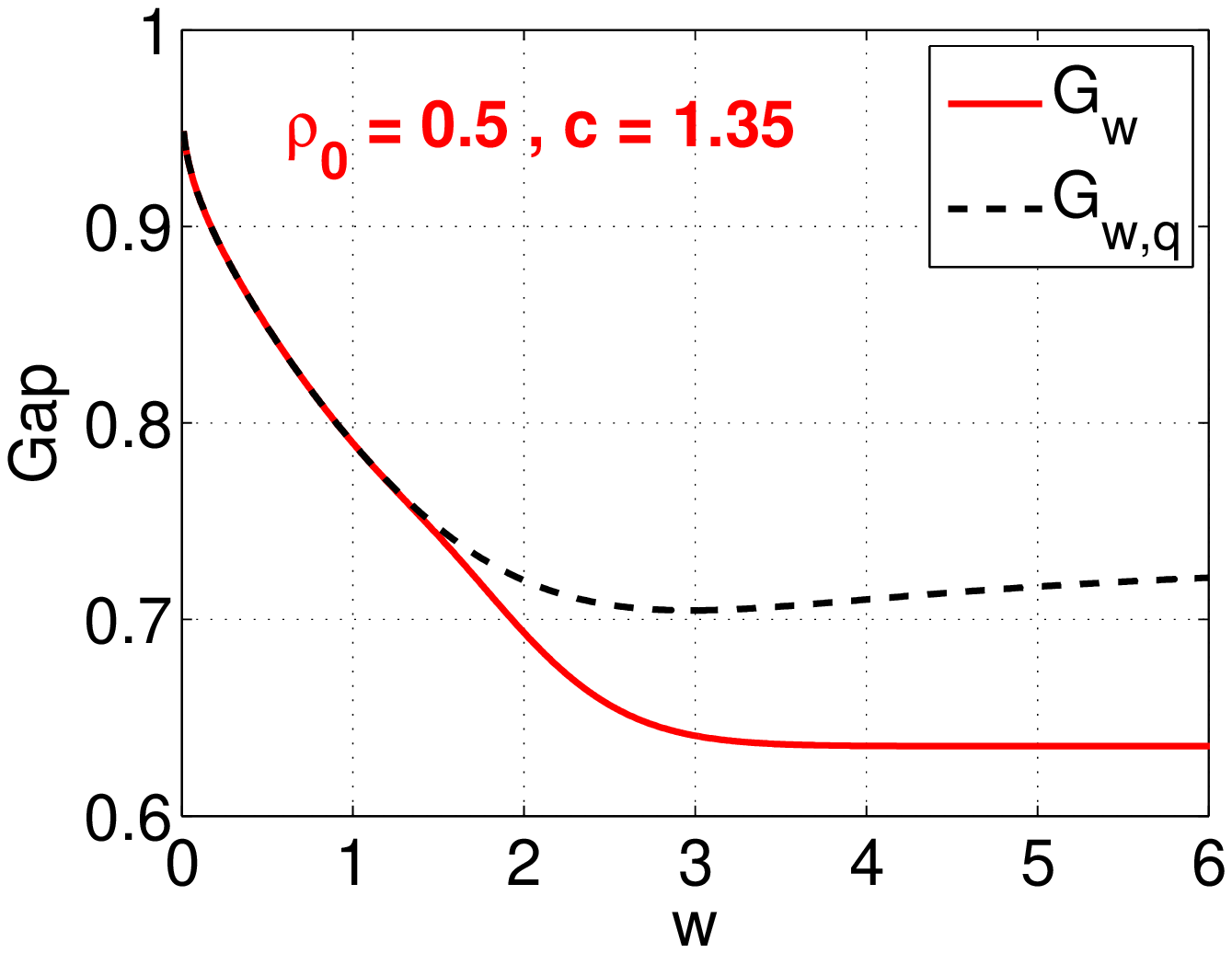}
\includegraphics[width = 2.2in]{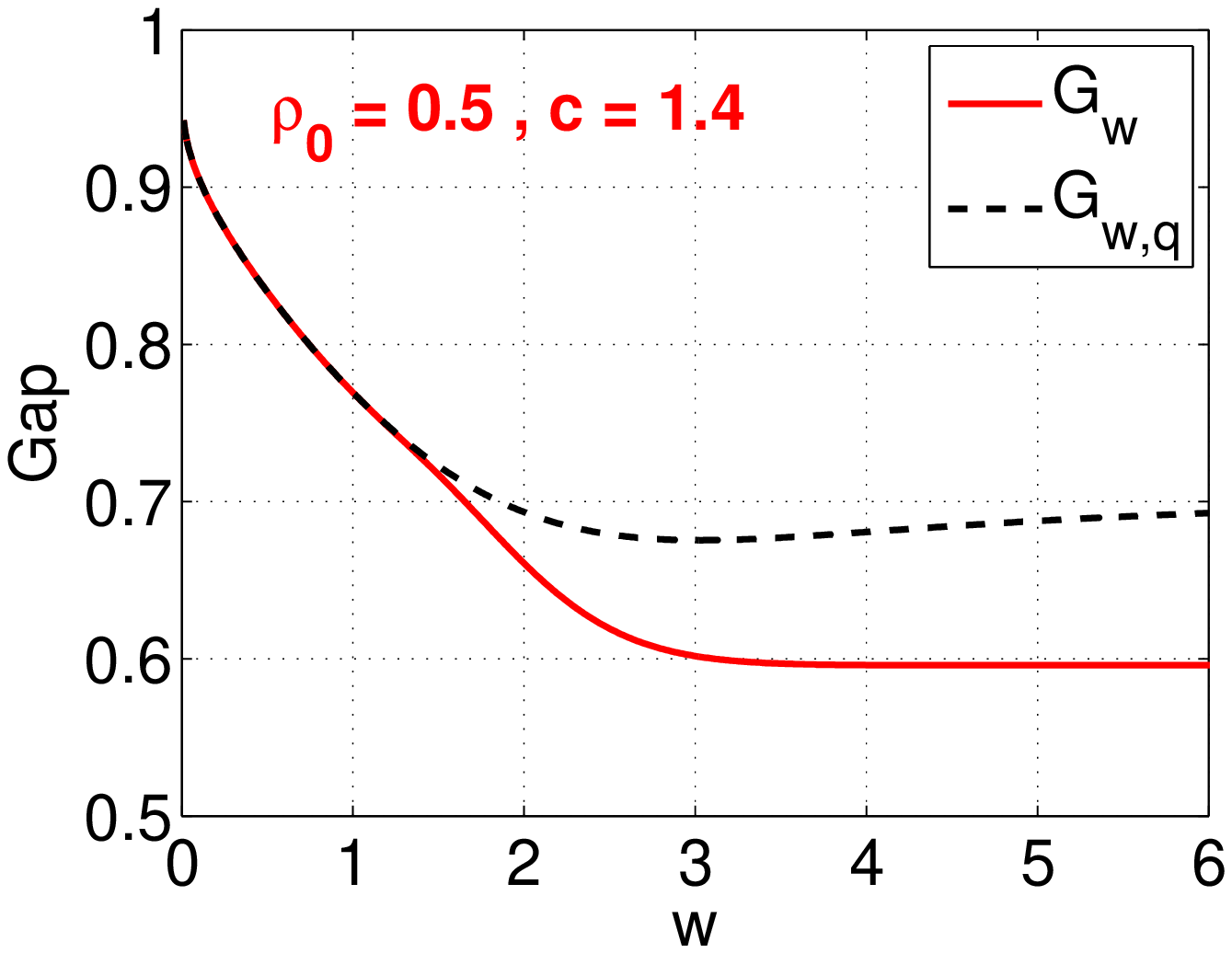}
}

\end{center}
\vspace{-.2in}
\caption{The gaps $G_w$ and $G_{w,q}$ as functions of $w$, for $\rho_0 = 0.5$ and a range of $c$ values.}\label{fig_GwqR05C}
\end{figure}

Figure~\ref{fig_GwqR099C}, Figure~\ref{fig_GwqR095C}, Figure~\ref{fig_GwqR09C}, and Figure~\ref{fig_GwqR05C} present $G_w$ and $G_{w,q}$ as functions of $w$, for $\rho_0 = 0.99$, $\rho_0 = 0.95$, $\rho_0 = 0.9$ and $\rho_0 = 0.5$, respectively. In each figure, we plot the curves for a wide range of $c$ values.  These figures illustrate where the optimum $w$ values are obtained. Clearly, in the high similarity region, the smallest $G$ values are obtained at low $w$ values, especially at small $c$. In the low (or moderate) similarity   region, the smallest $G$ values are usually attained at relatively large $w$.\\

\begin{figure}[h!]
\begin{center}
\mbox{
\includegraphics[width = 2.2in]{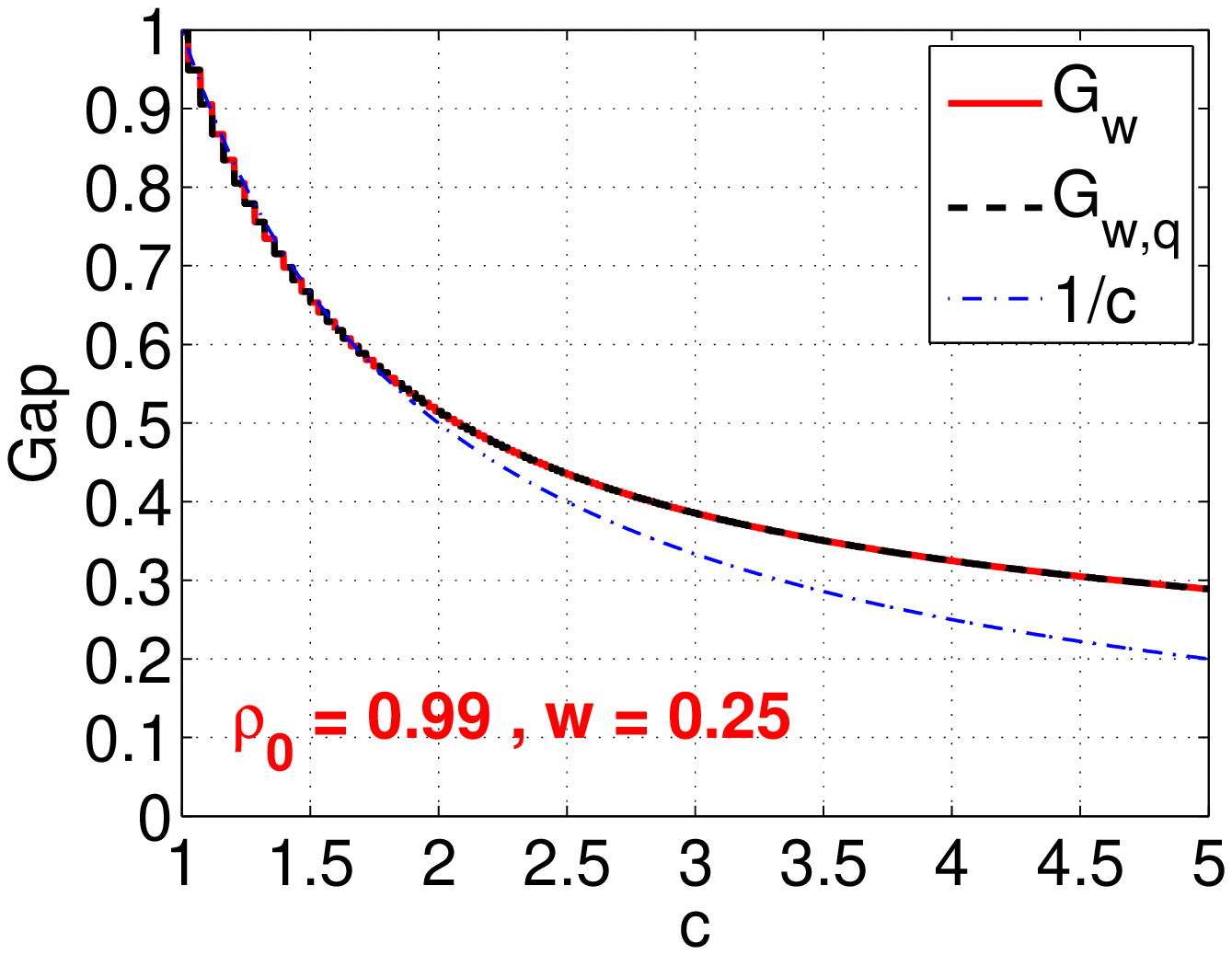}
\includegraphics[width = 2.2in]{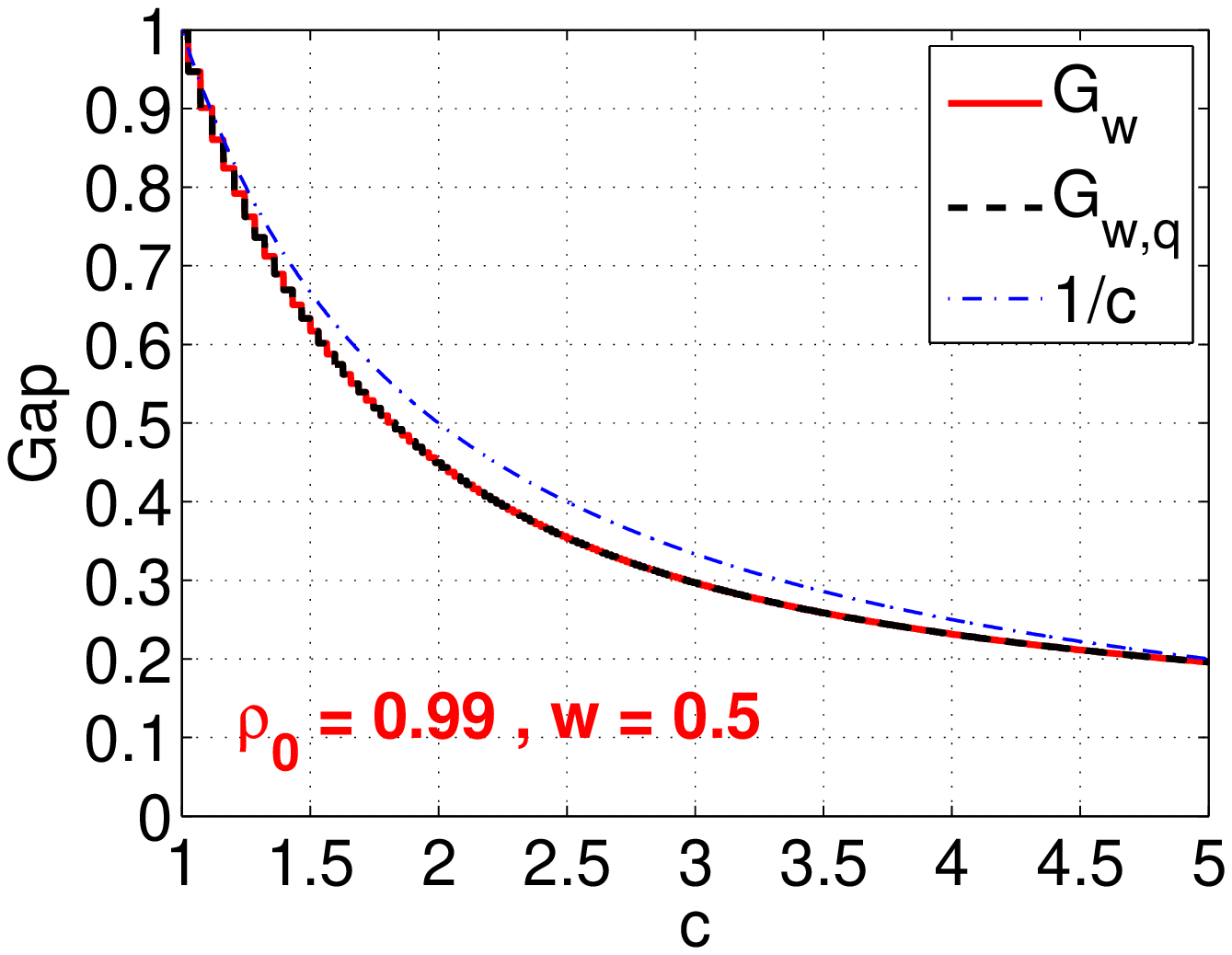}
\includegraphics[width = 2.2in]{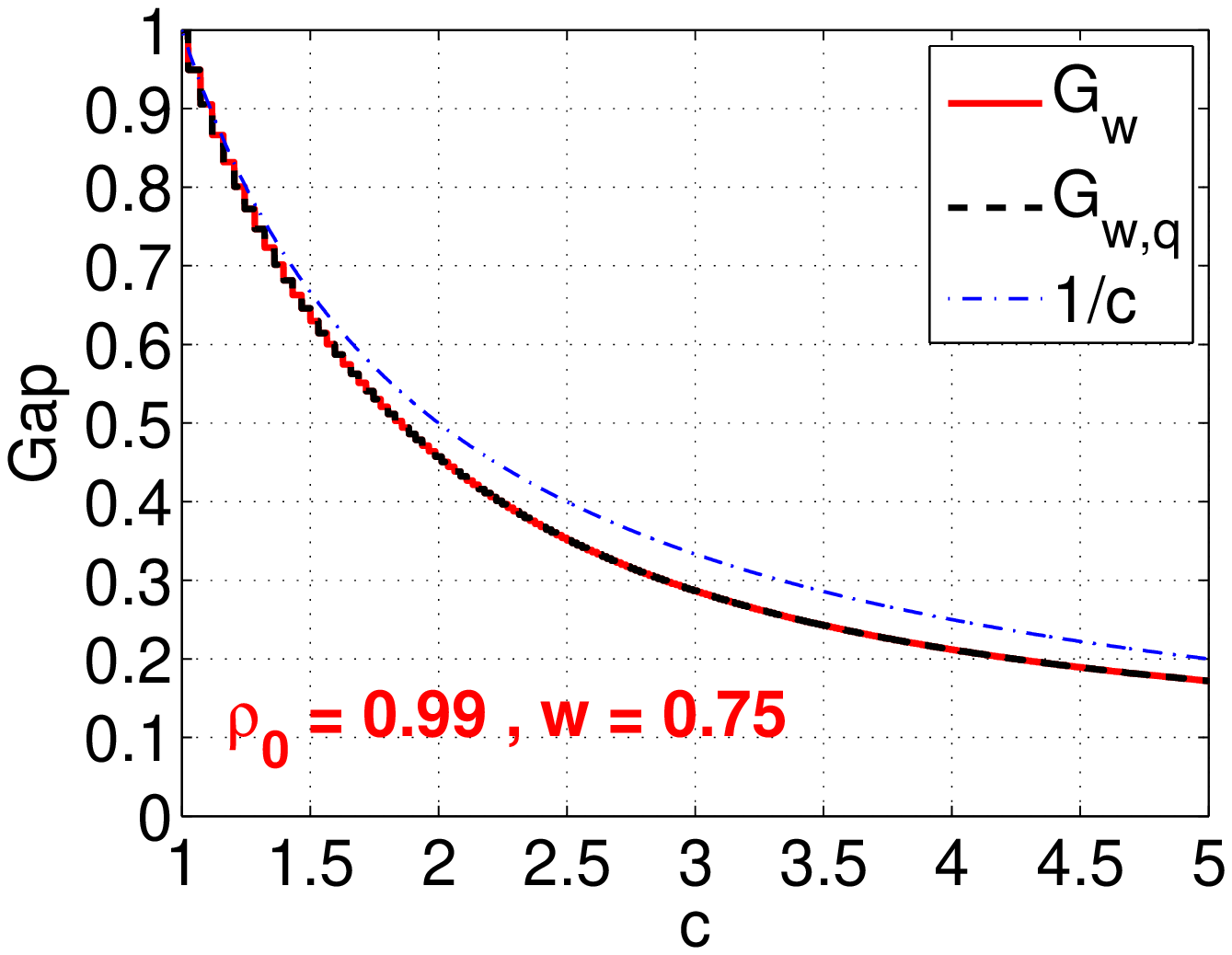}
}
\mbox{
\includegraphics[width = 2.2in]{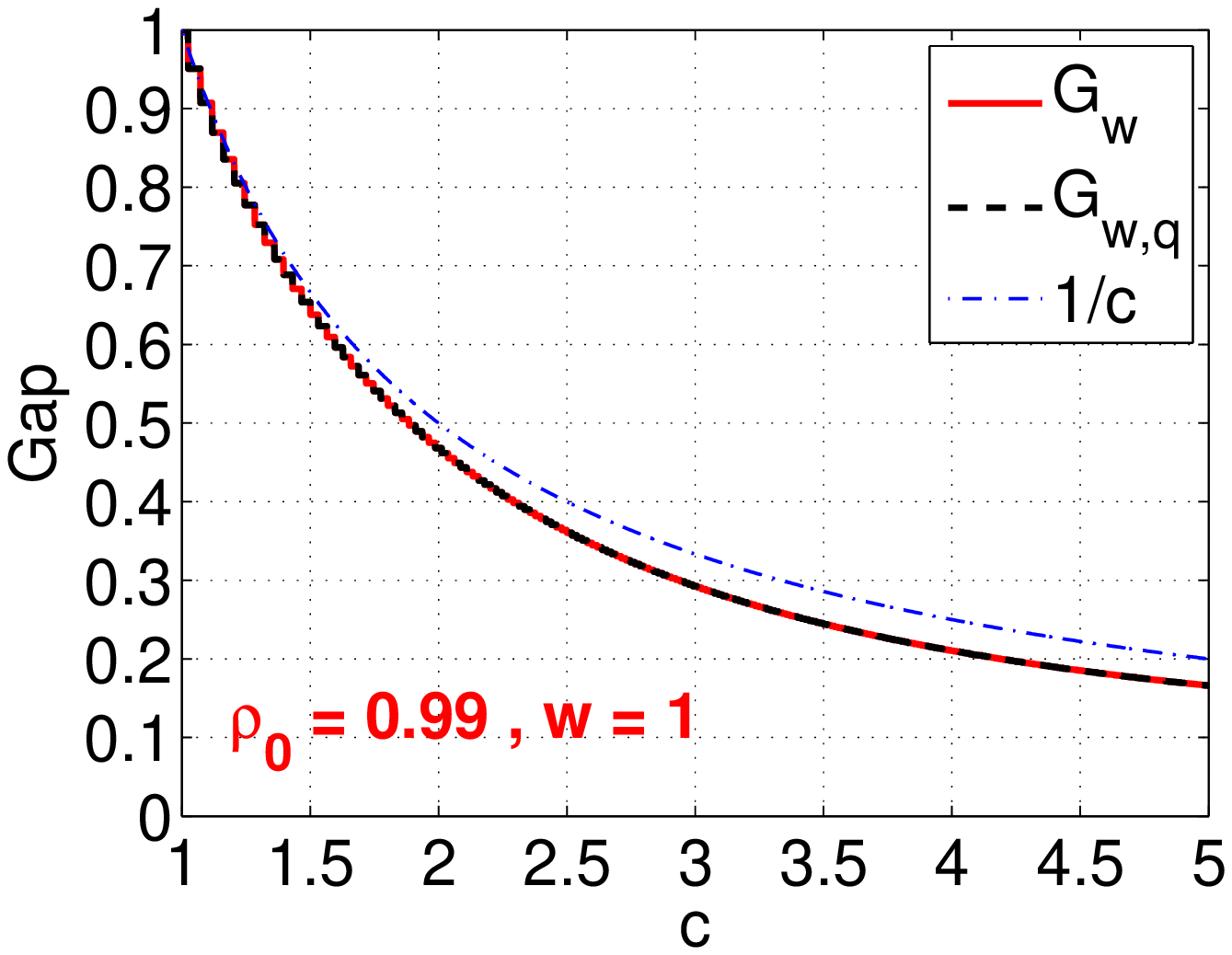}
\includegraphics[width = 2.2in]{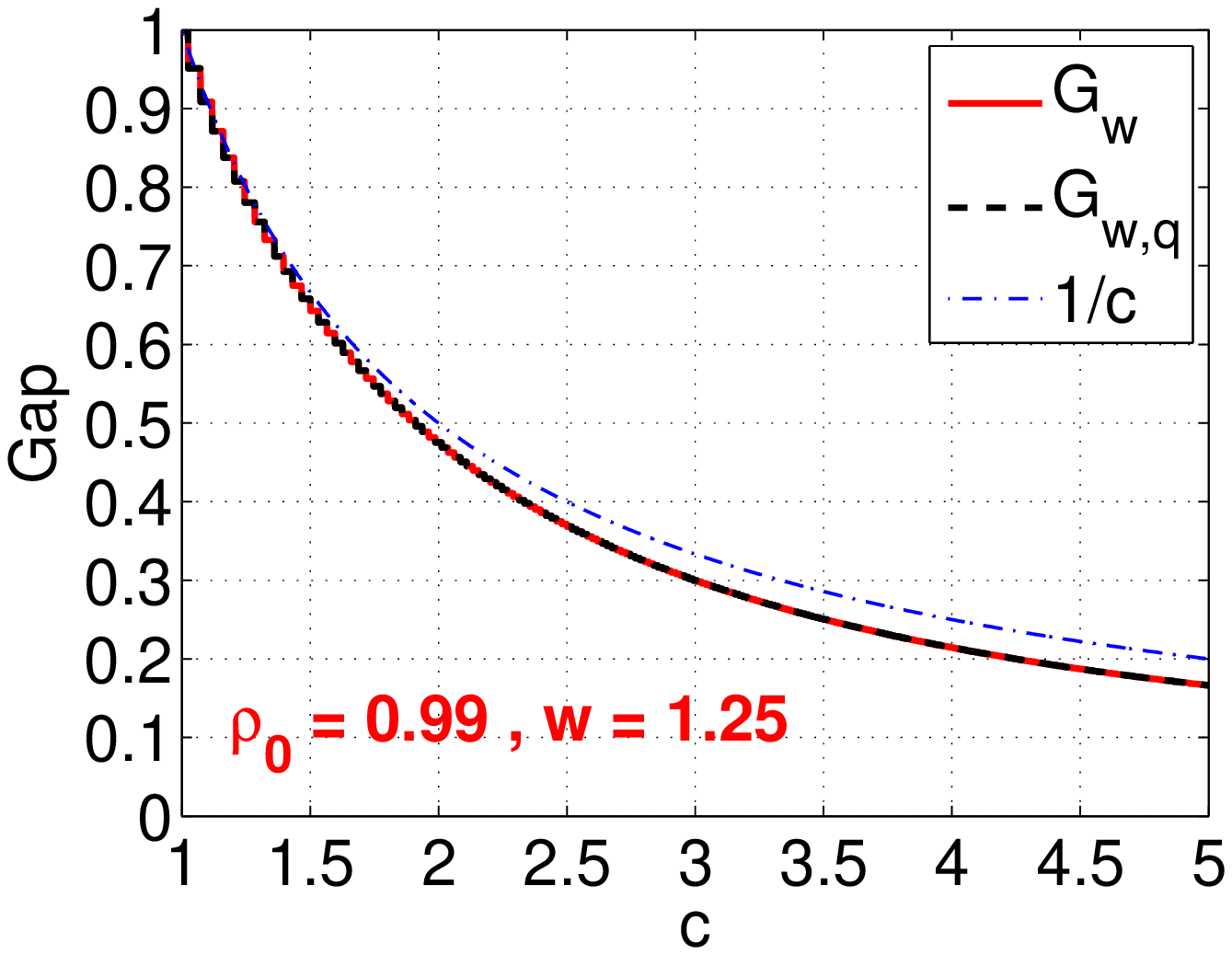}
\includegraphics[width = 2.2in]{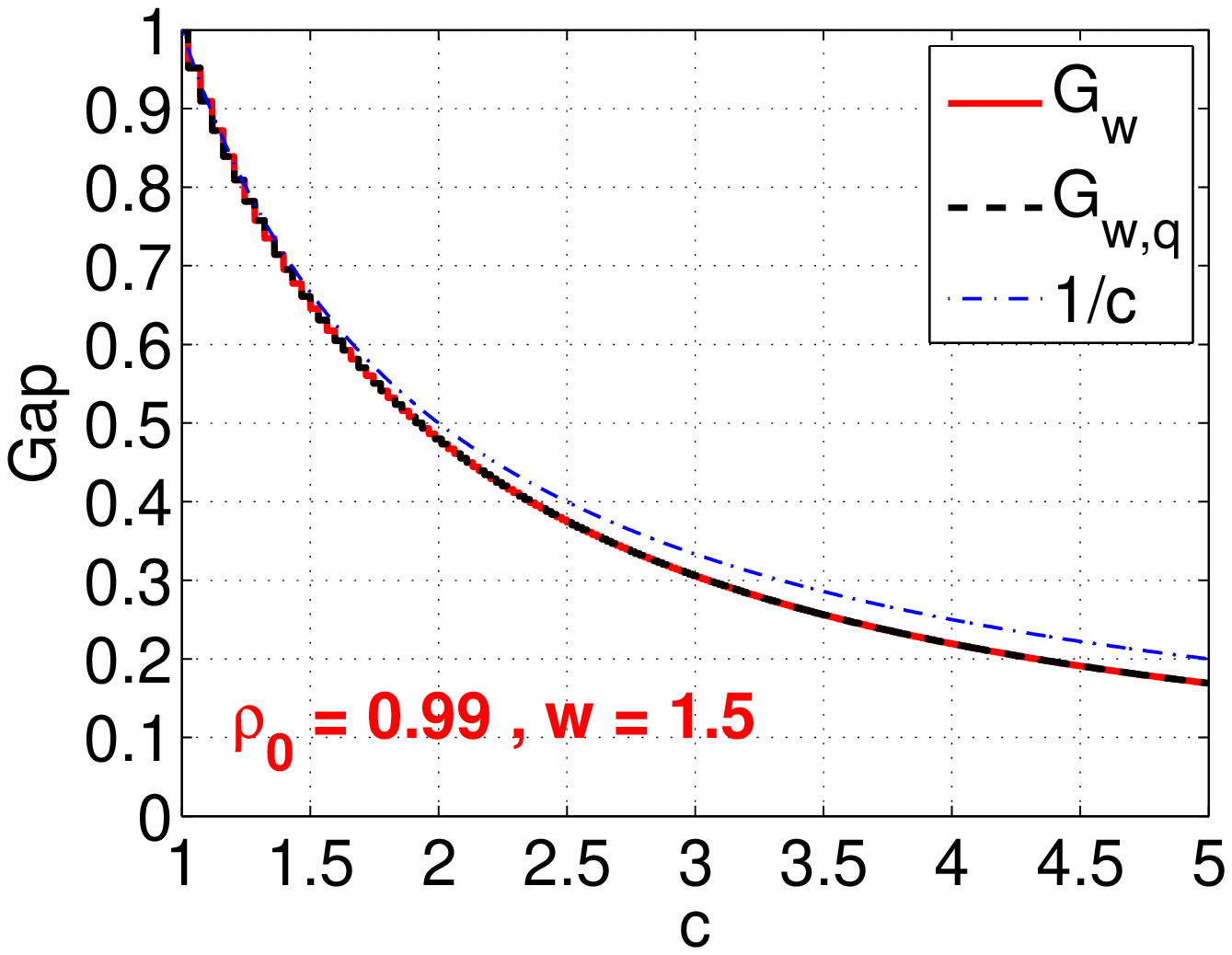}
}

\mbox{
\includegraphics[width = 2.2in]{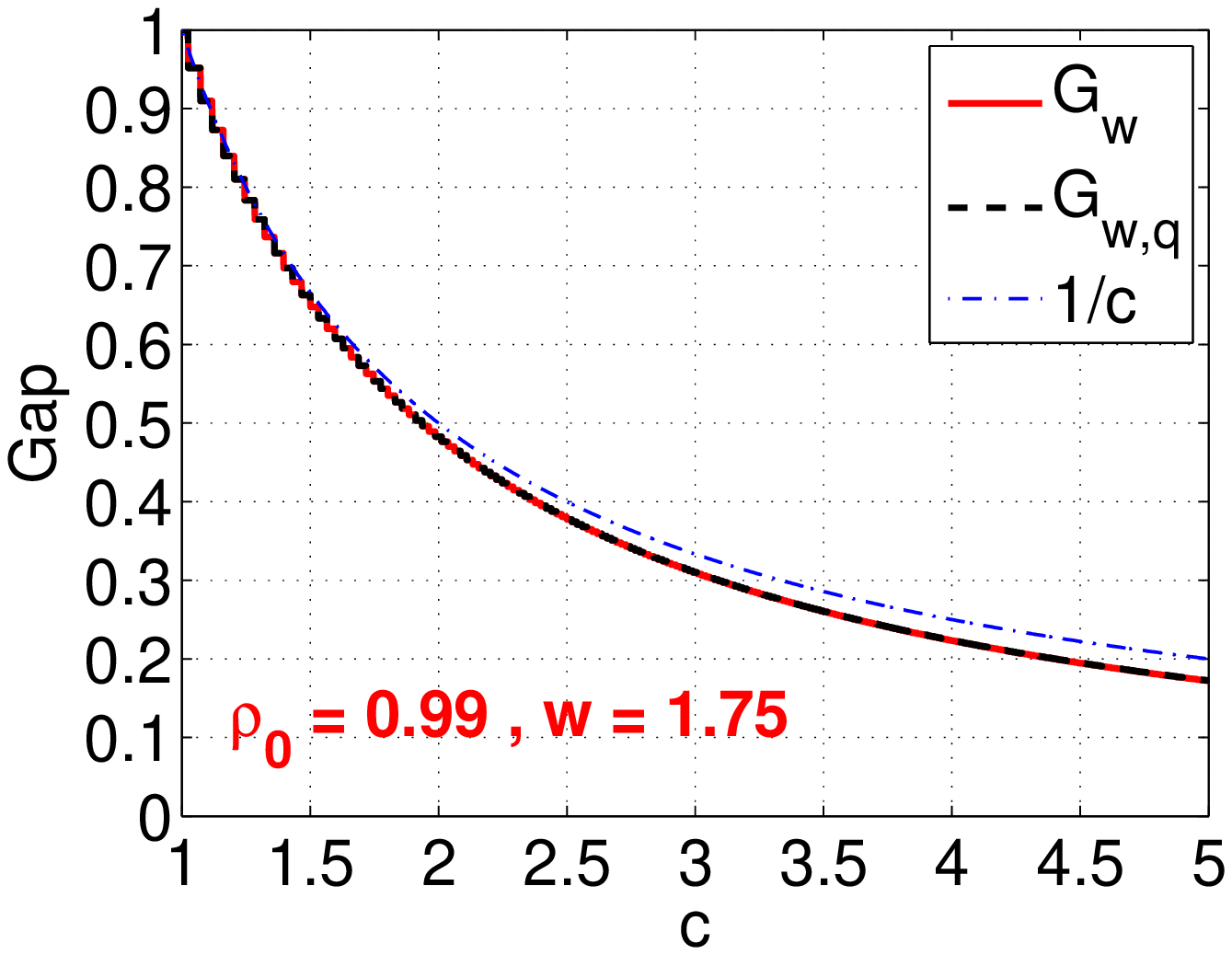}
\includegraphics[width = 2.2in]{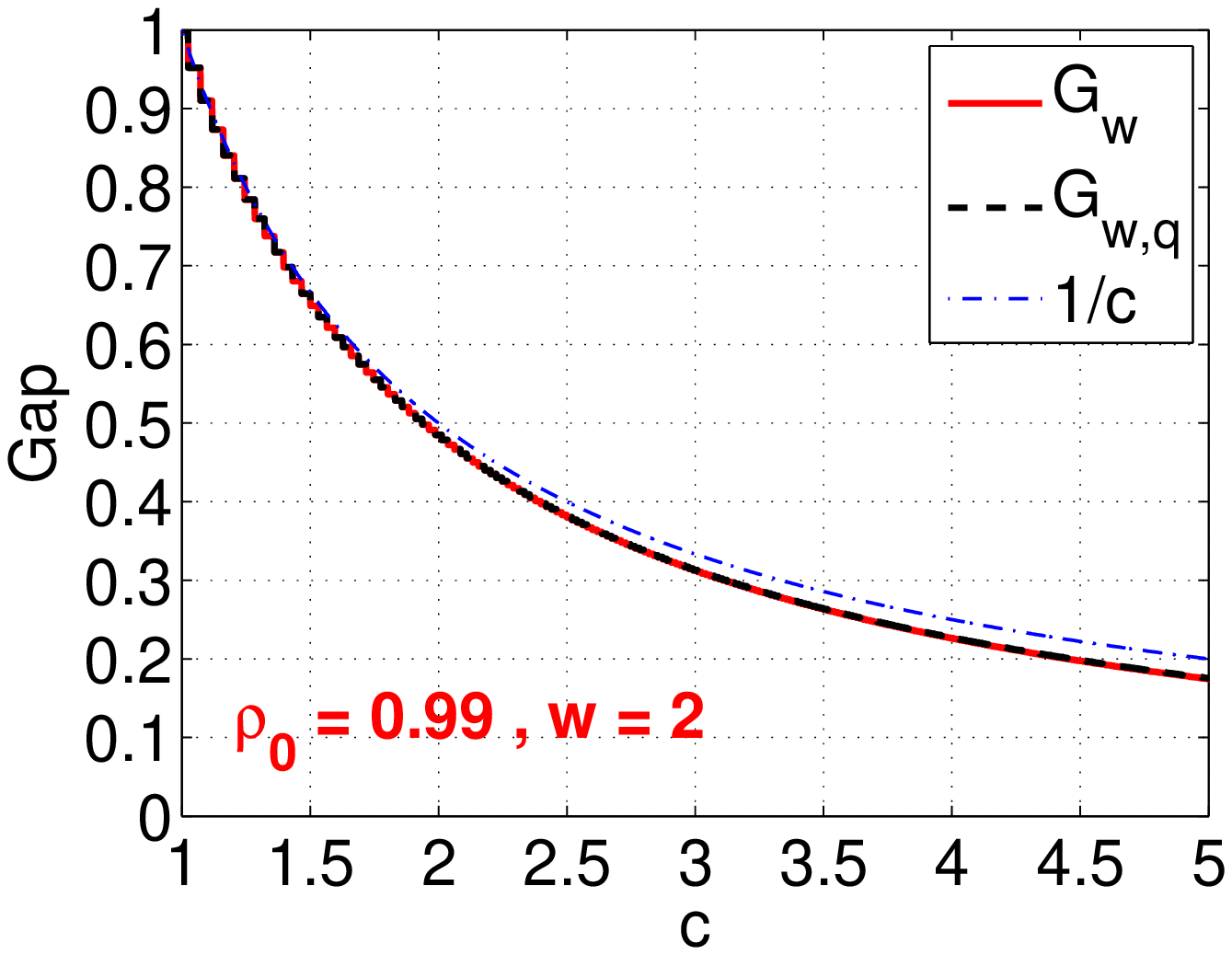}
\includegraphics[width = 2.2in]{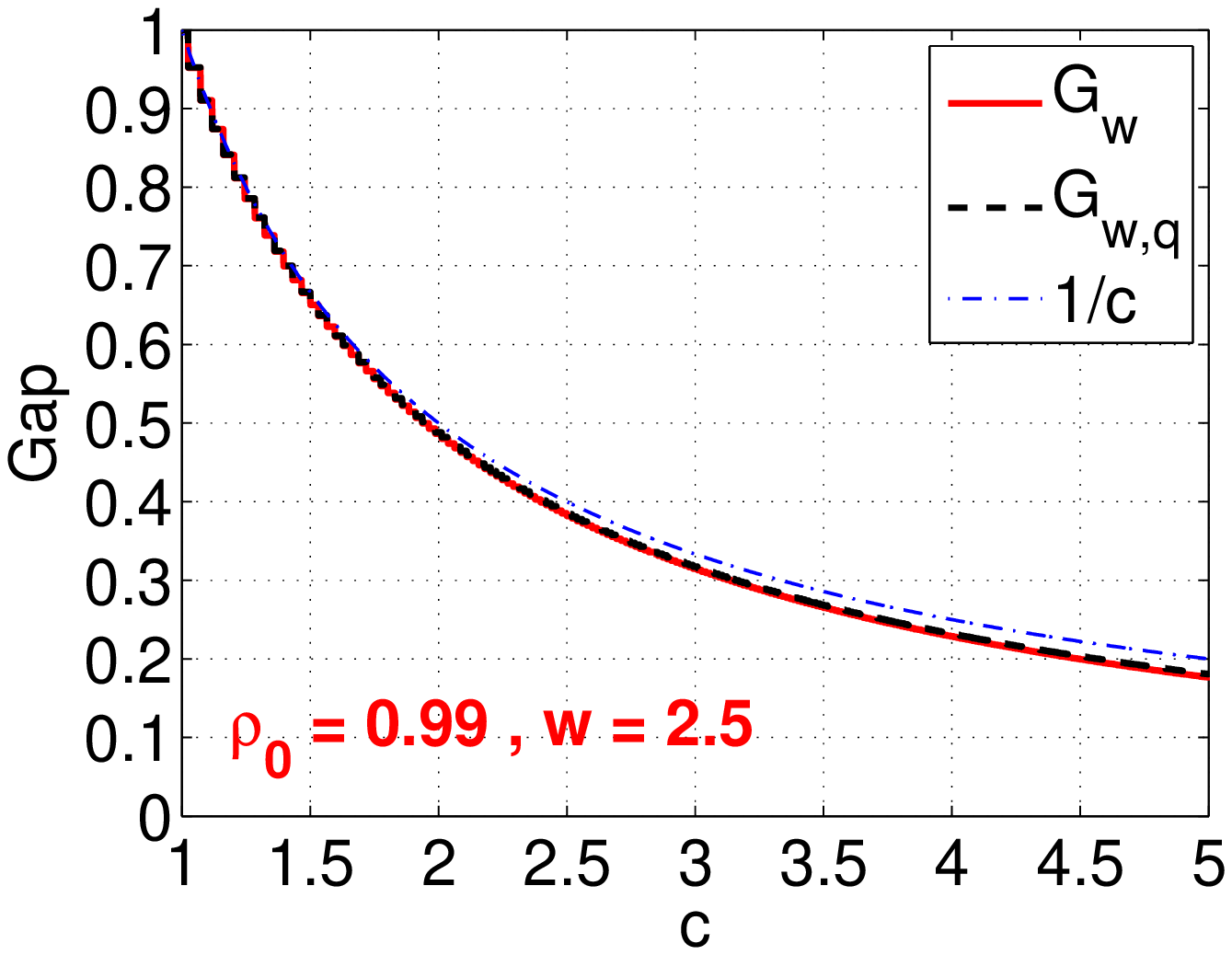}
}

\mbox{
\includegraphics[width = 2.2in]{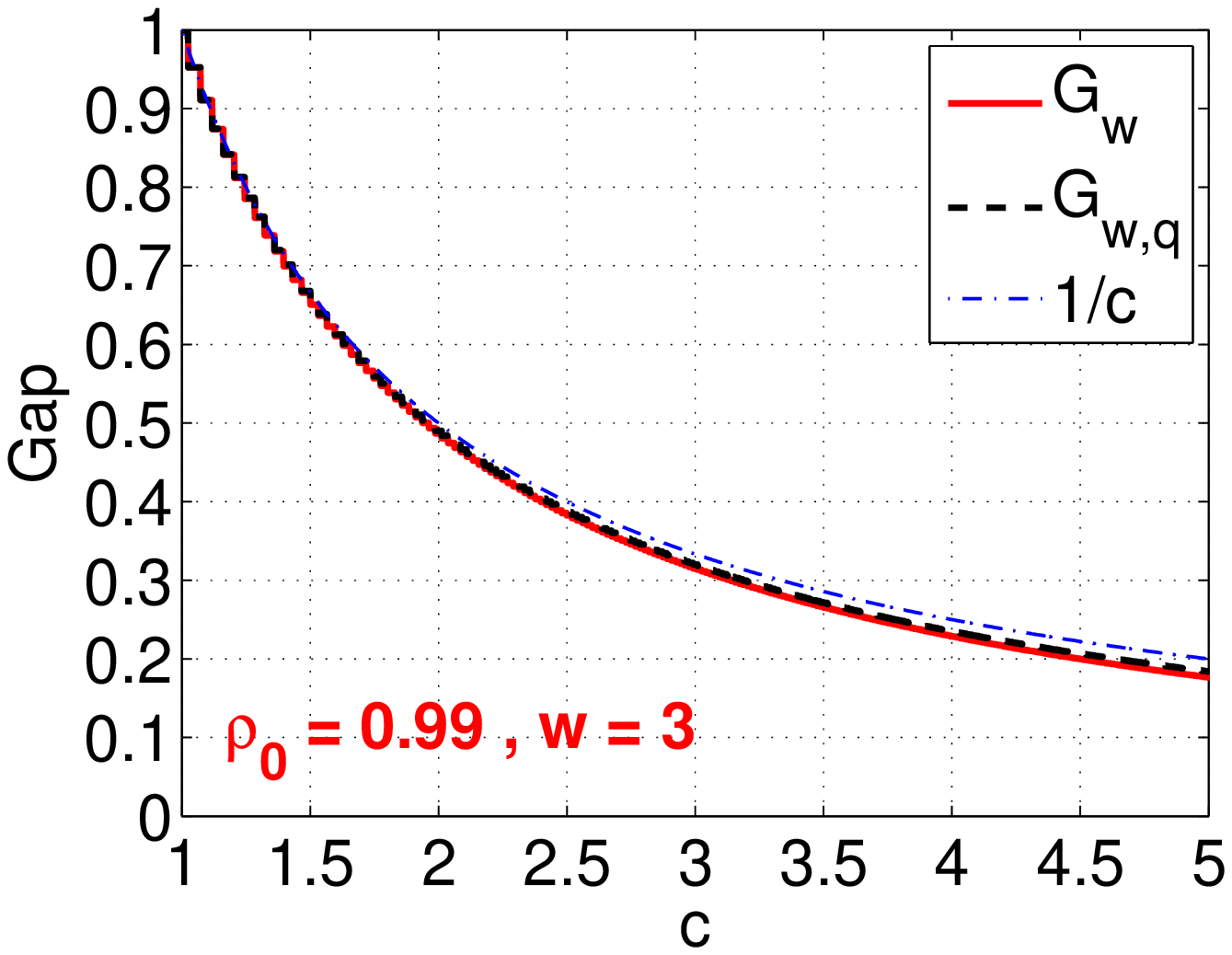}
\includegraphics[width = 2.2in]{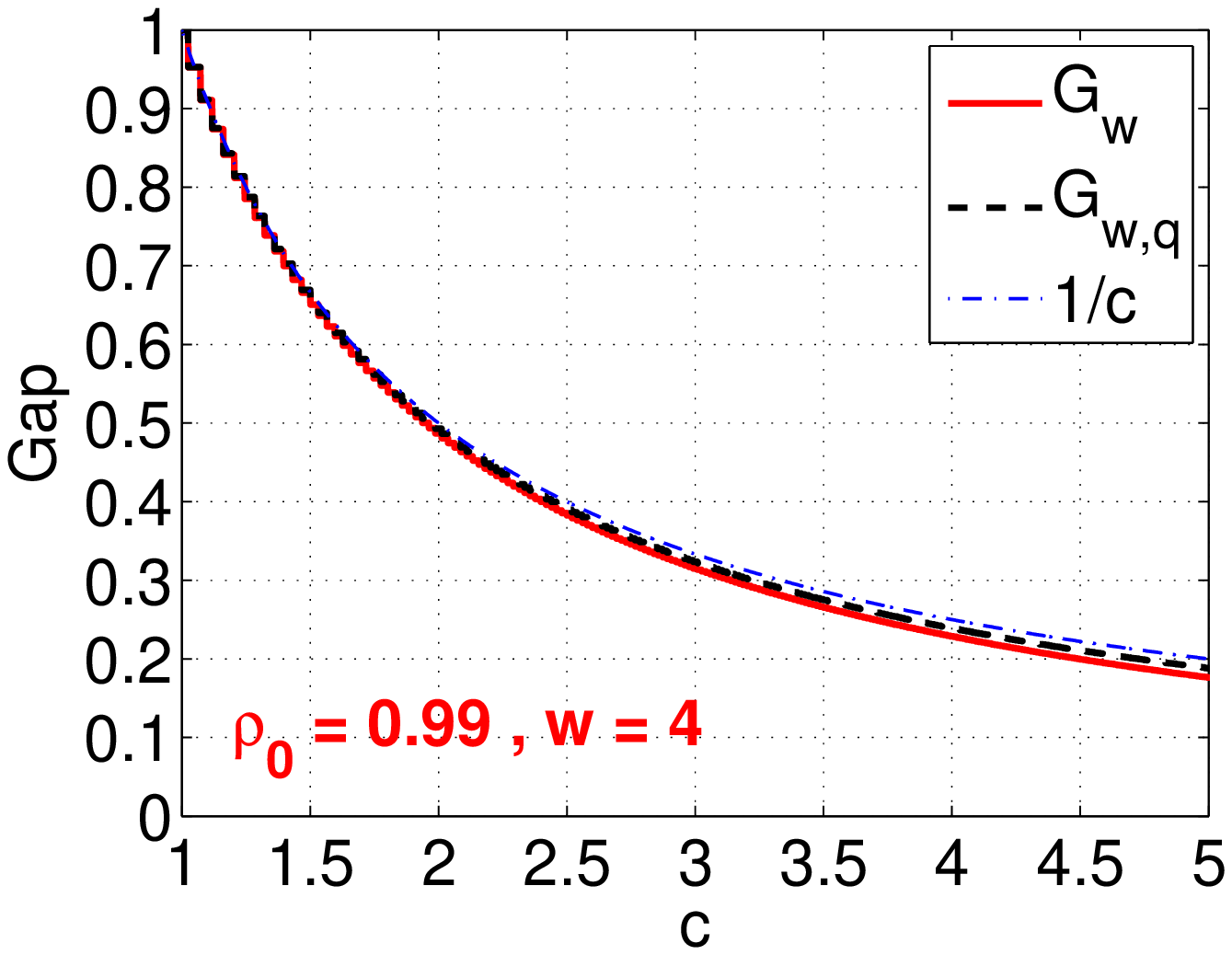}
\includegraphics[width = 2.2in]{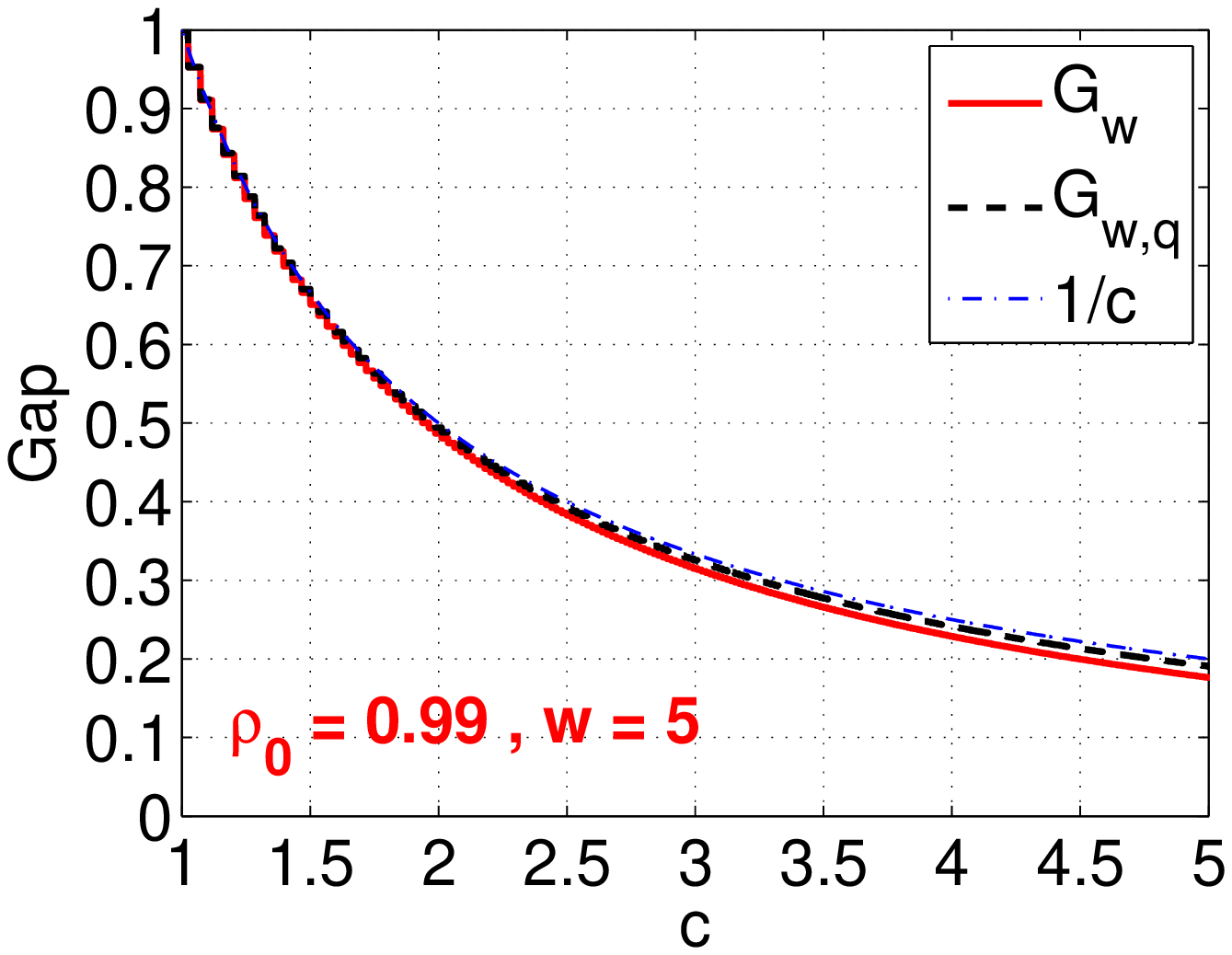}
}
\end{center}
\vspace{-.2in}
\caption{The gaps $G_w$ and $G_{w,q}$ as functions of $c$, for $\rho_0 = 0.99$. In each panel, we plot both $G_w$ and $G_{w,q}$ for a particular $w$ value. }\label{fig_GwqR099W}
\end{figure}

\begin{figure}[h!]
\begin{center}
\mbox{
\includegraphics[width = 2.2in]{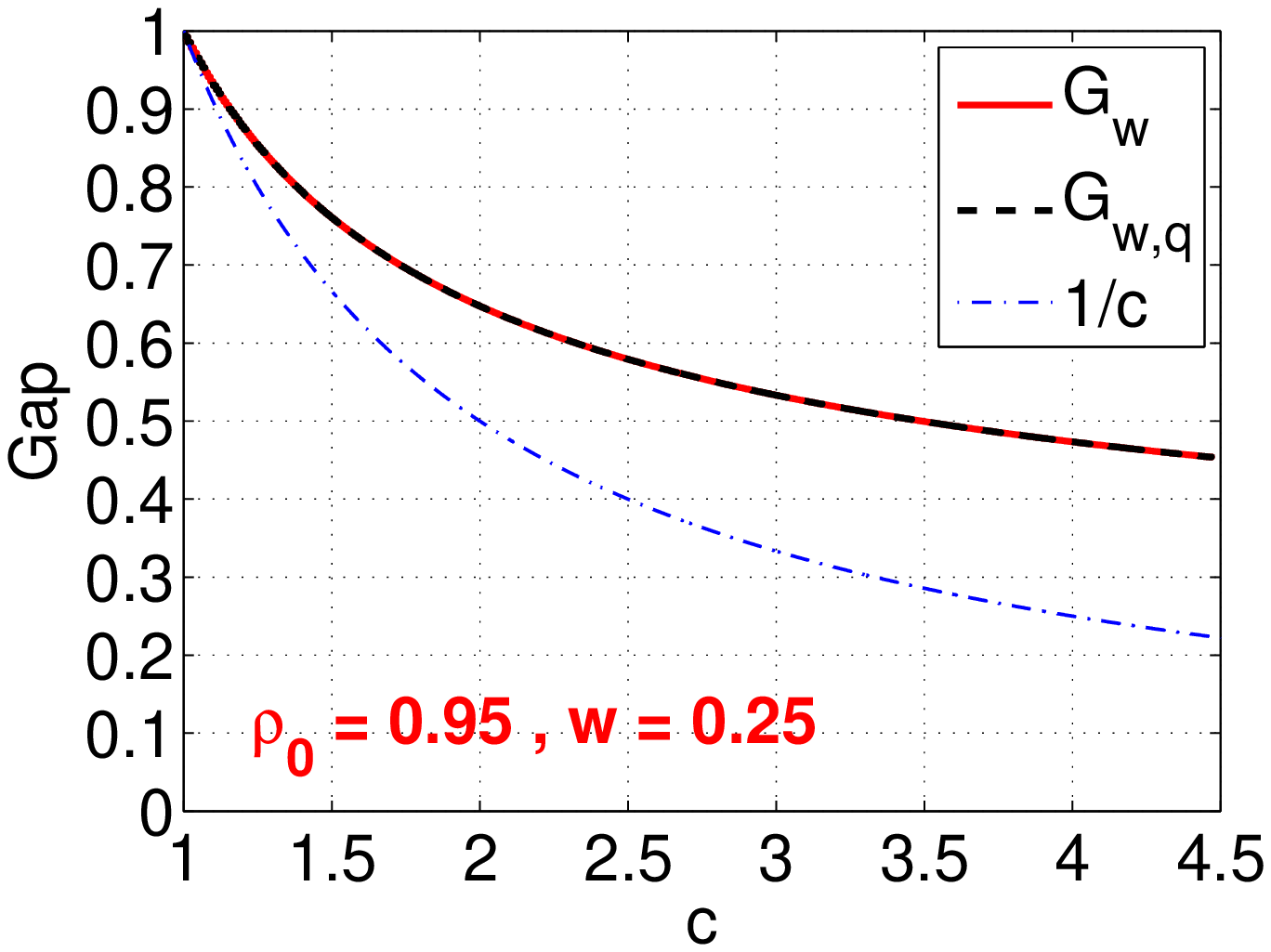}
\includegraphics[width = 2.2in]{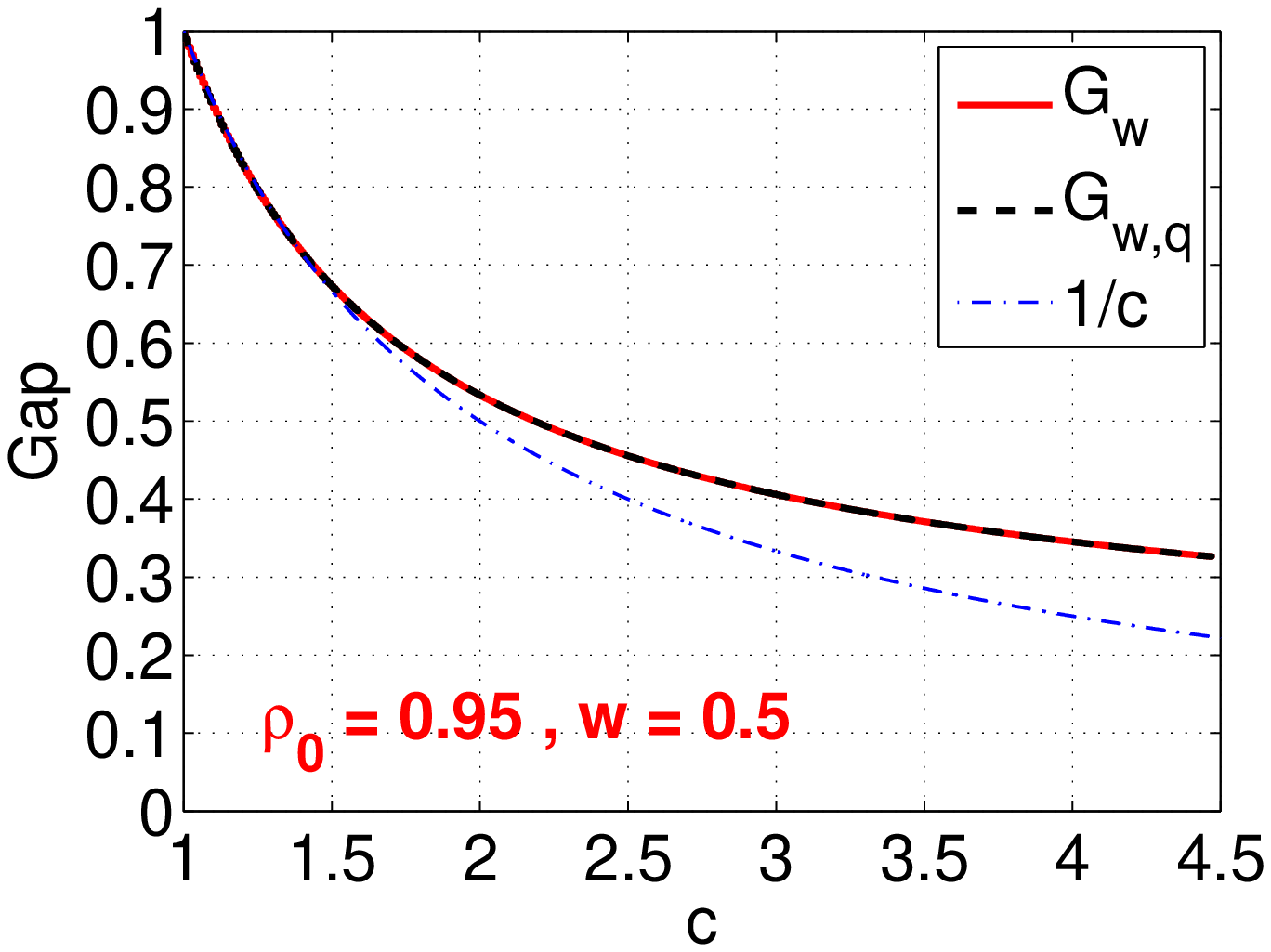}
\includegraphics[width = 2.2in]{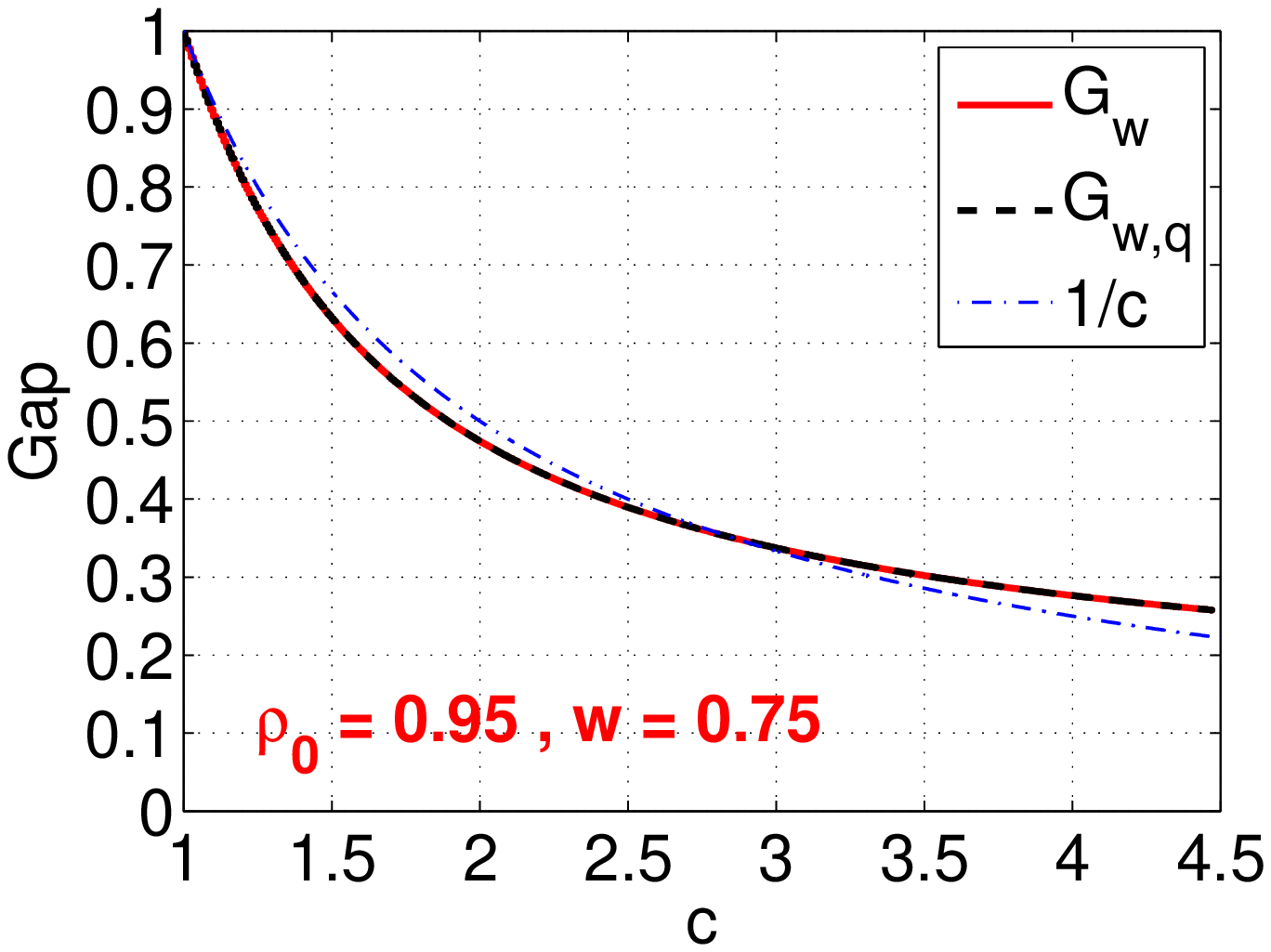}
}
\mbox{
\includegraphics[width = 2.2in]{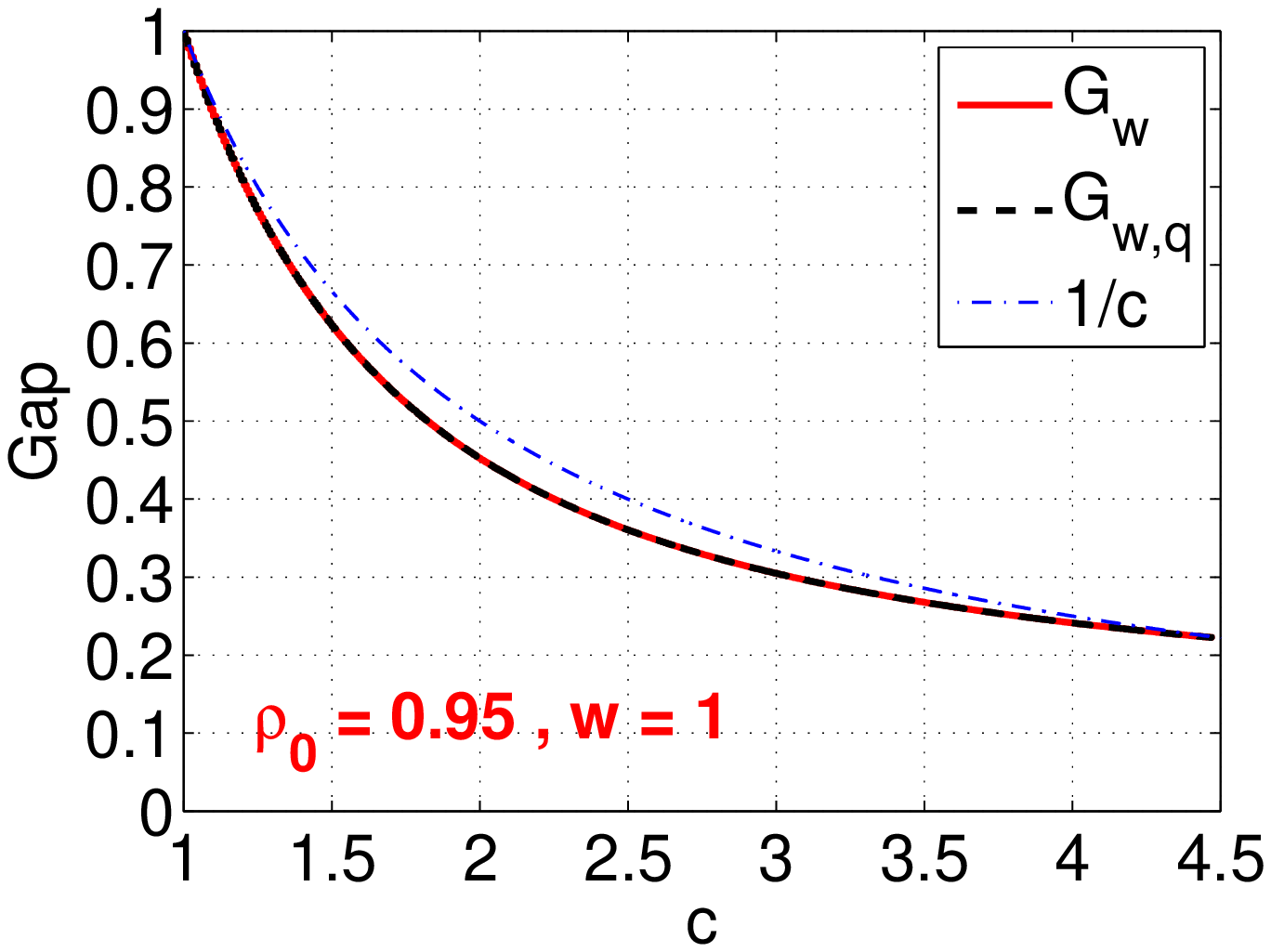}
\includegraphics[width = 2.2in]{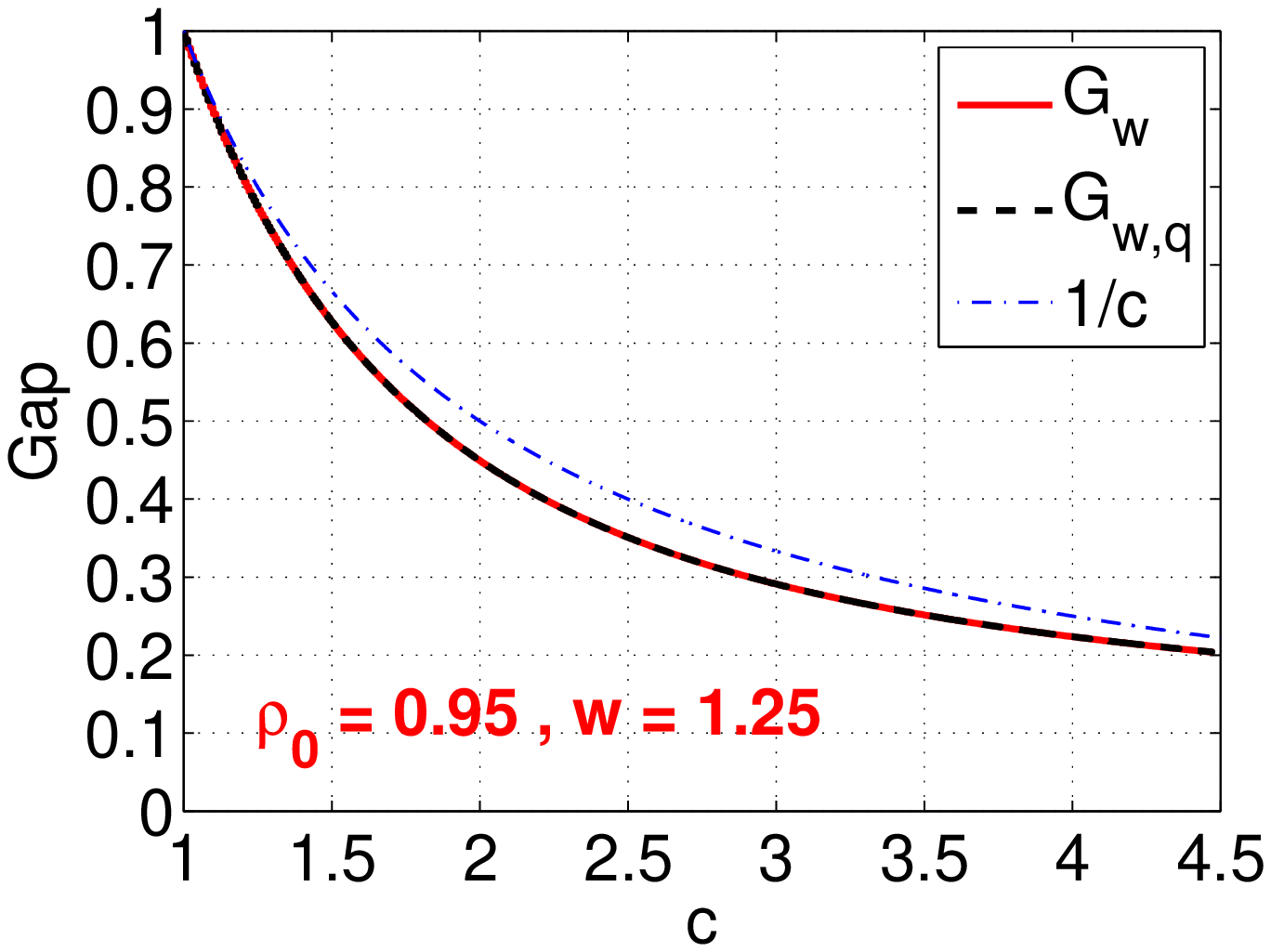}
\includegraphics[width = 2.2in]{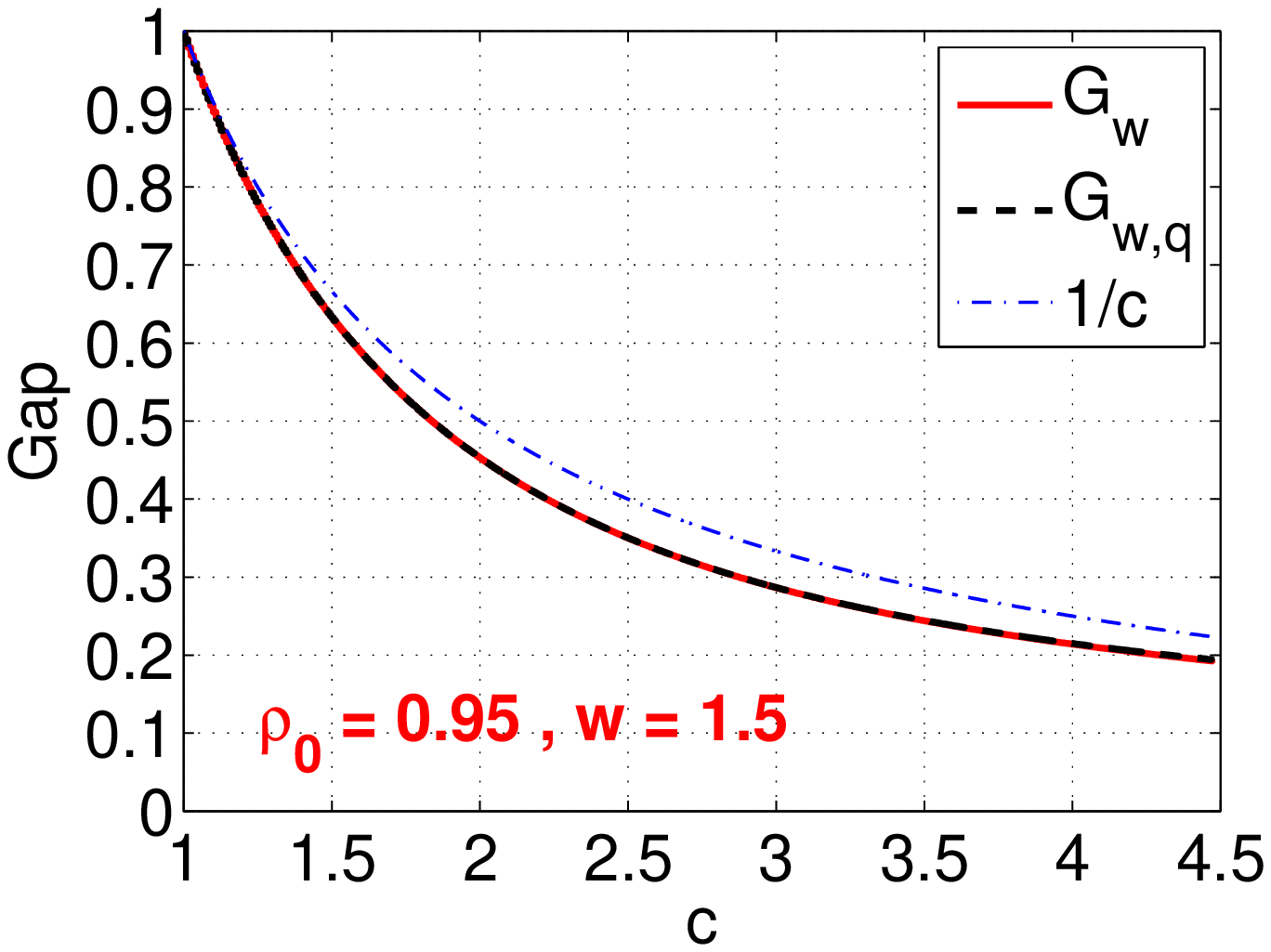}
}

\mbox{
\includegraphics[width = 2.2in]{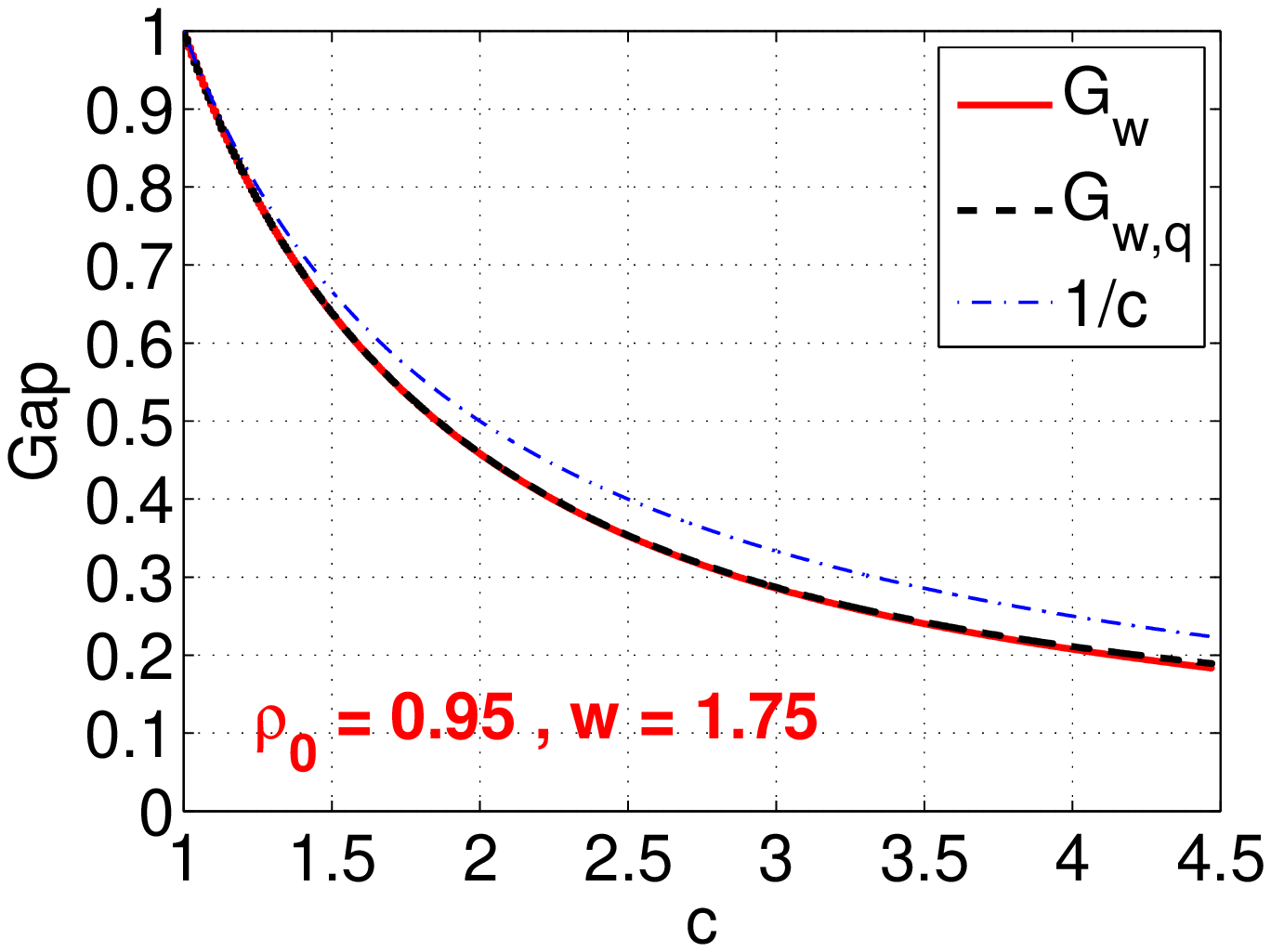}
\includegraphics[width = 2.2in]{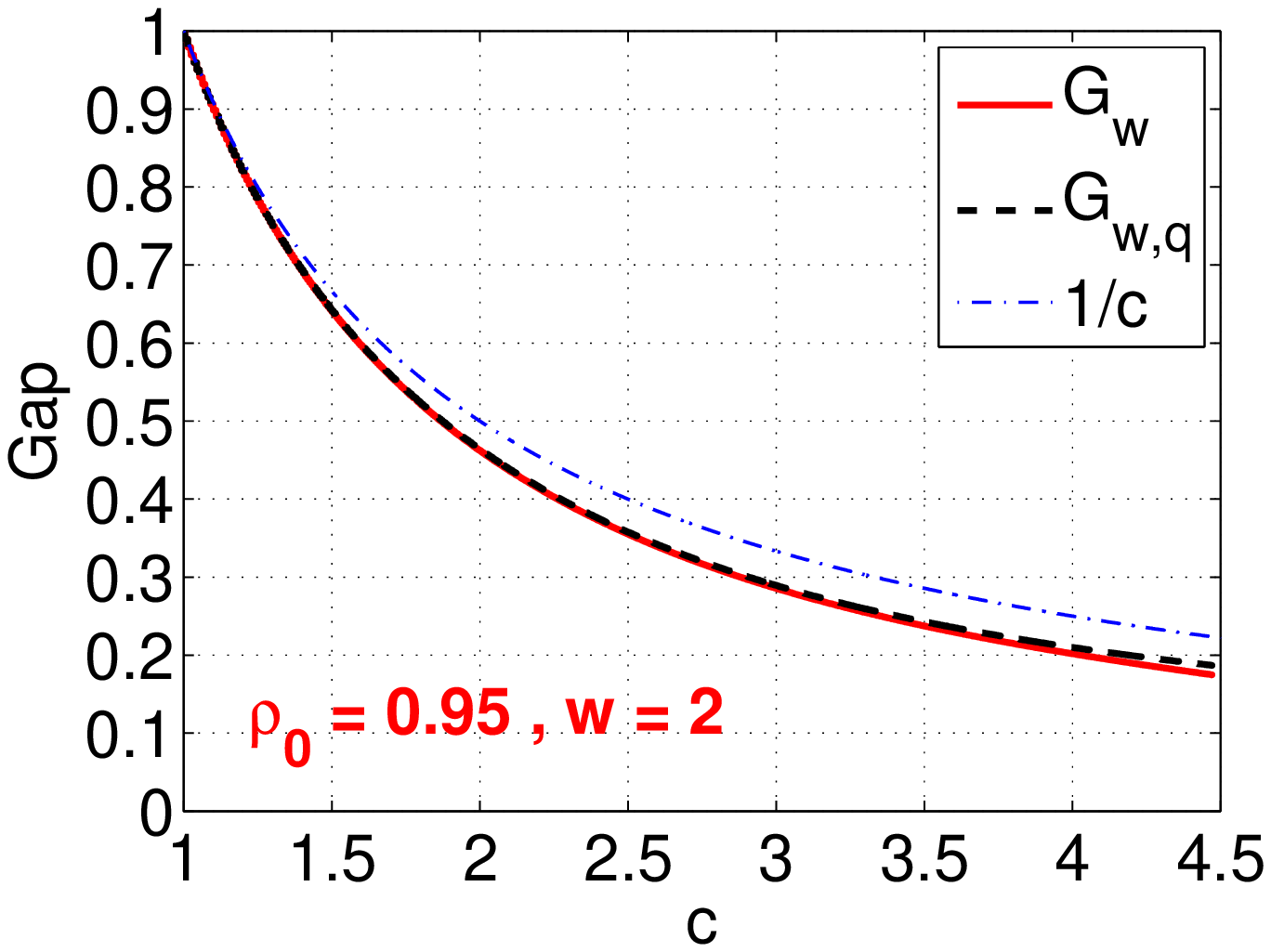}
\includegraphics[width = 2.2in]{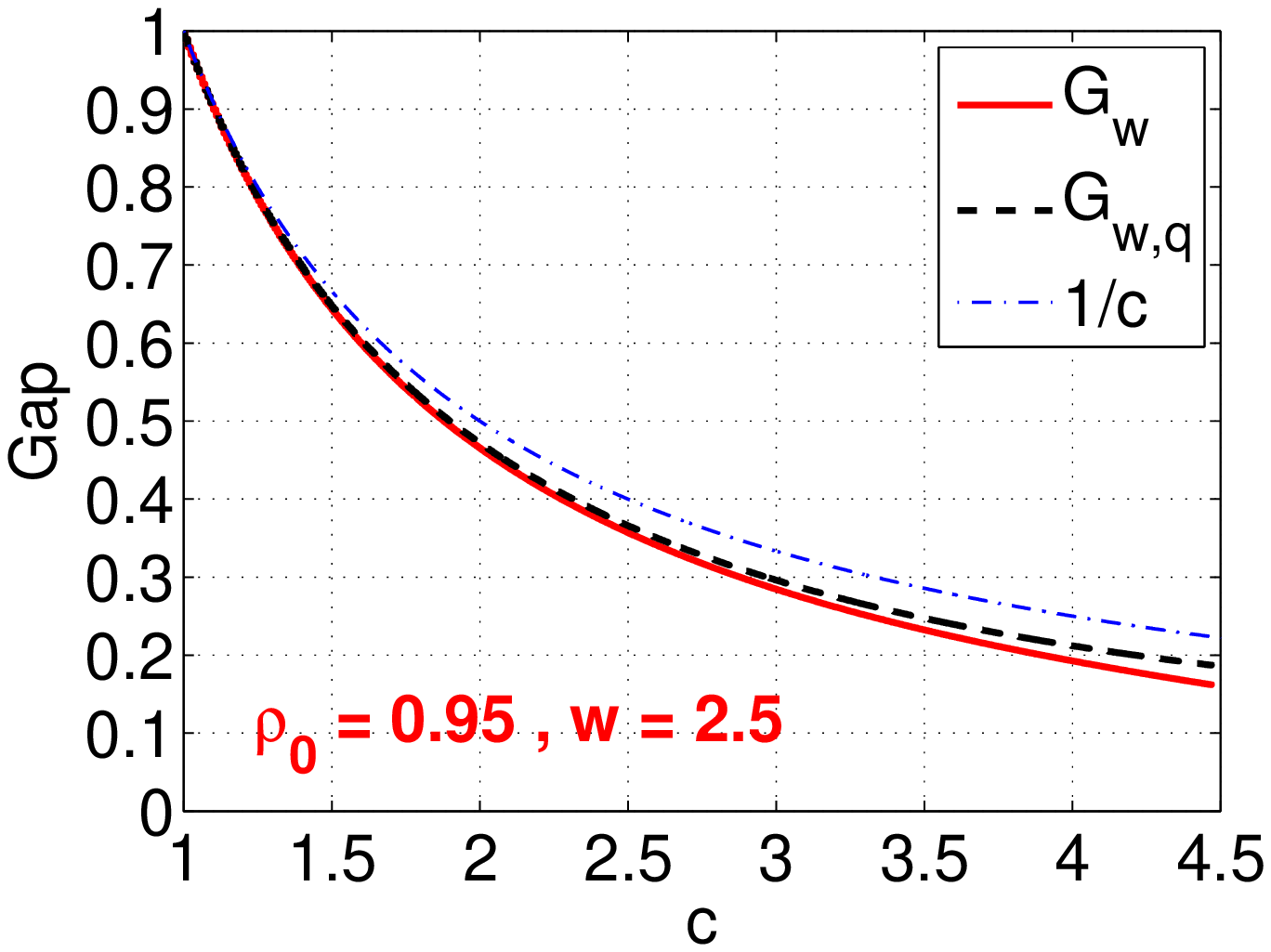}
}

\mbox{
\includegraphics[width = 2.2in]{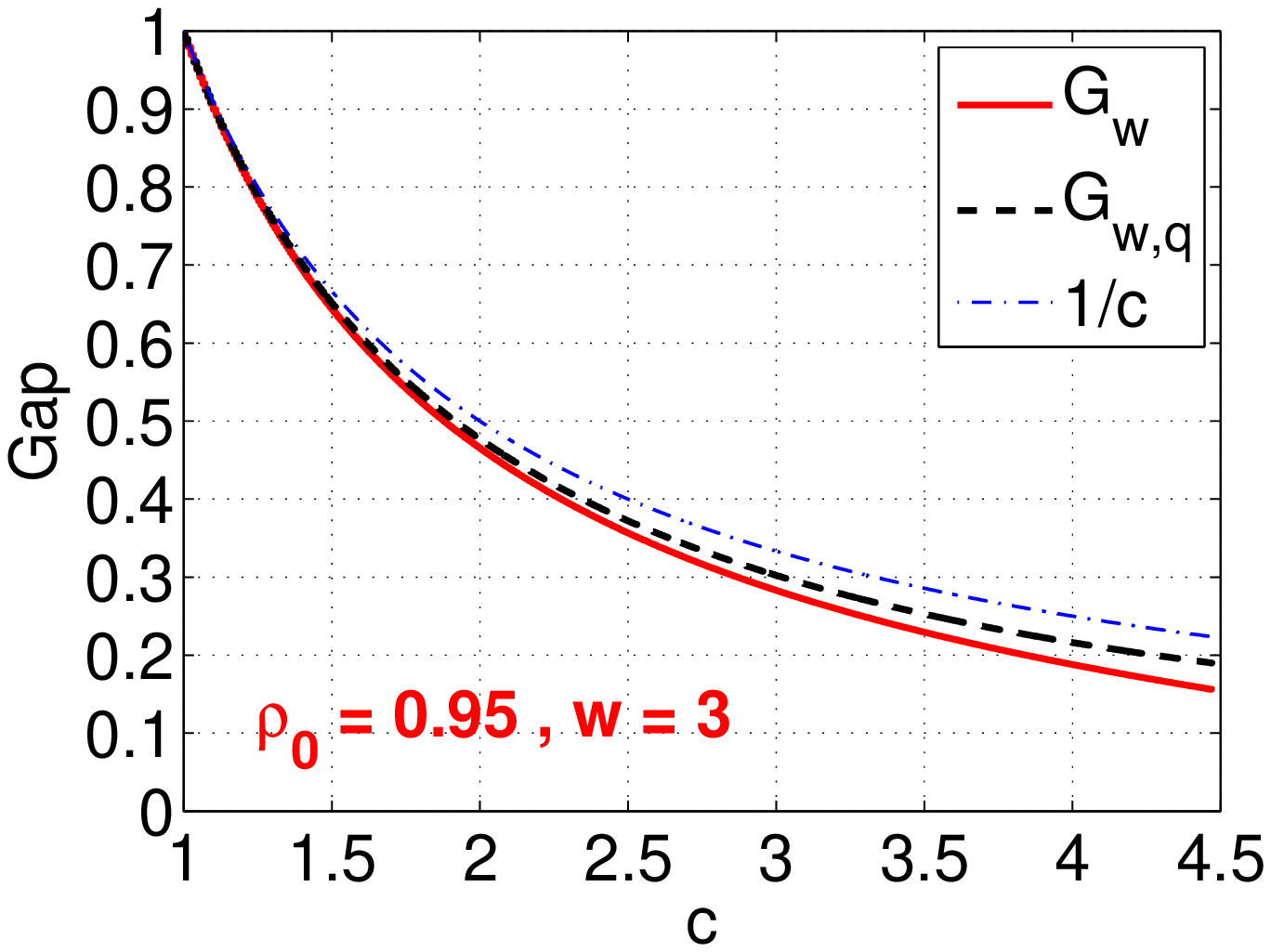}
\includegraphics[width = 2.2in]{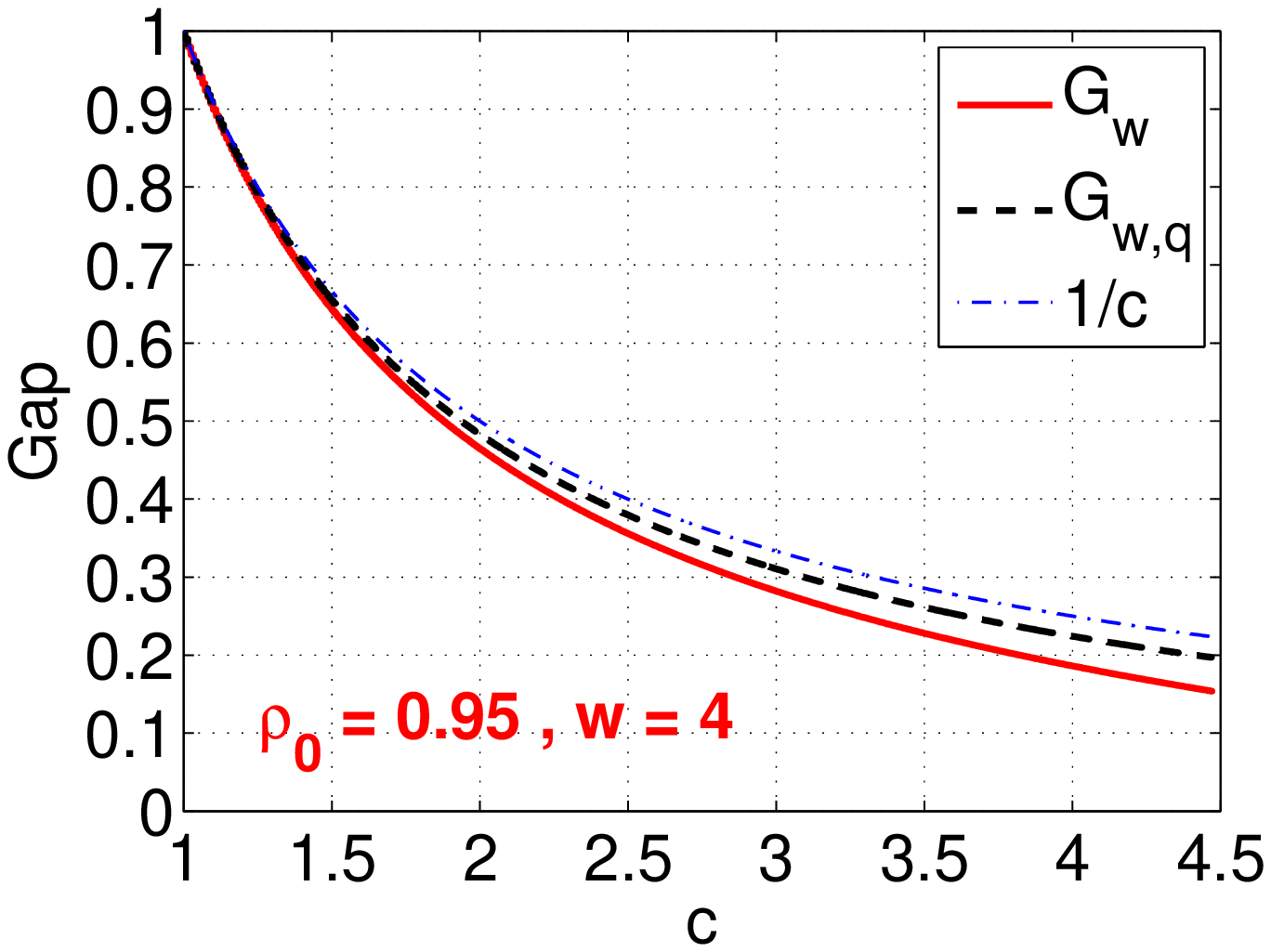}
\includegraphics[width = 2.2in]{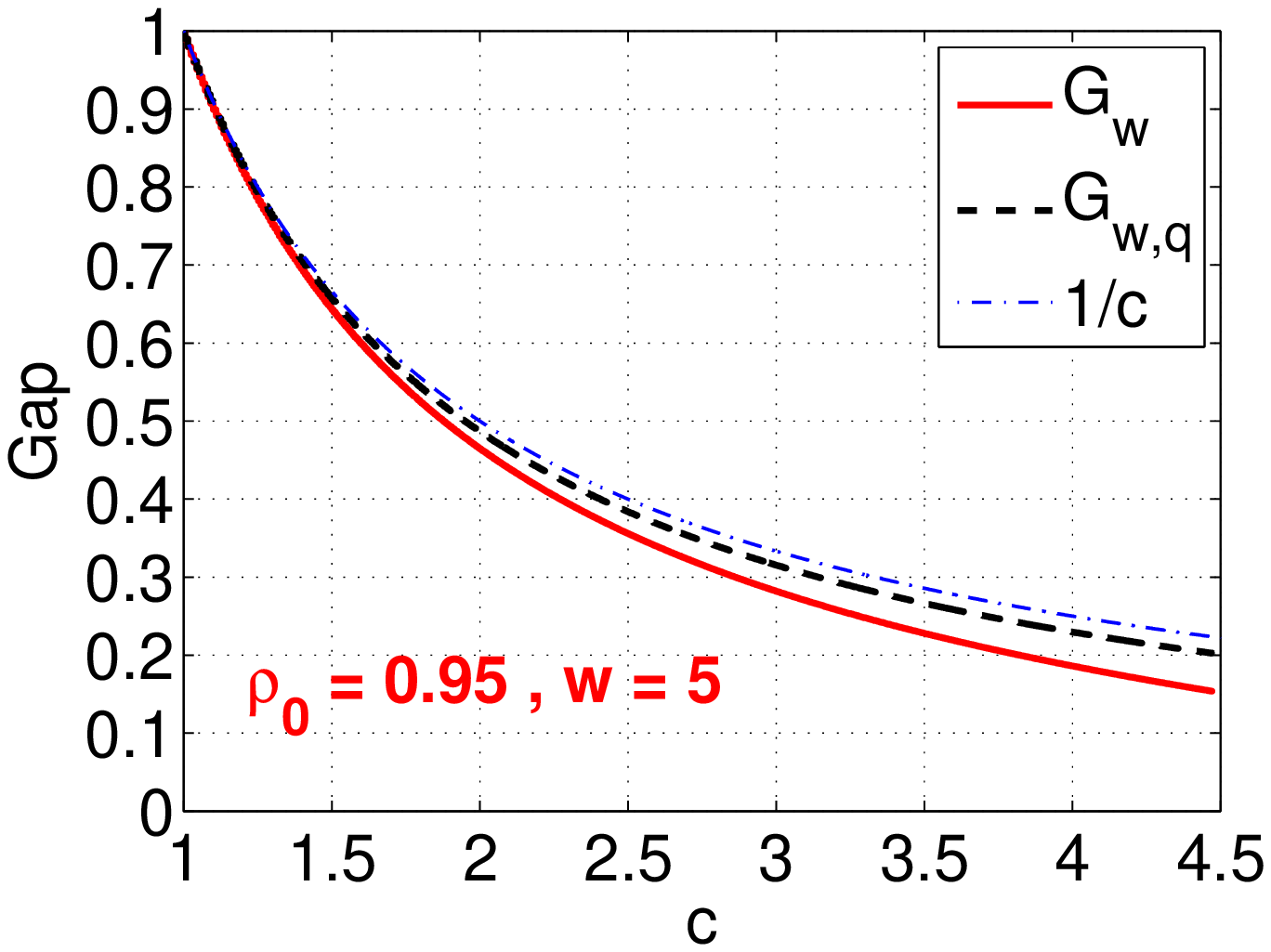}
}
\end{center}
\vspace{-.2in}
\caption{The gaps $G_w$ and $G_{w,q}$ as functions of $c$, for $\rho_0 = 0.95$. In each panel, we plot both $G_w$ and $G_{w,q}$ for a particular $w$ value. }\label{fig_GwqR095W}
\end{figure}

\begin{figure}[h!]
\begin{center}
\mbox{
\includegraphics[width = 2.2in]{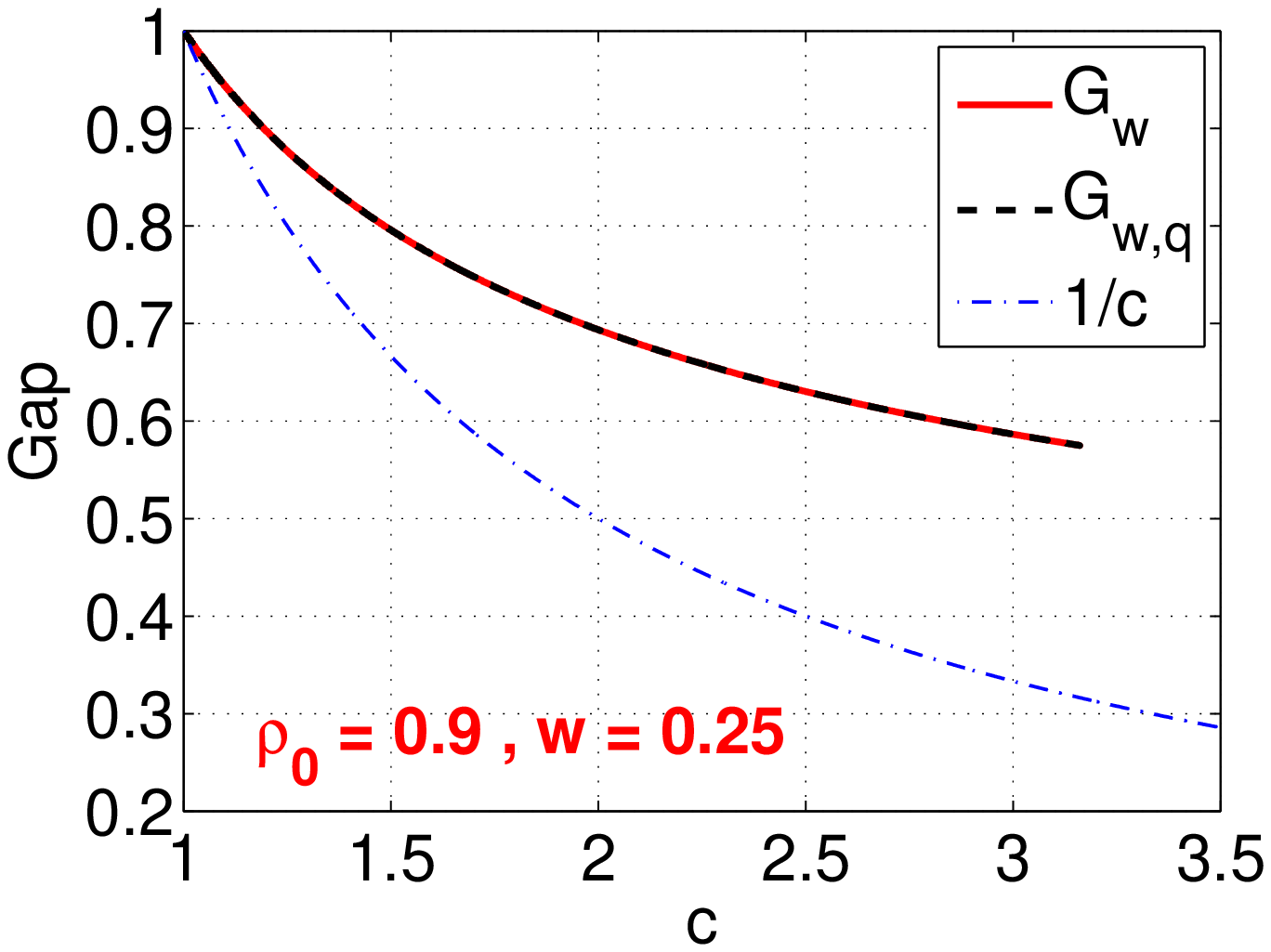}
\includegraphics[width = 2.2in]{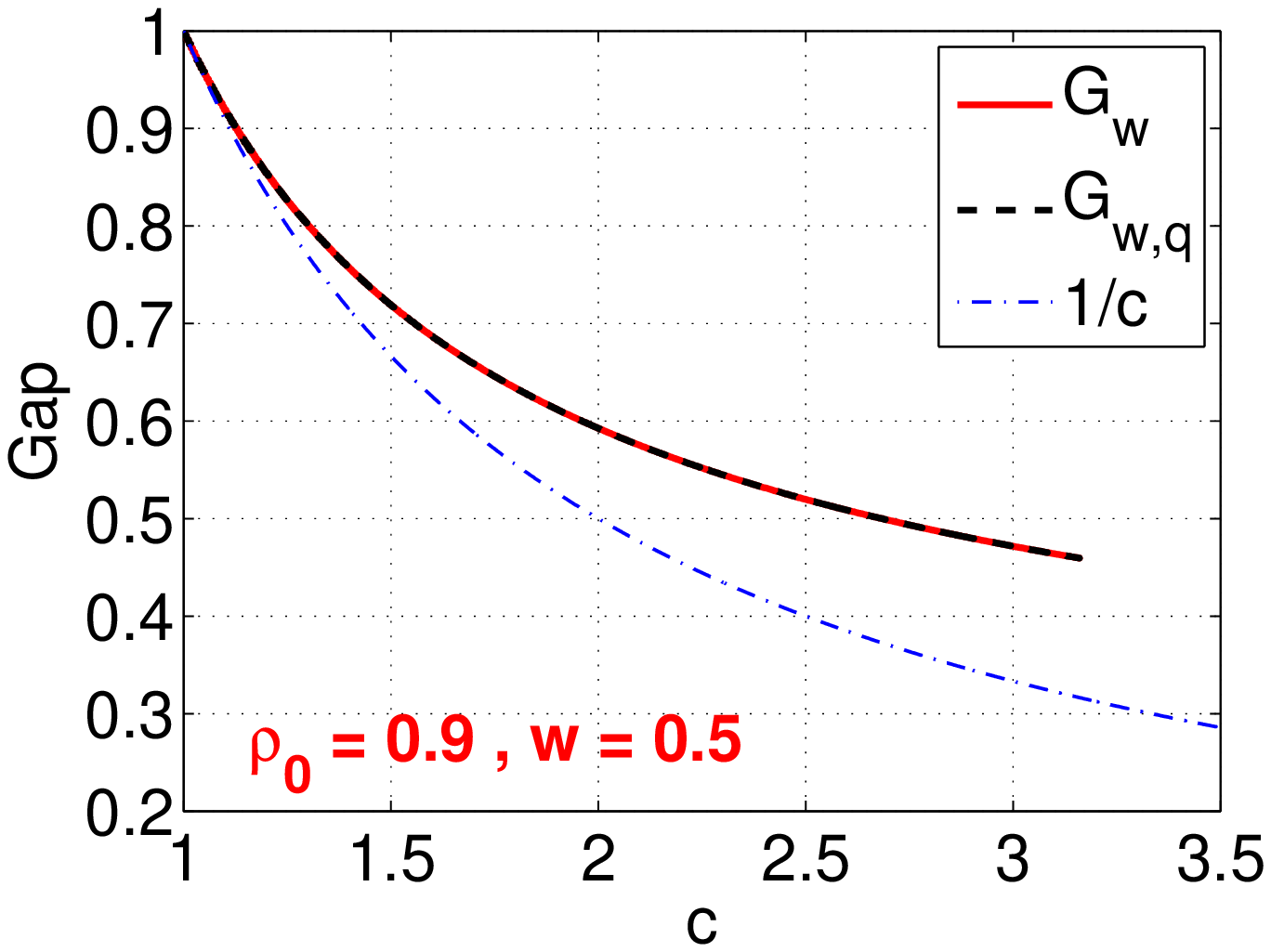}
\includegraphics[width = 2.2in]{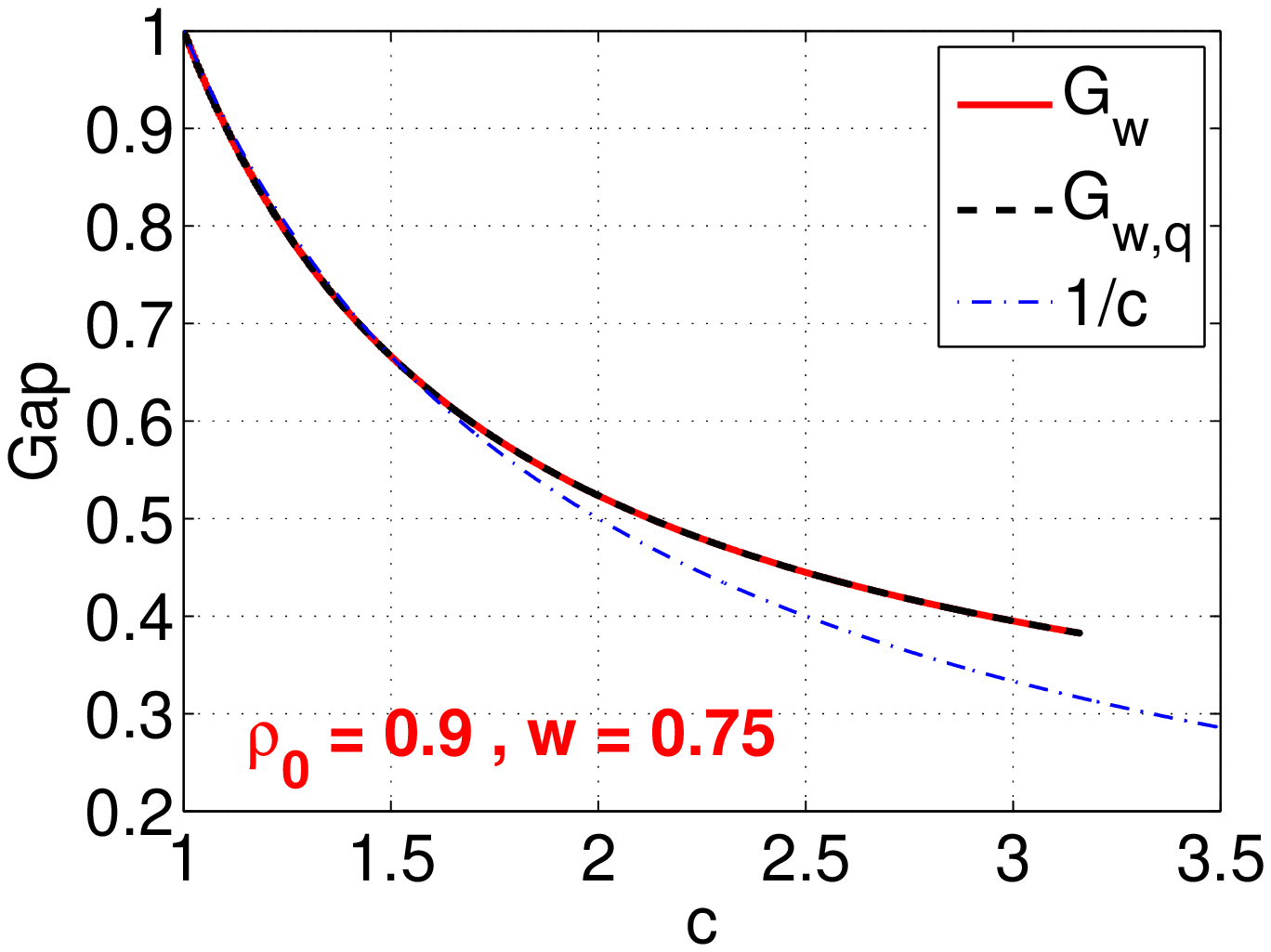}
}
\mbox{
\includegraphics[width = 2.2in]{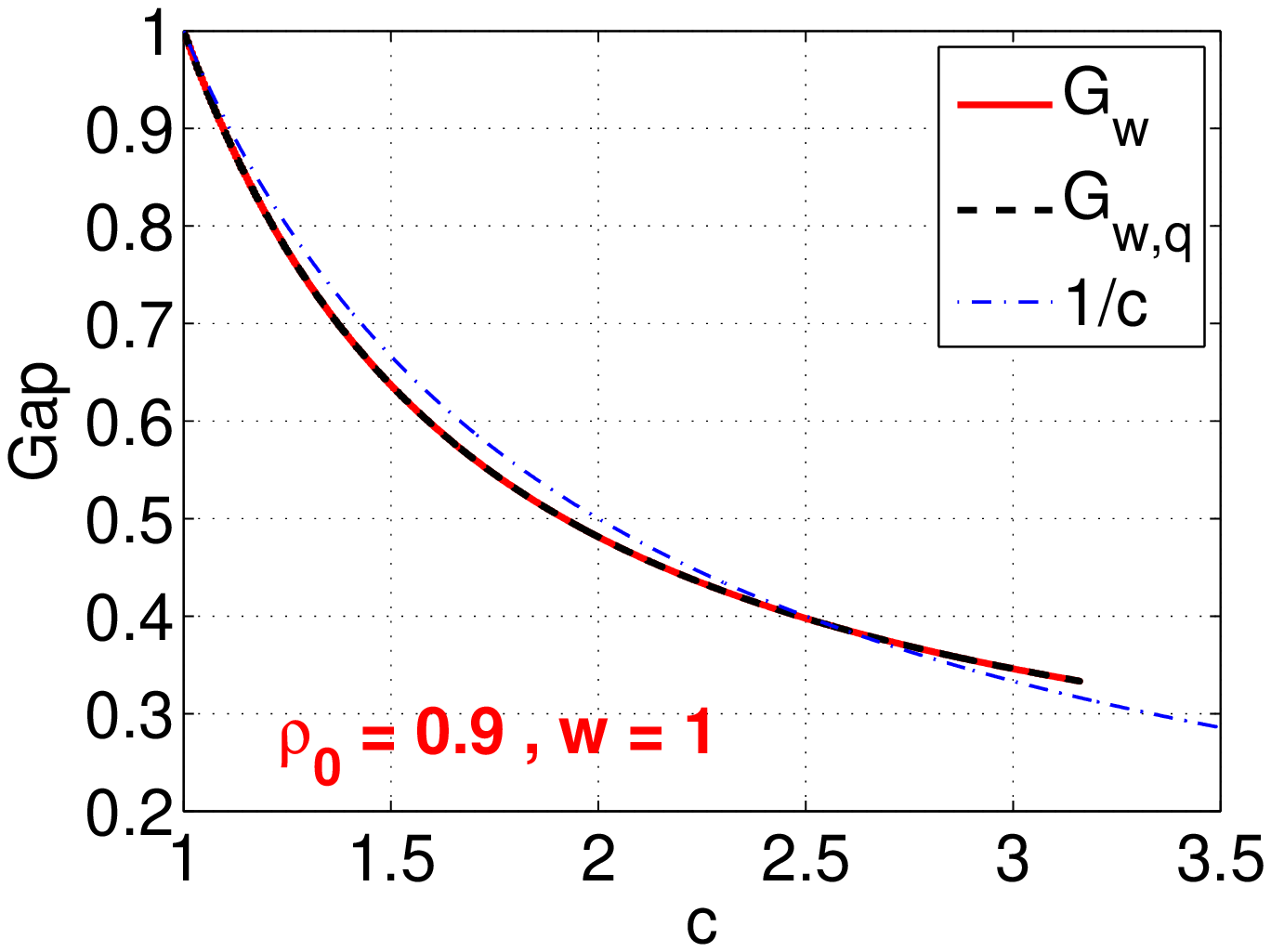}
\includegraphics[width = 2.2in]{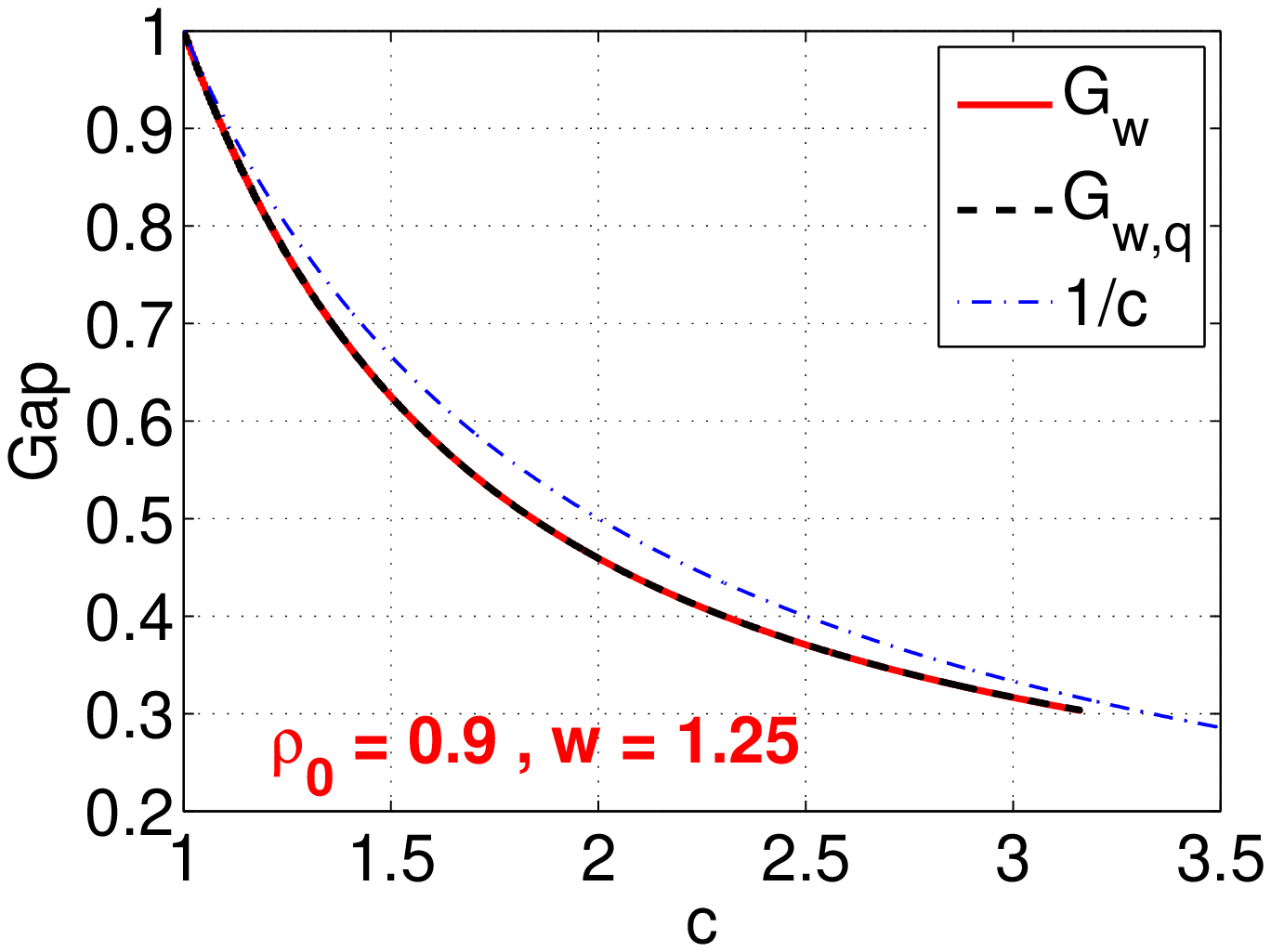}
\includegraphics[width = 2.2in]{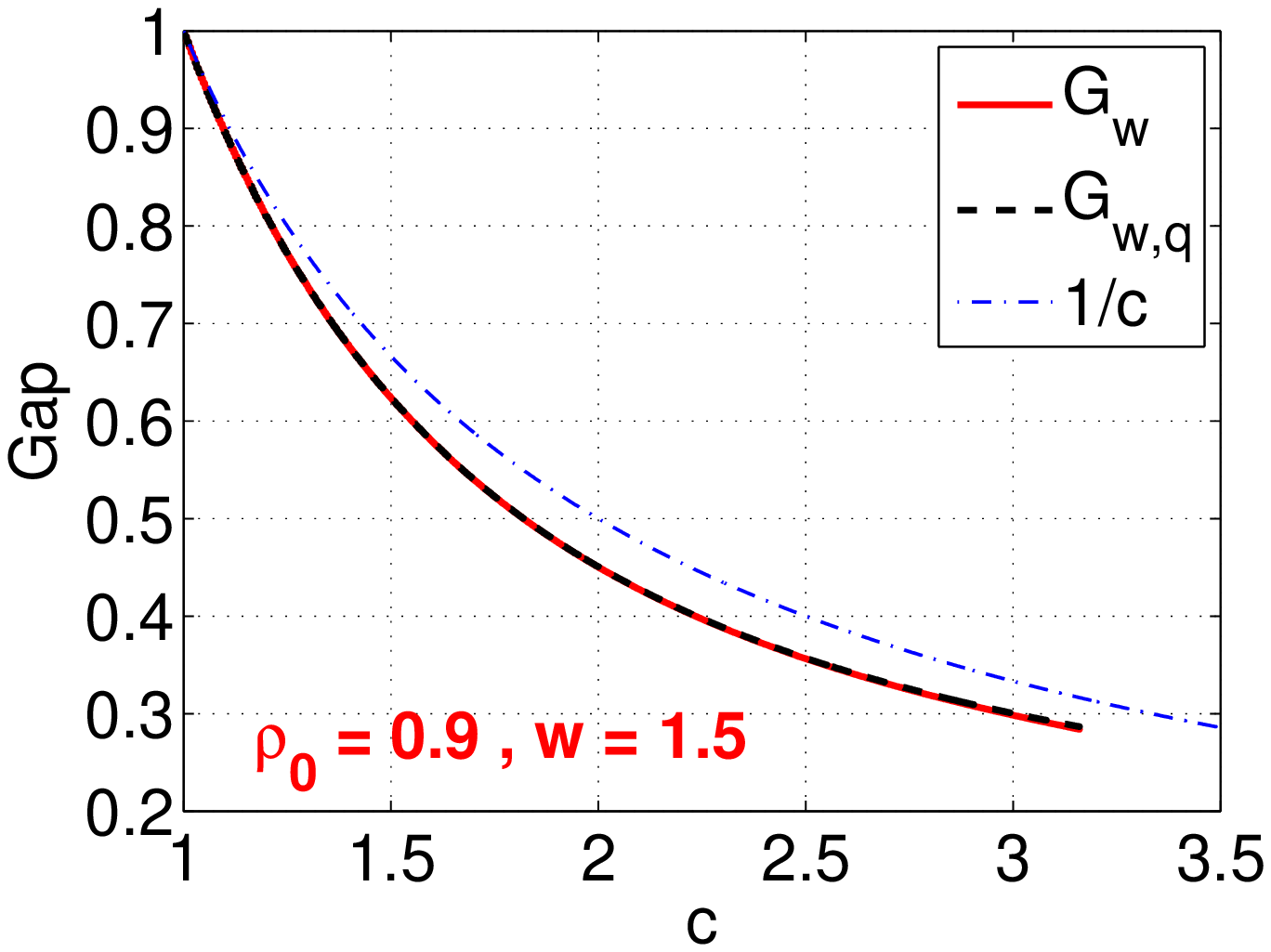}
}

\mbox{
\includegraphics[width = 2.2in]{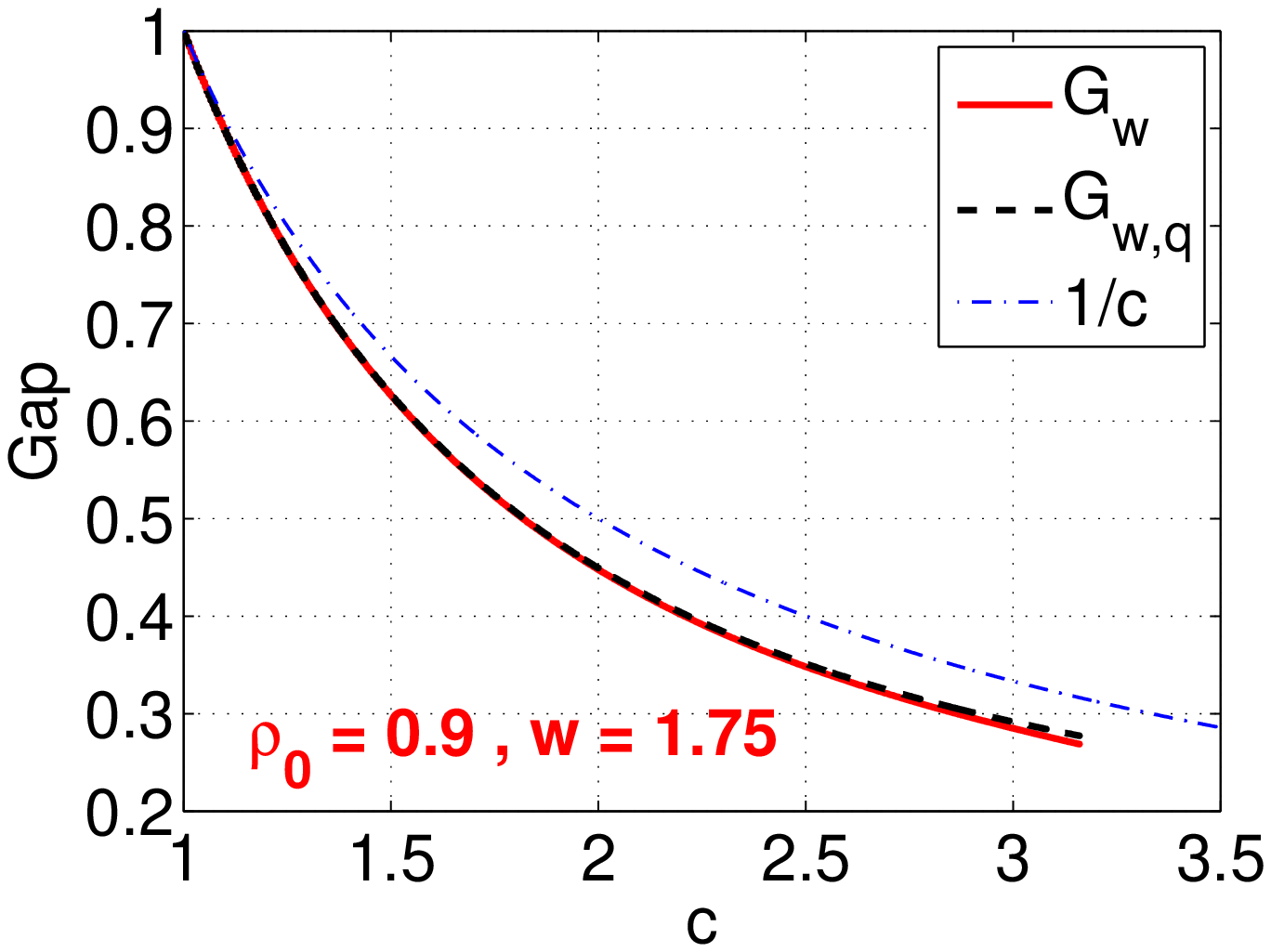}
\includegraphics[width = 2.2in]{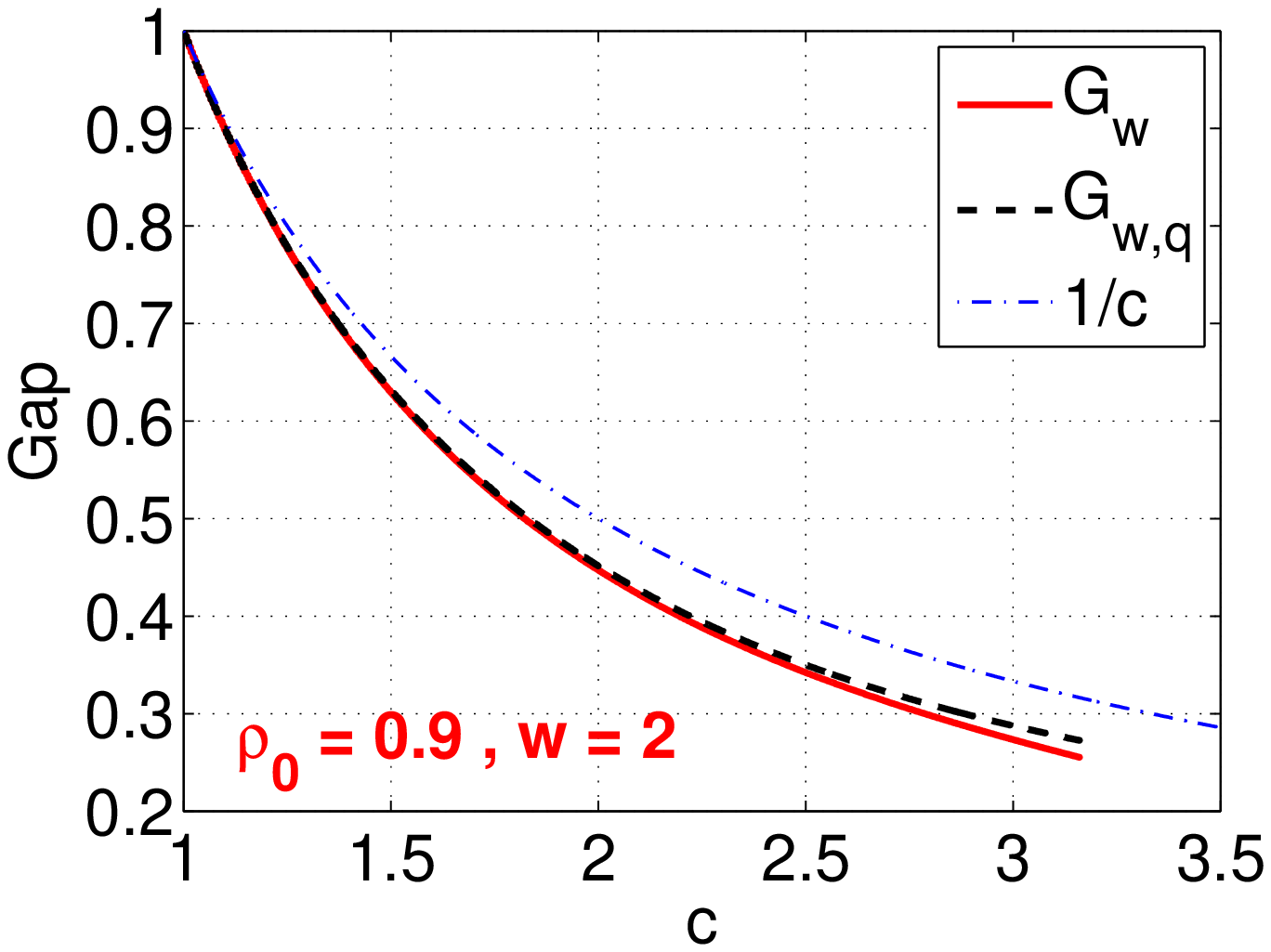}
\includegraphics[width = 2.2in]{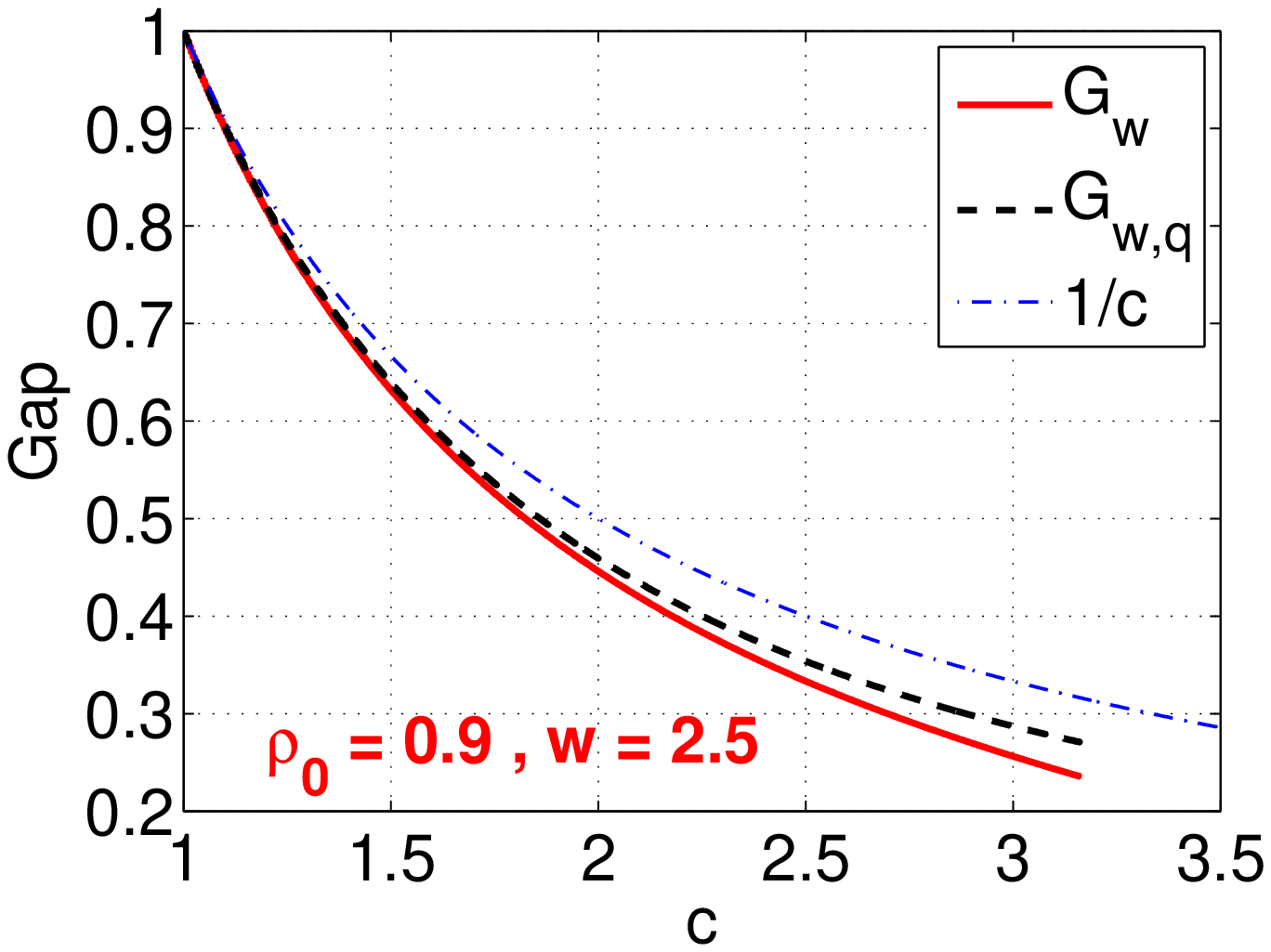}
}

\mbox{
\includegraphics[width = 2.2in]{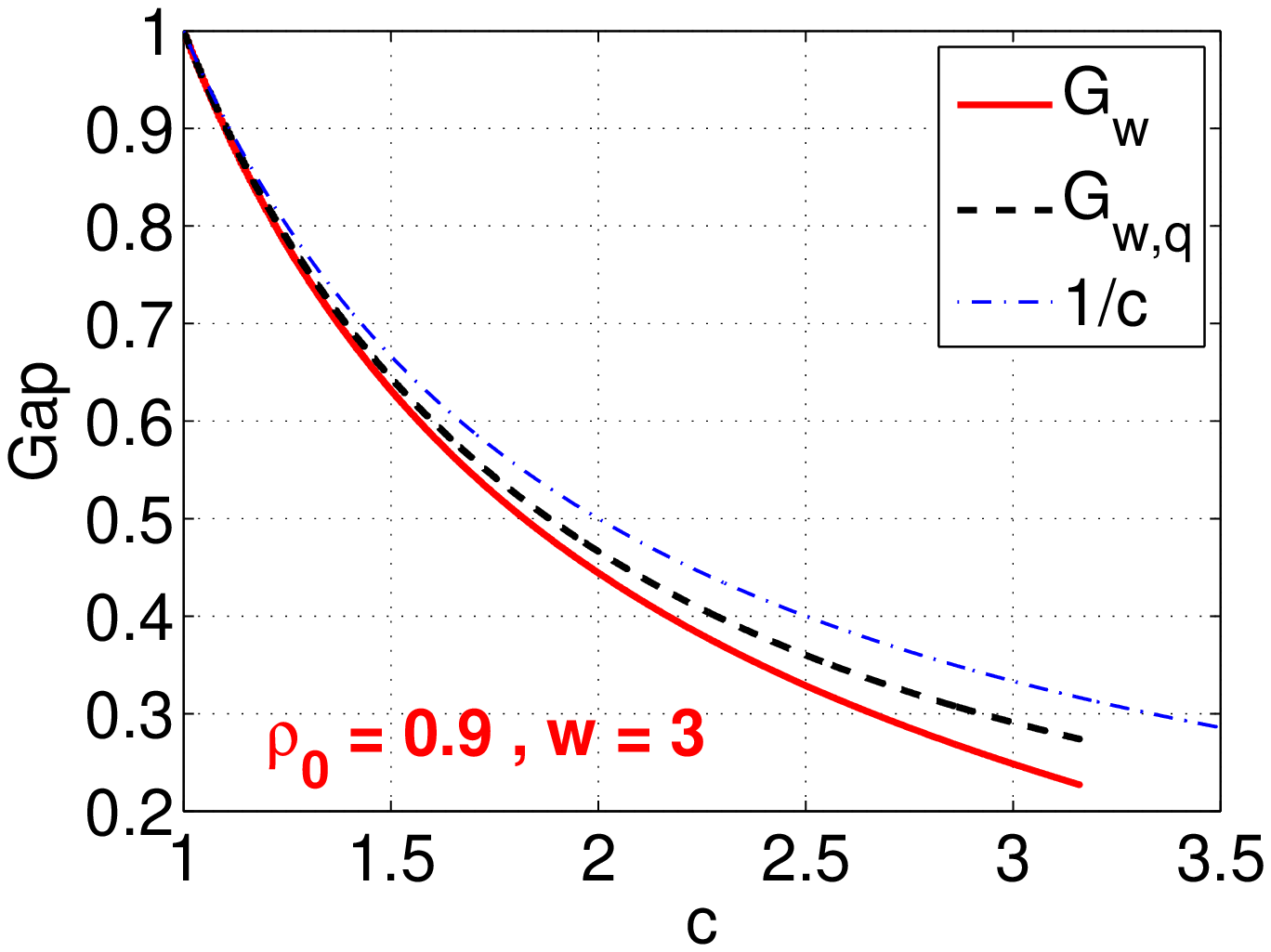}
\includegraphics[width = 2.2in]{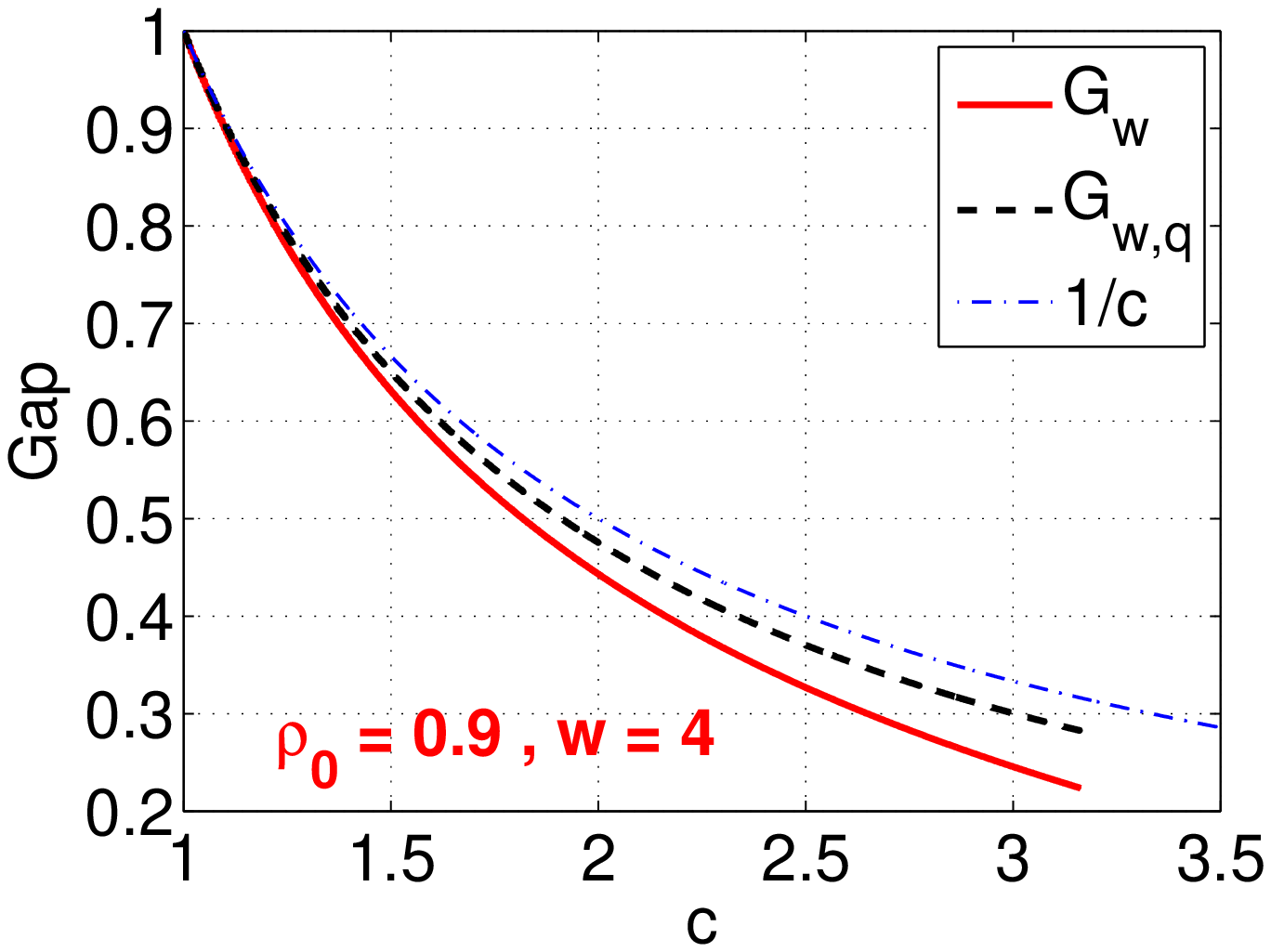}
\includegraphics[width = 2.2in]{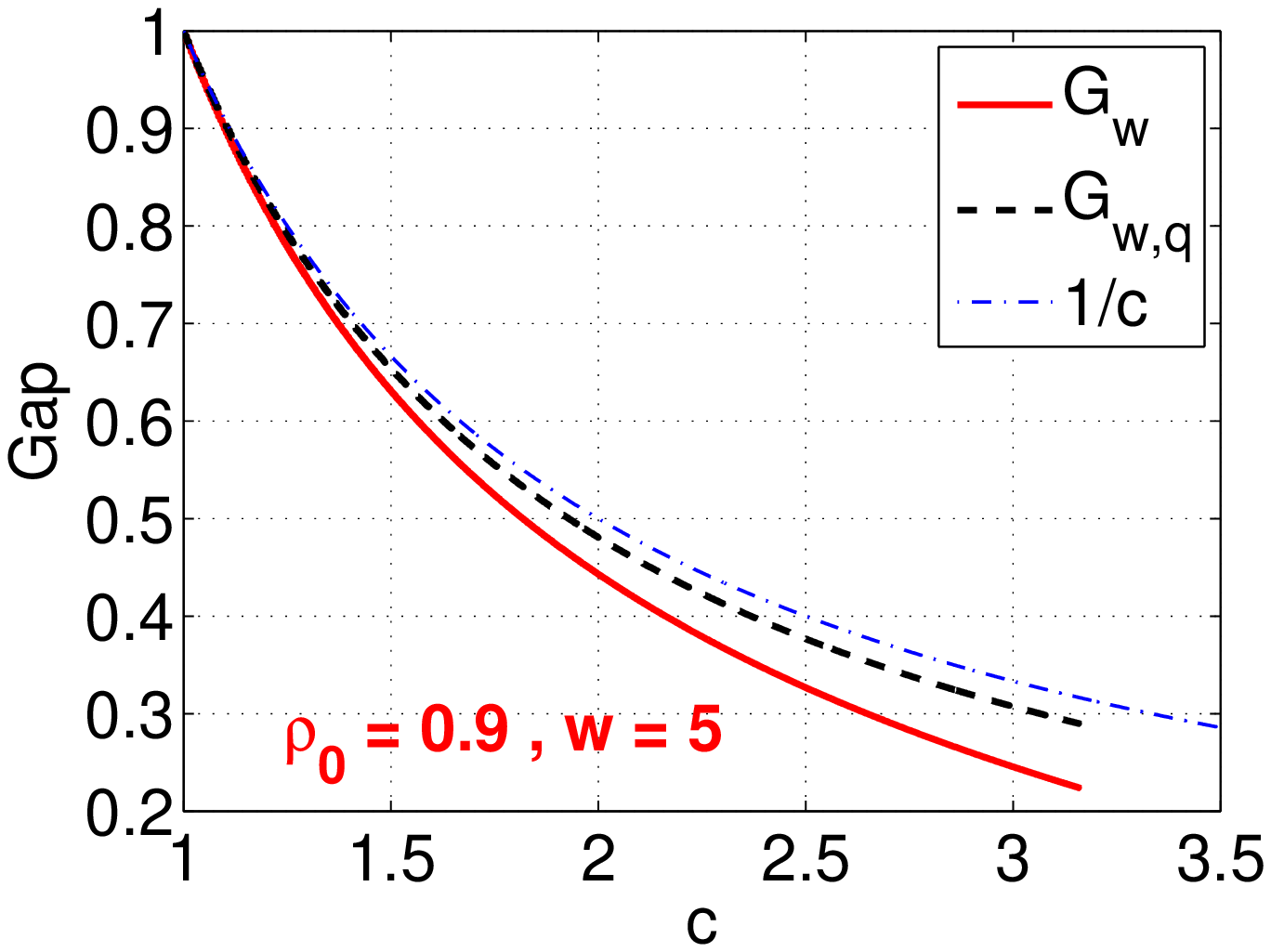}
}

\end{center}
\vspace{-.2in}
\caption{The gaps $G_w$ and $G_{w,q}$ as functions of $c$, for $\rho_0 = 0.9$. In each panel, we plot both $G_w$ and $G_{w,q}$ for a particular $w$ value. }\label{fig_GwqR09W}
\end{figure}

\begin{figure}[h!]
\begin{center}
\mbox{
\includegraphics[width = 2.2in]{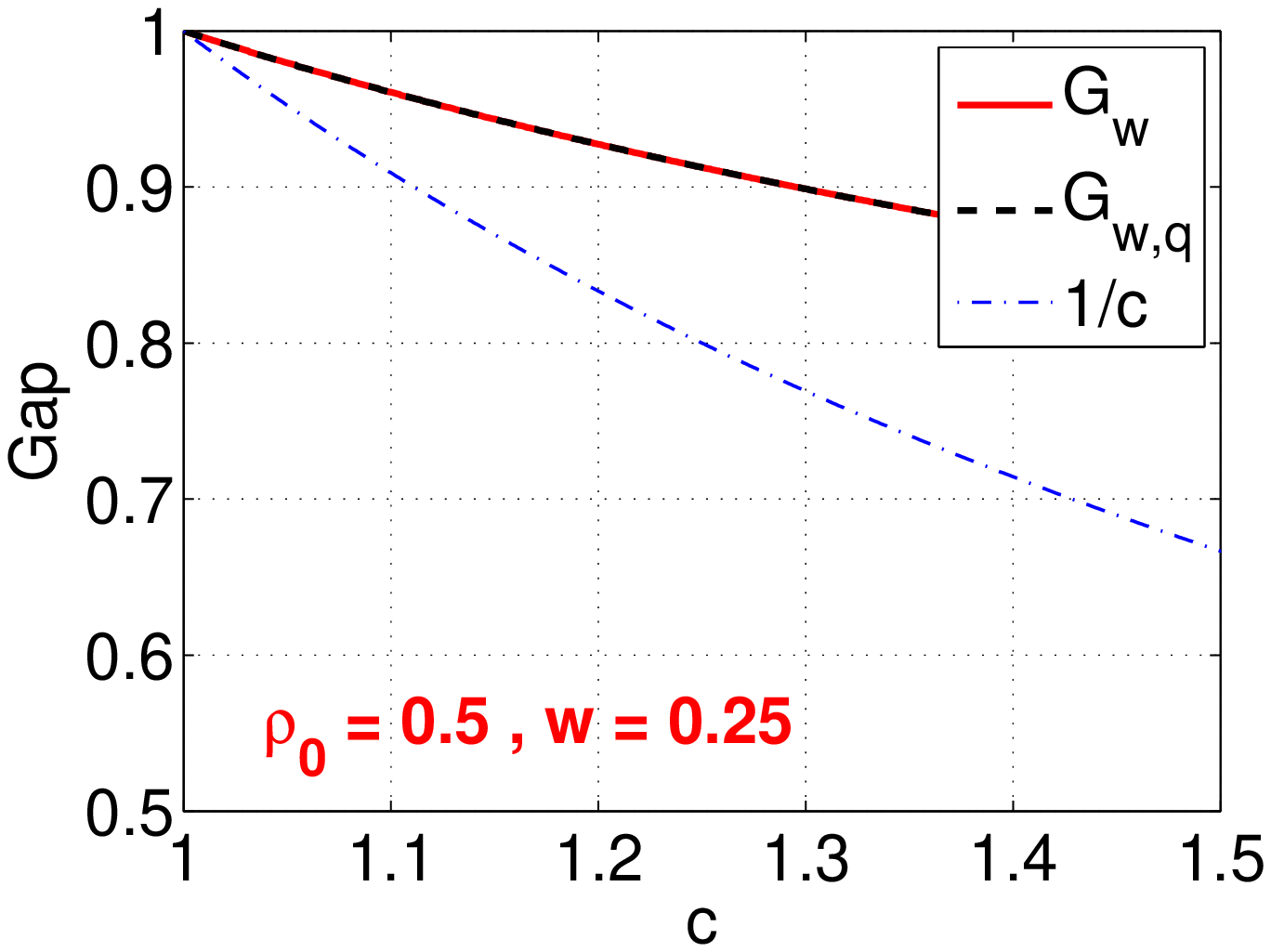}
\includegraphics[width = 2.2in]{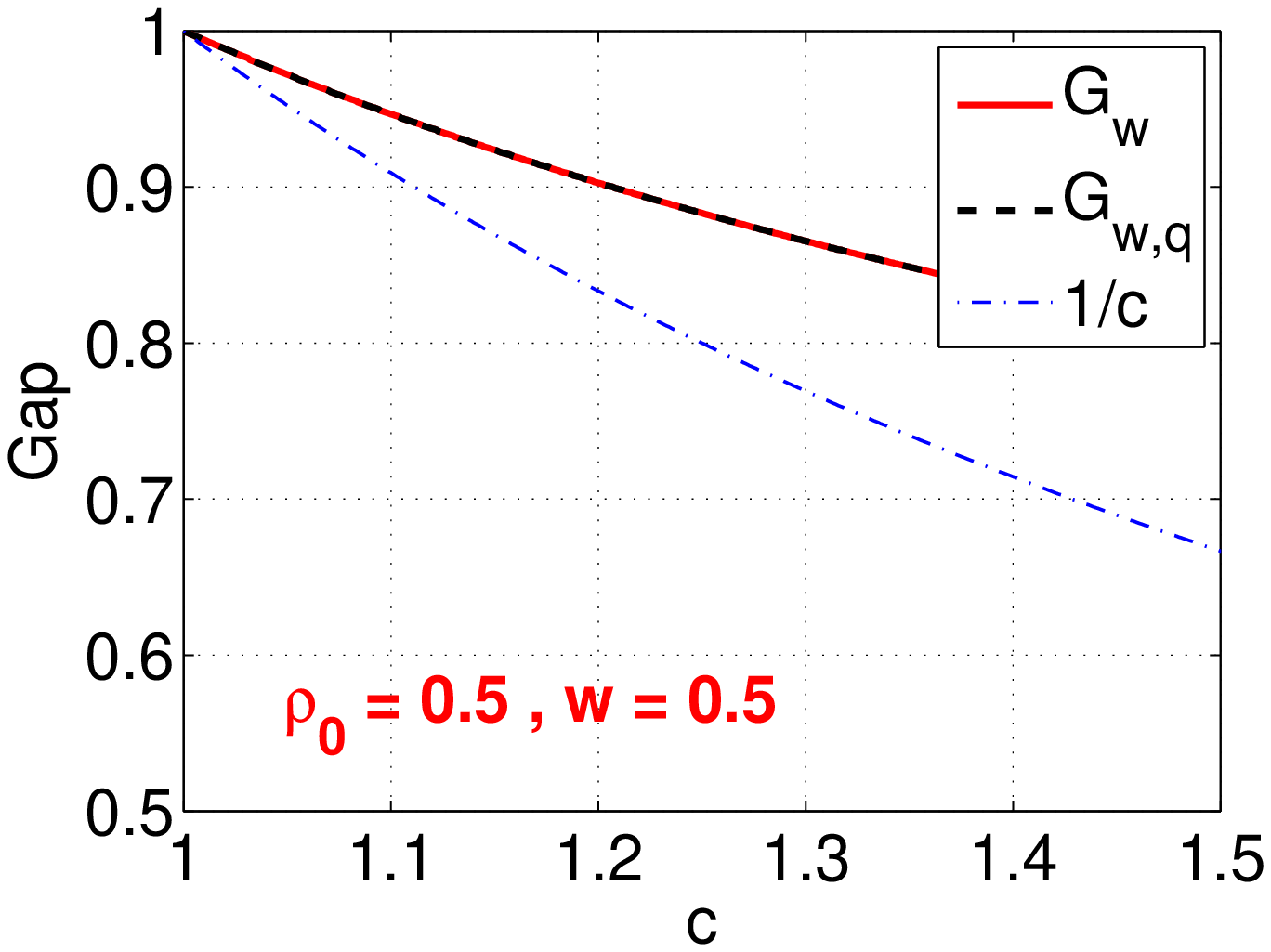}
\includegraphics[width = 2.2in]{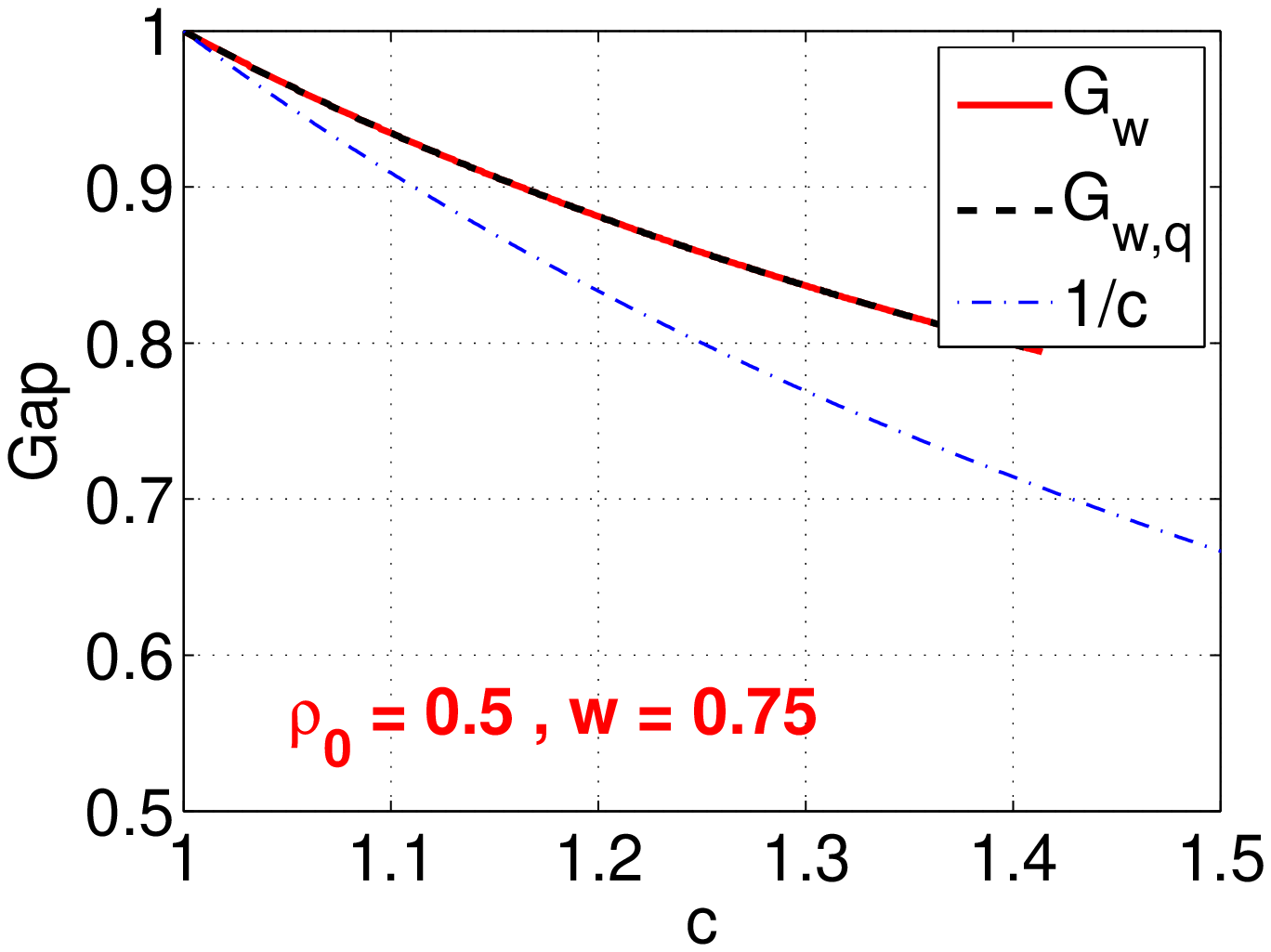}
}
\mbox{
\includegraphics[width = 2.2in]{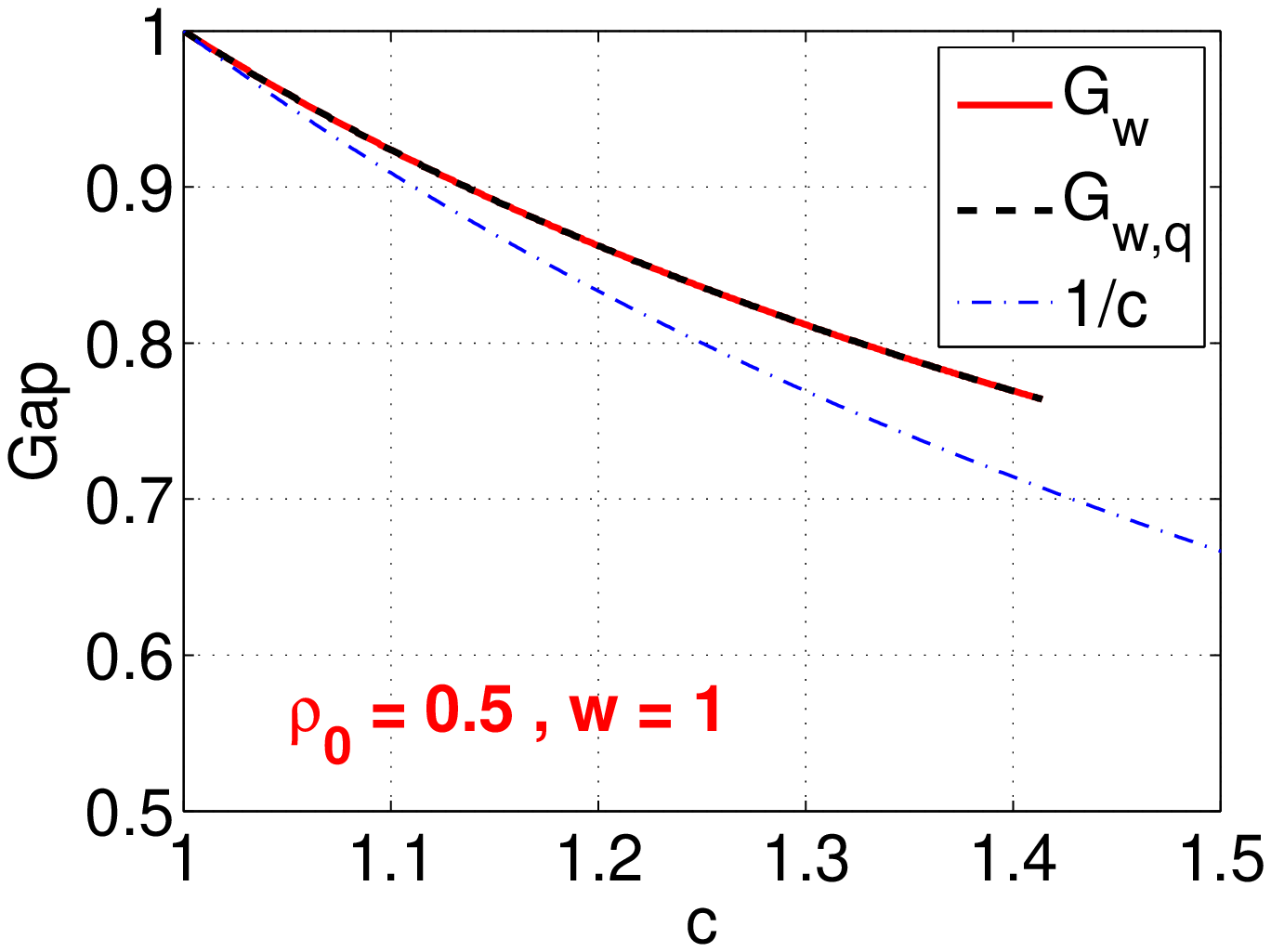}
\includegraphics[width = 2.2in]{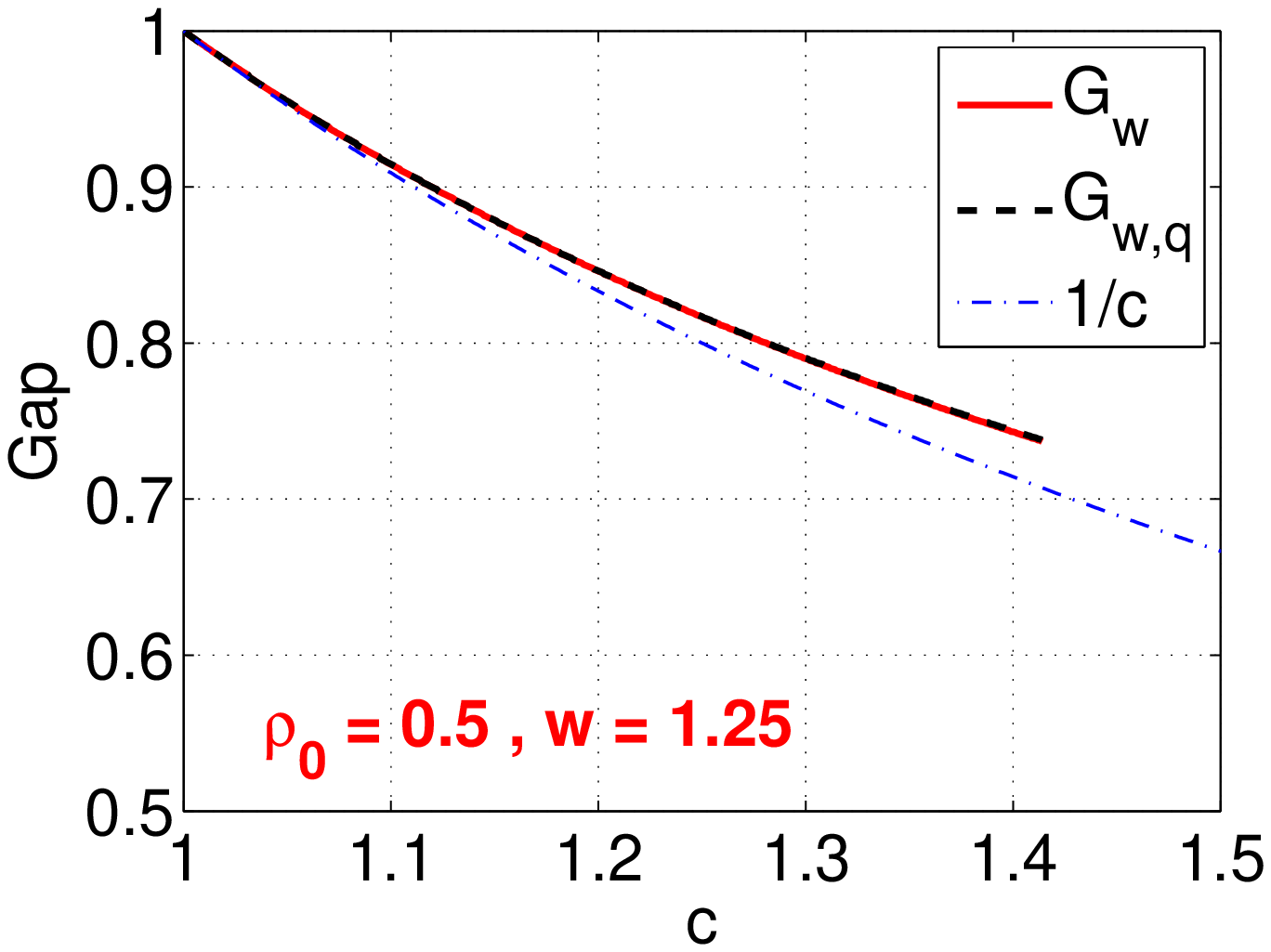}
\includegraphics[width = 2.2in]{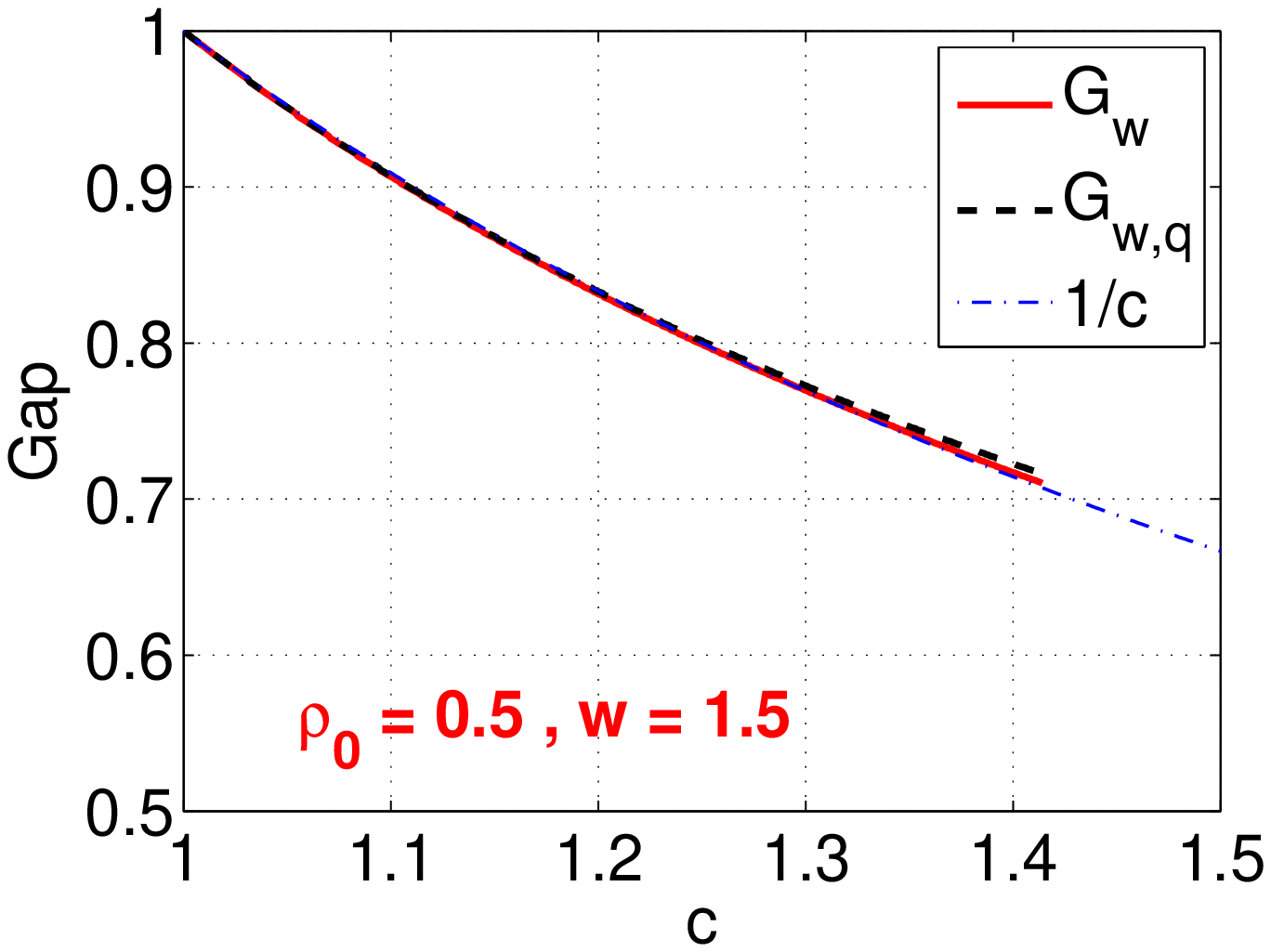}
}

\mbox{
\includegraphics[width = 2.2in]{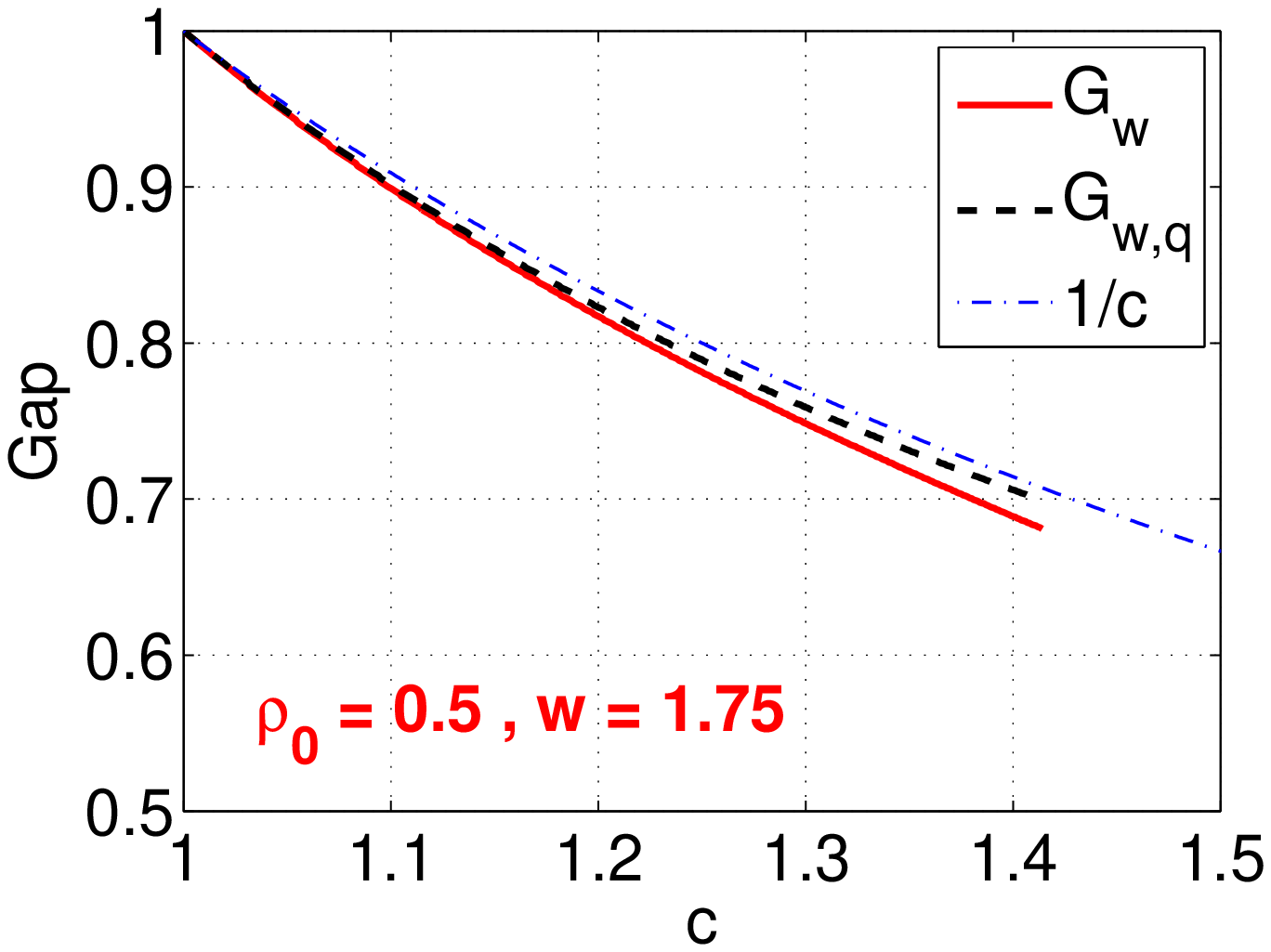}
\includegraphics[width = 2.2in]{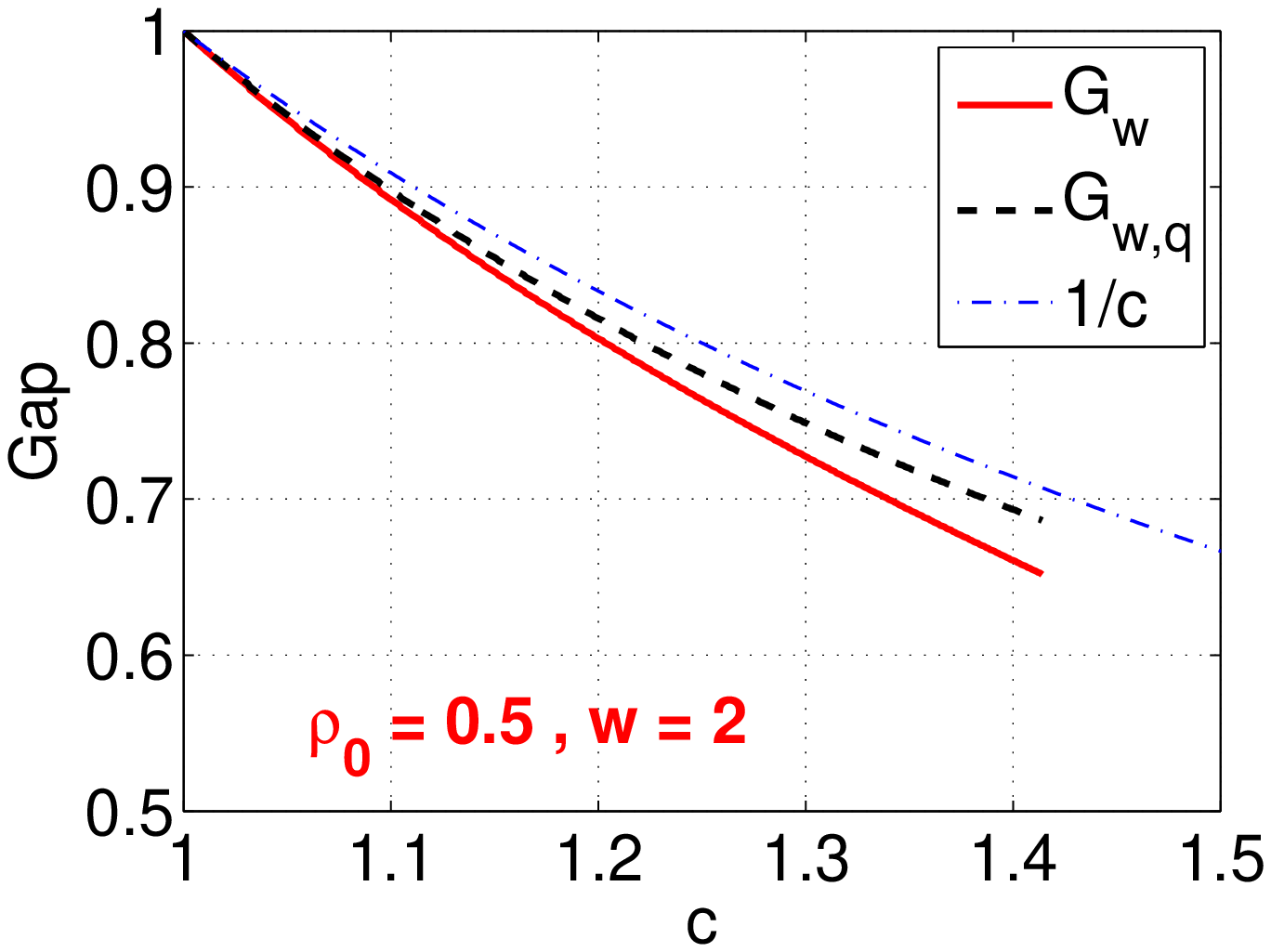}
\includegraphics[width = 2.2in]{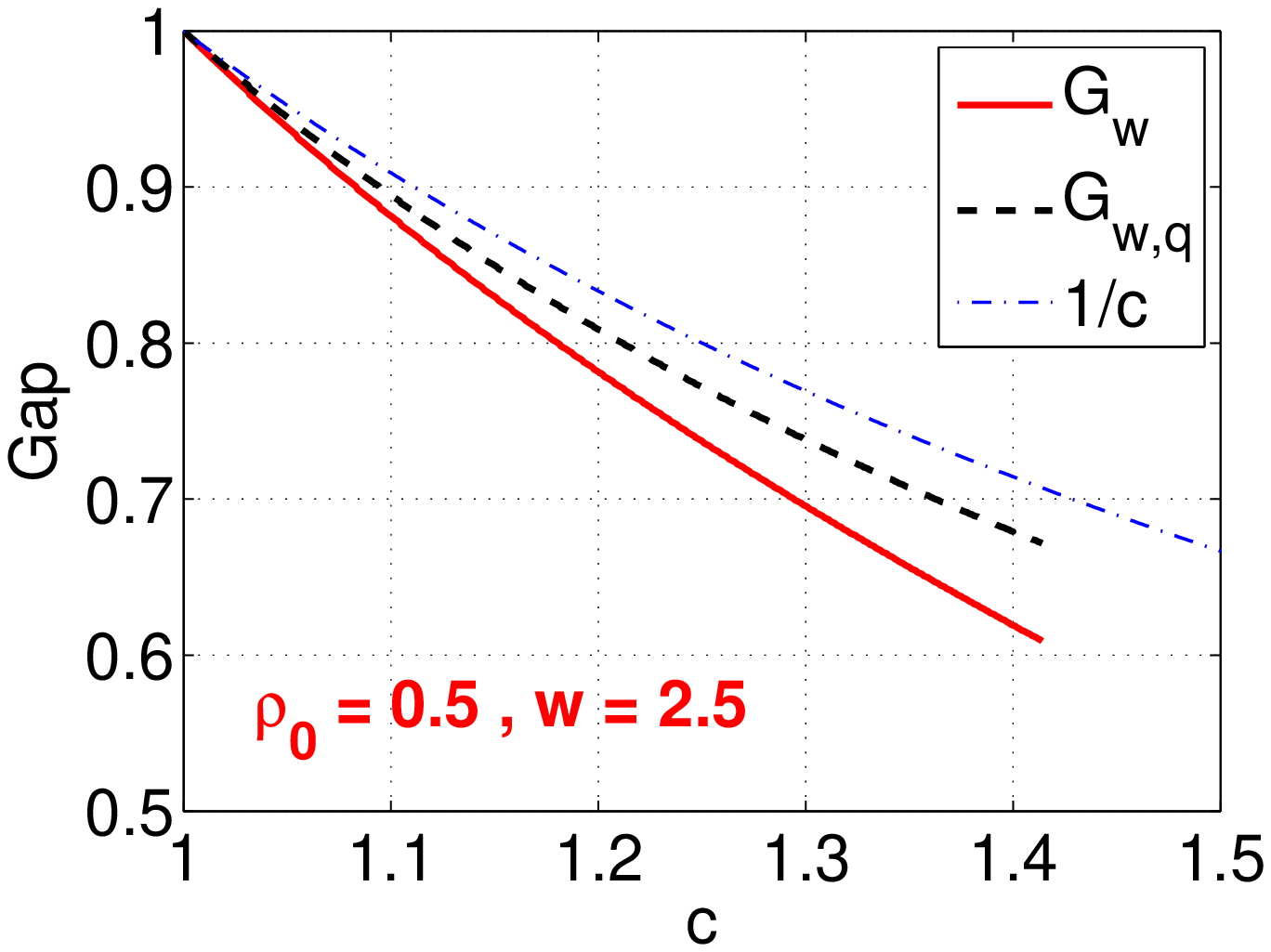}
}

\mbox{
\includegraphics[width = 2.2in]{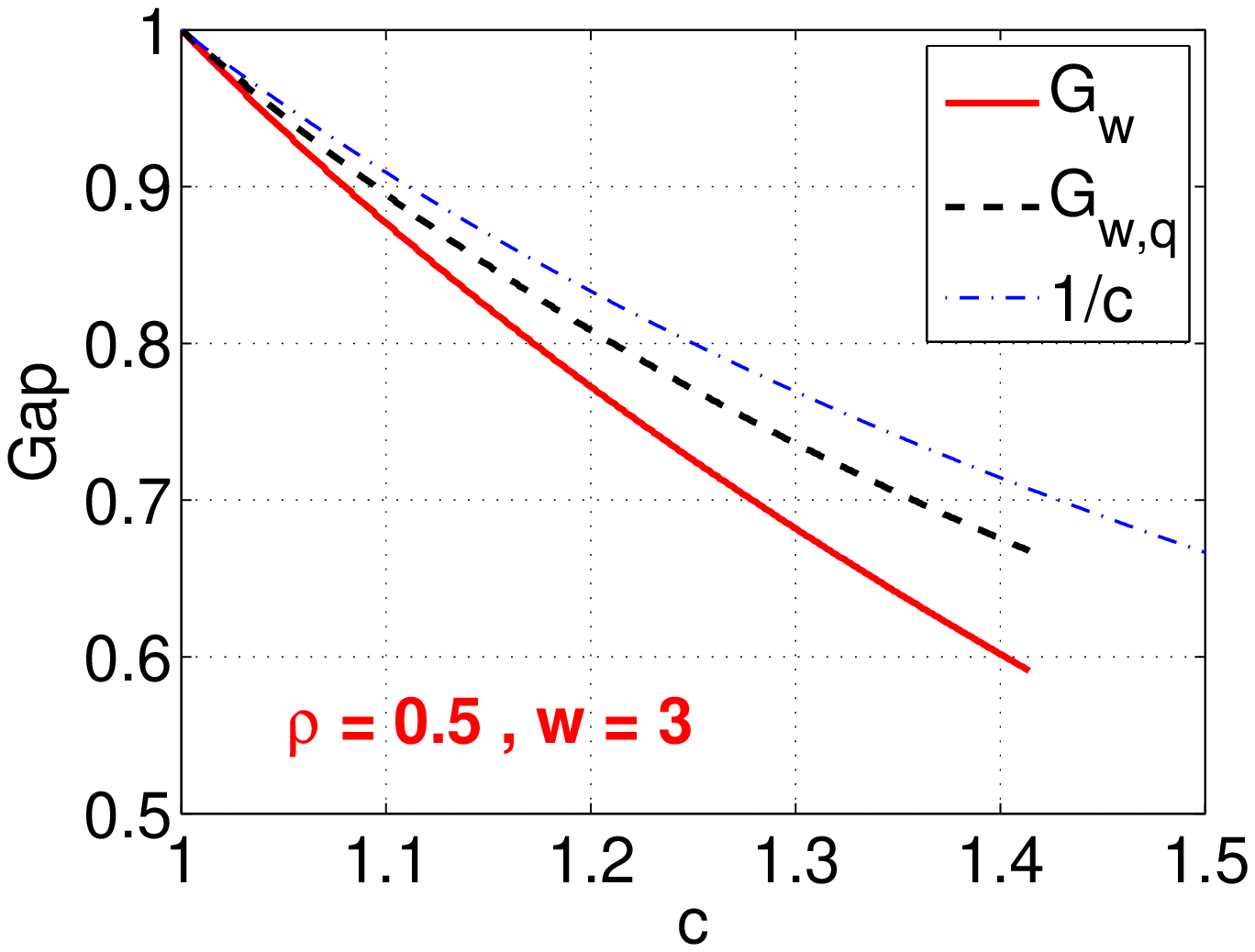}
\includegraphics[width = 2.2in]{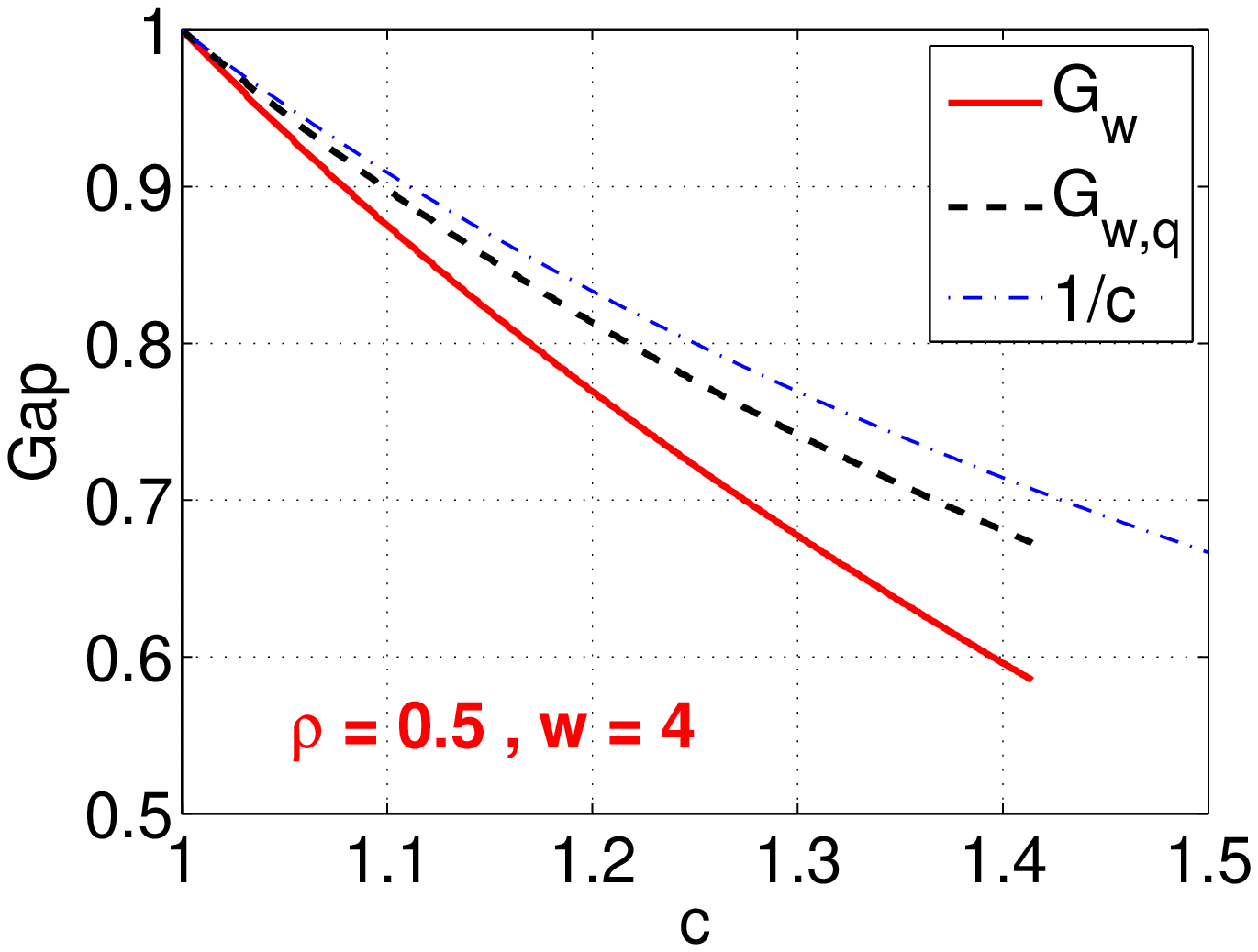}
\includegraphics[width = 2.2in]{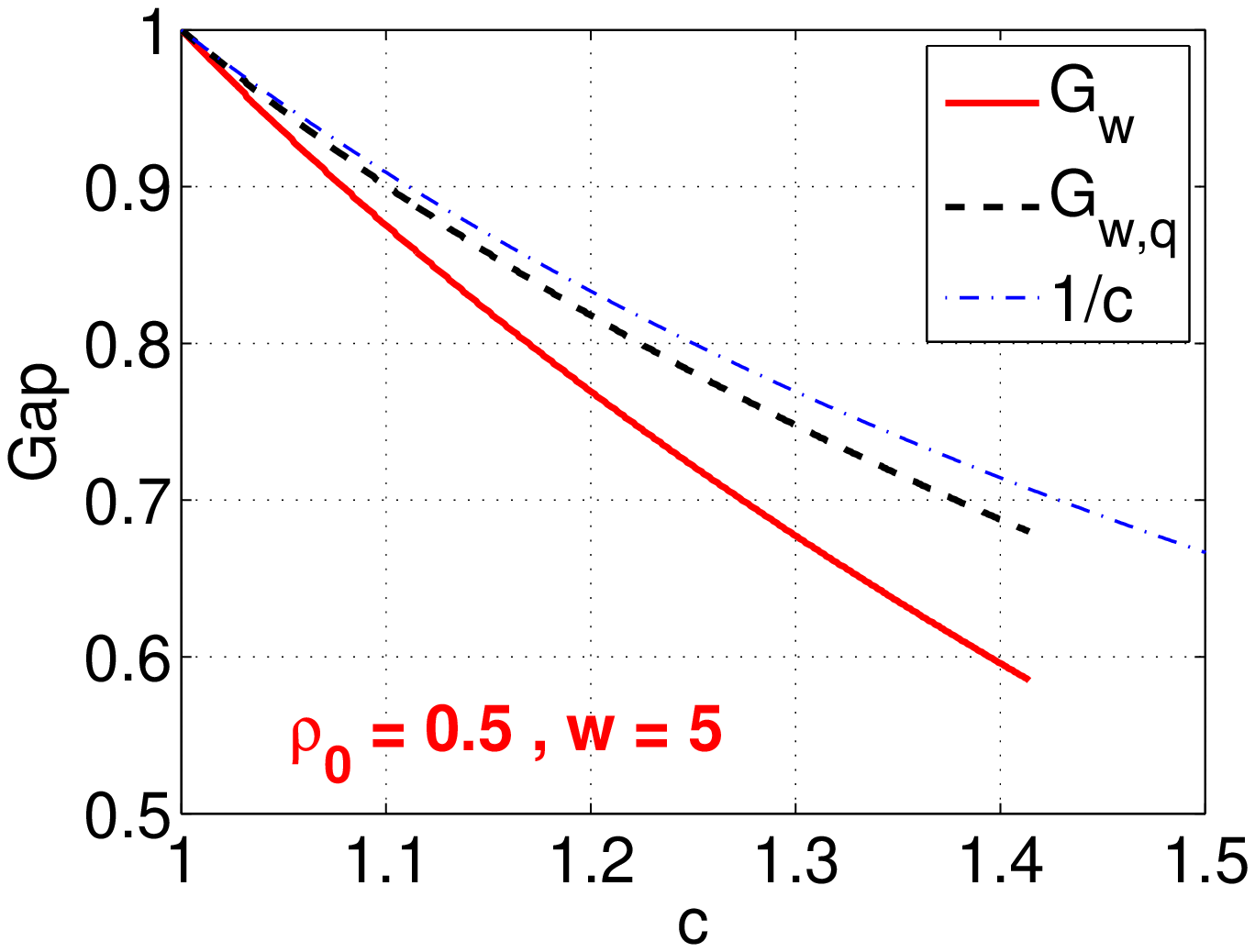}
}

\end{center}
\vspace{-.2in}
\caption{The gaps $G_w$ and $G_{w,q}$ as functions of $c$, for $\rho_0 = 0.5$. In each panel, we plot both $G_w$ and $G_{w,q}$ for a particular $w$ value. }\label{fig_GwqR05W}
\end{figure}

In practice, we normally have to  pre-specify a $w$, for all $c$ and $\rho_0$ values. In other words, the ``optimum'' $G$ values presented in Figure~\ref{fig_GwqOpt} are in general not attainable. Therefore,  Figure~\ref{fig_GwqR099W}, Figure~\ref{fig_GwqR095W}, Figure~\ref{fig_GwqR09W}, and Figure~\ref{fig_GwqR05W} present $G_w$ and $G_{w,q}$ as functions of $c$, for $\rho_0 = 0.99$, $\rho_0 = 0.95$, $\rho_0 = 0.9$ and $\rho_0 = 0.5$, respectively. In each figure, we plot the curves for a wide range of $w$ values. These figures again confirm that $G_w$ is smaller than $G_{w,q}$.

\newpage\clearpage

\section{Optimal Gaps}

\begin{figure}[h!]
\begin{center}
\mbox{
\includegraphics[width = 2.7in]{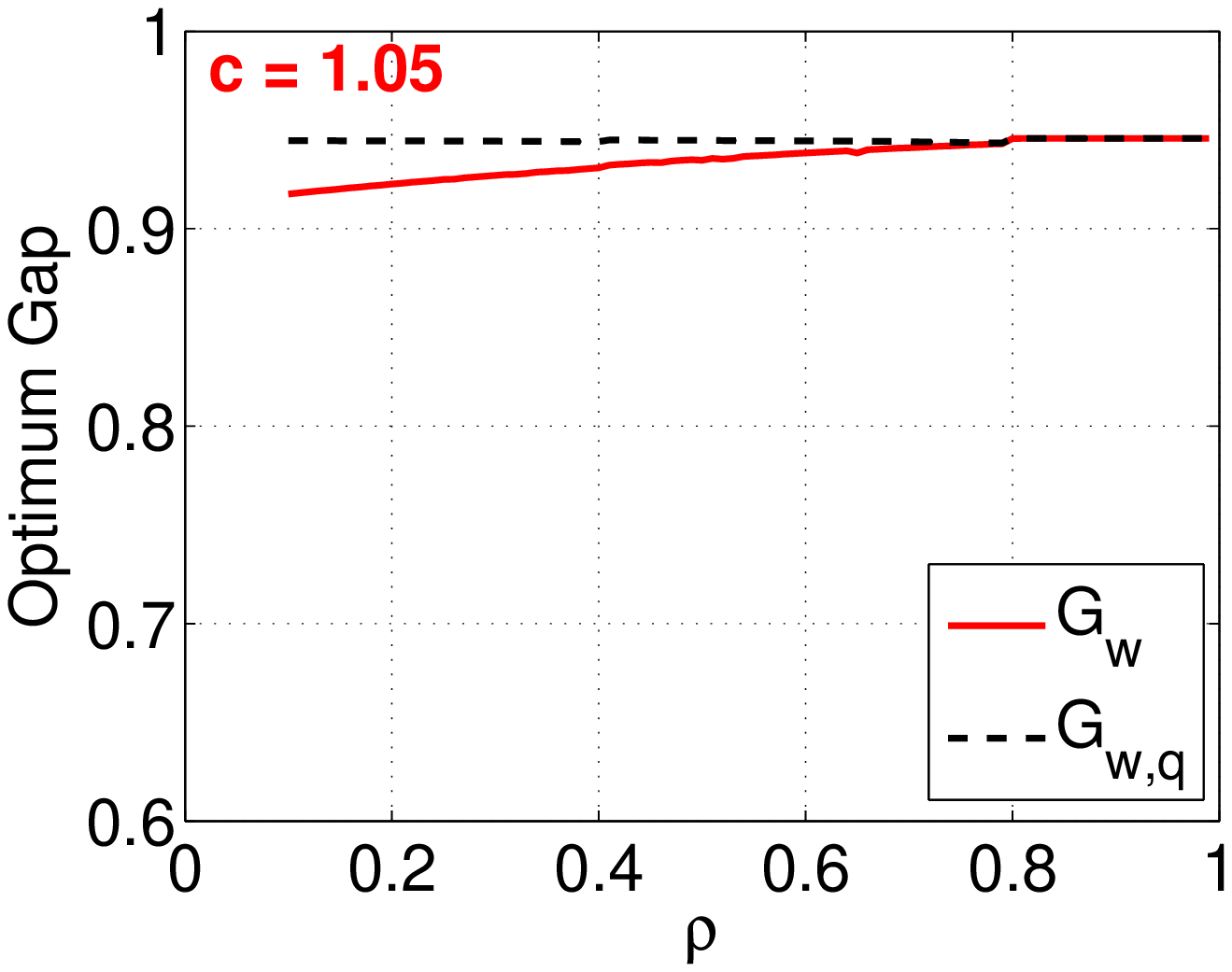}
\includegraphics[width = 2.7in]{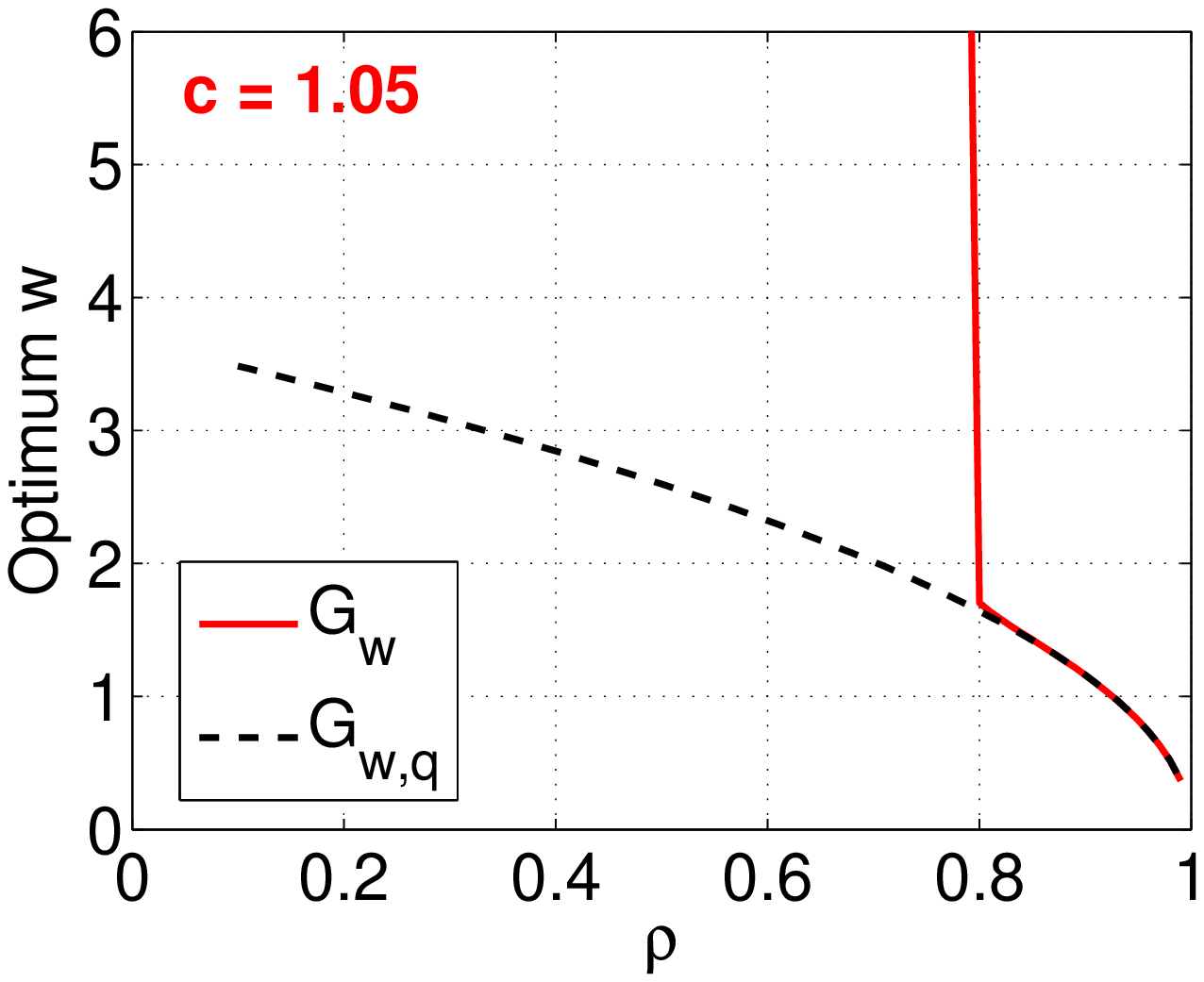}
}

\mbox{
\includegraphics[width = 2.7in]{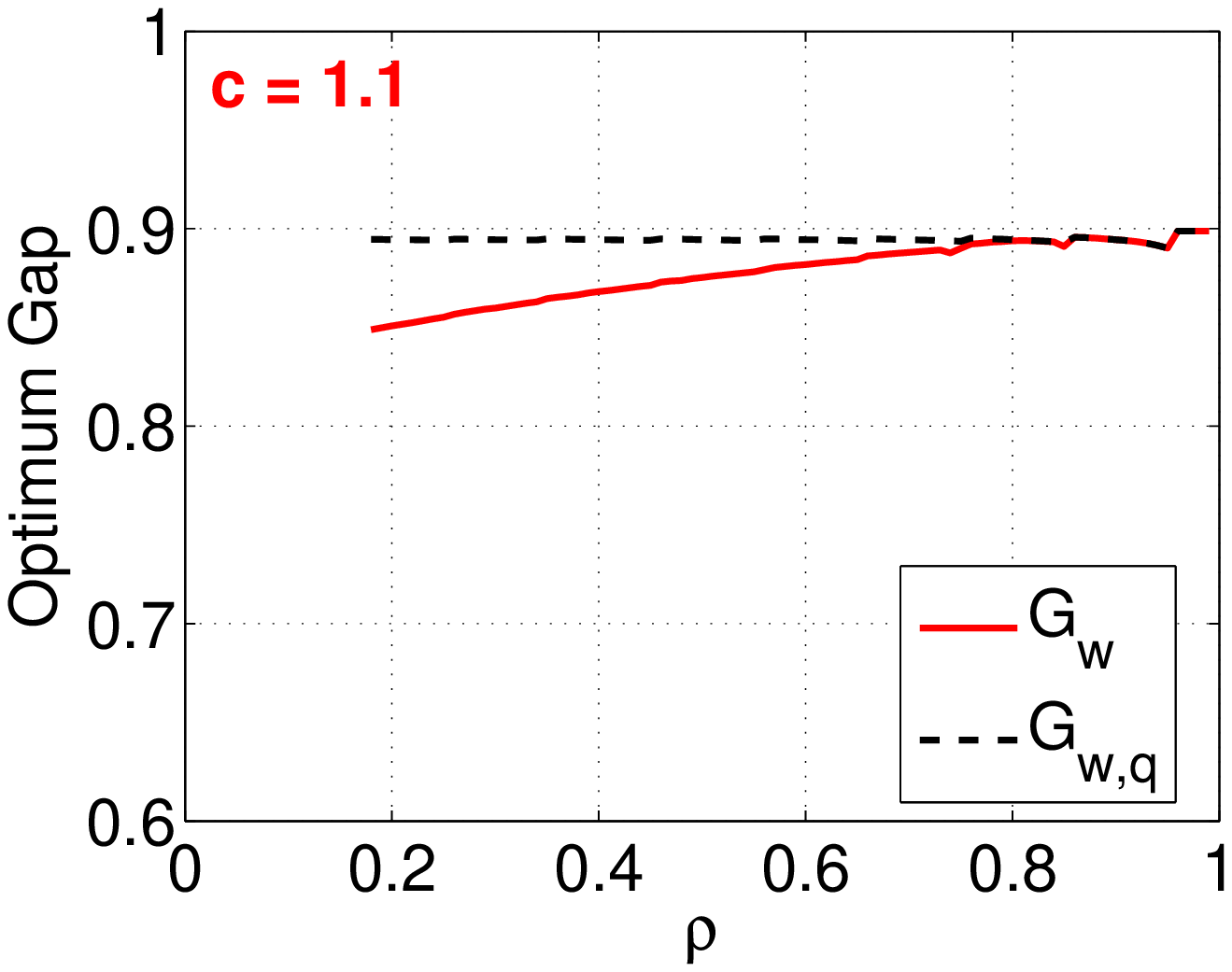}
\includegraphics[width = 2.7in]{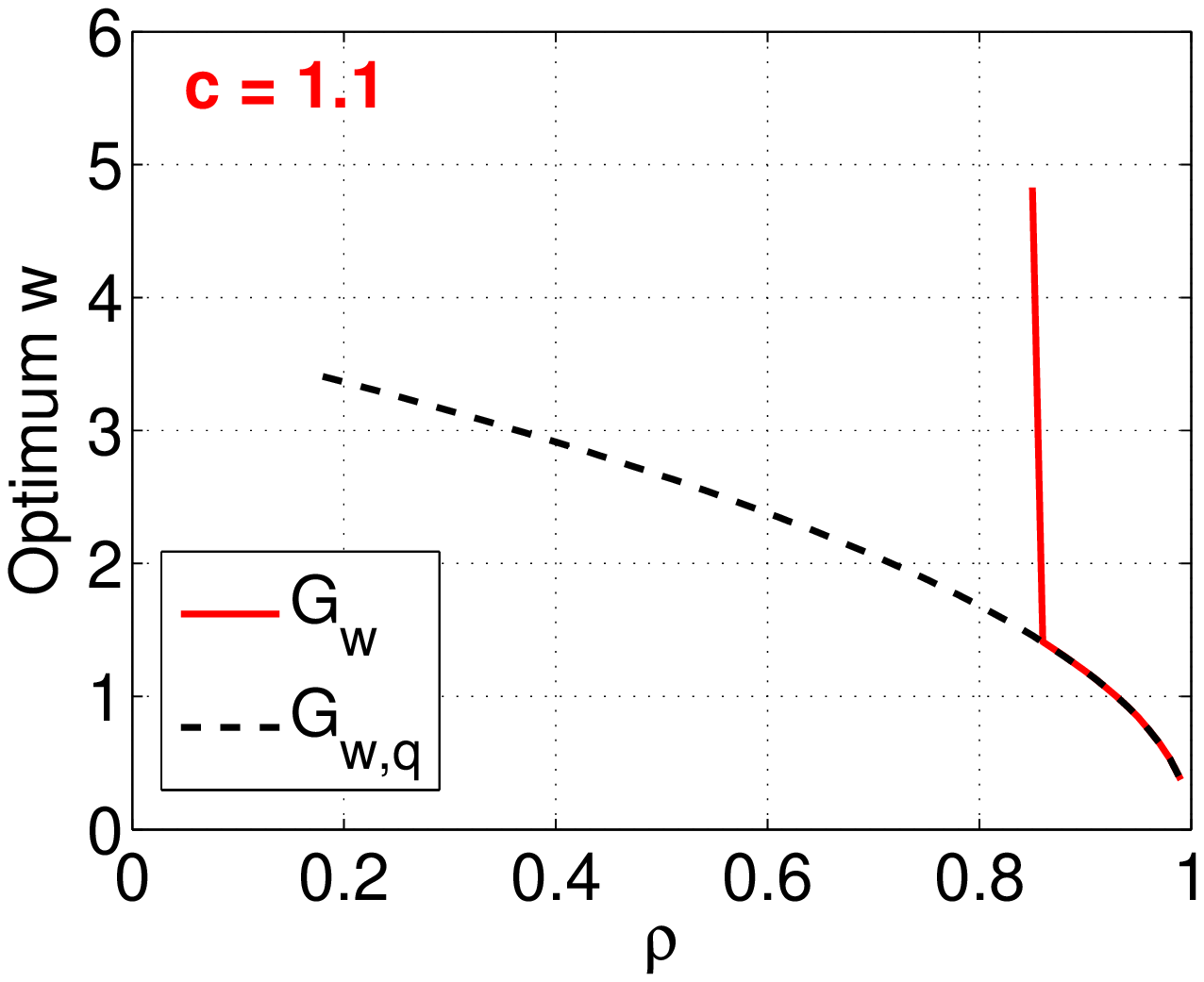}
}

\mbox{
\includegraphics[width = 2.7in]{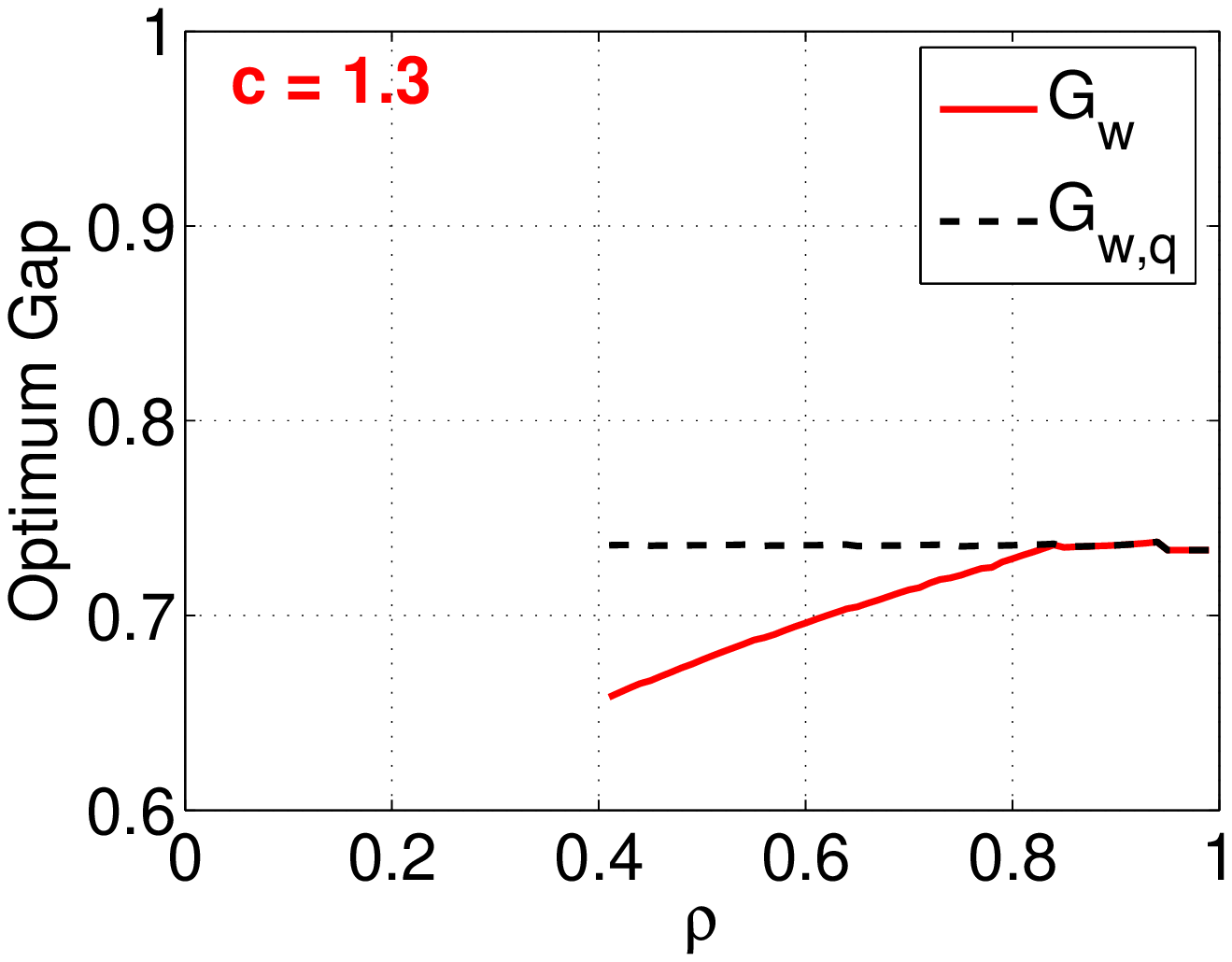}
\includegraphics[width = 2.7in]{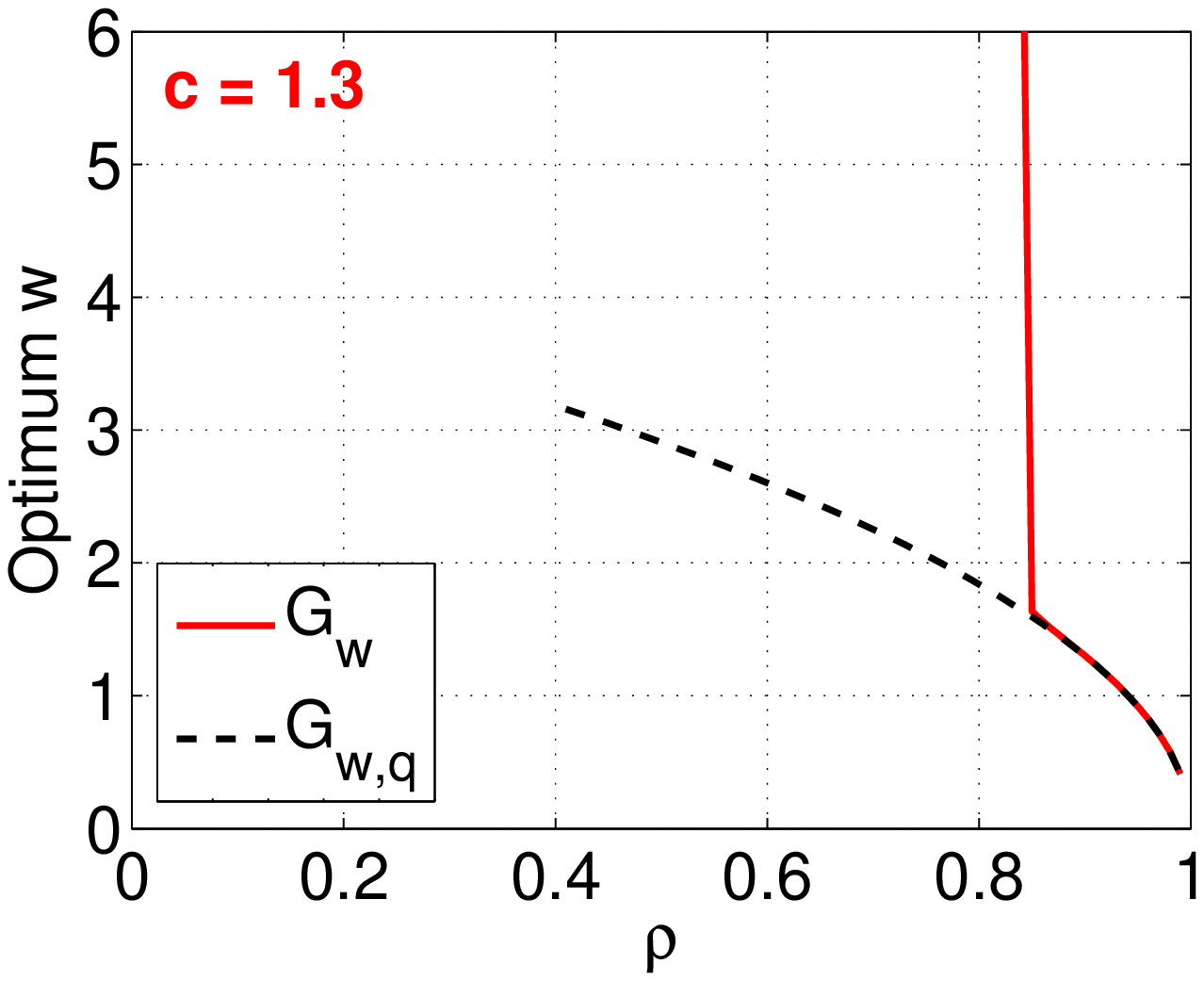}
}
\end{center}
\vspace{-0.2in}
\caption{\textbf{Left panels}: the optimal (smallest) gaps at given $c$ values and the entire range of $\rho$. We can see that $G_{w,q}$ is always larger than $G_w$, confirming that it is better to use $h_w$ instead of $h_{w,q}$. \textbf{Right panels}: the optimal values of $w$ at which the optimal gaps are attained. When the target similarity $\rho$ is very high, it is best to use a relatively small $w$. When the target similarity is not that high, if we use $h_w$, it is best to use $w>3$.   }\label{fig_OptGW1}
\end{figure}

\begin{figure}
\begin{center}

\mbox{
\includegraphics[width = 2.7in]{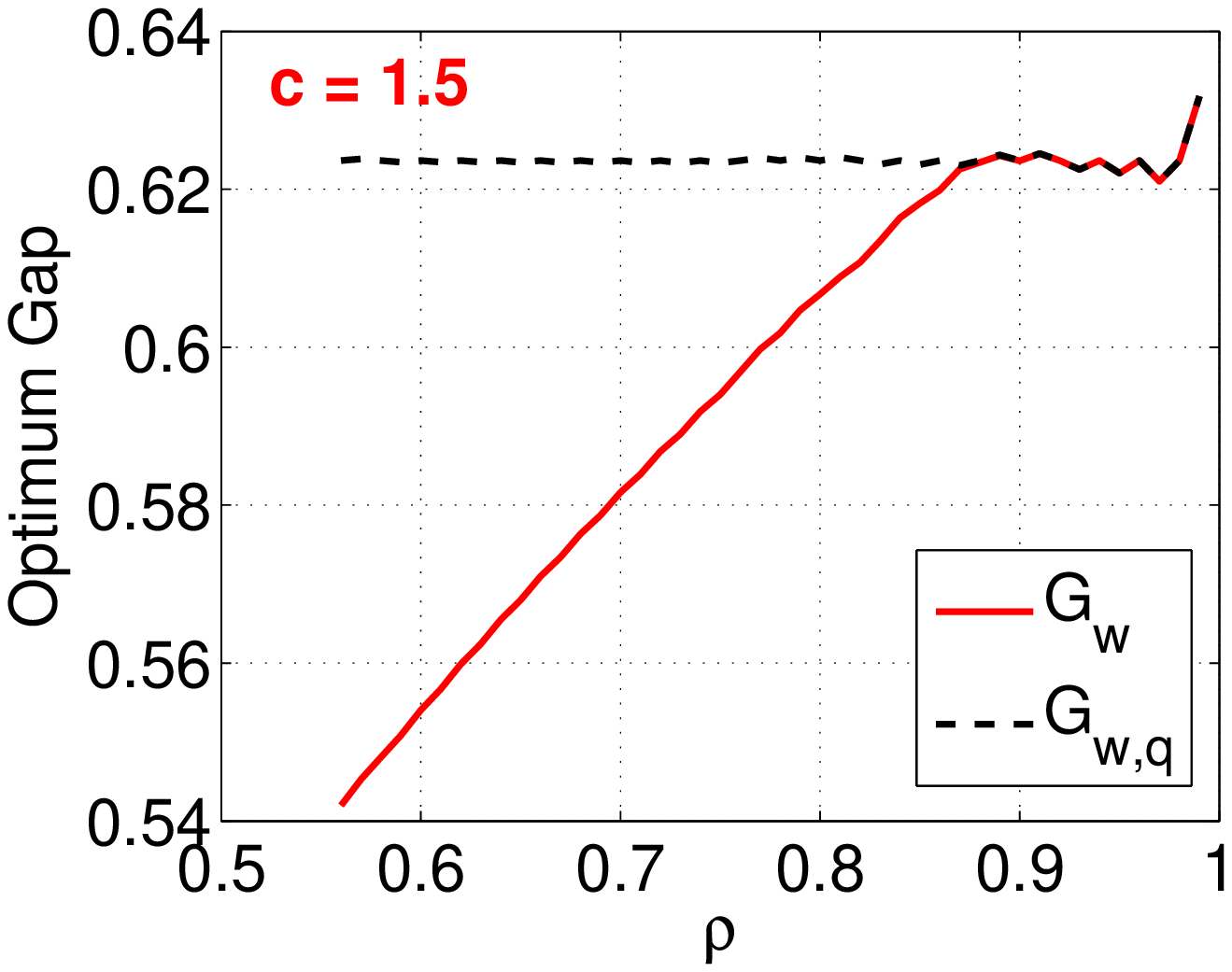}
\includegraphics[width = 2.7in]{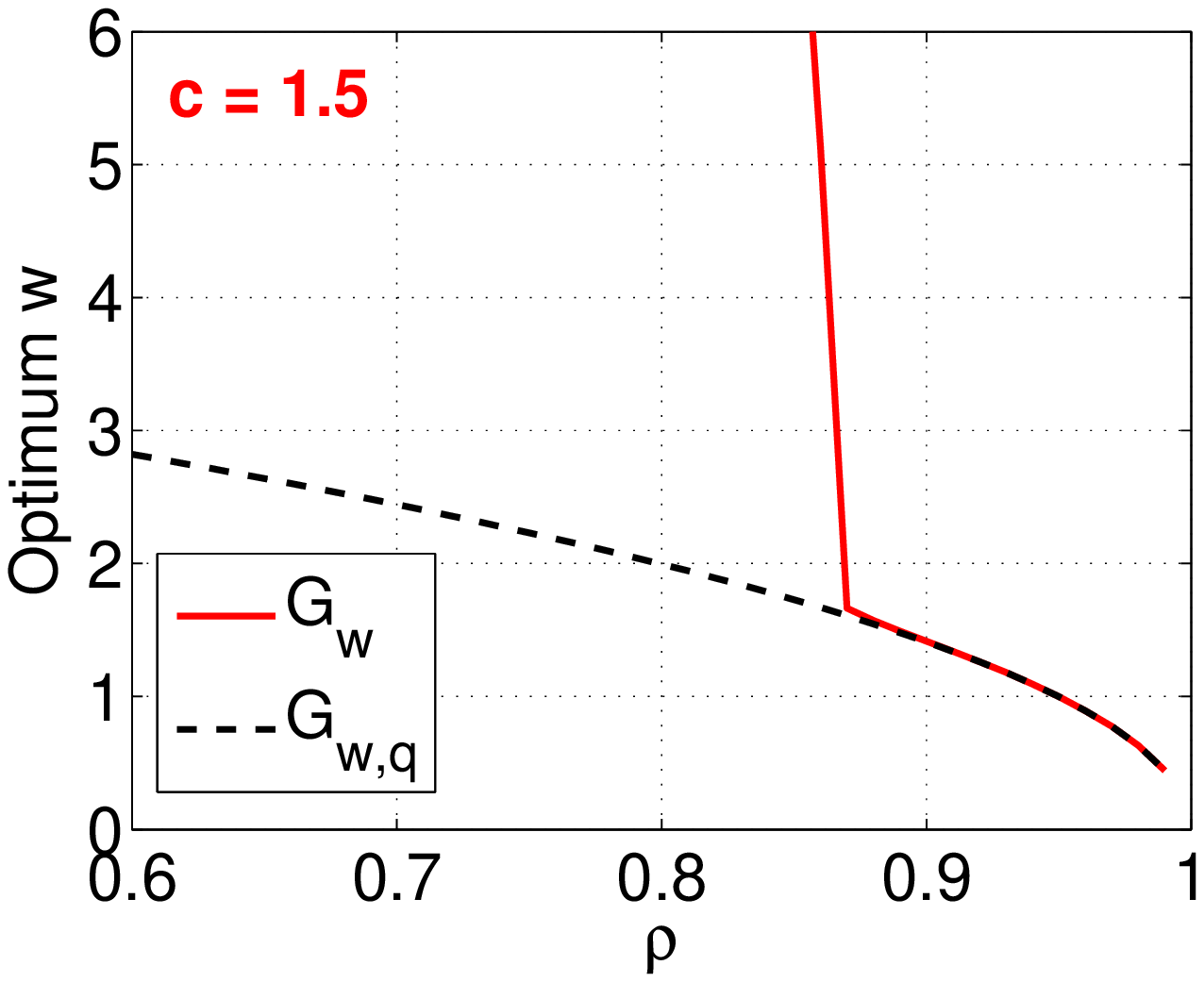}
}

\mbox{
\includegraphics[width = 2.7in]{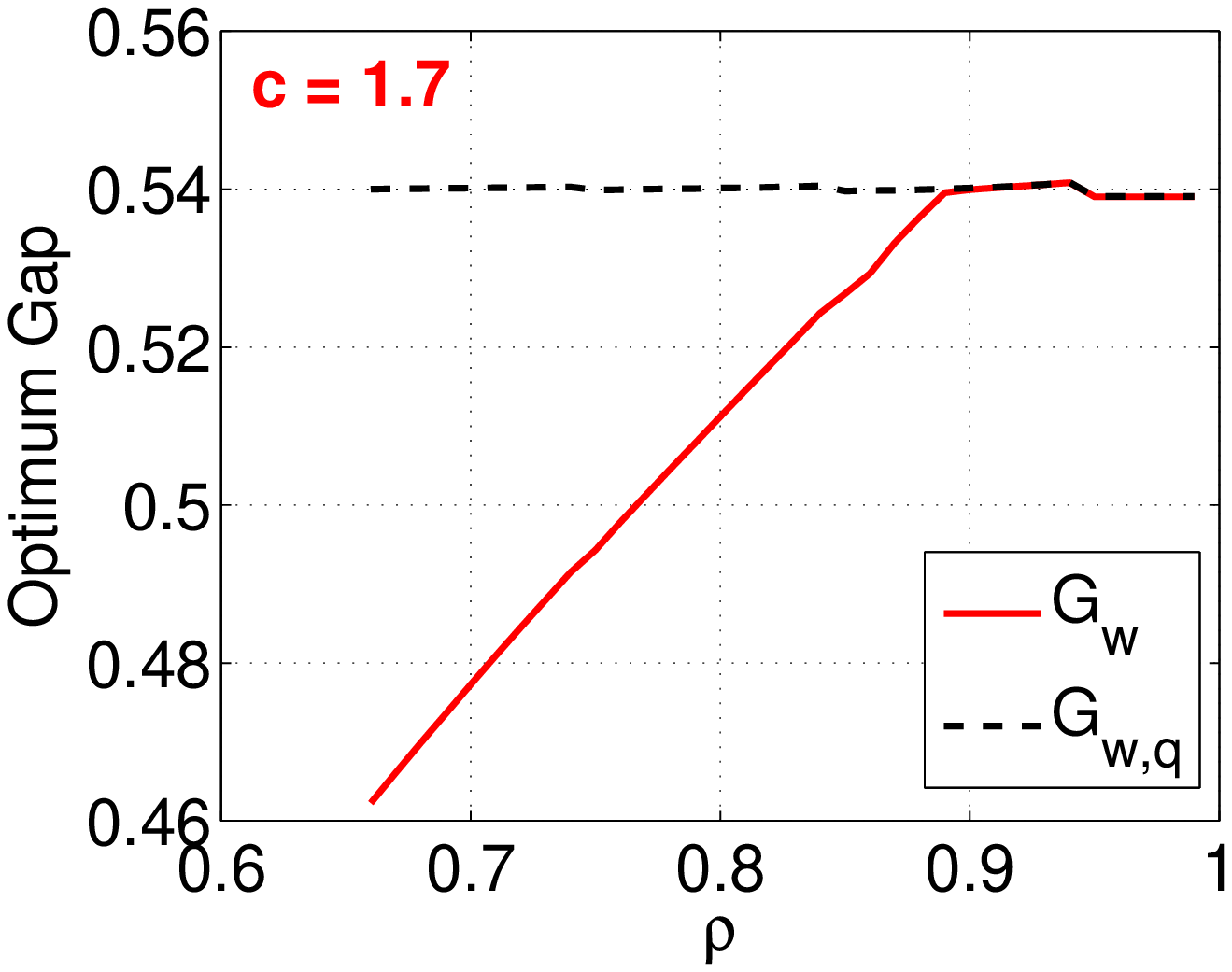}
\includegraphics[width = 2.7in]{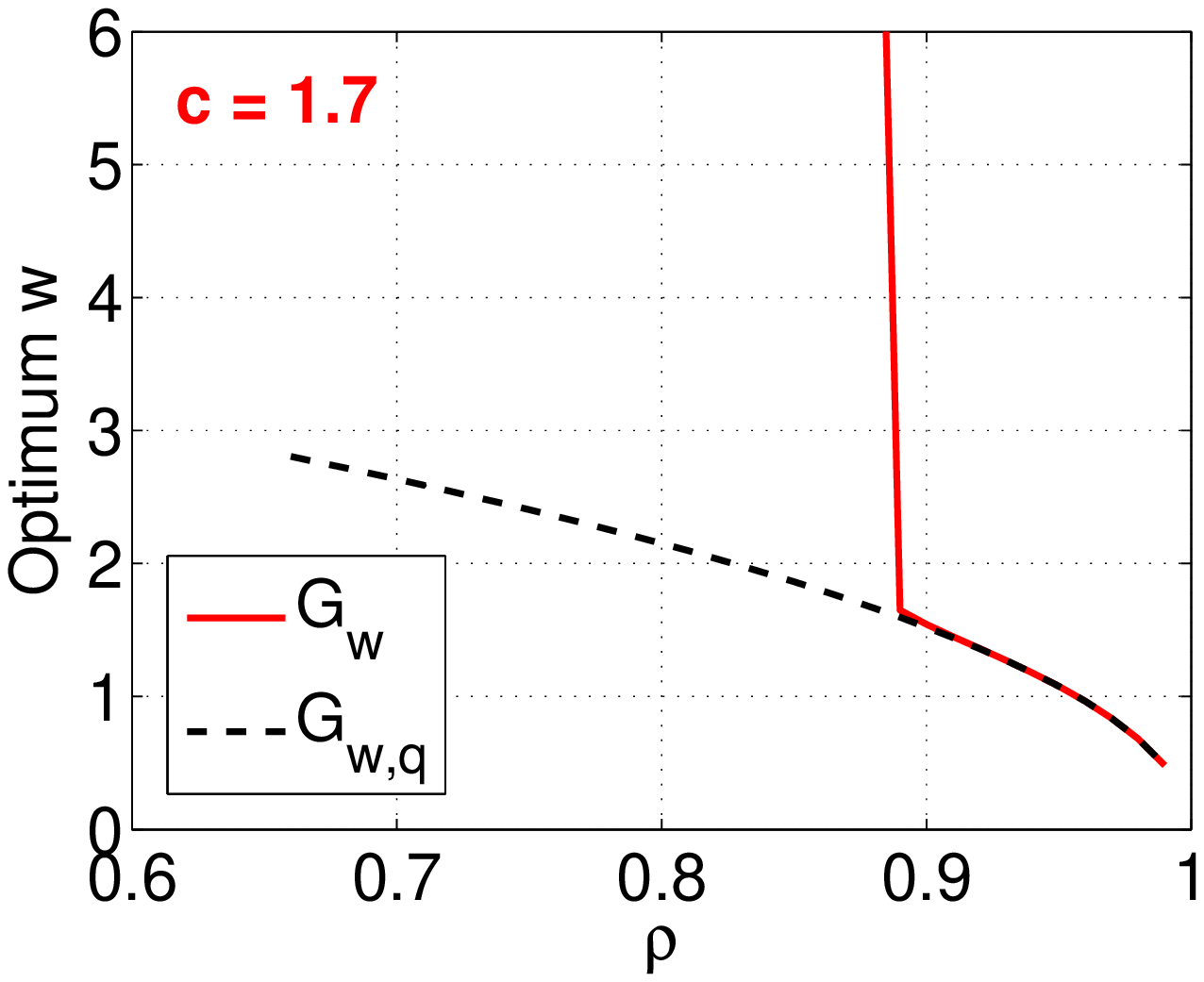}
}

\mbox{
\includegraphics[width = 2.7in]{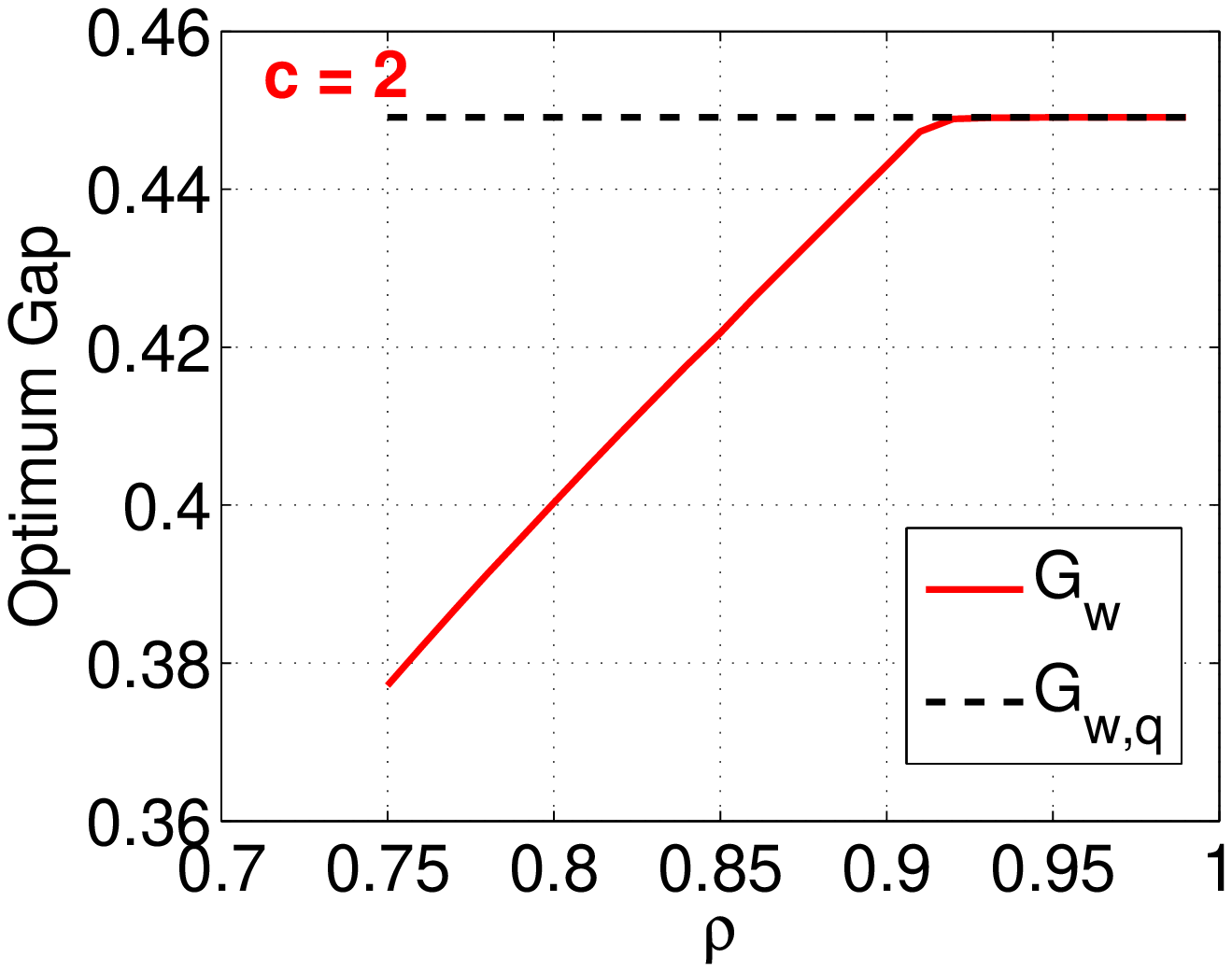}
\includegraphics[width = 2.7in]{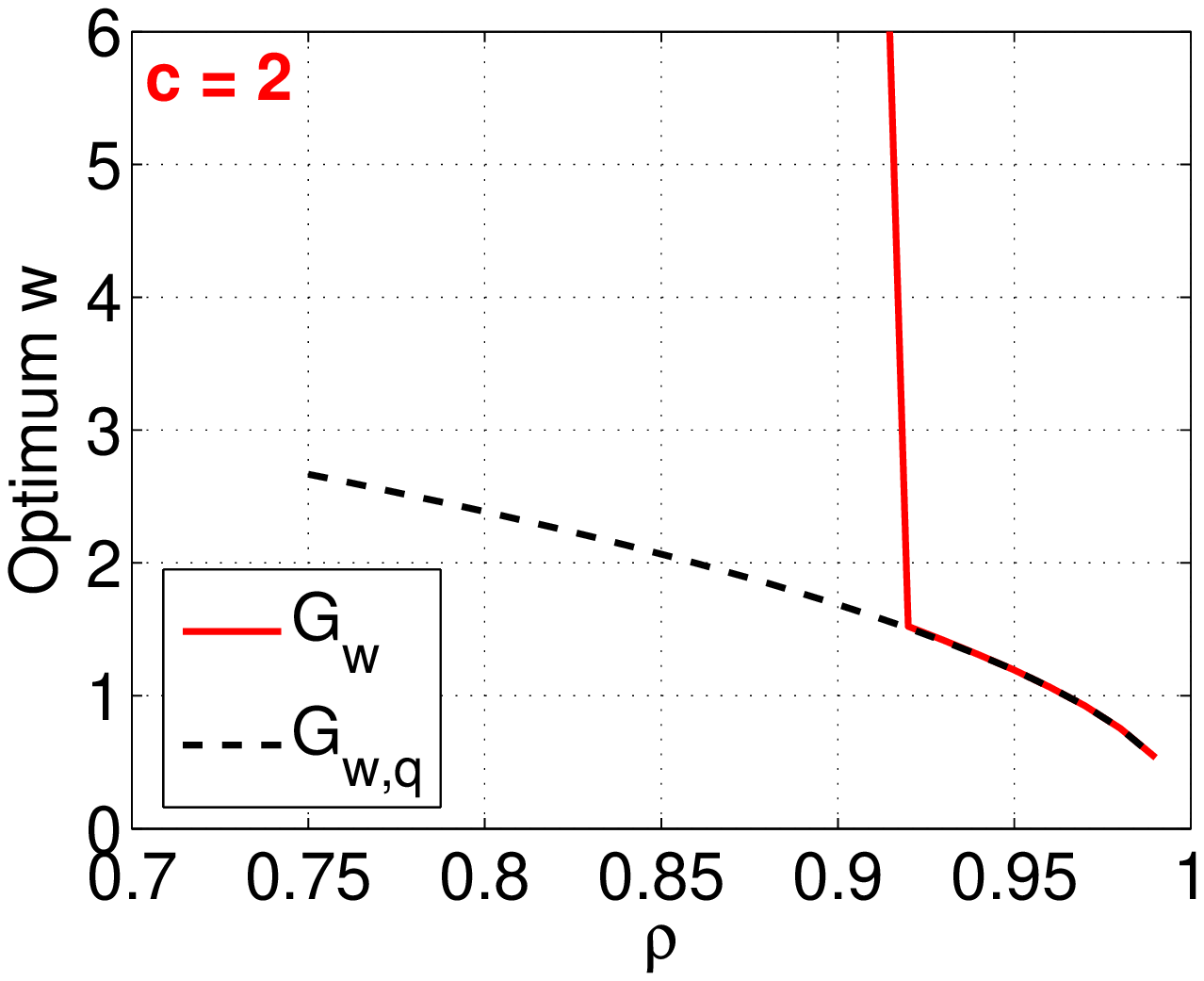}
}

\end{center}
\vspace{-.2in}
\caption{\textbf{Left panels}: the optimal (smallest) gaps at given $c$ values and the entire range of $\rho$. We can see that $G_{w,q}$ is always larger than $G_w$, confirming that it is better to use $h_w$ instead of $h_{w,q}$. \textbf{Right panels}: the optimal values of $w$ at which the optimal gaps are attained. When the target similarity $\rho$ is very high, it is best to use a relatively small $w$. When the target similarity is not that high, if we use $h_w$, it is best to use $w>3$.   }\label{fig_OptGW2}
\end{figure}

To view the optimal gaps more clearly, Figure~\ref{fig_OptGW1} and Figure~\ref{fig_OptGW2} plot the best gaps (left panels) and the optimal $w$ values (right panels) at which the best gaps are attained, for selected values of $c$ and the entire range of $\rho$. The results can be summarized as follows
\begin{itemize}
\item At any $\rho$ and $c$, the optimal gap $G_{w,q}$ is always at least as large as the optimal gap $G_{w}$.  At relatively low similarities, the optimal $G_{w,q}$  can be substantially larger than the optimal $G_{w}$.
\item When the target similarity level $\rho$ is high (e.g., $\rho>0.85$), for both schemes $h_w$ and $h_{w,q}$, the optimal $w$ values are relatively low, for example, $w=1\sim 1.5$ when $0.85<\rho<0.9$. In this region, both $h_{w,q}$ and $h_{w}$ behavior similarly.
\item When the target similarity level $\rho$ is not so high, for $h_w$, it is best to use a large value of $w$, in particular $w\geq 2\sim3$. In comparison, for $h_{w,q}$, the optimal $w$ values grow smoothly with decreasing $\rho$.
\end{itemize}

These plots again confirm the previous comparisons: (i) we should always replace $h_{w,q}$ with $h_{w}$; (ii) if we use $h_w$ and target at very high similarity, a good choice of $w$ might be $w=1\sim 1.5$; (iii) if we use $h_{w}$ and the target similarity is not too high, then we can safely use $w=2\sim3$. \\

We should also mention that,  although the optimal $w$ values for $h_w$ appear to exhibit a ``jump'' in the right panels of  Figure~\ref{fig_OptGW1} and Figure~\ref{fig_OptGW2}, the choice of $w$ does not influence the performance much, as shown in previous plots. In Figures~\ref{fig_GwqR099C} to~\ref{fig_GwqR05C}, we have seen that even when the optimal $w$ appear to approach ``$\infty$'', the actual gaps are not much difference between $w=3$ and $w\gg3$. In the real-data evaluations in the next section, we will see the same phenomenon for $h_w$.\\

Note that the Gaussian density decays rapidly at the tail, for example, $1-\Phi(6) = 9.9\times10^{-10}$. If we choose $w=1.5$, or 2, or 3, then we just need a small number of bits to code each hashed value.

\newpage\clearpage

\section{An Experimental Study}

Two datasets, {\em Peekaboom} and {\em Youtube}, are used in our experiments for validating the theoretical results. {\em Peekaboom} is  a standard image retrieval dataset, which is divided into two subsets, one with 1998 data points and another with 55599 data points. We use the larger subset for building hash tables and the smaller subset for query data points. The reported experimental results are averaged over all query data points.

Available in the UCI repository, {\em Youtube} is a multi-view dataset. For simplicity, we only use the largest set of audio features. The original training set, with 97934 data points, is used for building hash tables. 5000 data points, randomly selected from the original test set, are used as query data points. \\

We use the standard $(K,L)$-LSH  implementation~\cite{Proc:Indyk_STOC98}. We generate $K\times L$ independent hash functions $h^{i,j}$, $i=1$ to $K$, $j=1$ to $L$.  For each hash table $j$, $j = 1$ to $L$, we concatenate $K$ hash functions $<h^{1,j}, h^{2,j}, h^{3,j}, ..., h^{K,j}>$. For each data point, we compute the hash values and place them  (in fact, their pointers) into the appropriate buckets of the hash table $i$. In the query phase, we compute the hash value of the query data points using the same hash functions to find the bucket in which the query data point belongs to and only search for near neighbor among the data points in that bucket of hash table $i$. We repeat the process for each hash table and the final retrieved data points are the union of the retrieved data points in all the hash tables. Ideally, the  number of retrieved data points will be substantially smaller than the total number of data points. We use the term {\em fraction  retrieved} to indicate the ratio of the number of retrieved data points over the total number of data points. A smaller value of {\em fraction retrieved} would be more desirable.

To thoroughly evaluate the two coding schemes, we conduct extensive experiments on the two datasets, by using many combinations of $K$ (from 3 to 40) and $L$ (from 1 to 200). At each choice of $(K,L)$, we vary $w$  from 0.5 to 5. Thus, the total number of combinations is  large, and the experiments are very time-consuming.\\

There are many ways to evaluate the performance of an LSH scheme. We could specify a threshold of similarity and only count the retrieved data points whose (exact) similarity is above the threshold as ``true positives''. To avoid specifying a threshold and consider the fact that in practice people often would like to retrieve the   top-$T$ nearest neighbors, we take a simple approach by computing the {\em recall} based on top-$T$ neighbors. For example, suppose the  number of retrieved data points is 120, among which 70 data points belong to the top-$T$. Then the recall value would be $70/T = 70\%$ if $T=100$.  Ideally, we hope the recalls would be as high as possible and in the meanwhile we hope to keep the fraction retrieved as low as possible. \\

\begin{figure}
\begin{center}
\mbox{
\includegraphics[width = 2.35in]{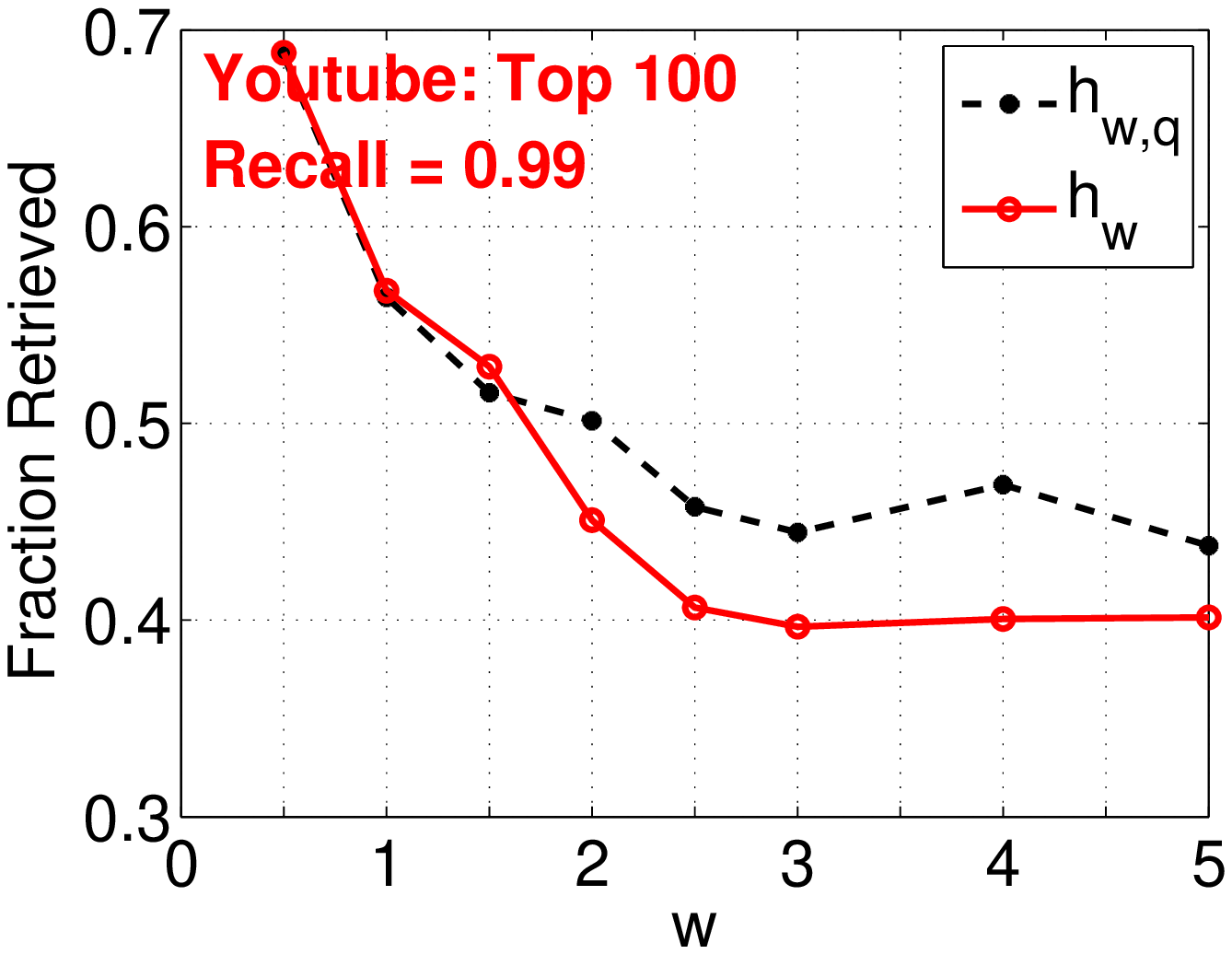}\hspace{-0.15in}
\includegraphics[width = 2.35in]{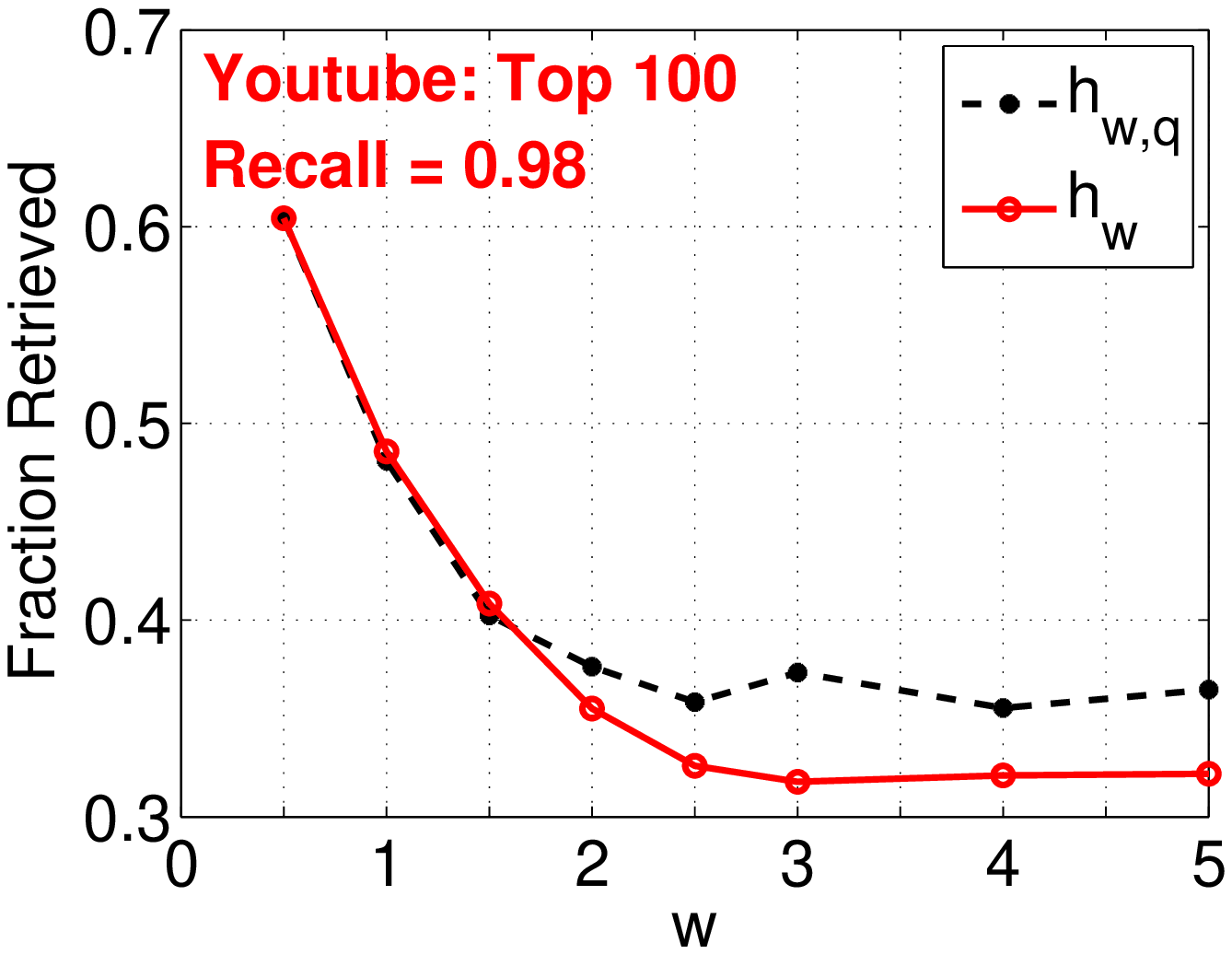}\hspace{-0.15in}
\includegraphics[width = 2.35in]{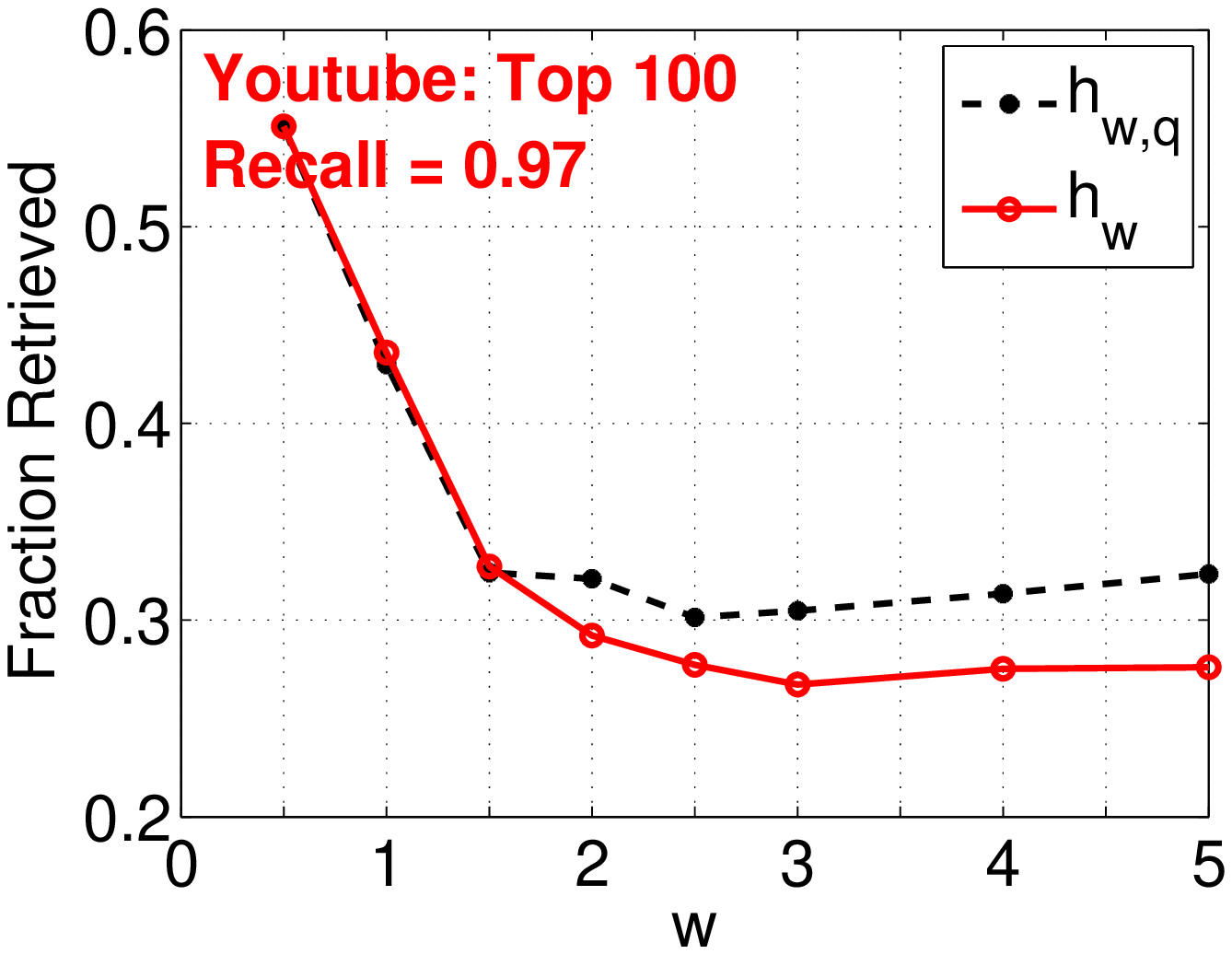}
}

\mbox{
\includegraphics[width = 2.35in]{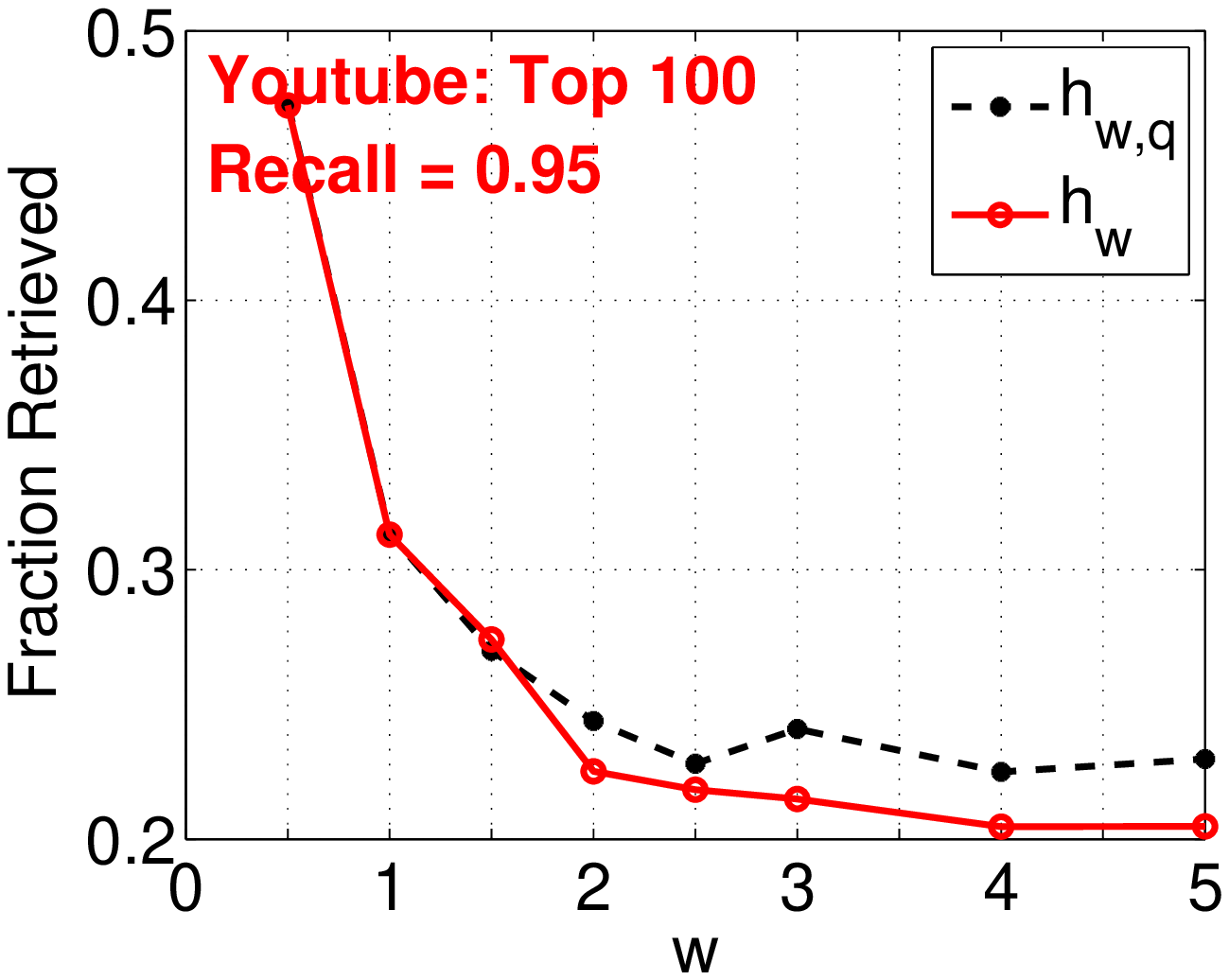}\hspace{-0.15in}
\includegraphics[width = 2.35in]{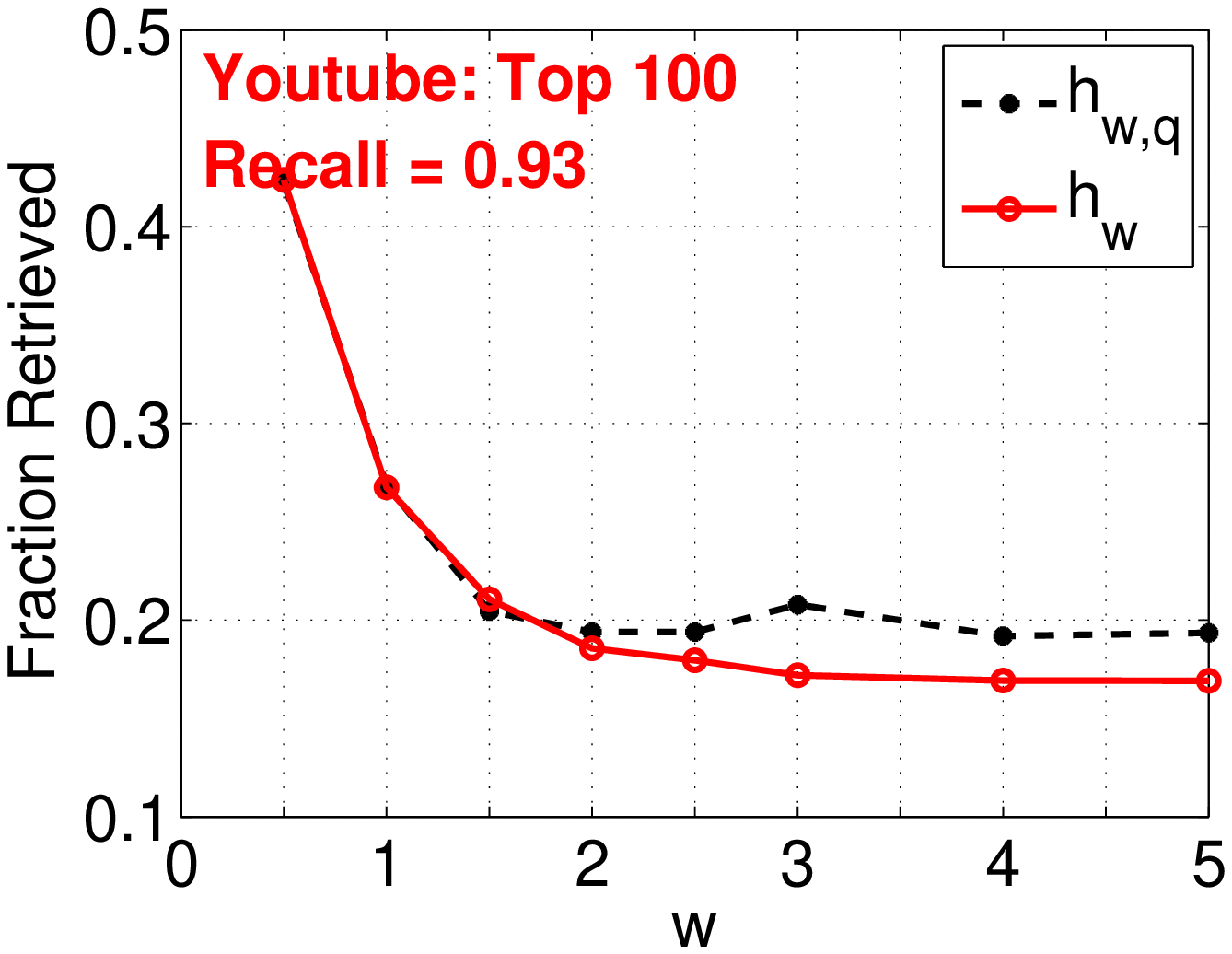}\hspace{-0.15in}
\includegraphics[width = 2.35in]{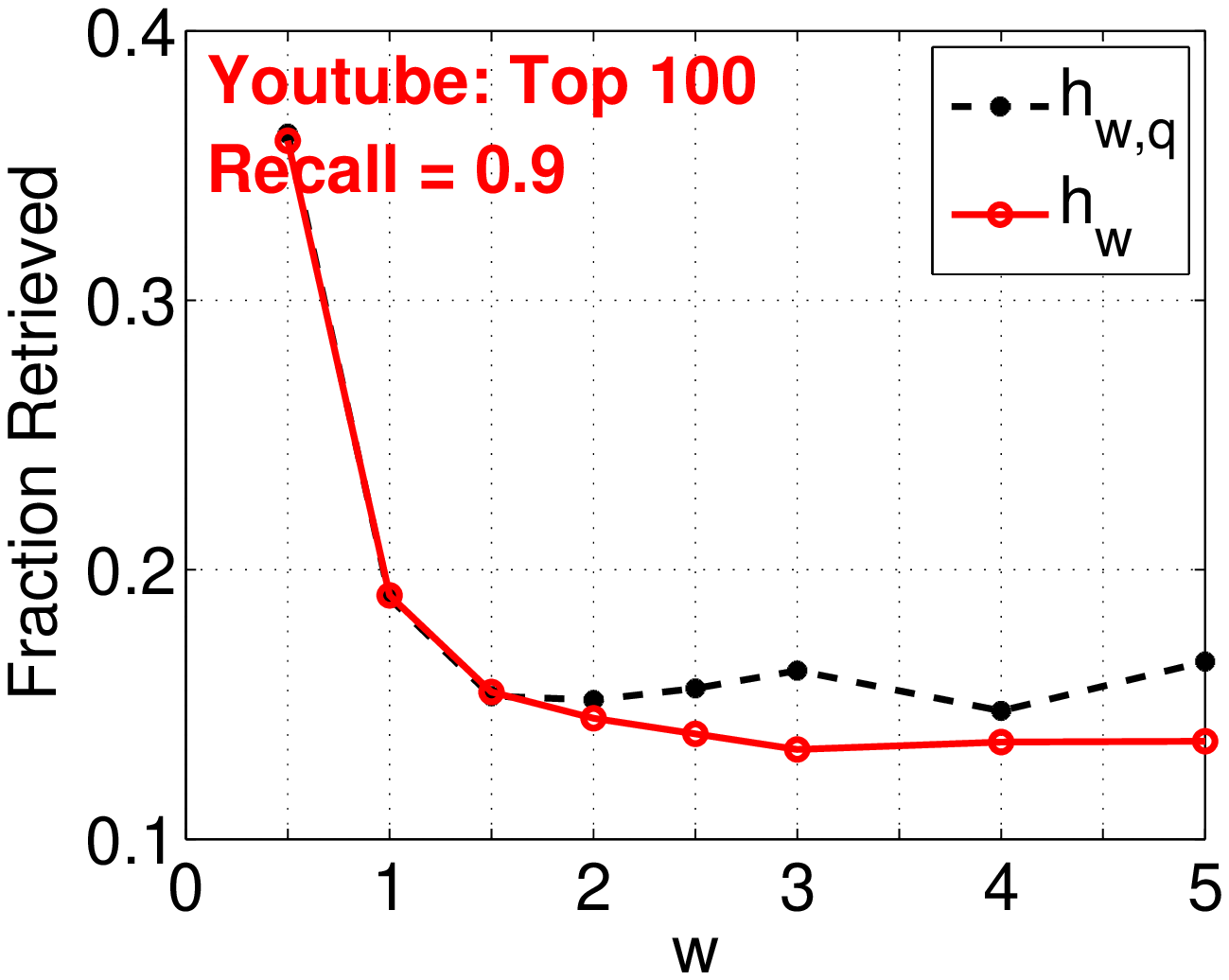}
}

\mbox{
\includegraphics[width = 2.35in]{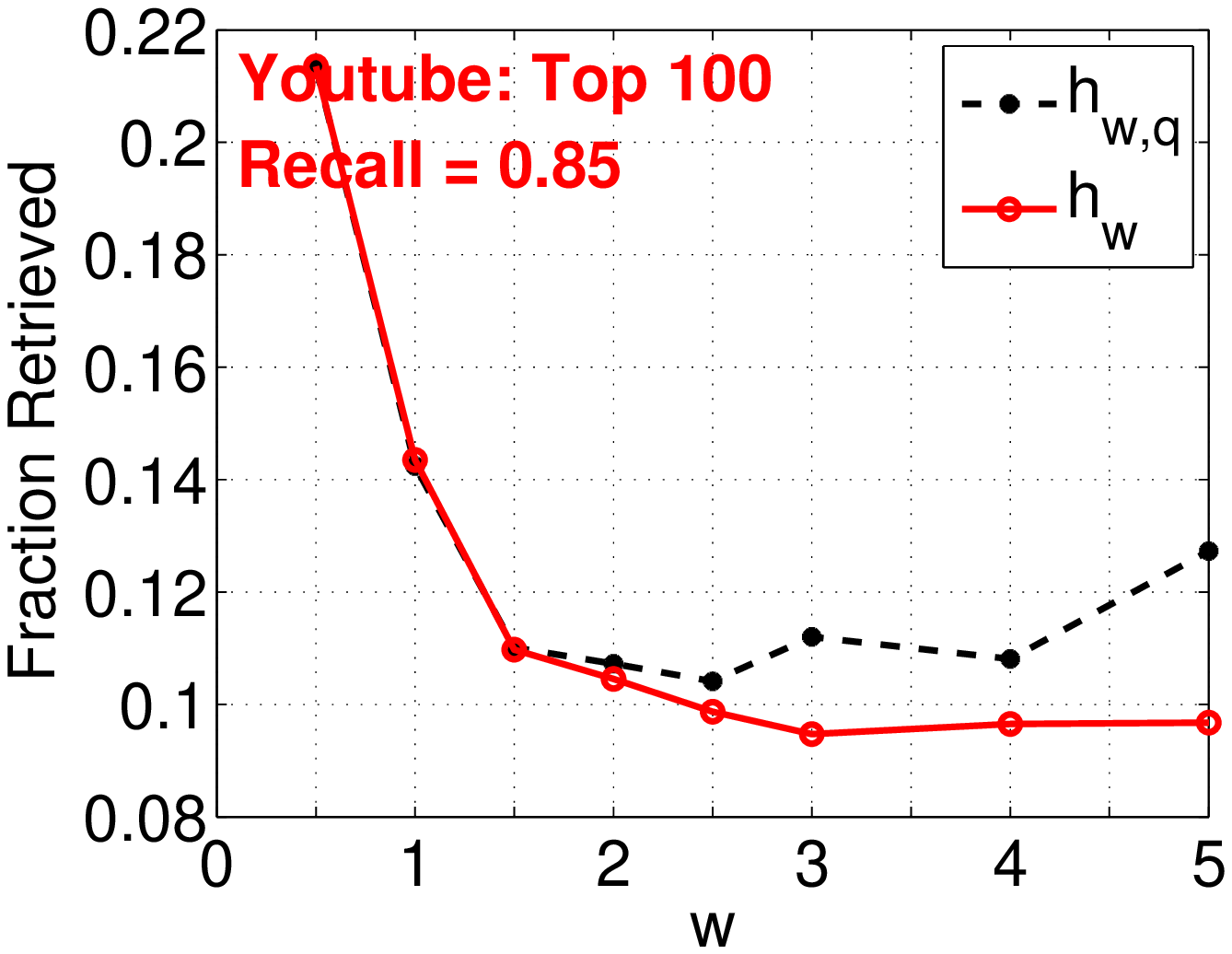}\hspace{-0.15in}
\includegraphics[width = 2.35in]{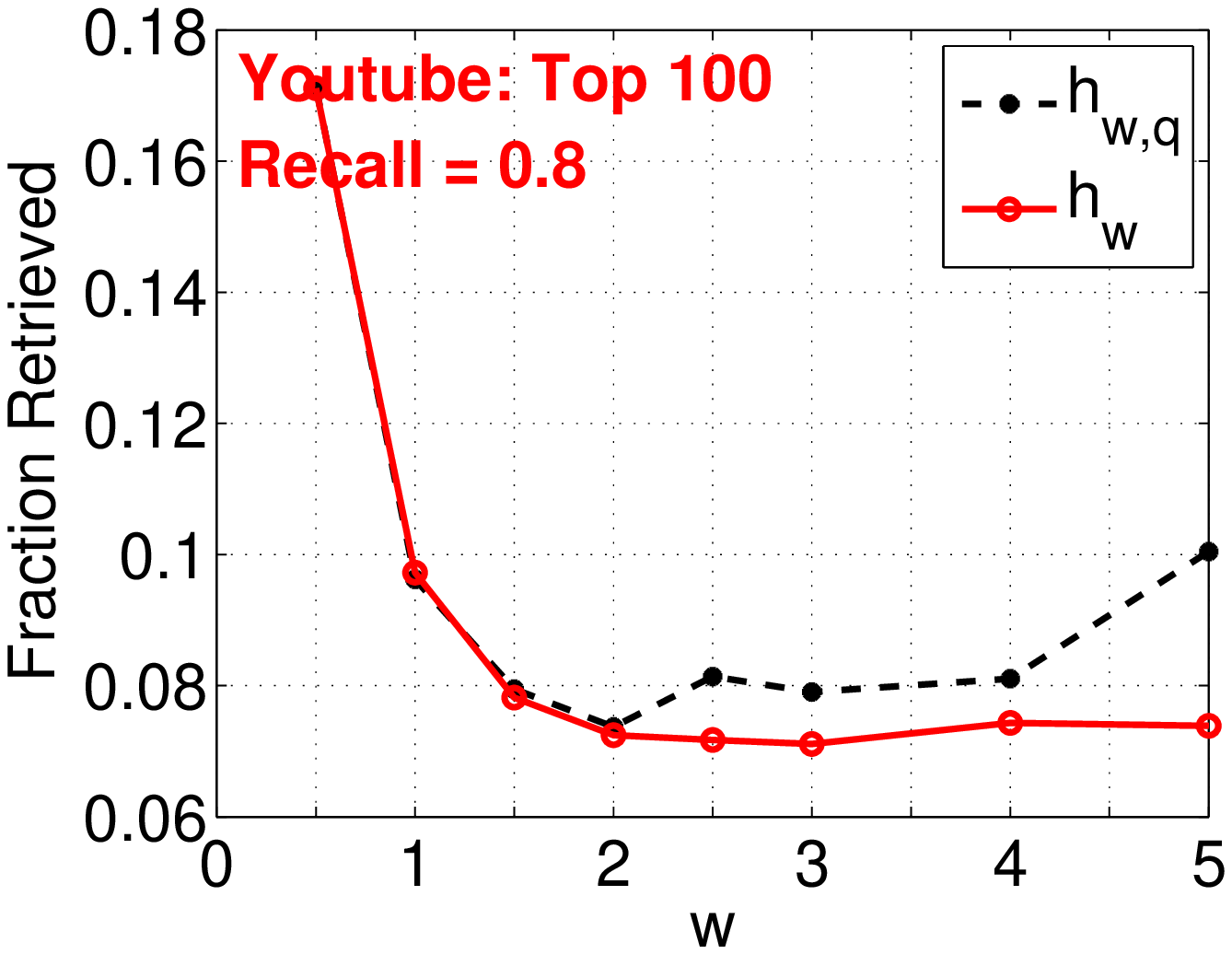}\hspace{-0.15in}
\includegraphics[width = 2.35in]{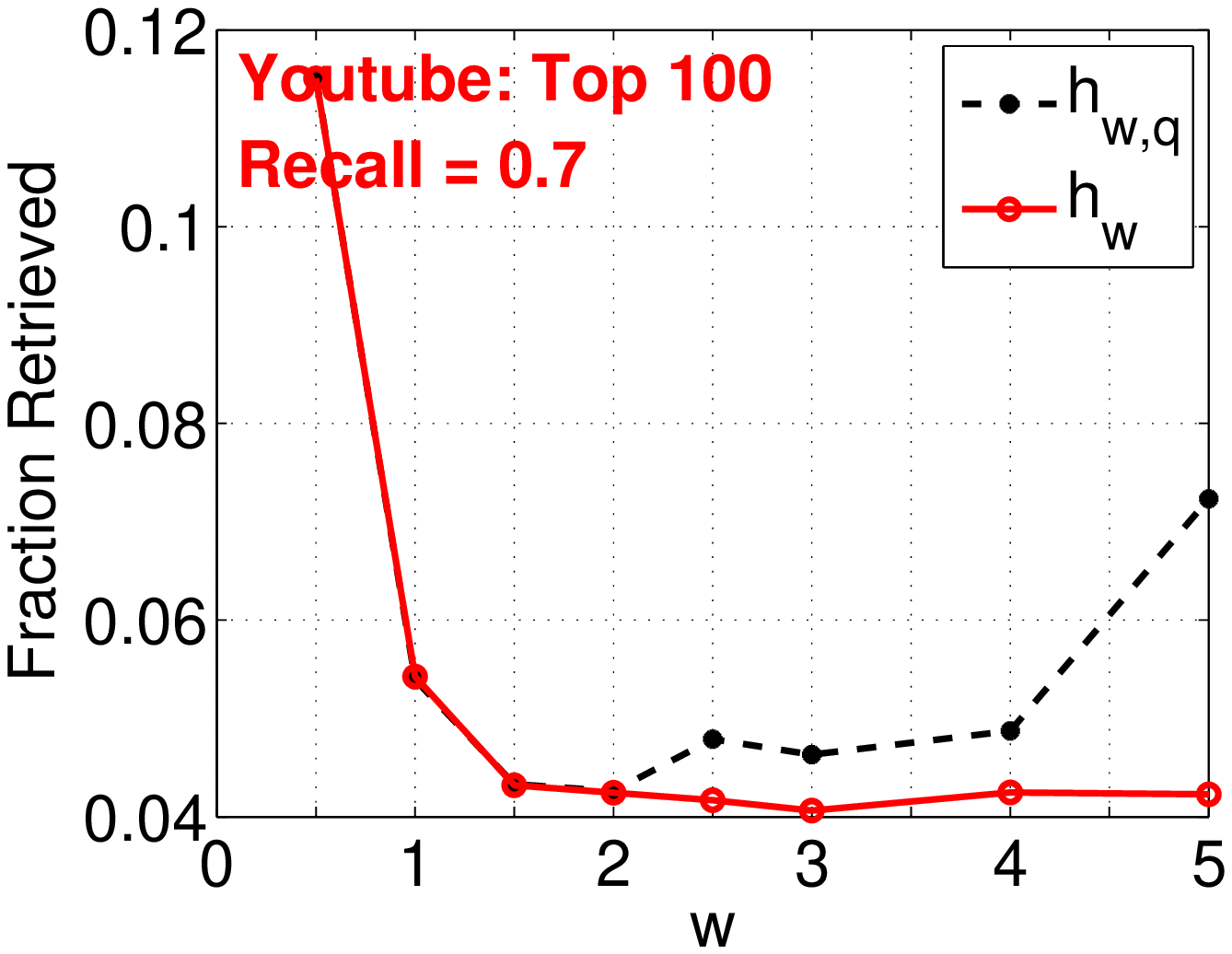}
}

\mbox{
\includegraphics[width = 2.35in]{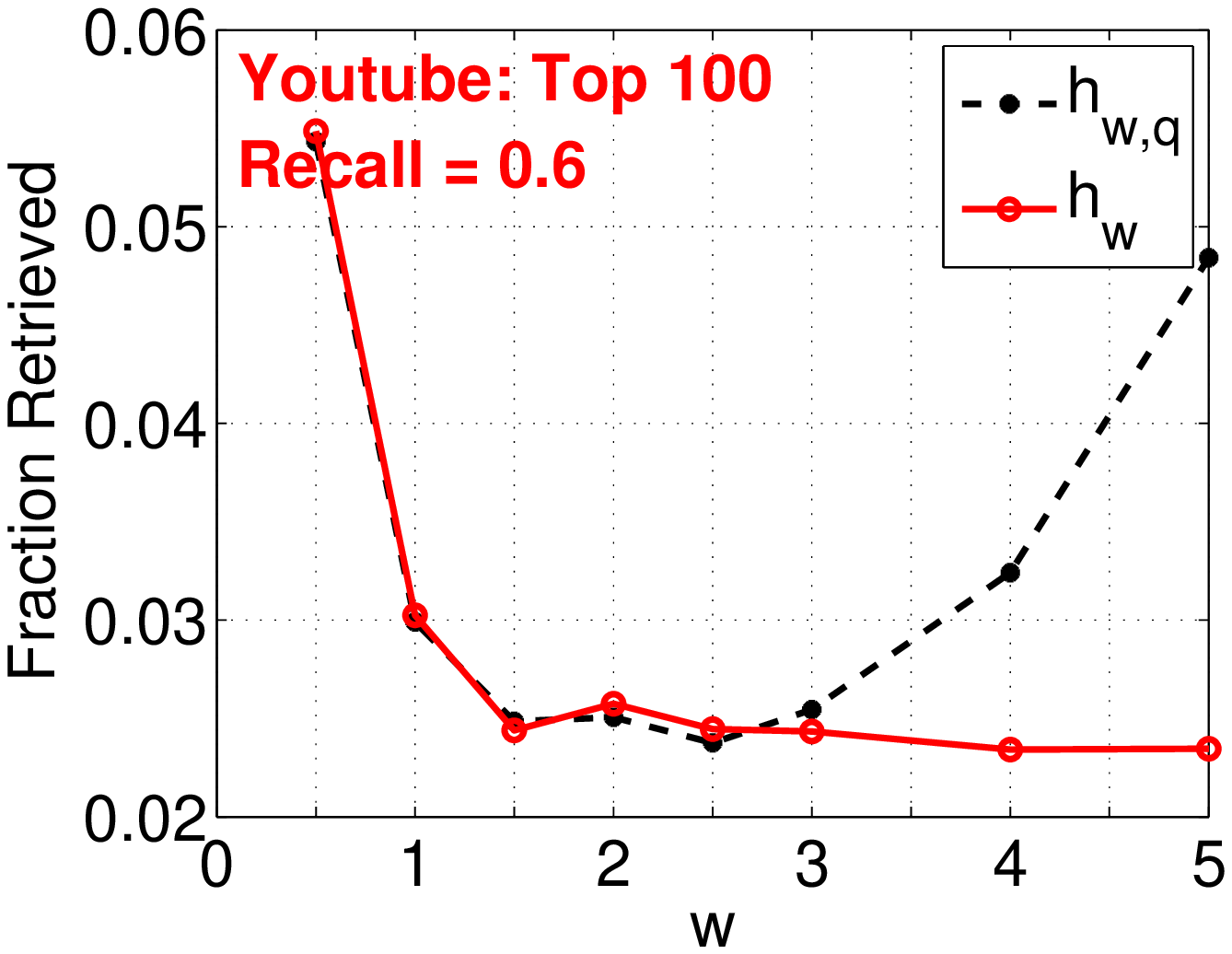}\hspace{-0.15in}
\includegraphics[width = 2.35in]{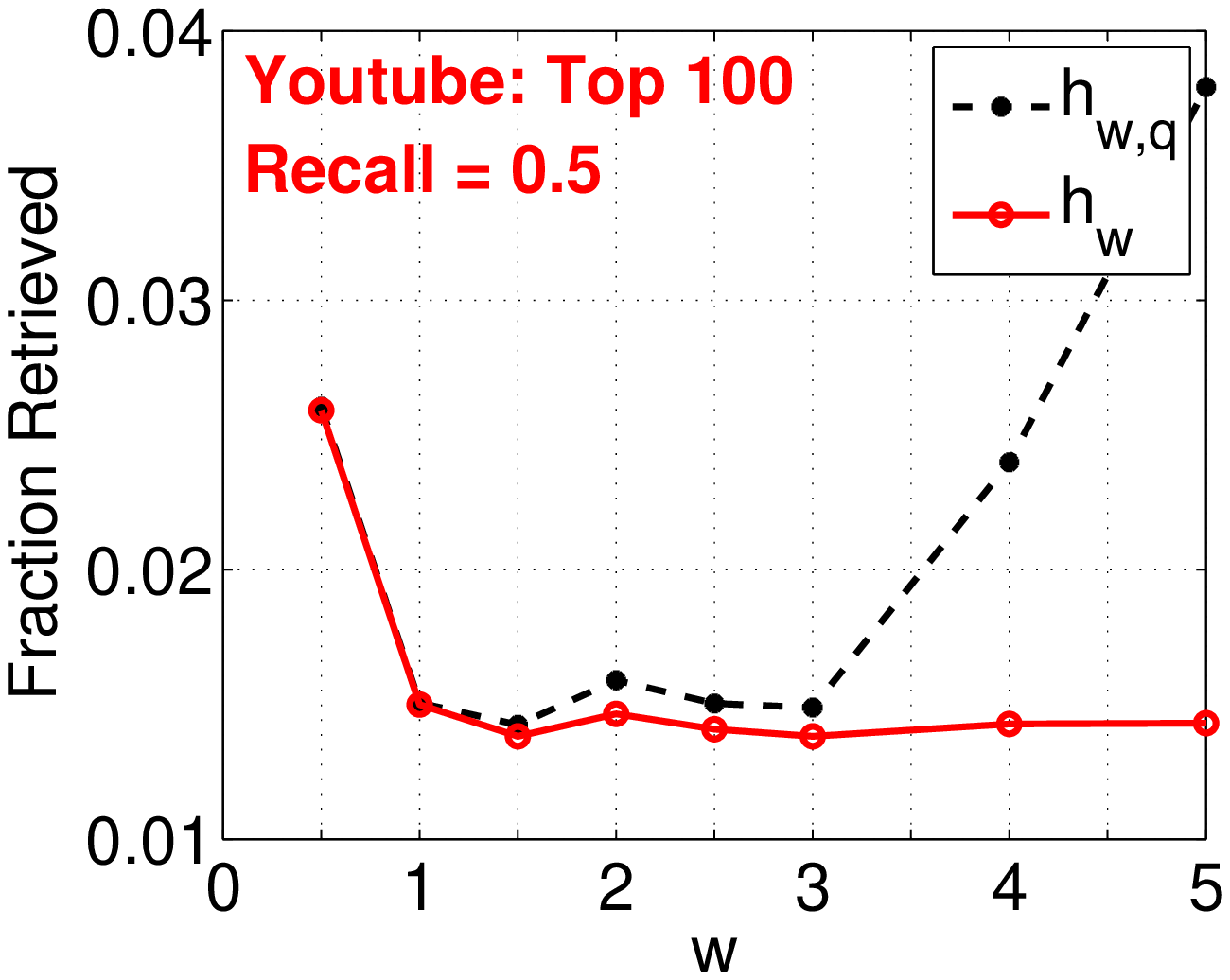}\hspace{-0.15in}
\includegraphics[width = 2.35in]{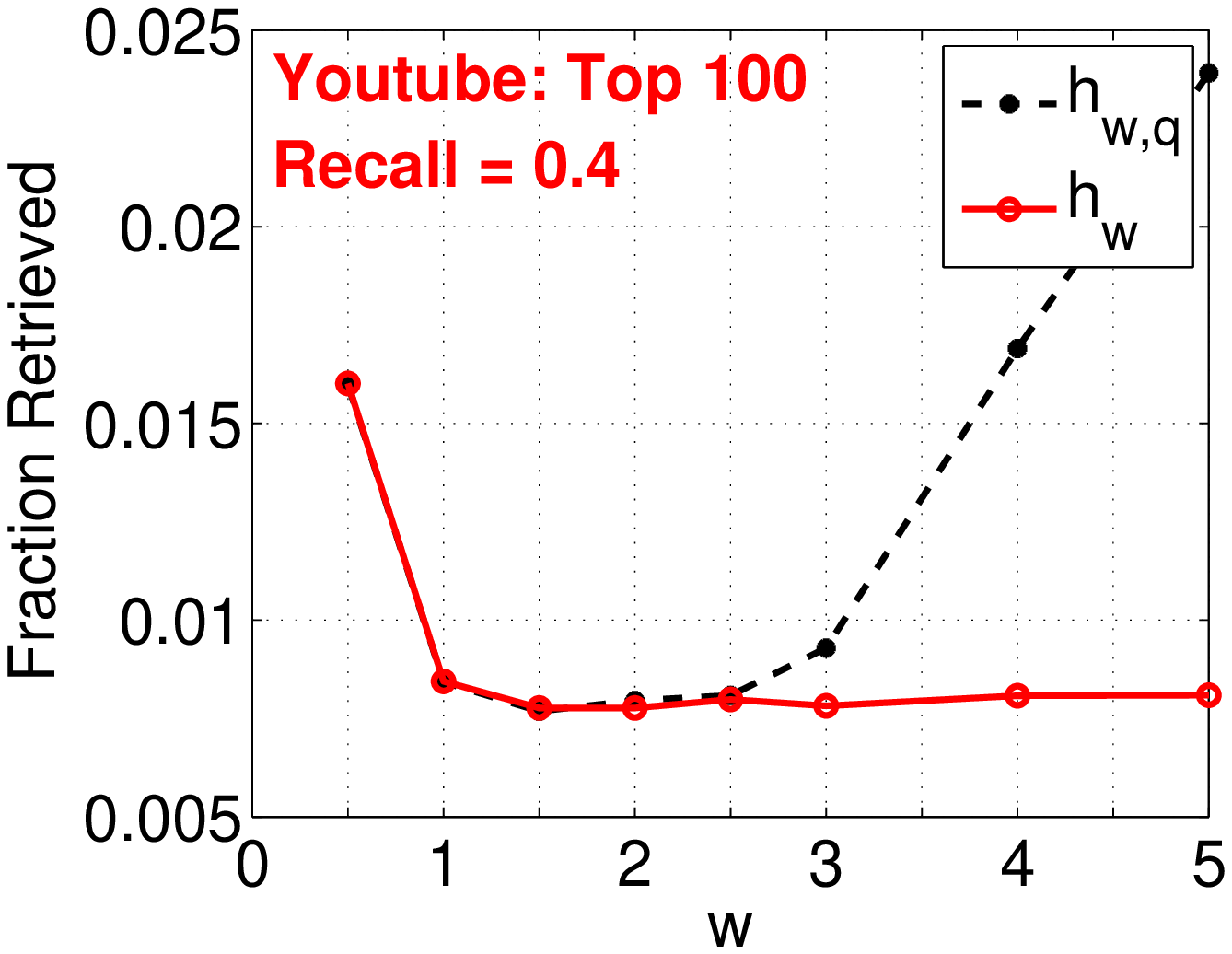}
}

\mbox{
\includegraphics[width = 2.35in]{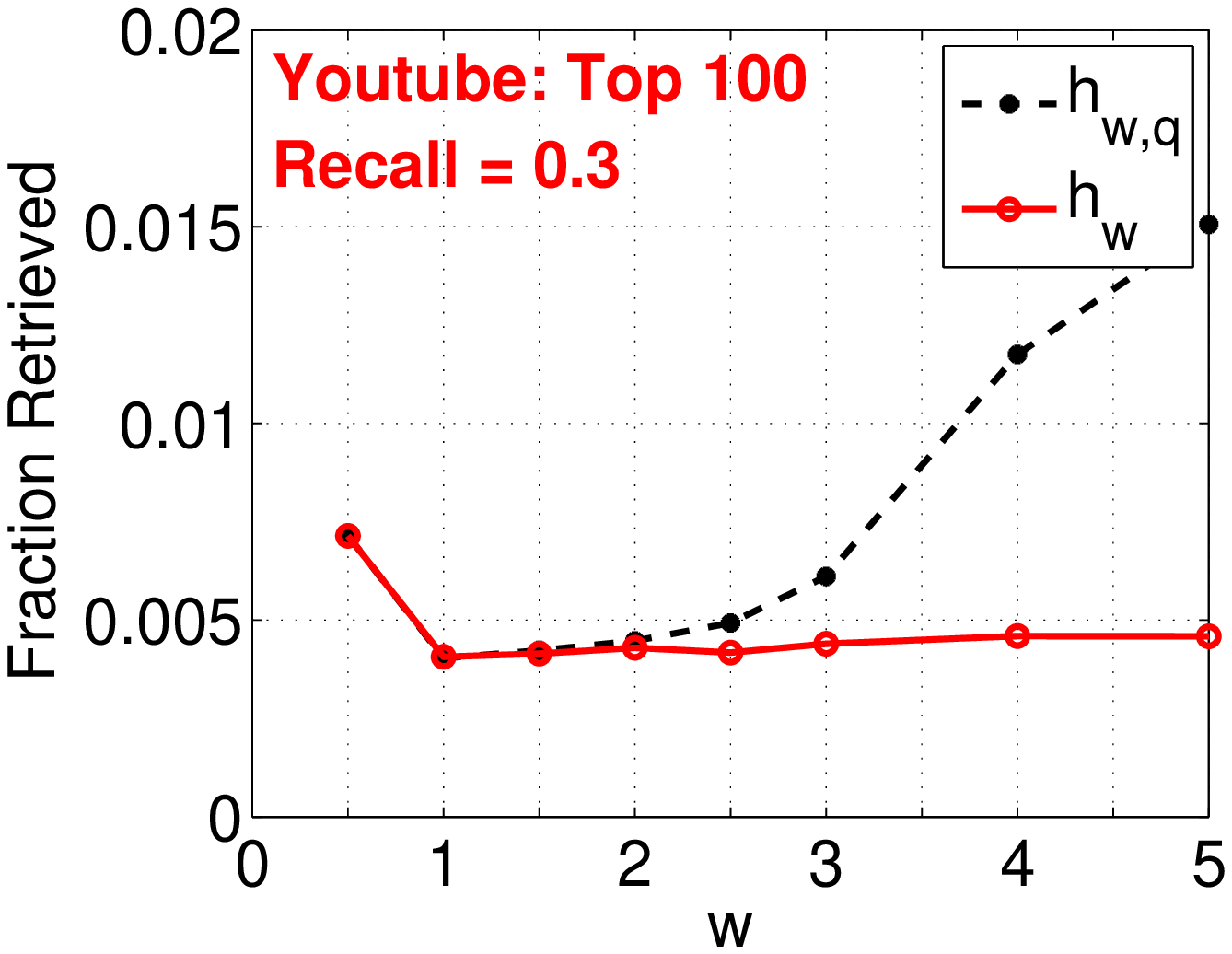}\hspace{-0.15in}
\includegraphics[width = 2.35in]{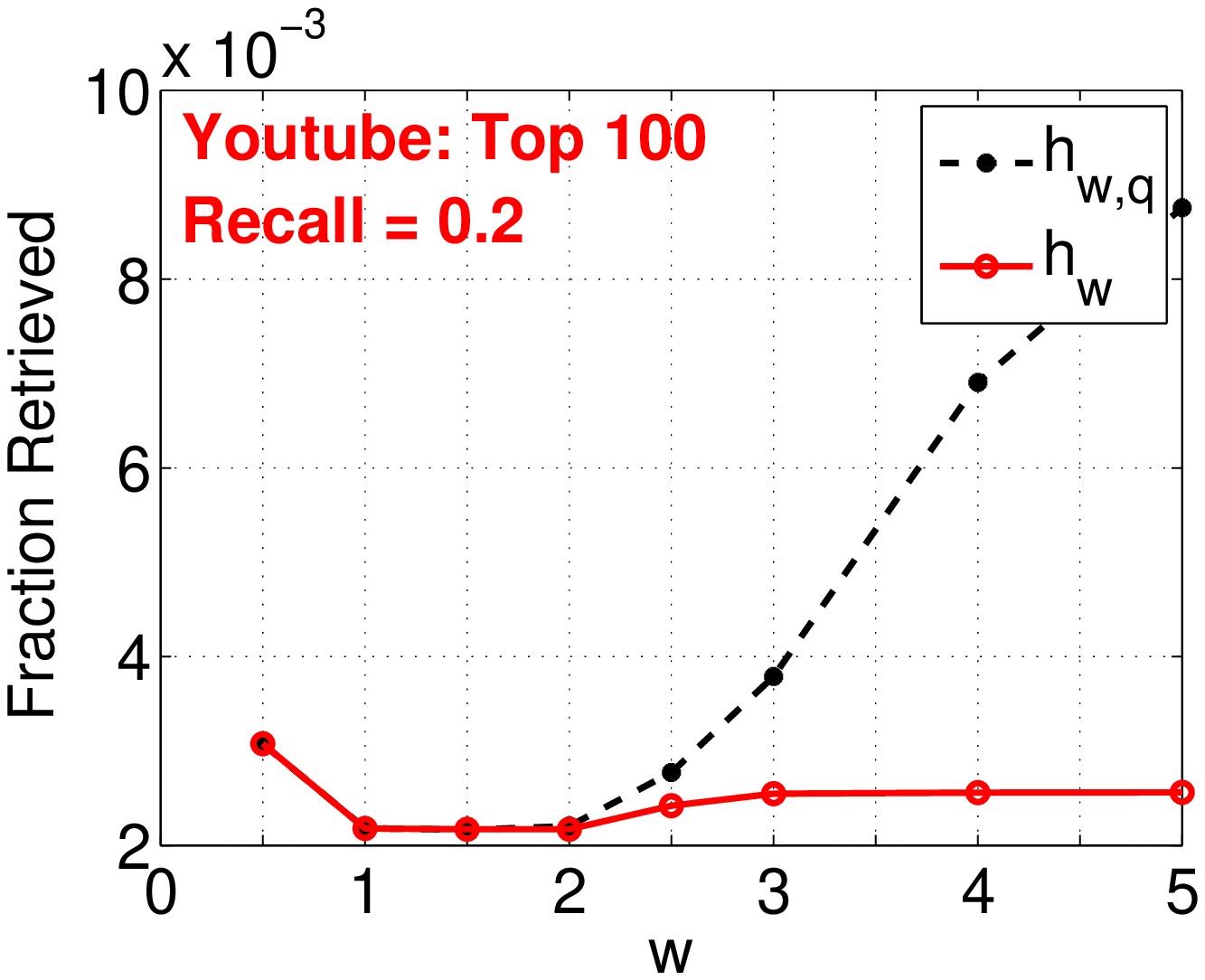}\hspace{-0.15in}
\includegraphics[width = 2.35in]{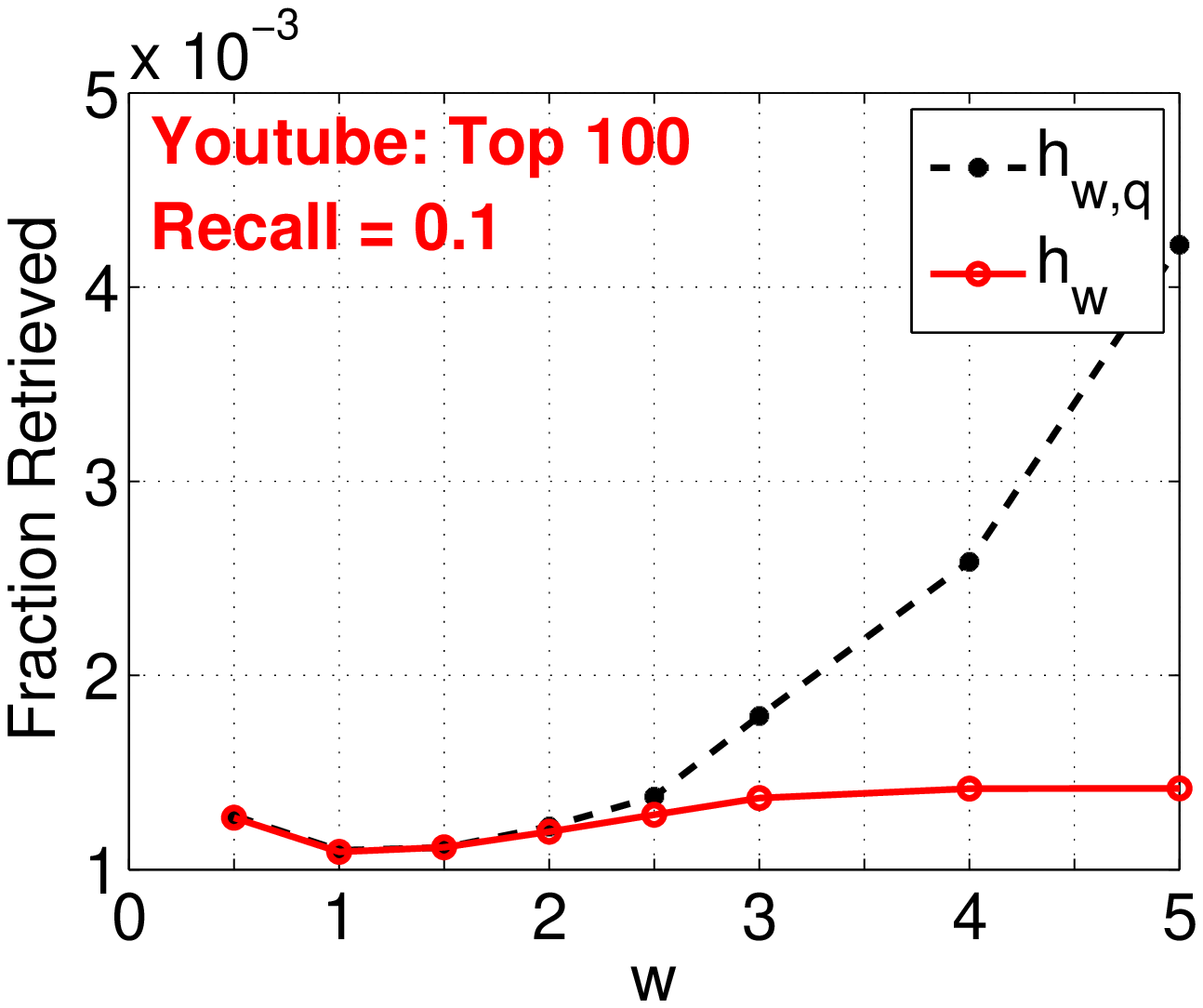}
}

\end{center}
\vspace{-.3in}
\caption{ \textbf{Youtube Top 100} . In each panel, we plot the optimal {\em fraction retrieved} at a target {\em recall} value (for top-100) with respect to $w$ for both coding schemes $h_w$ and $h_{w,q}$.  Lower is better.}\label{fig_YoutubeRecallvsWT100}
\end{figure}

Figure~\ref{fig_YoutubeRecallvsWT100} presents the  results on {\em Youtube} for $T=100$ and  target recalls from 0.1 to 0.99.  In every panel, we set a target recall threshold. At every bin width $w$, we find the smallest {\em fraction retrieved} over a wide range of LSH parameters, $K$ and $L$. Note that, if the target recall is  high (e.g., 0.95), we basically have to effectively lower the target threshold $\rho$, so that we do not have to go down the re-ranked list too far. The plots show that, for high target recalls, we need to use relatively large $w$ (e.g., $w\geq 2\sim3$), and for low target recalls, we should use a relatively small $w$ (e.g., $w=1.5$). \\

\begin{figure}
\begin{center}
\mbox{
\includegraphics[width = 2.7in]{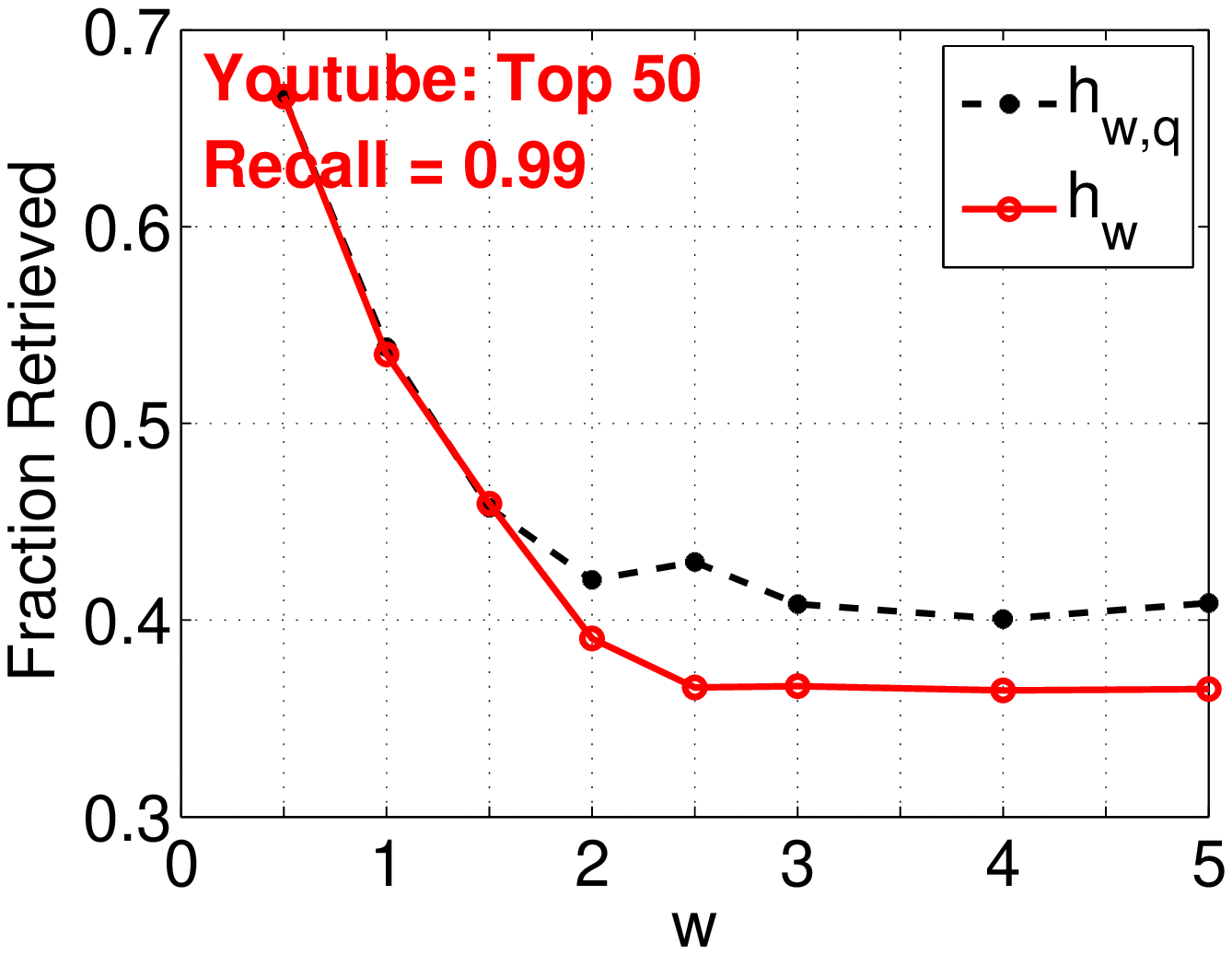}
\includegraphics[width = 2.7in]{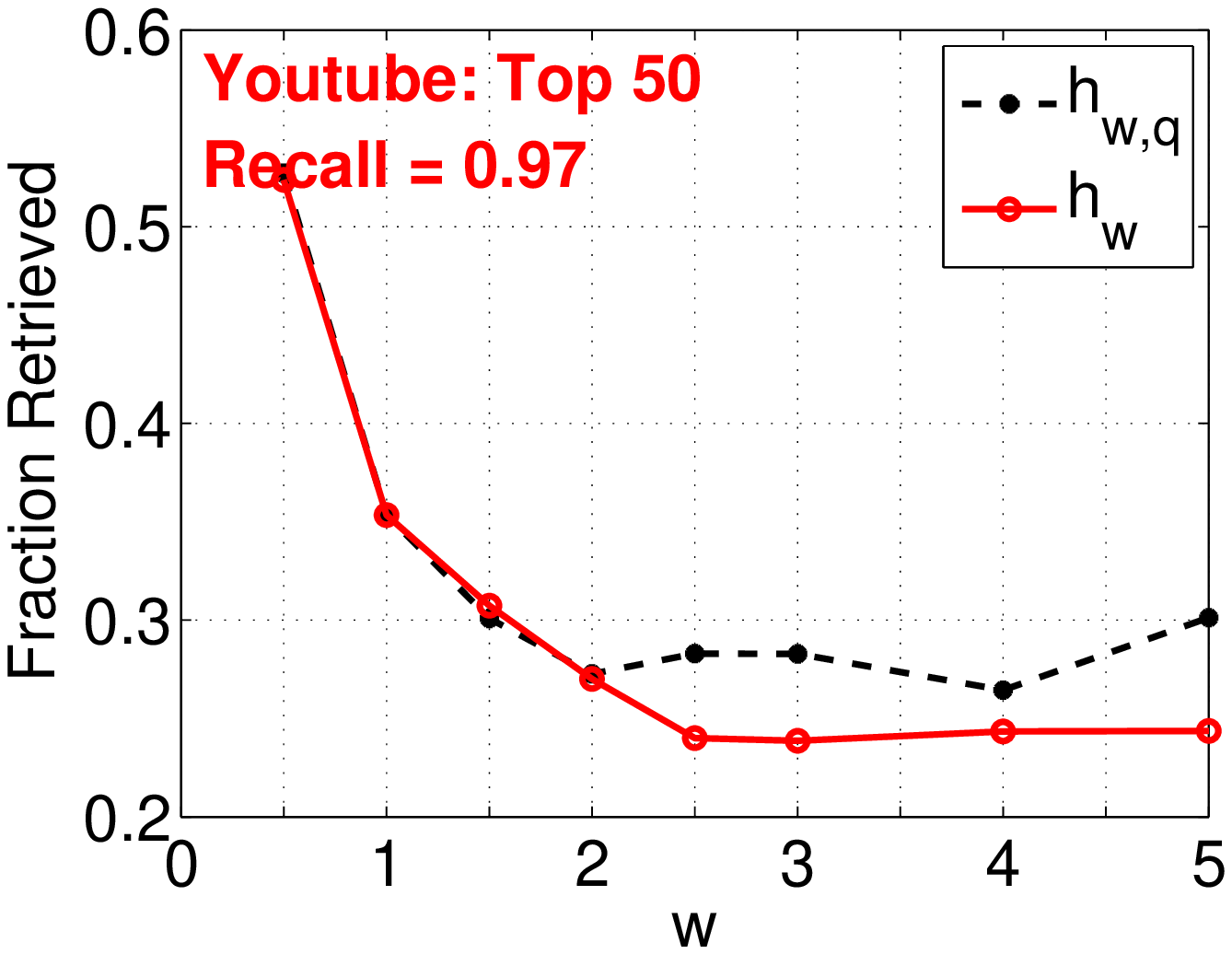}
}
\mbox{
\includegraphics[width = 2.7in]{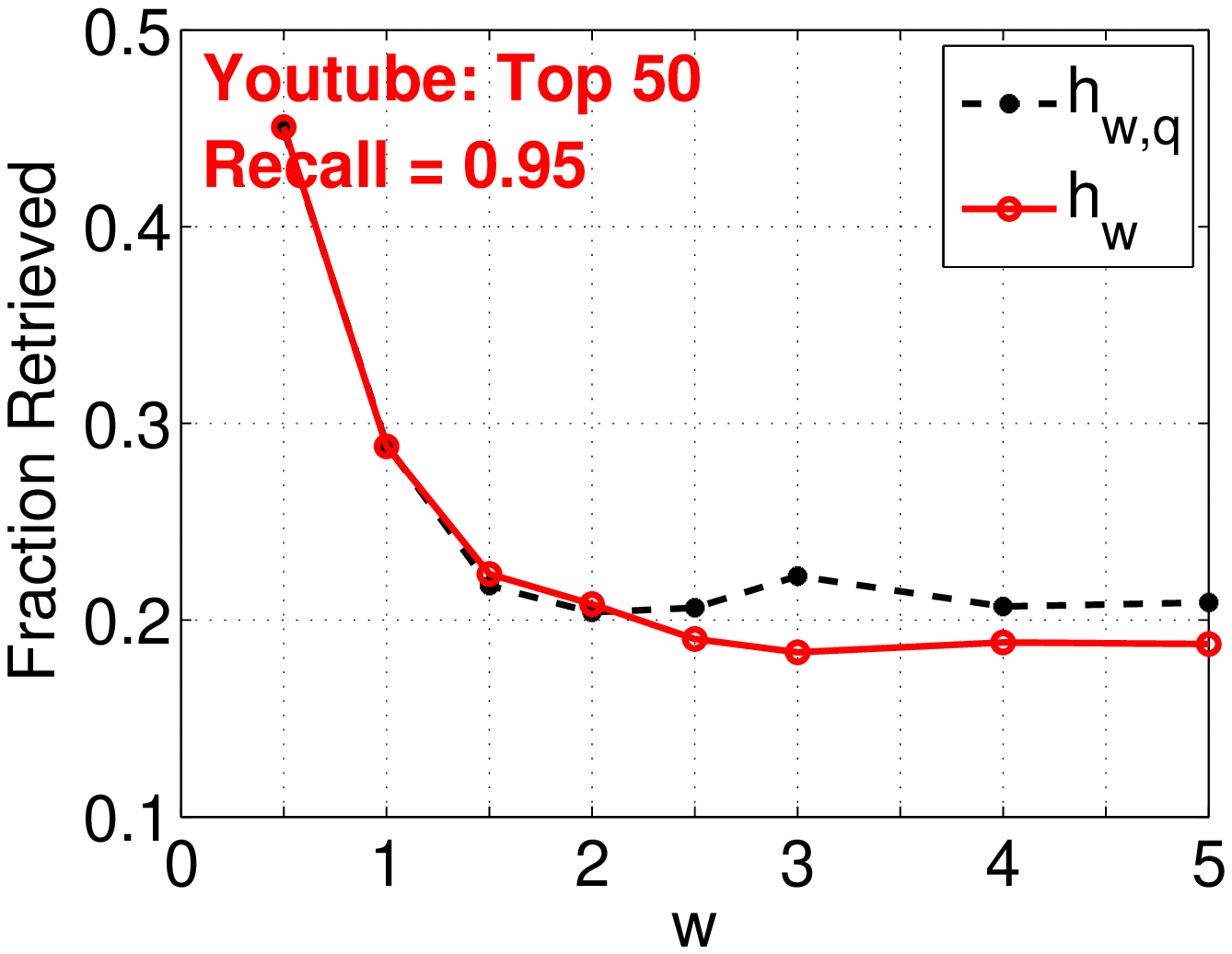}
\includegraphics[width = 2.7in]{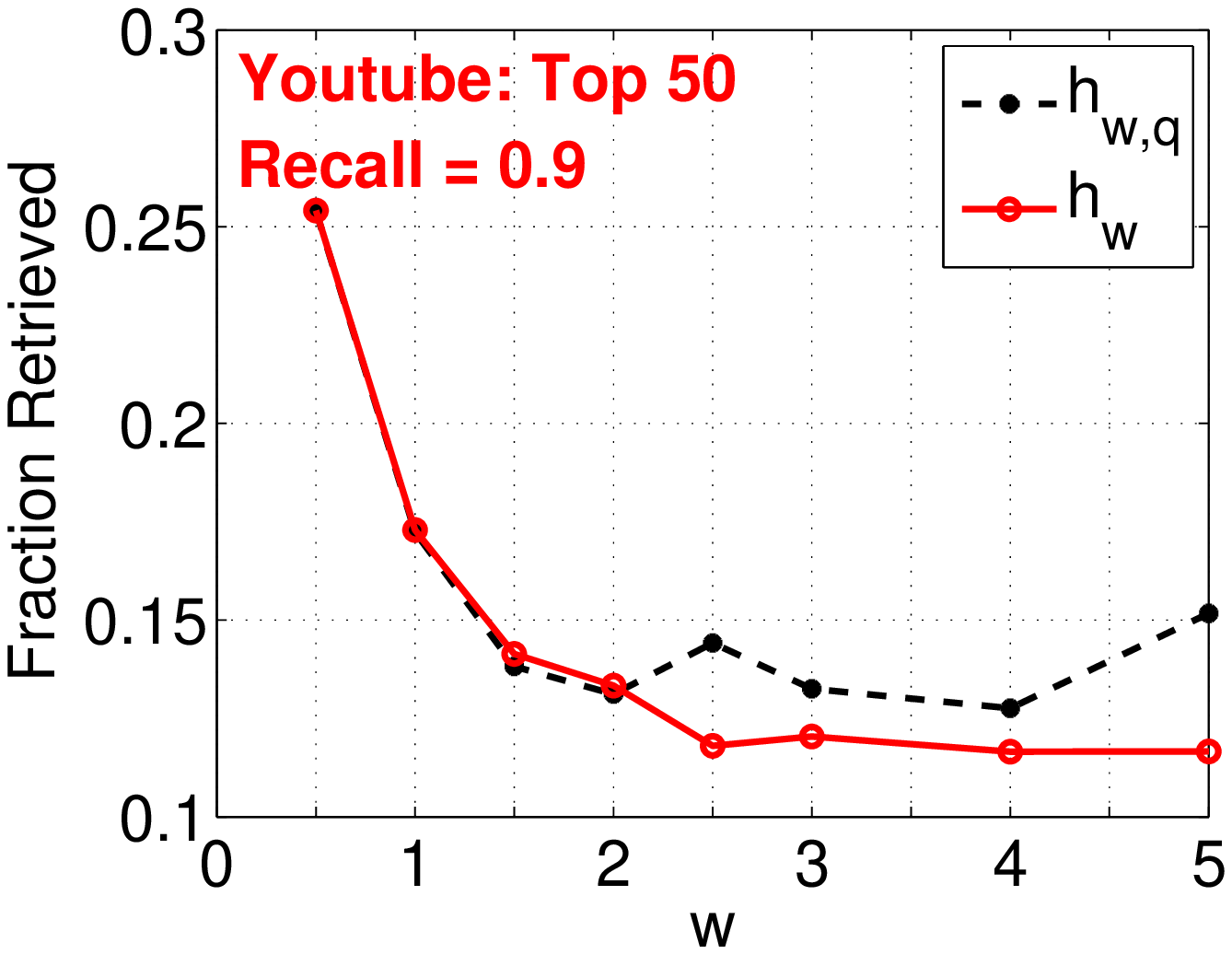}
}

\mbox{
\includegraphics[width = 2.7in]{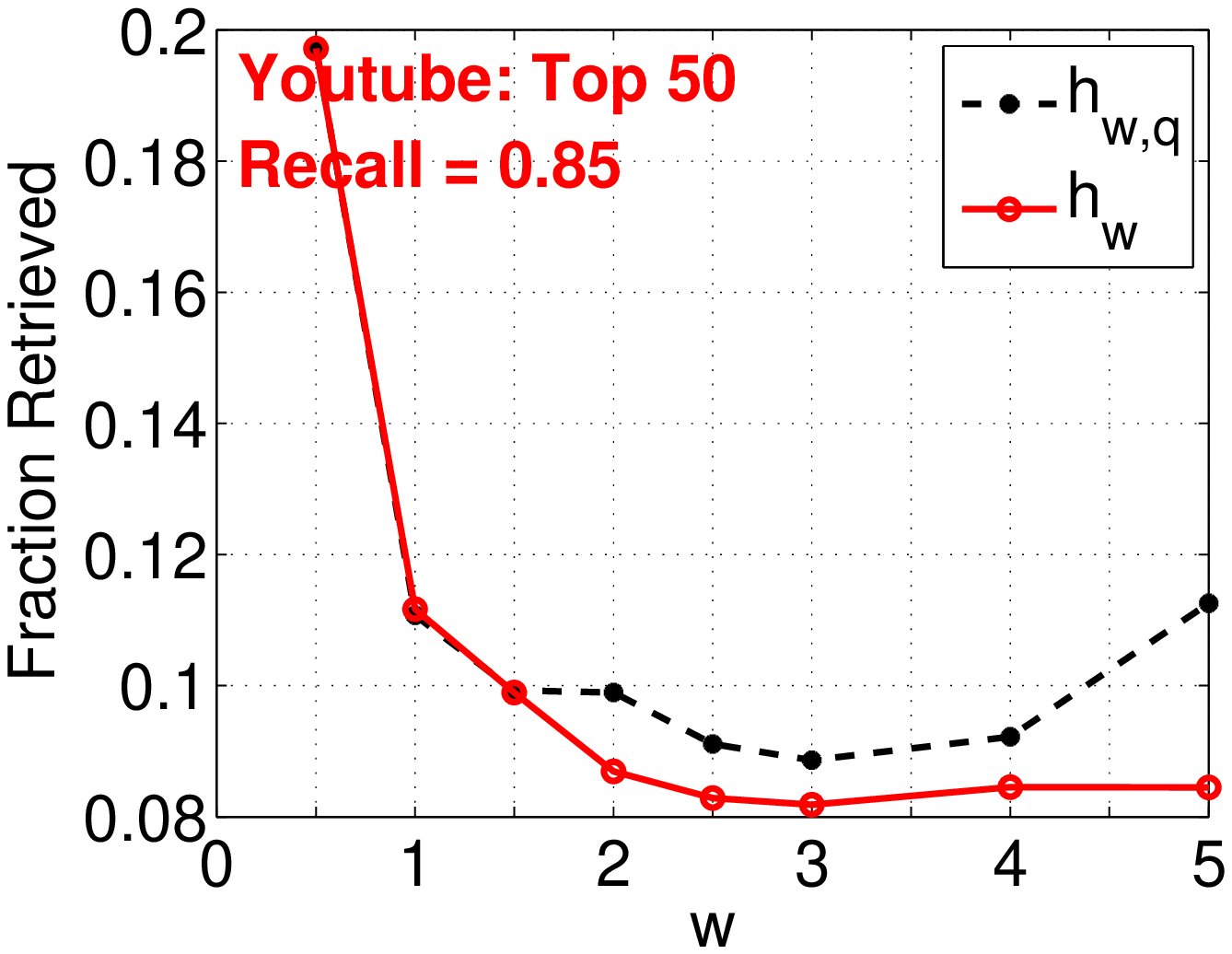}
\includegraphics[width = 2.7in]{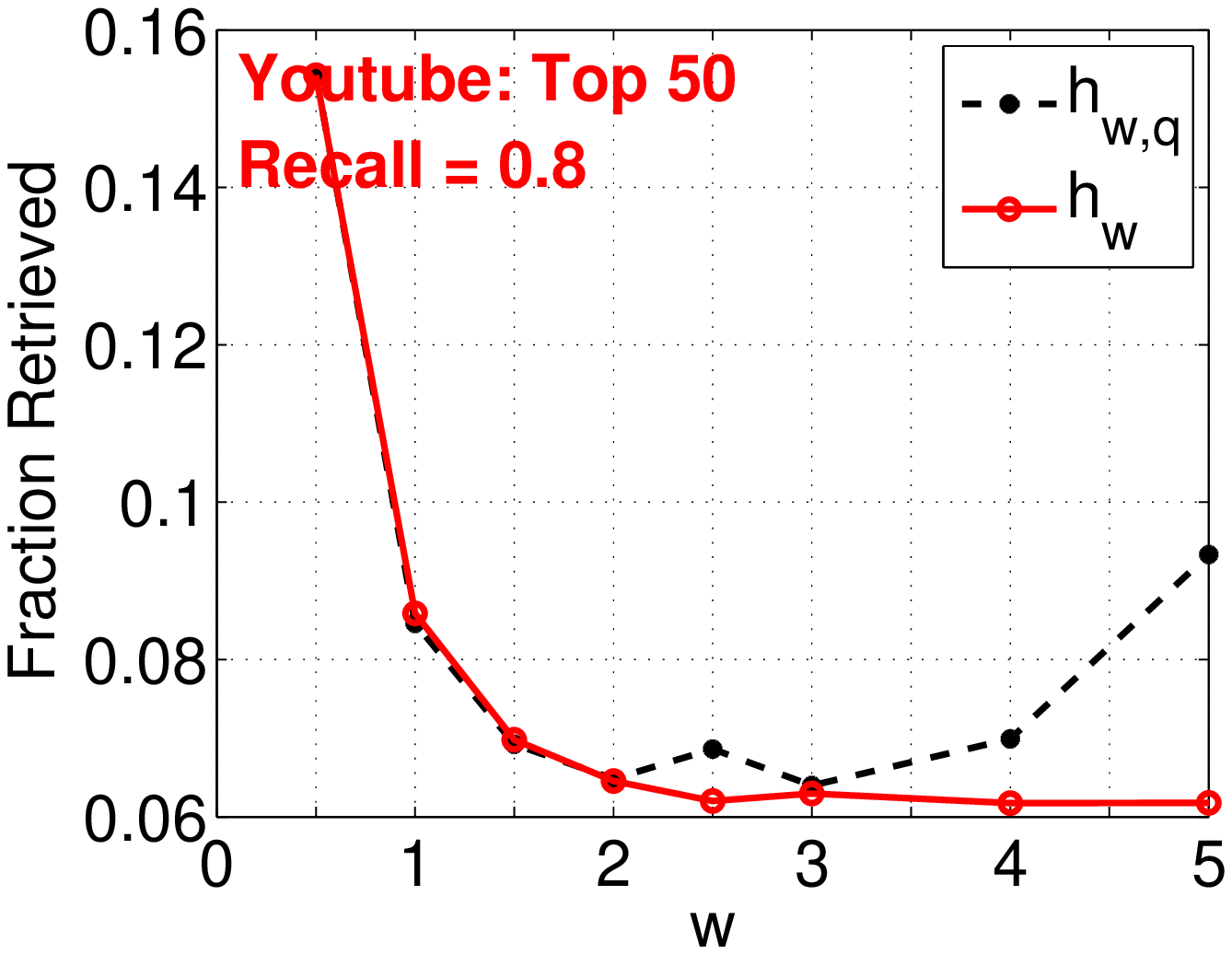}
}

\mbox{
\includegraphics[width = 2.7in]{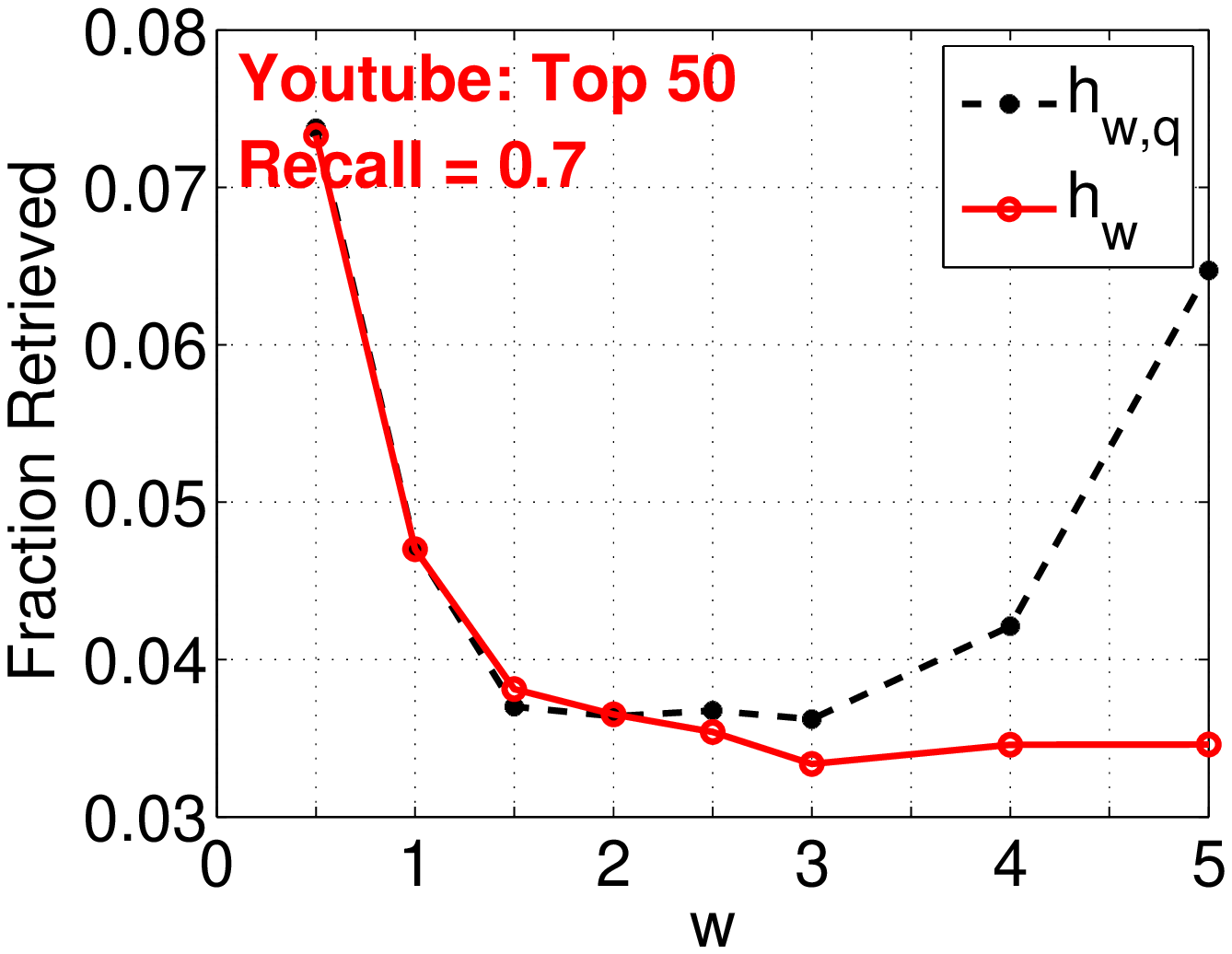}
\includegraphics[width = 2.7in]{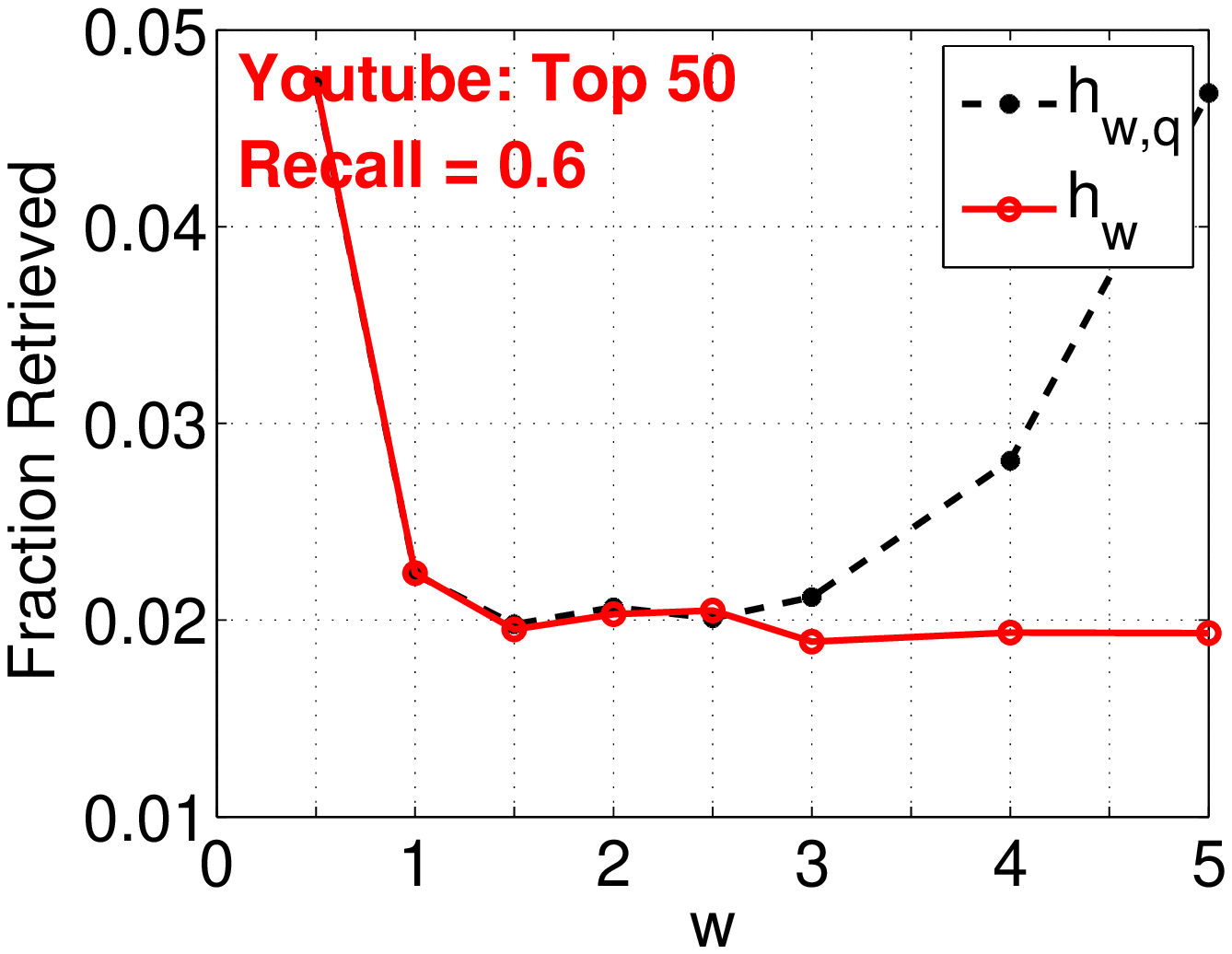}
}

\end{center}
\vspace{-.2in}
\caption{ \textbf{Youtube Top 50} . In each panel, we plot the optimal {\em fraction retrieved} at a target {\em recall} value (for top-50) with respect to $w$ for both coding schemes $h_w$ and $h_{w,q}$. }\label{fig_YoutubeRecallvsWT50}
\end{figure}

\begin{figure}
\begin{center}
\mbox{
\includegraphics[width = 2.7in]{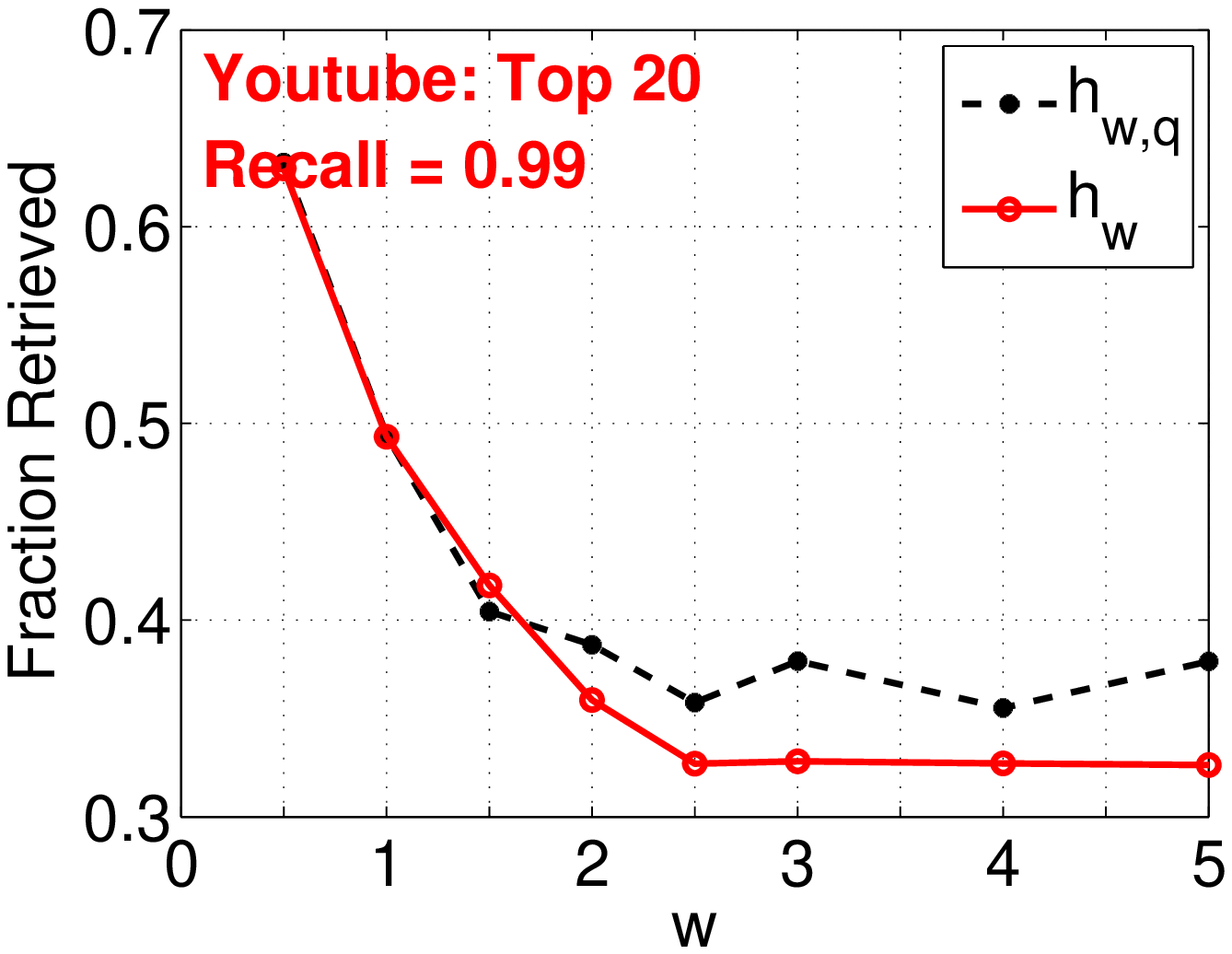}
\includegraphics[width = 2.7in]{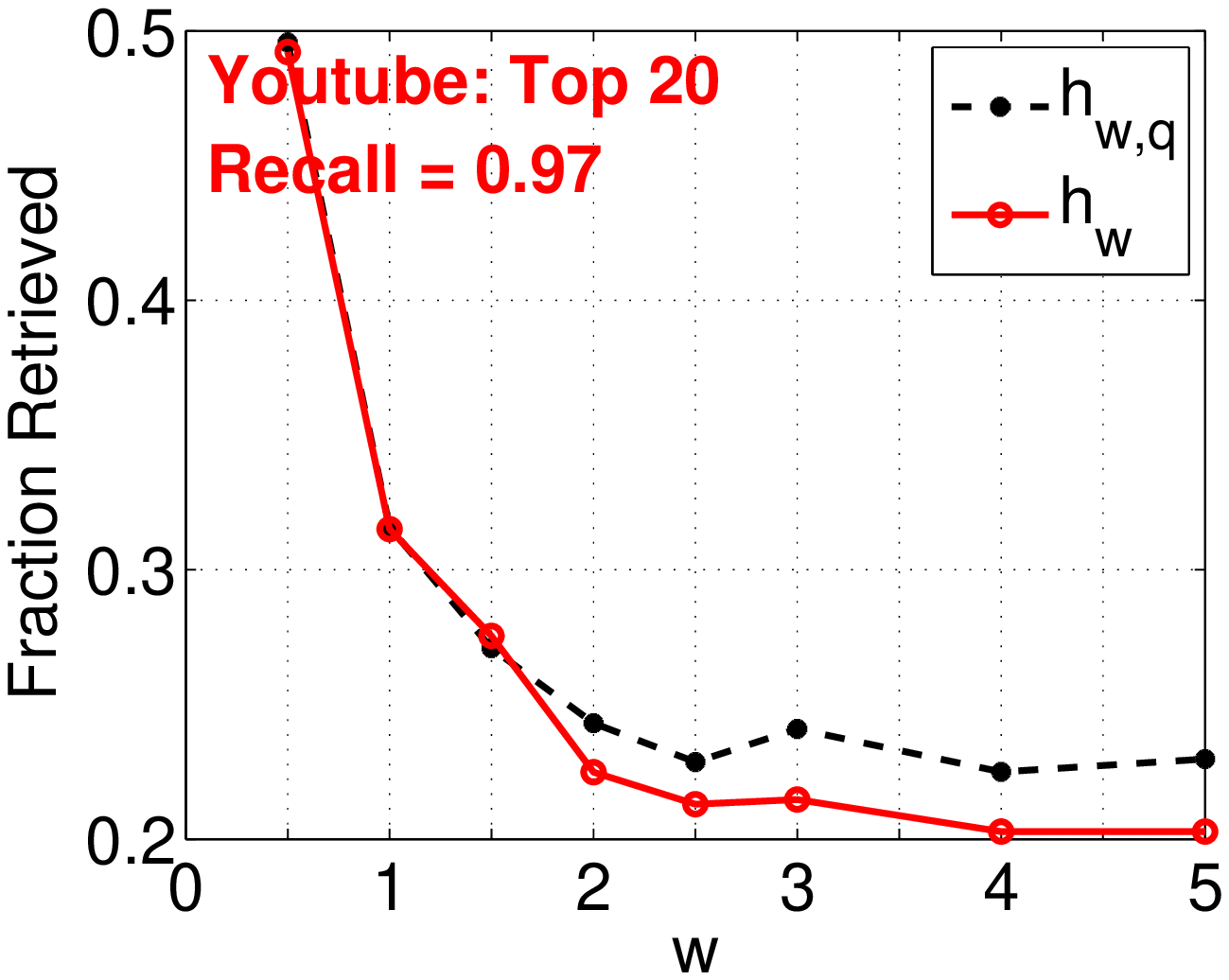}
}
\mbox{
\includegraphics[width = 2.7in]{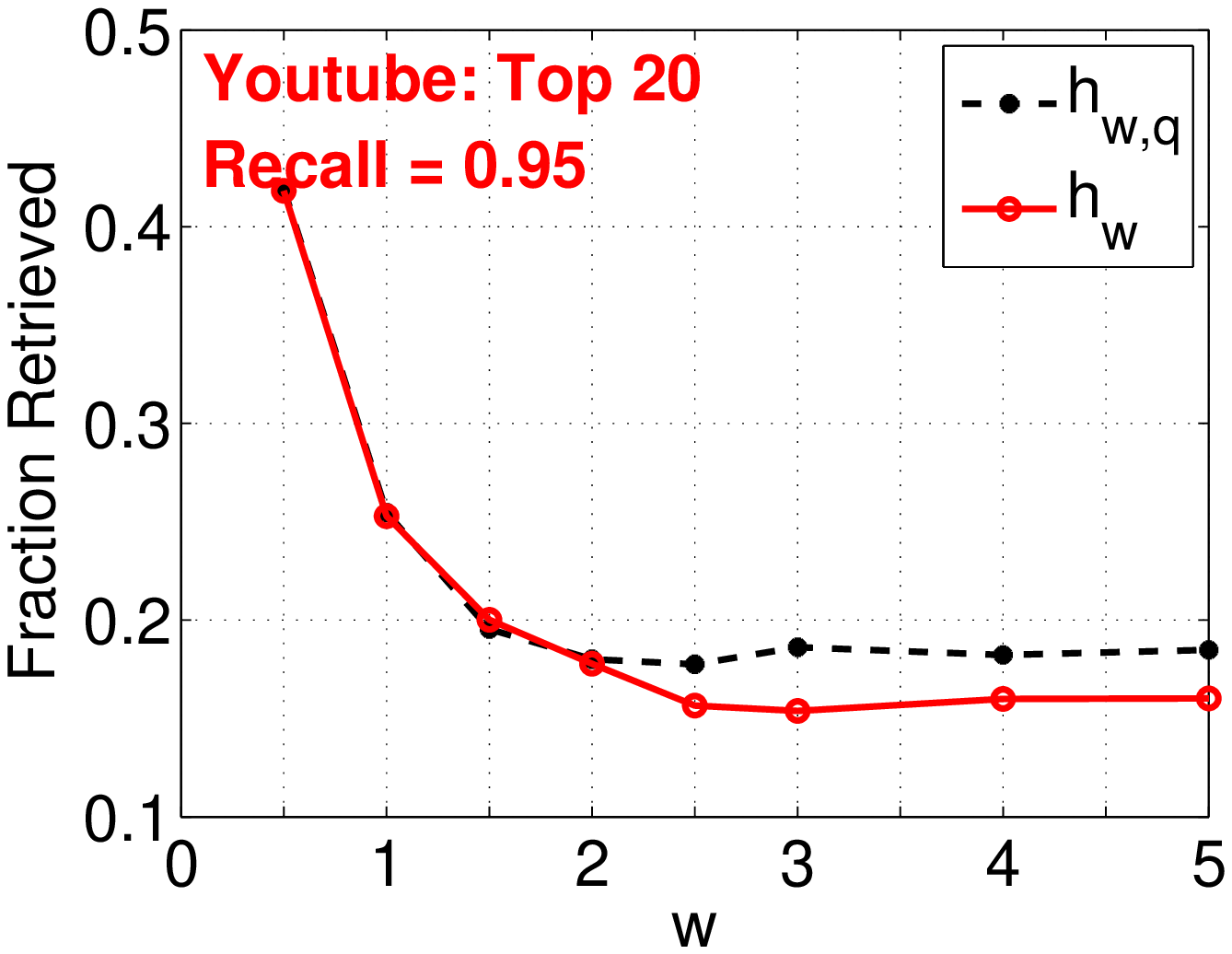}
\includegraphics[width = 2.7in]{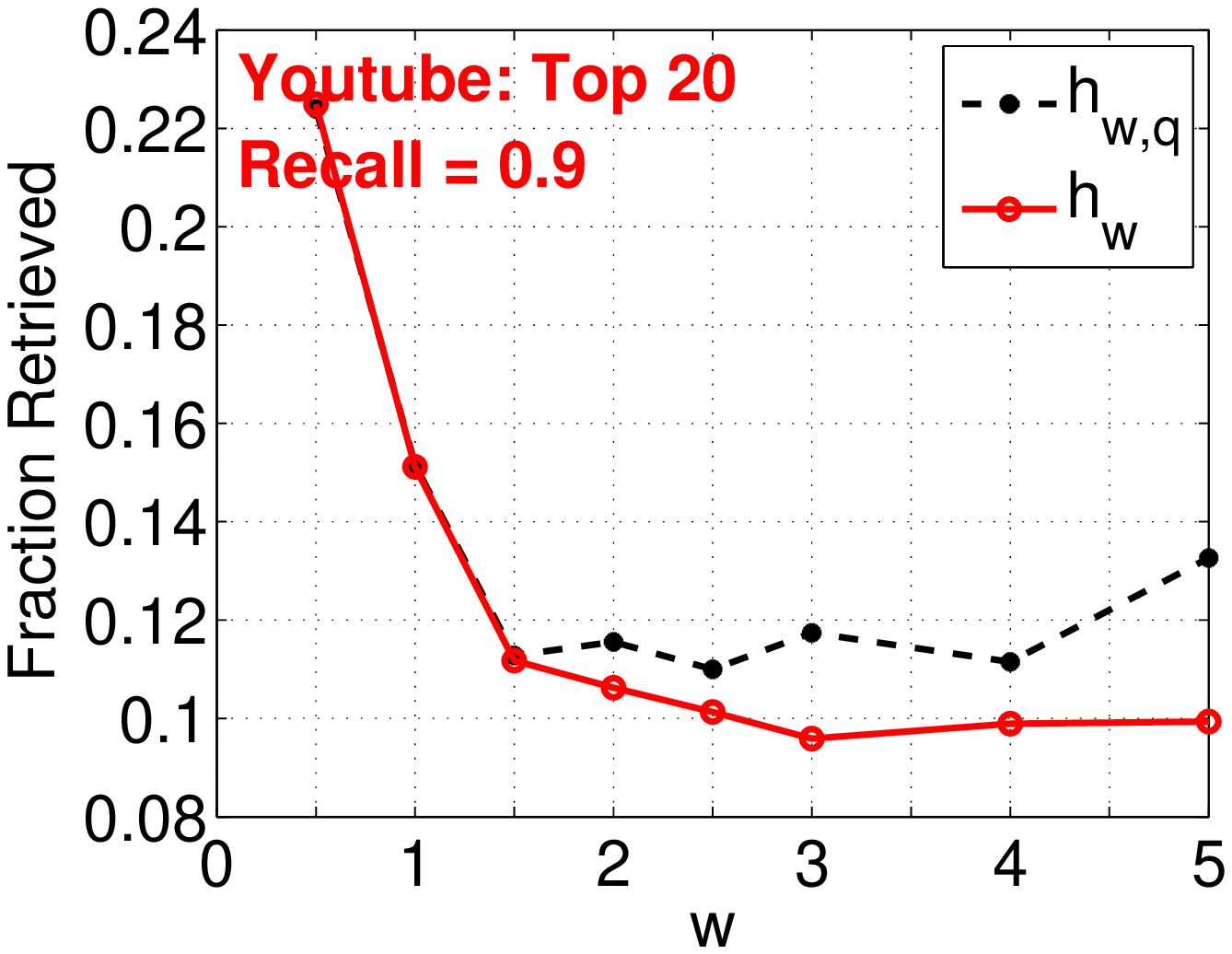}
}

\mbox{
\includegraphics[width = 2.7in]{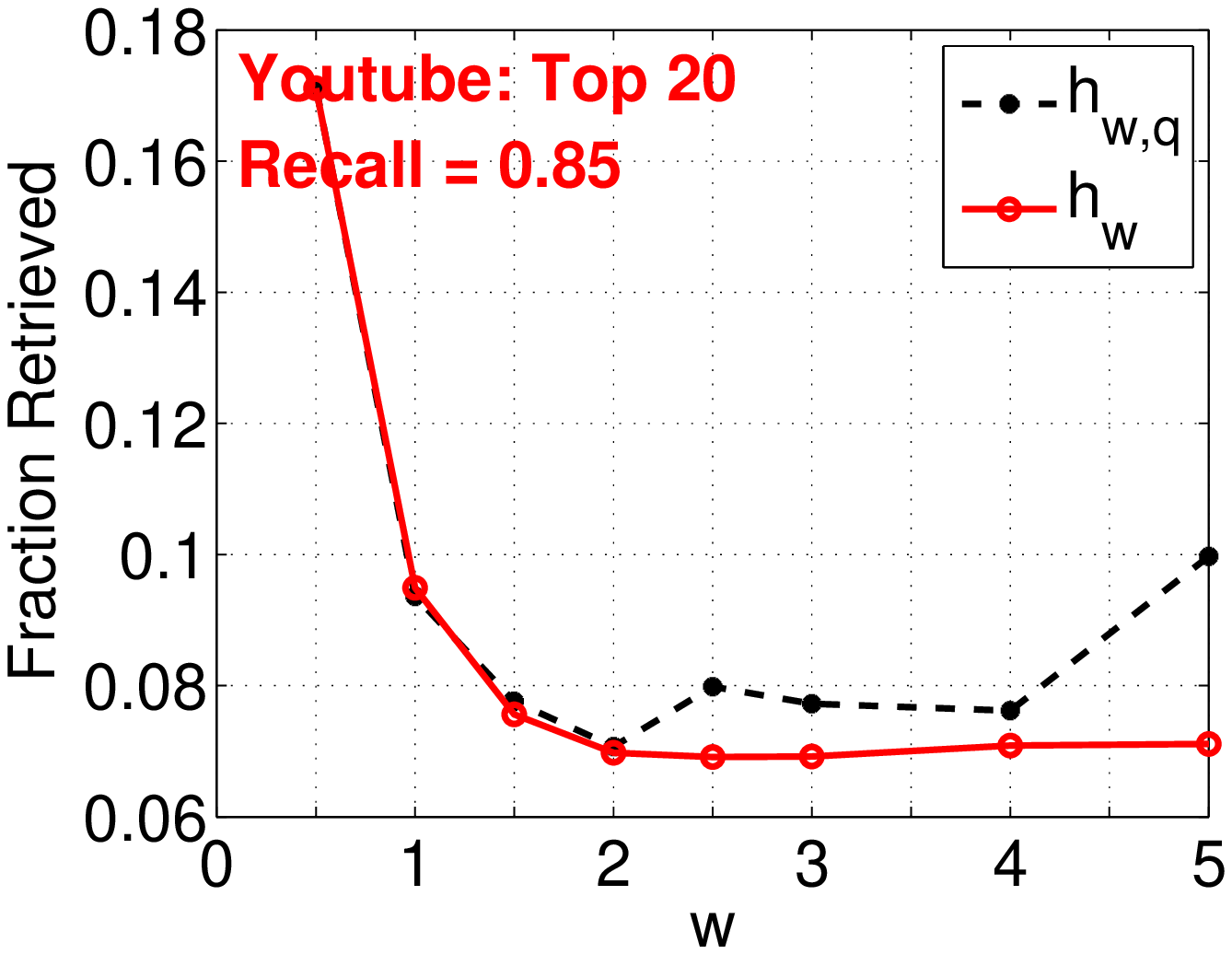}
\includegraphics[width = 2.7in]{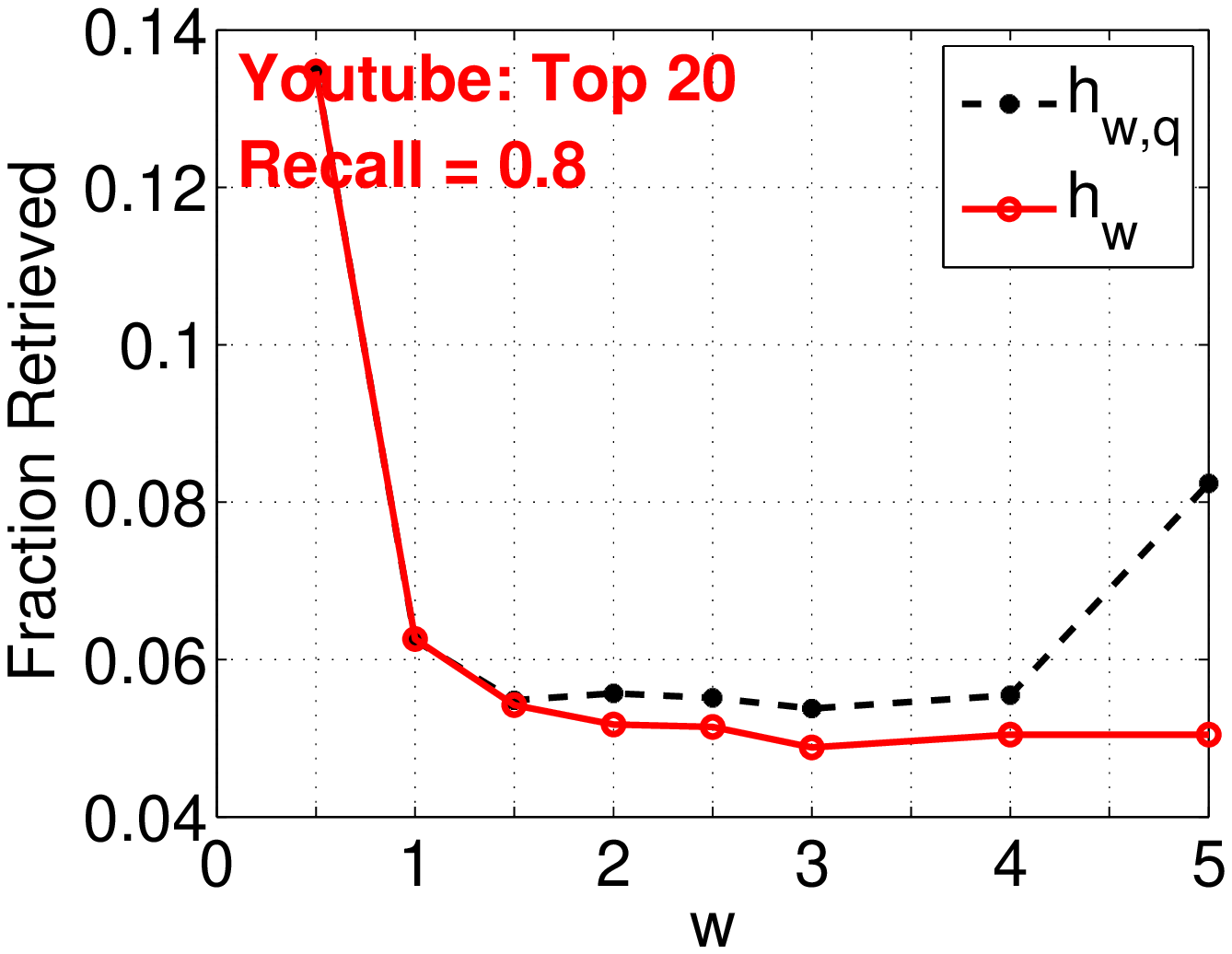}
}

\mbox{
\includegraphics[width = 2.7in]{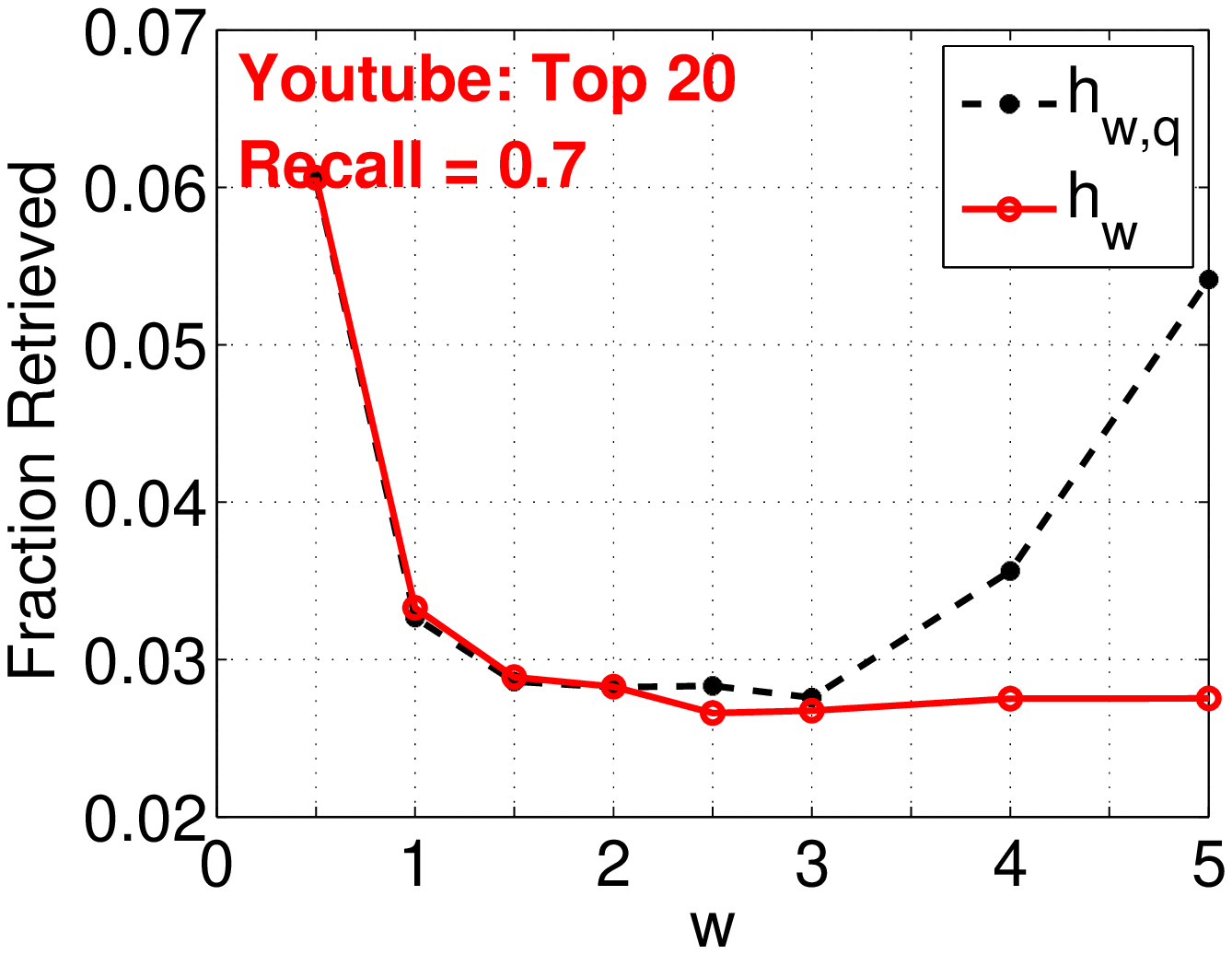}
\includegraphics[width = 2.7in]{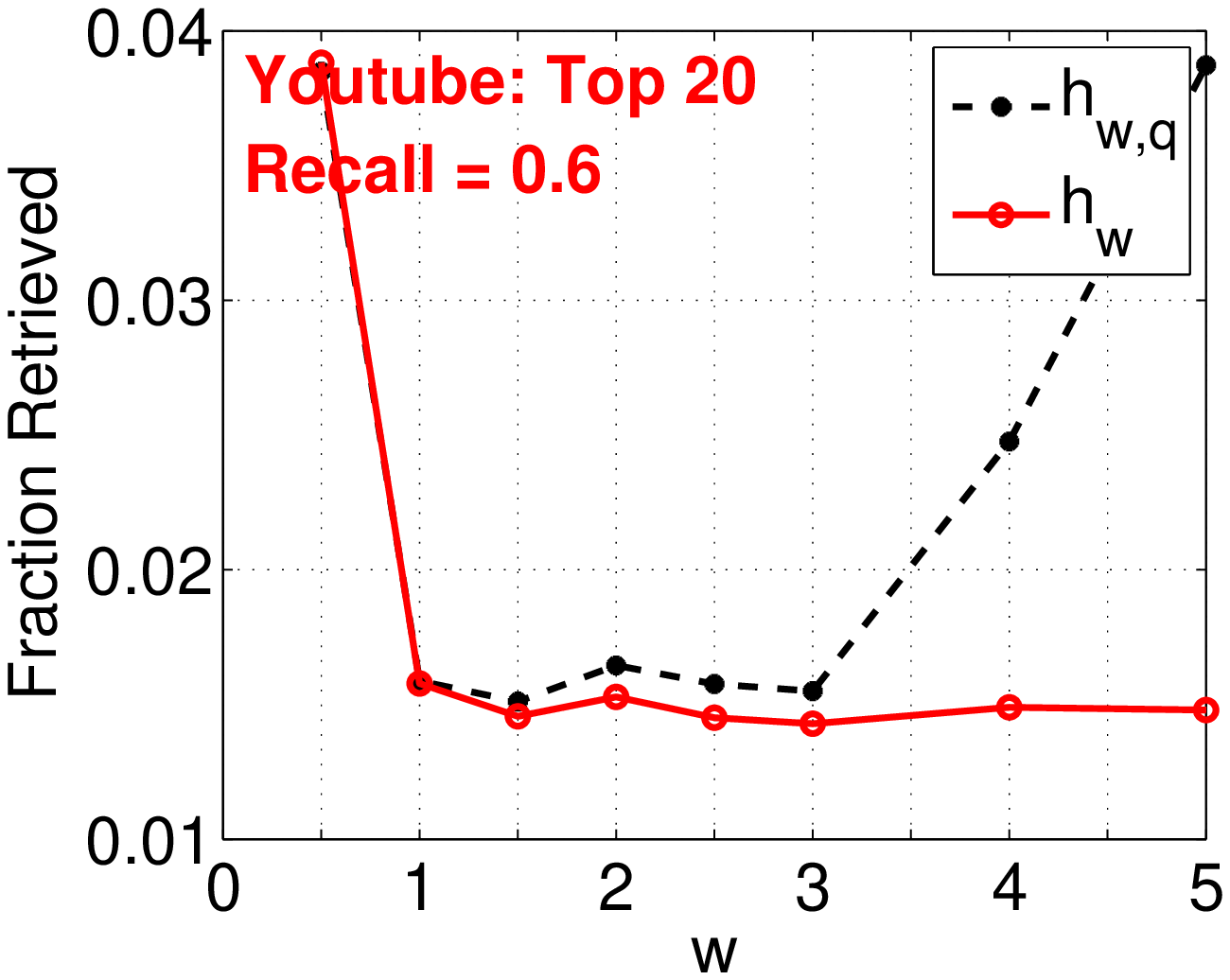}
}

\end{center}
\vspace{-.2in}
\caption{ \textbf{Youtube Top 20} . In each panel, we plot the optimal {\em fraction retrieved} at a target {\em recall} value (for top-20) with respect to $w$ for both coding schemes $h_w$ and $h_{w,q}$. }\label{fig_YoutubeRecallvsWT20}
\end{figure}

\begin{figure}
\begin{center}
\mbox{
\includegraphics[width = 2.7in]{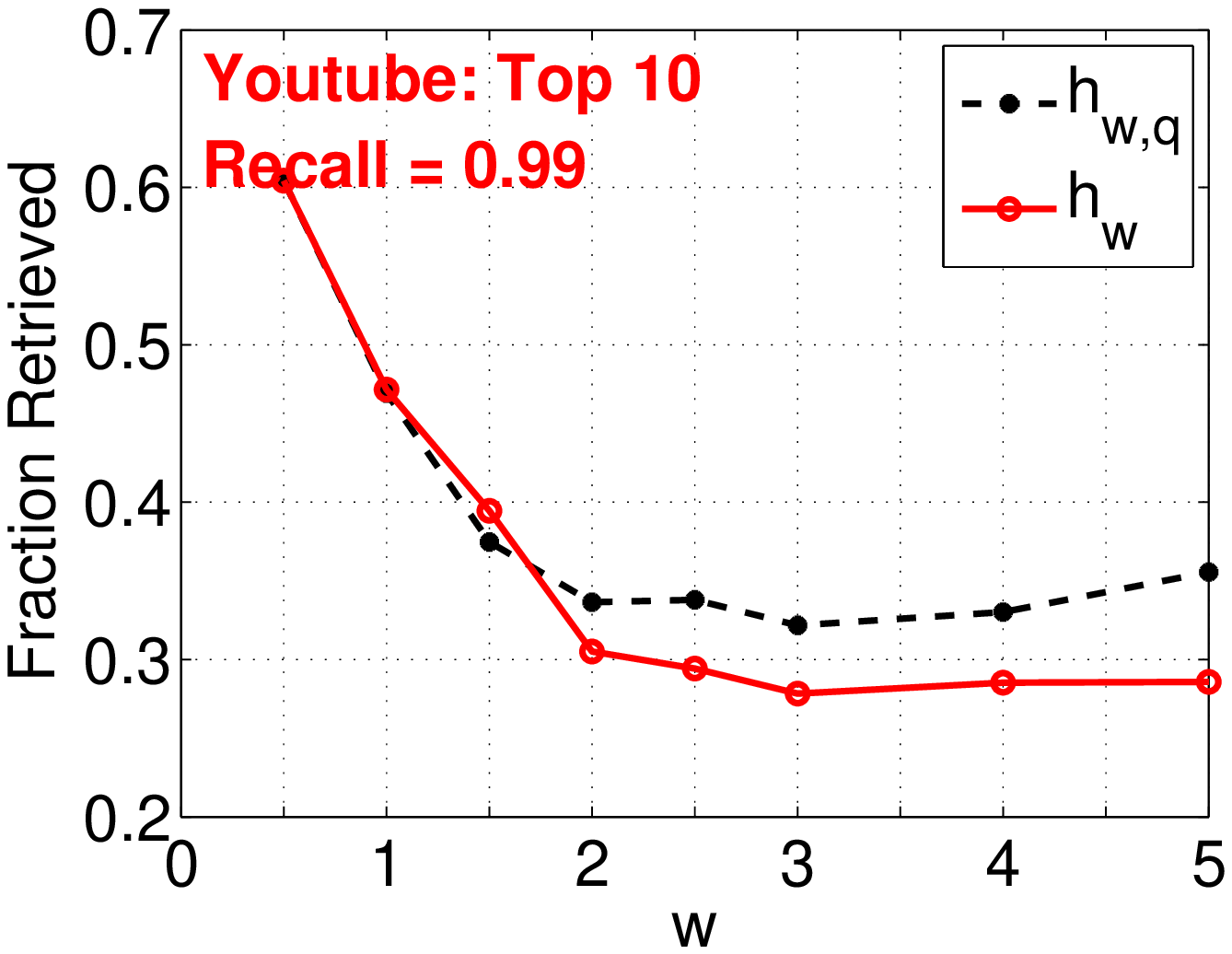}
\includegraphics[width = 2.7in]{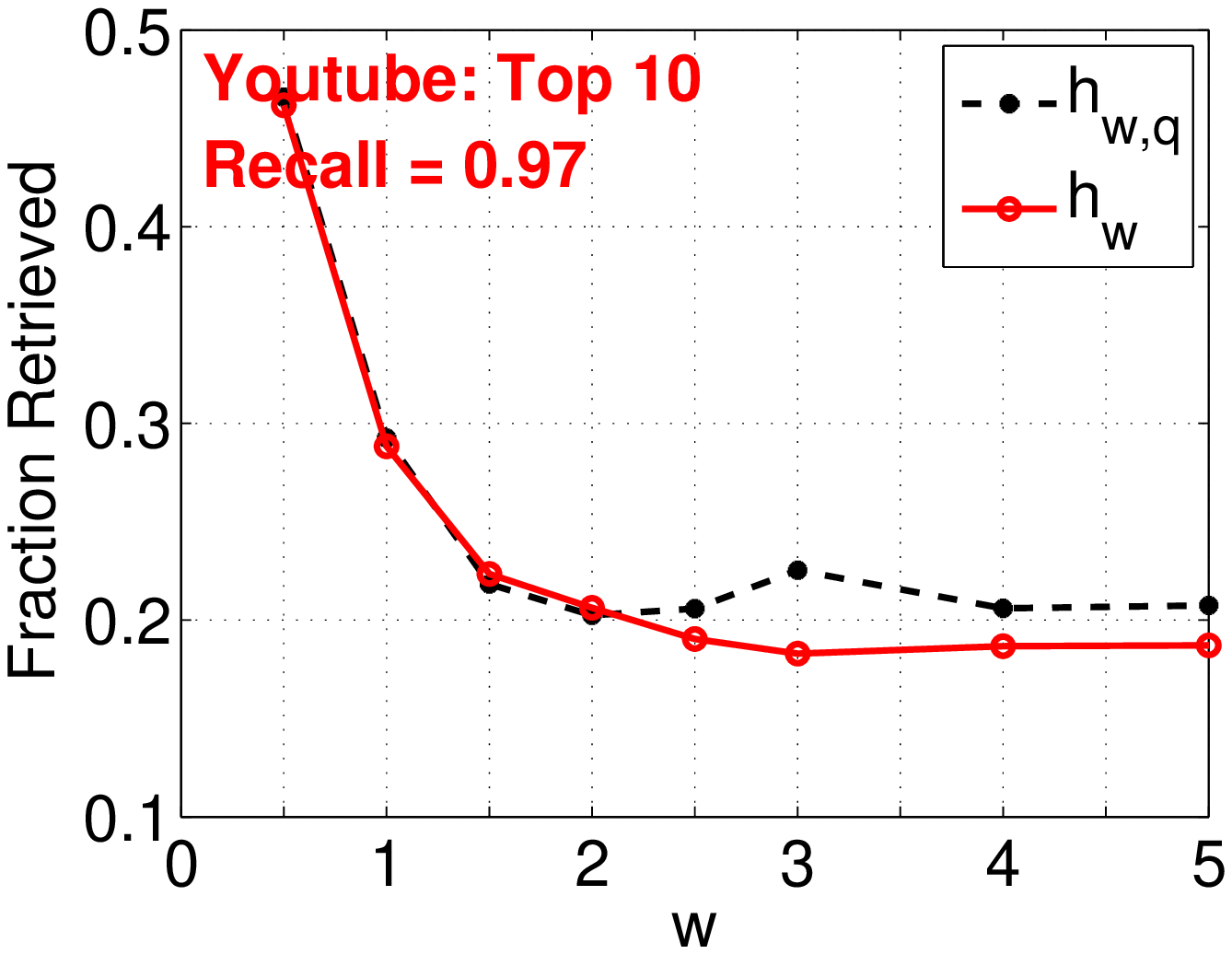}
}
\mbox{
\includegraphics[width = 2.7in]{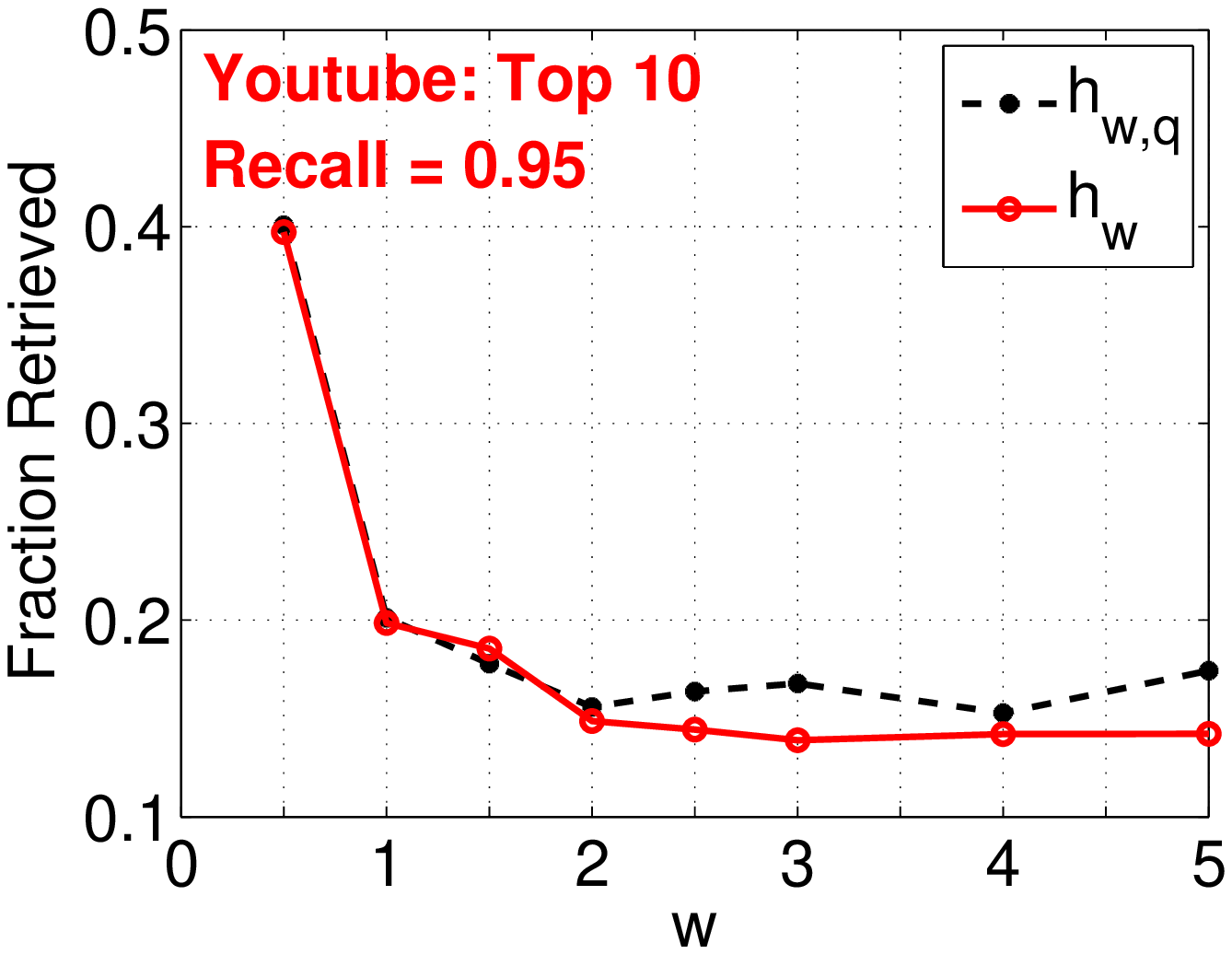}
\includegraphics[width = 2.7in]{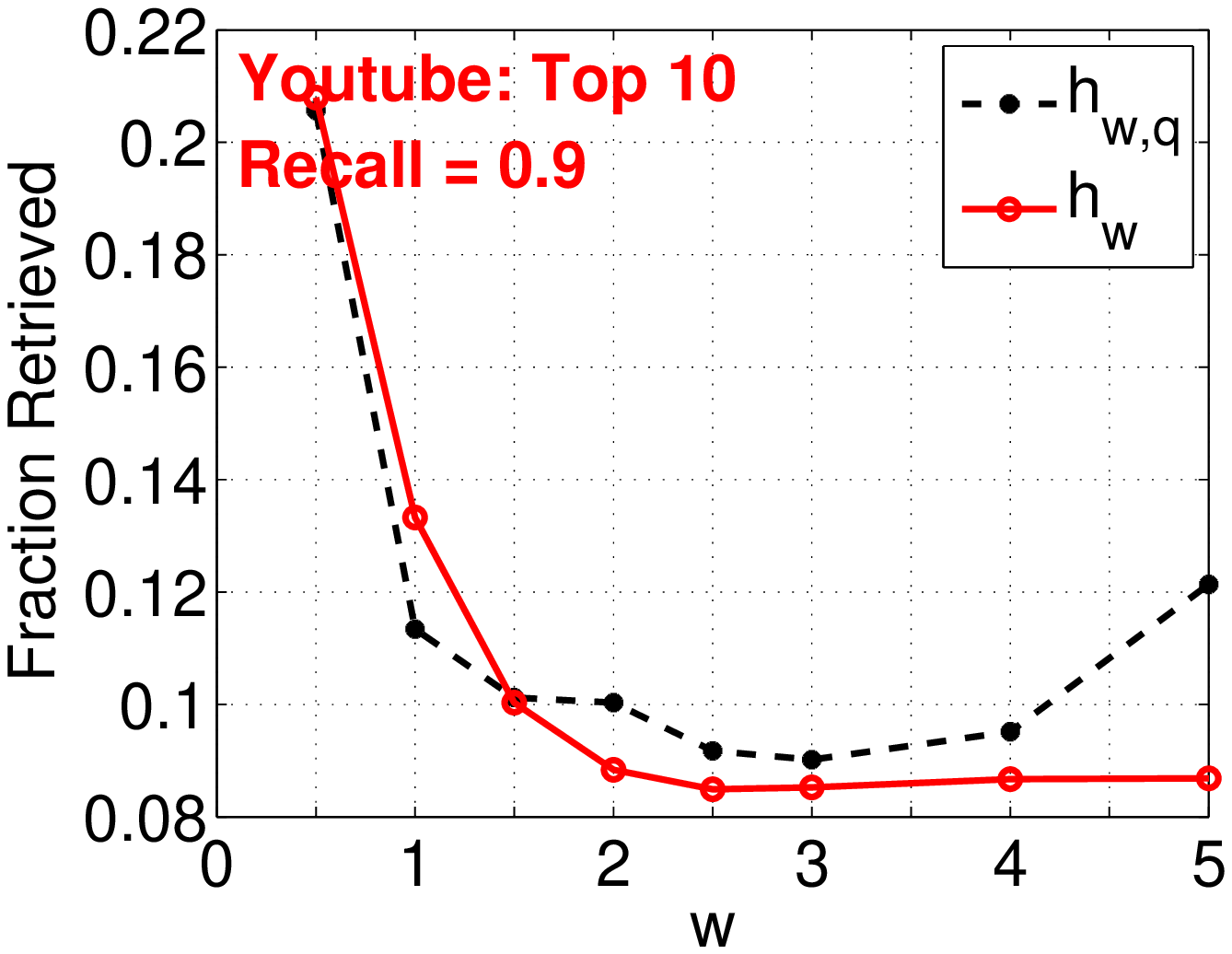}
}

\mbox{
\includegraphics[width = 2.7in]{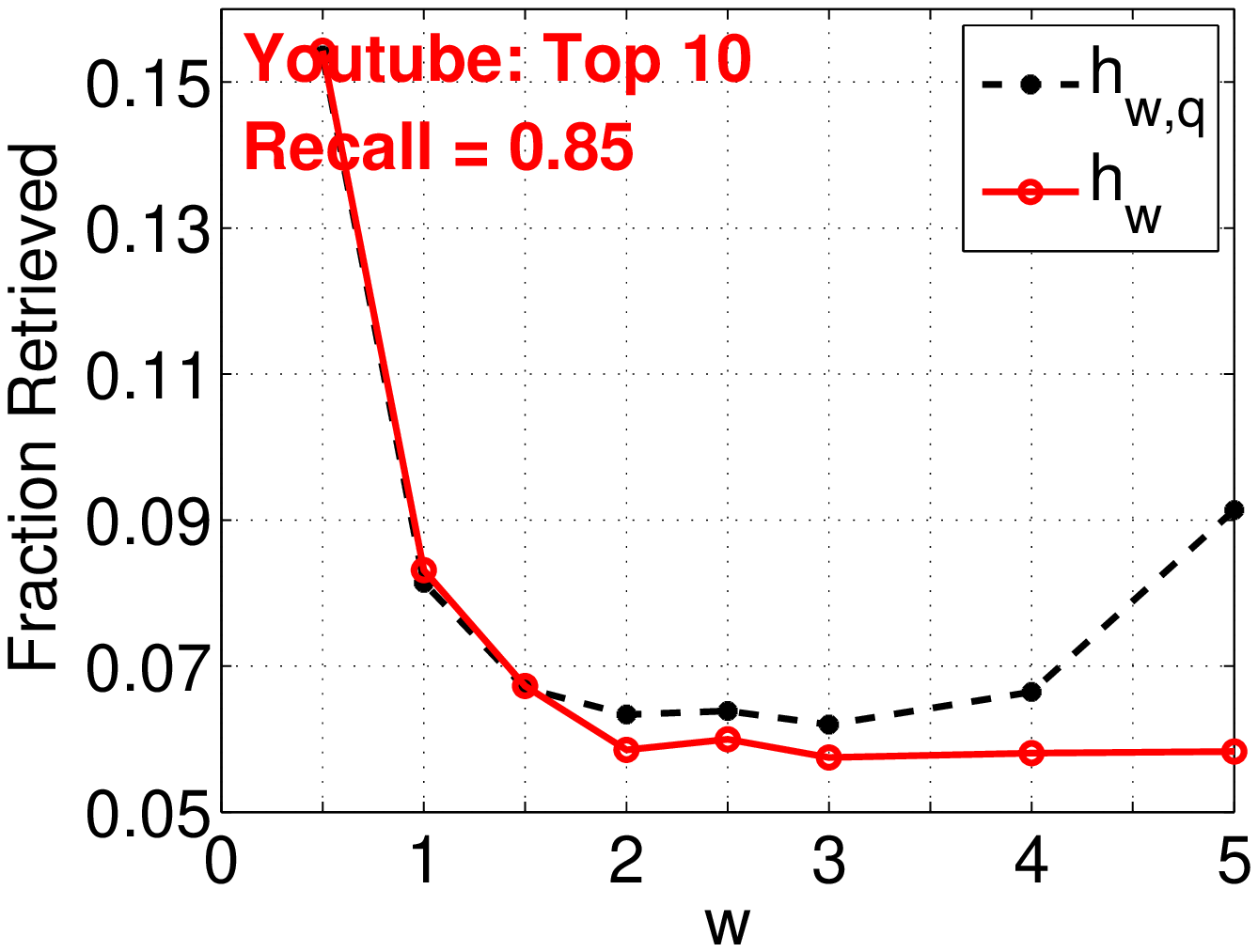}
\includegraphics[width = 2.7in]{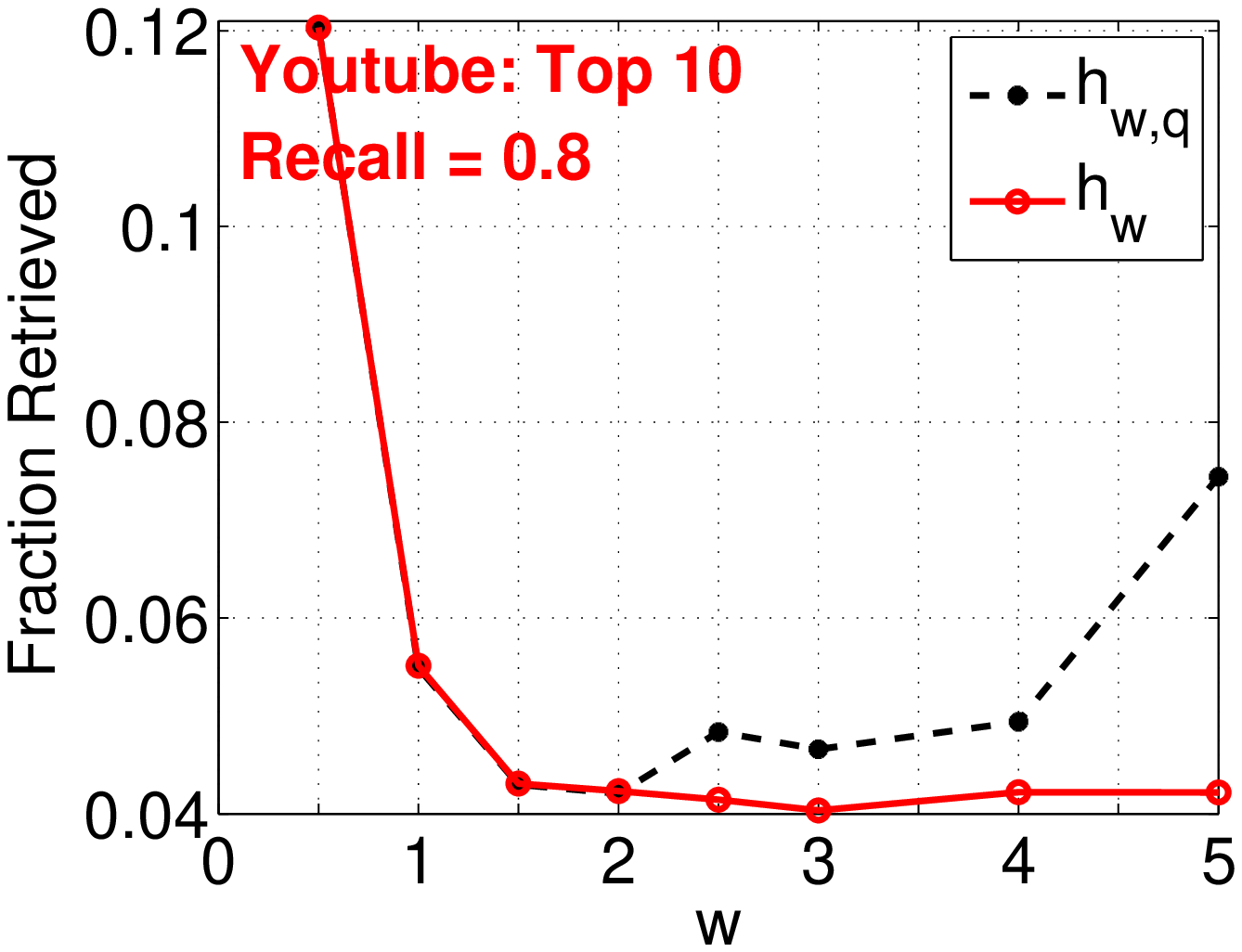}
}

\mbox{
\includegraphics[width = 2.7in]{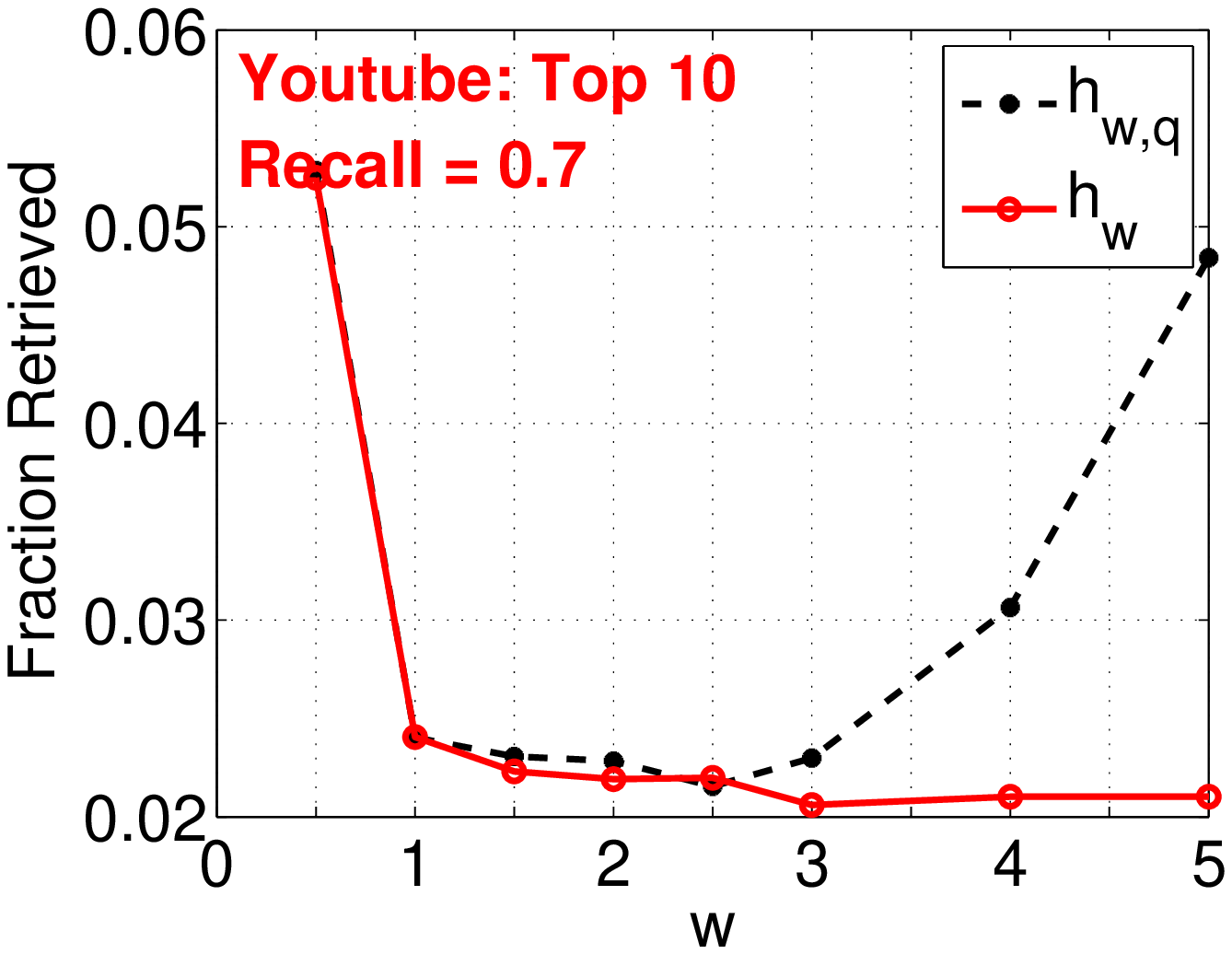}
\includegraphics[width = 2.7in]{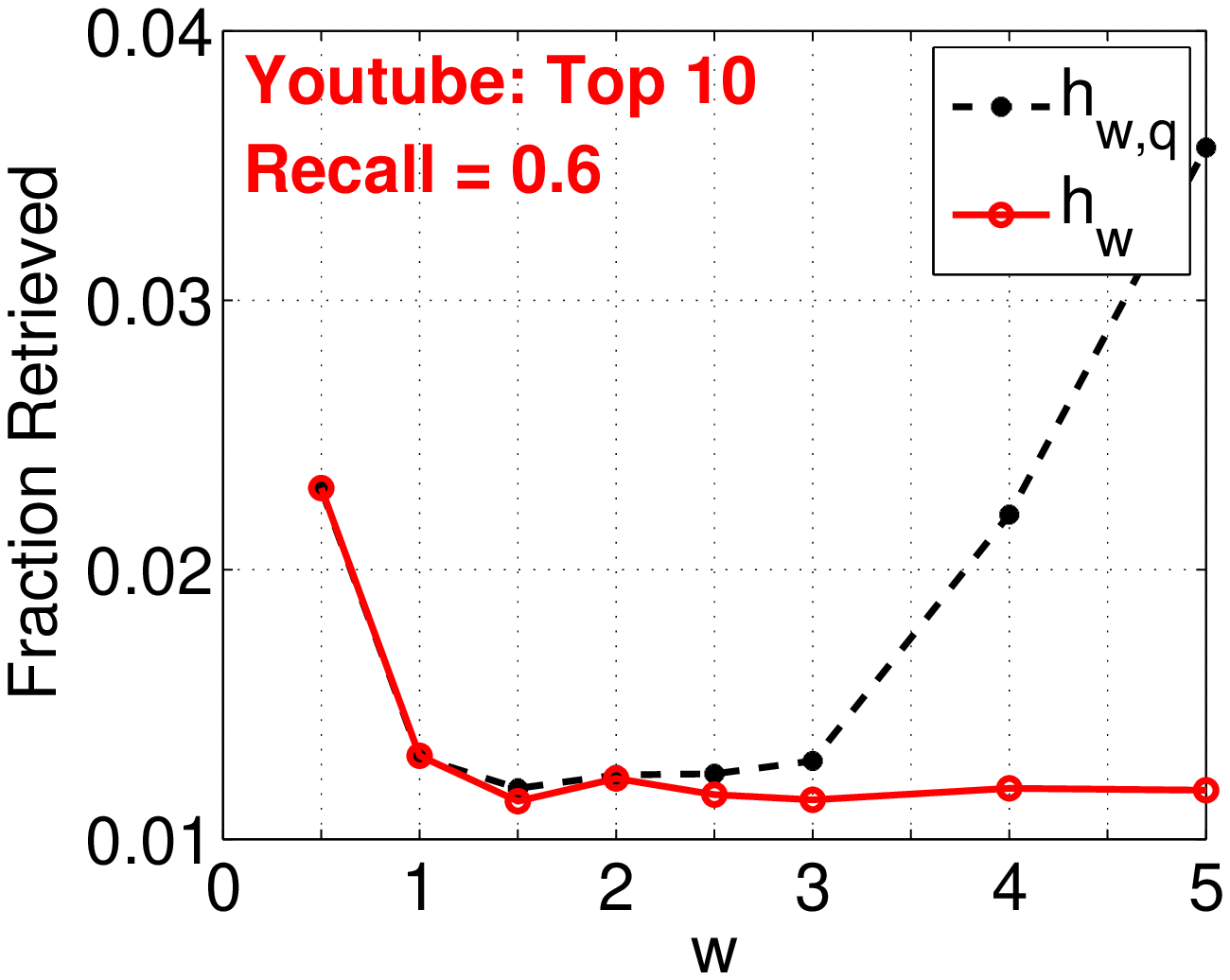}
}

\end{center}
\vspace{-.2in}
\caption{ \textbf{Youtube Top 10} . In each panel, we plot the optimal {\em fraction retrieved} at a target {\em recall} value (for top-10) with respect to $w$ for both coding schemes $h_w$ and $h_{w,q}$. }\label{fig_YoutubeRecallvsWT10}
\end{figure}

\begin{figure}
\begin{center}
\mbox{
\includegraphics[width = 2.7in]{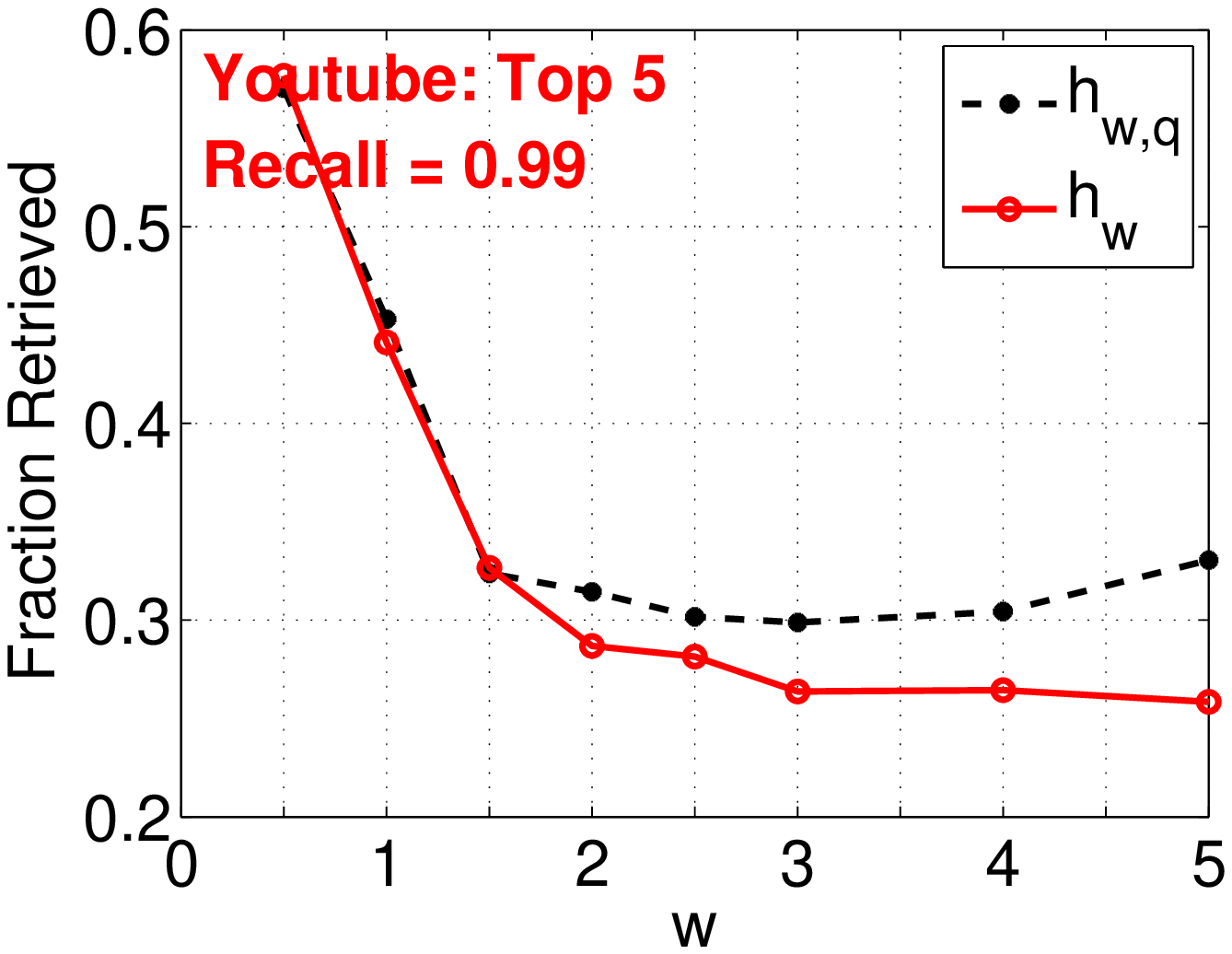}
\includegraphics[width = 2.7in]{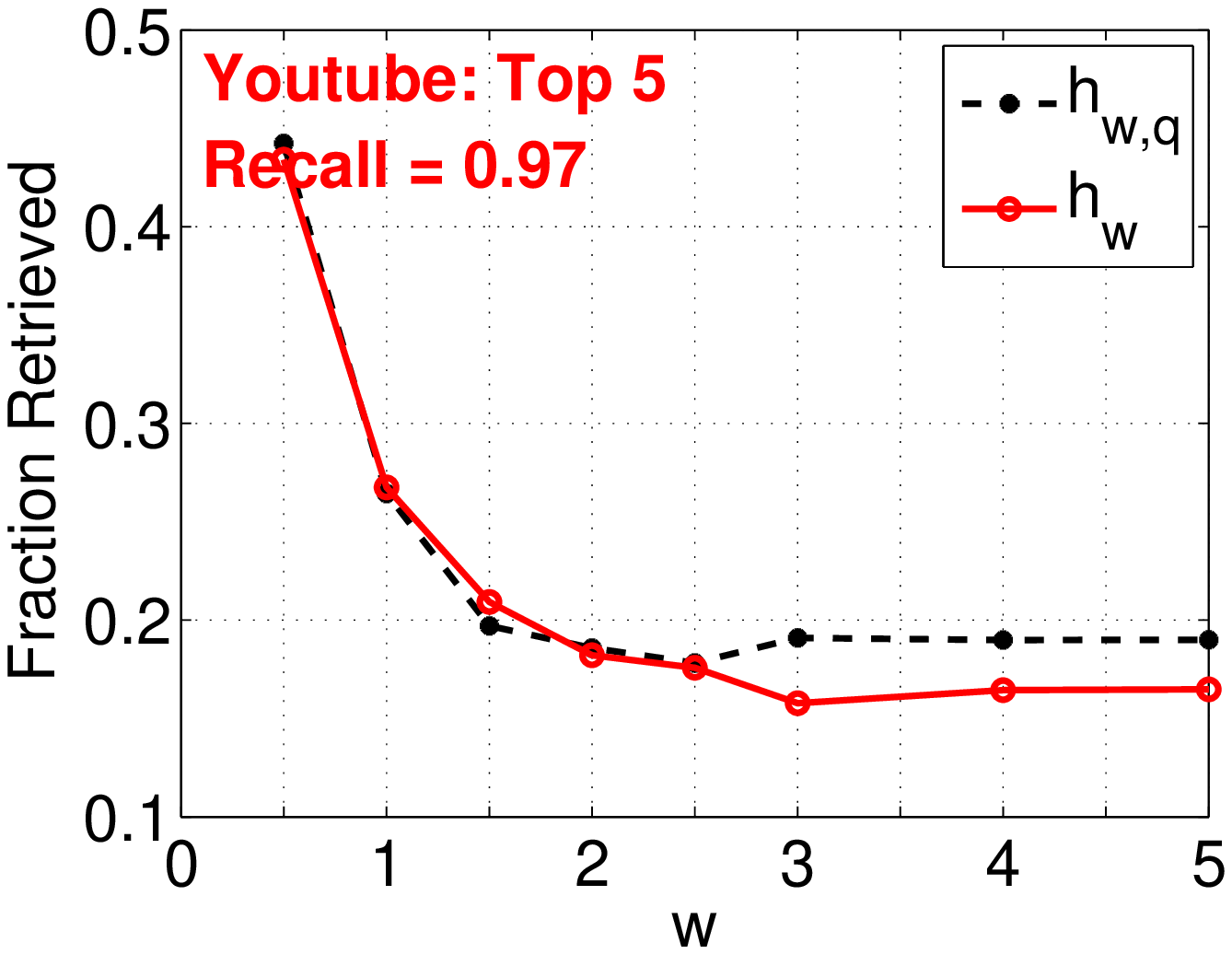}
}
\mbox{
\includegraphics[width = 2.7in]{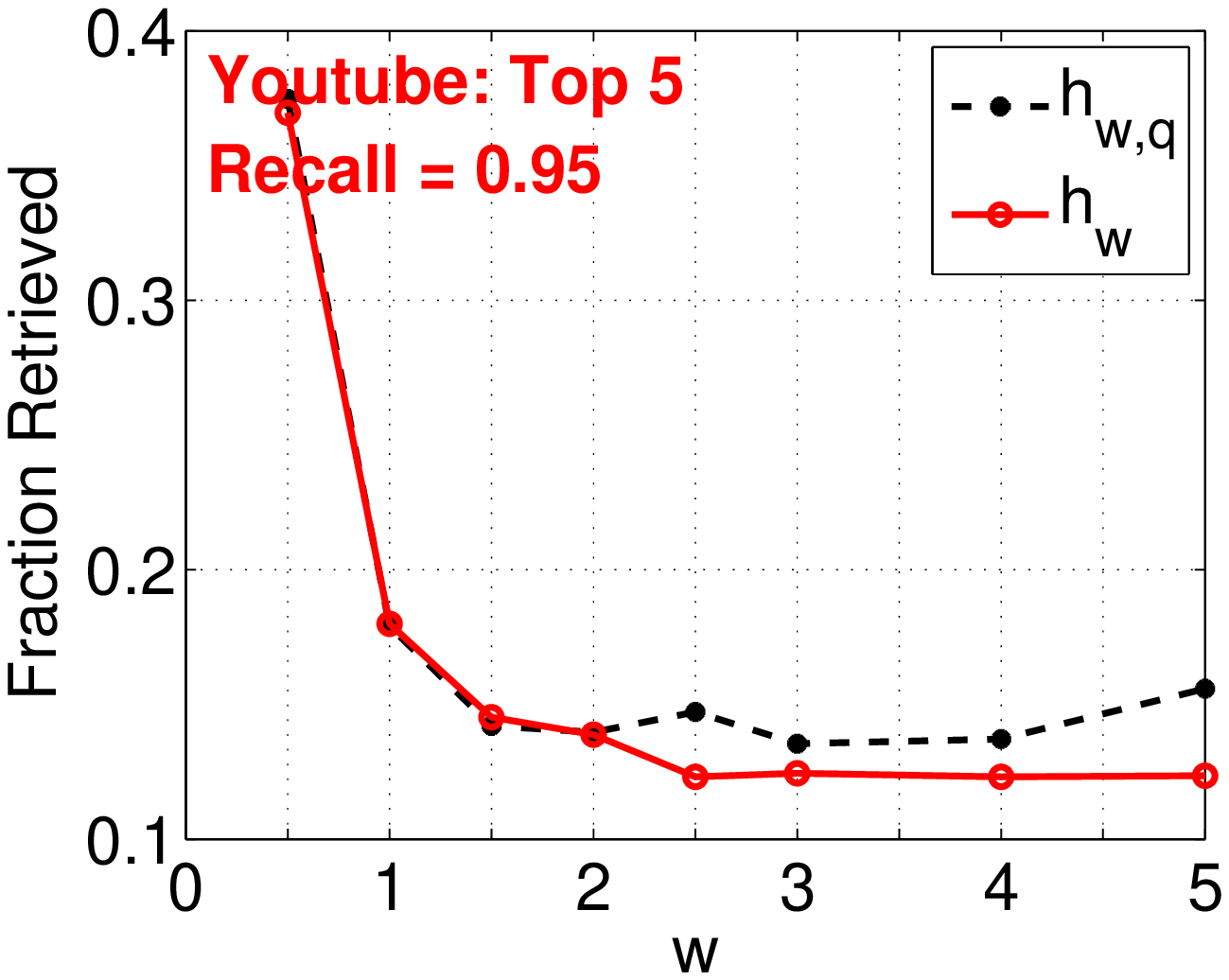}
\includegraphics[width = 2.7in]{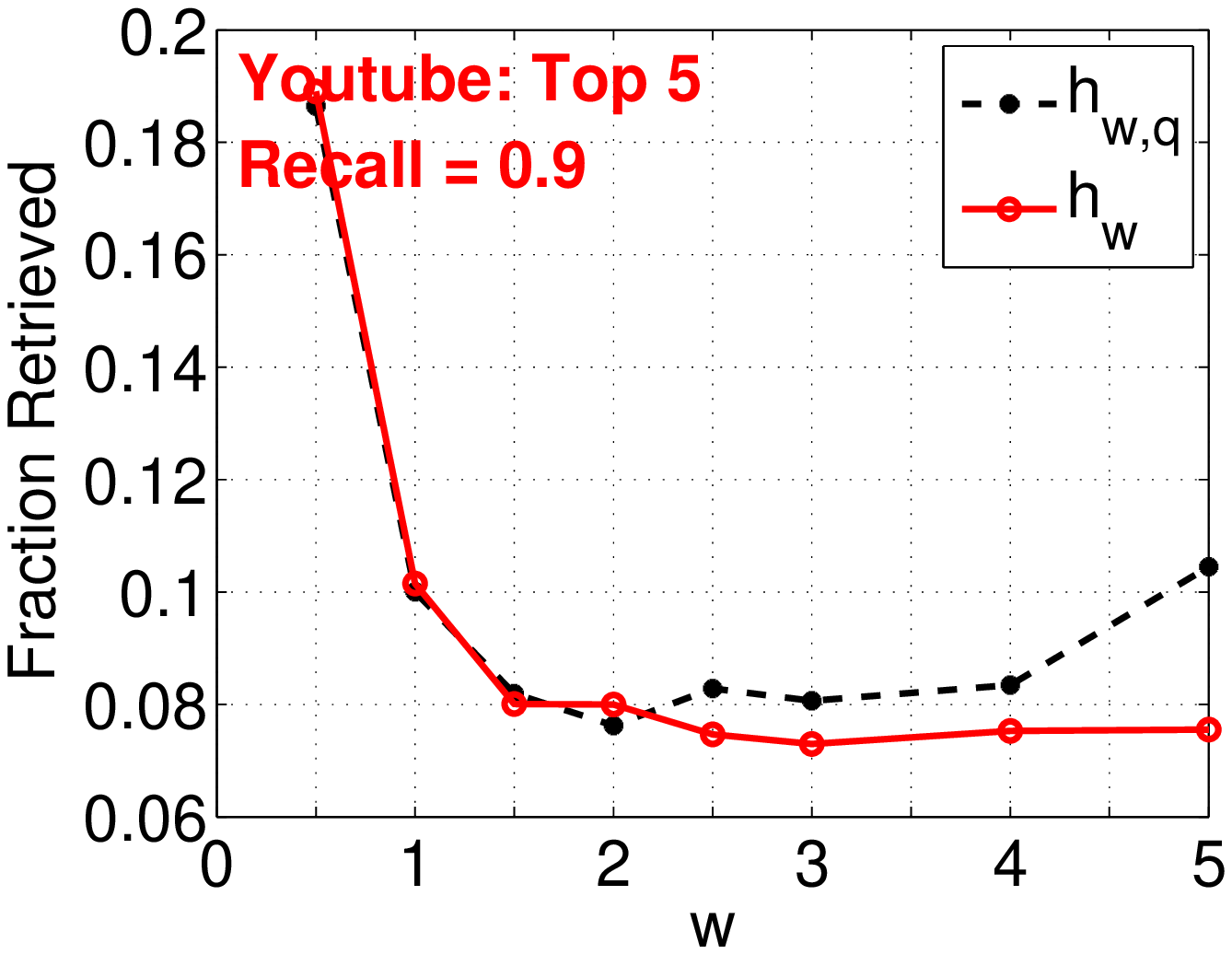}
}

\mbox{
\includegraphics[width = 2.7in]{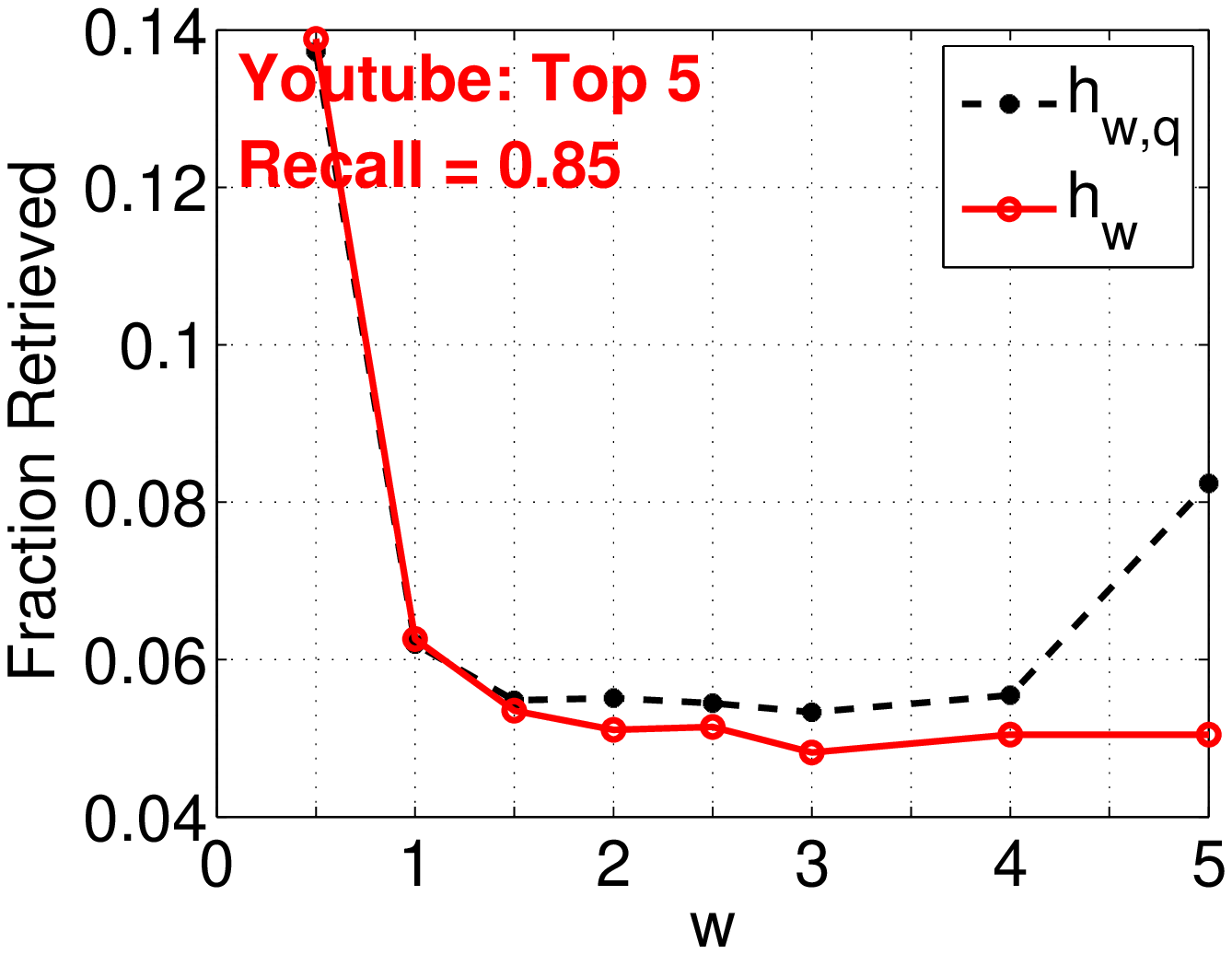}
\includegraphics[width = 2.7in]{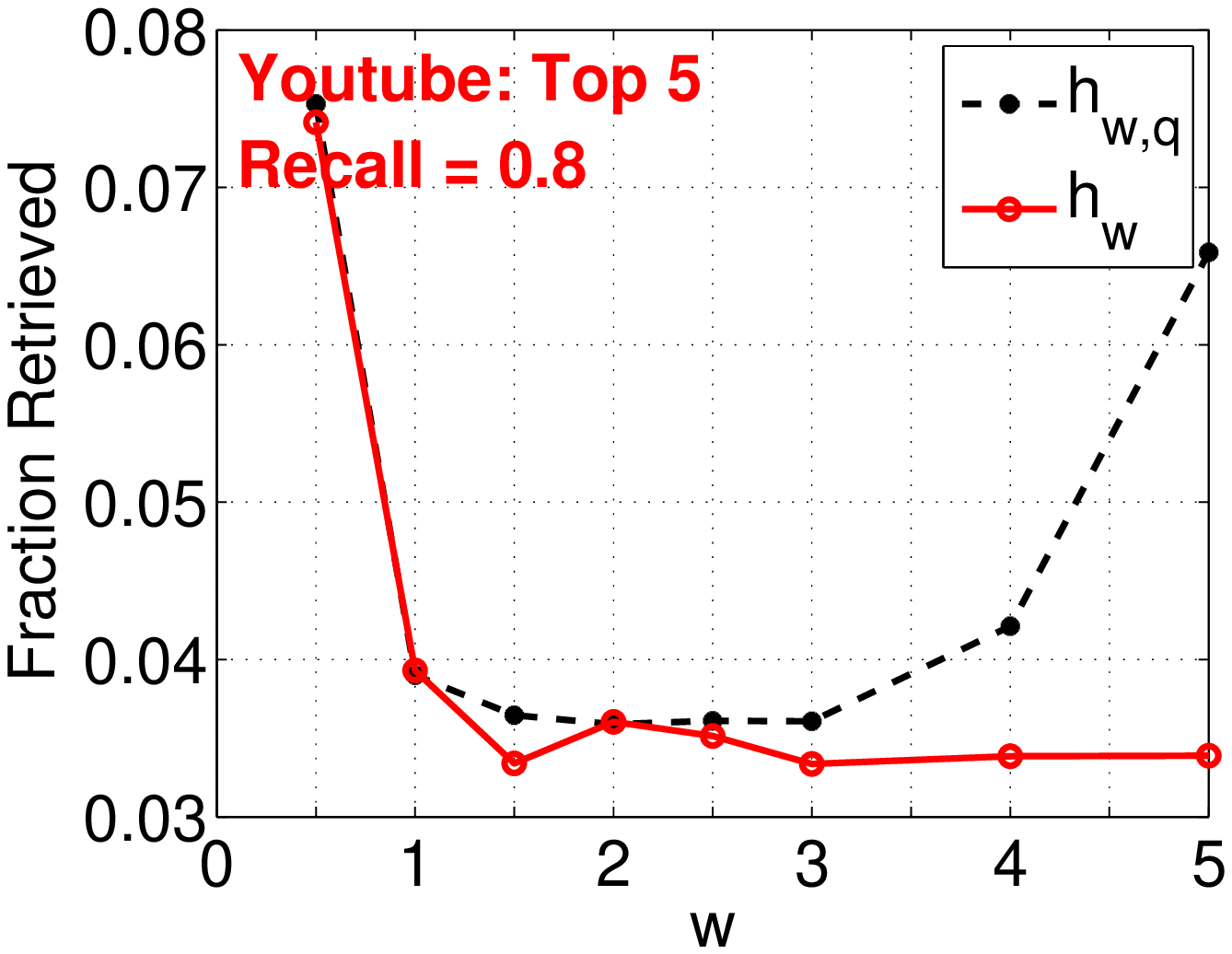}
}

\mbox{
\includegraphics[width = 2.7in]{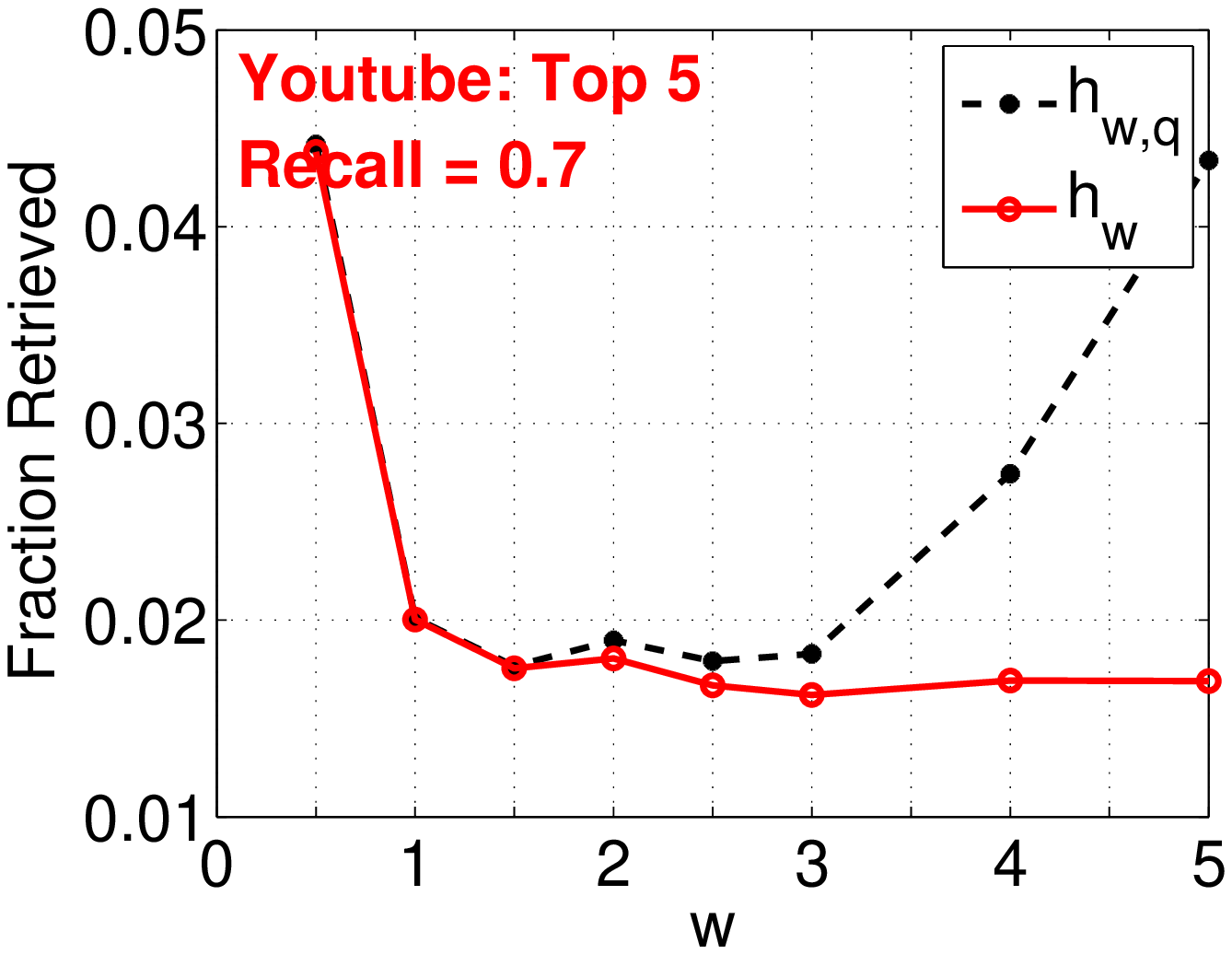}
\includegraphics[width = 2.7in]{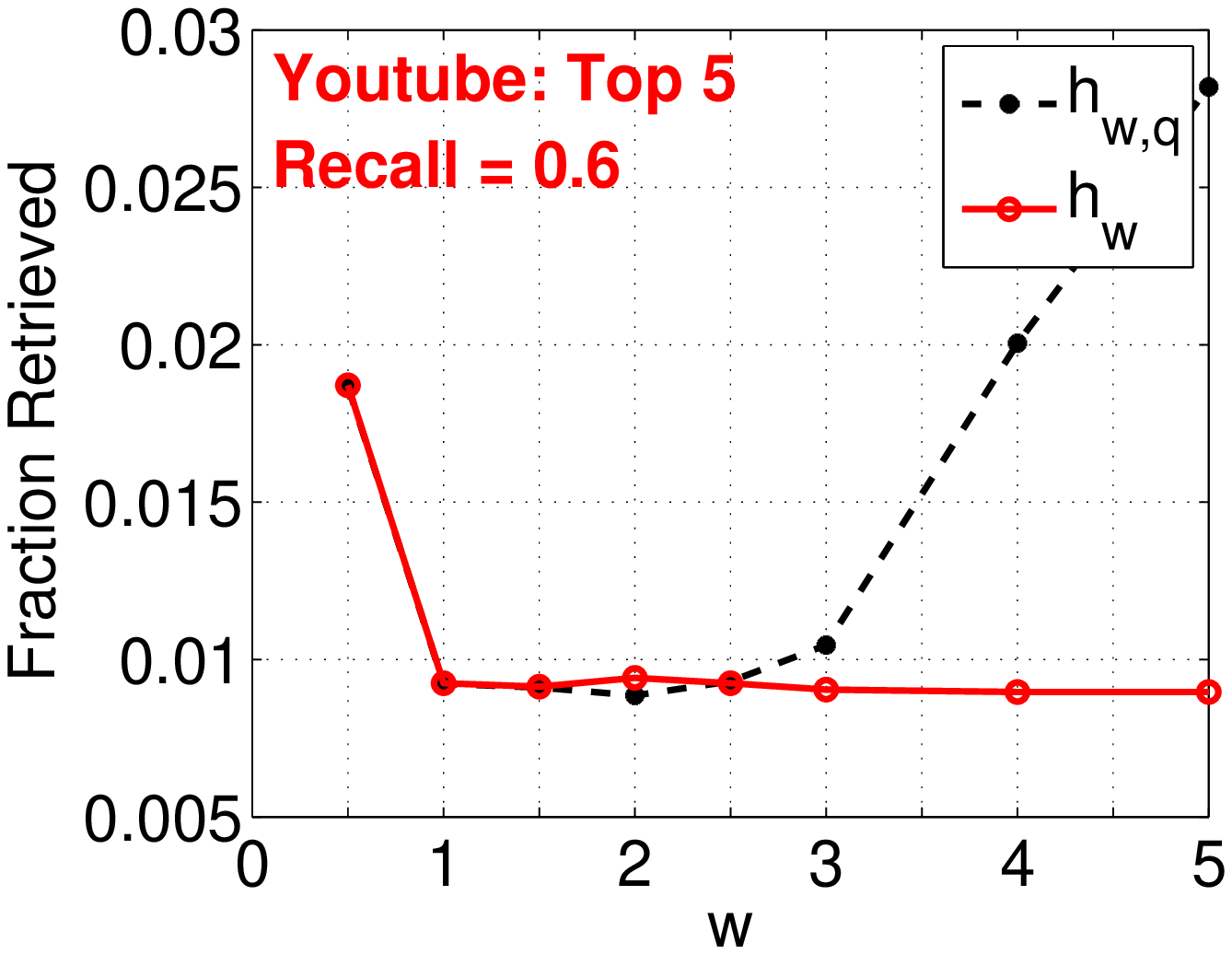}
}

\end{center}
\vspace{-.2in}
\caption{ \textbf{Youtube Top 5} . In each panel, we plot the optimal {\em fraction retrieved} at a target {\em recall} value (for top-5) with respect to $w$ for both coding schemes $h_w$ and $h_{w,q}$. }\label{fig_YoutubeRecallvsWT5}
\end{figure}

\begin{figure}
\begin{center}
\mbox{
\includegraphics[width = 2.7in]{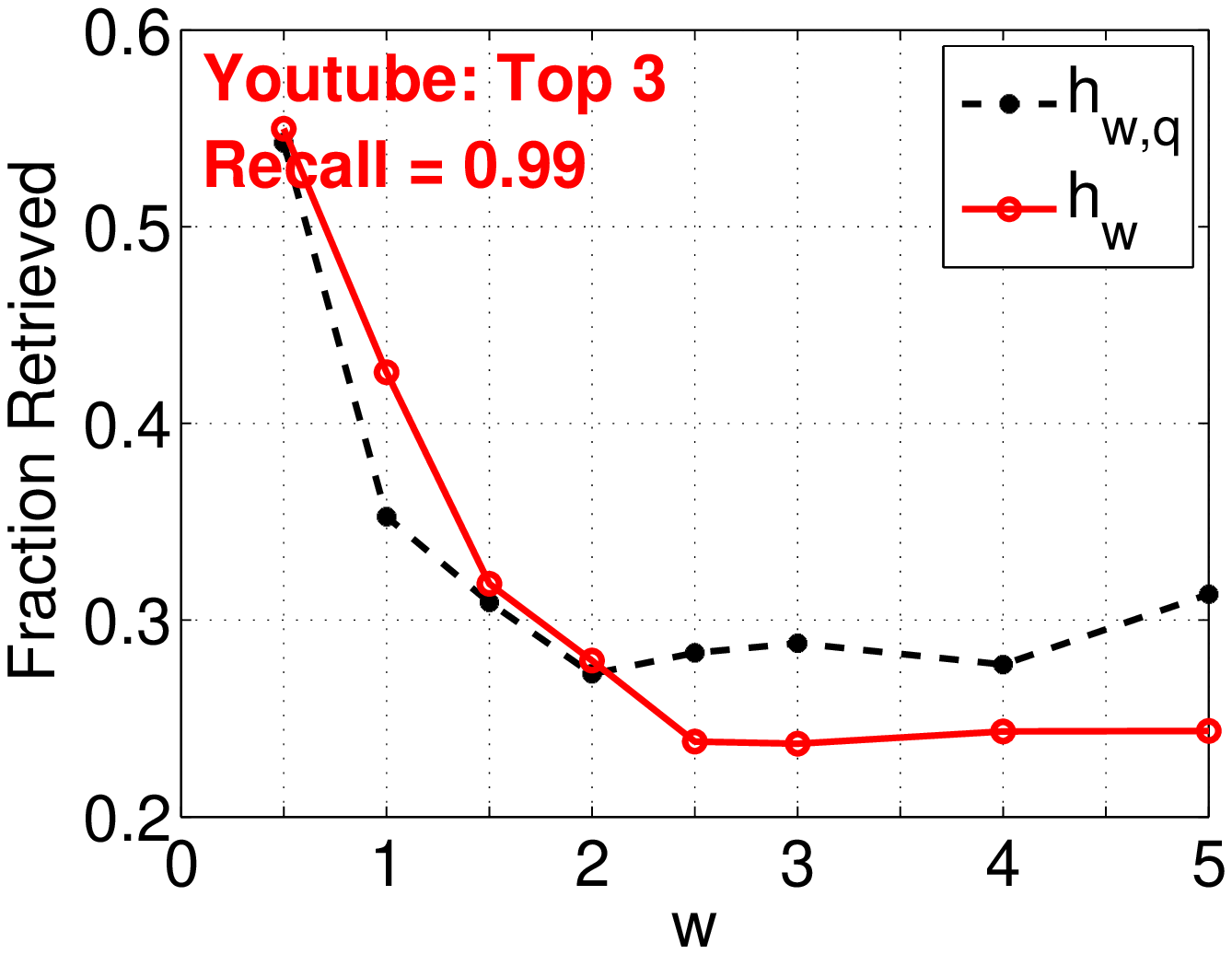}
\includegraphics[width = 2.7in]{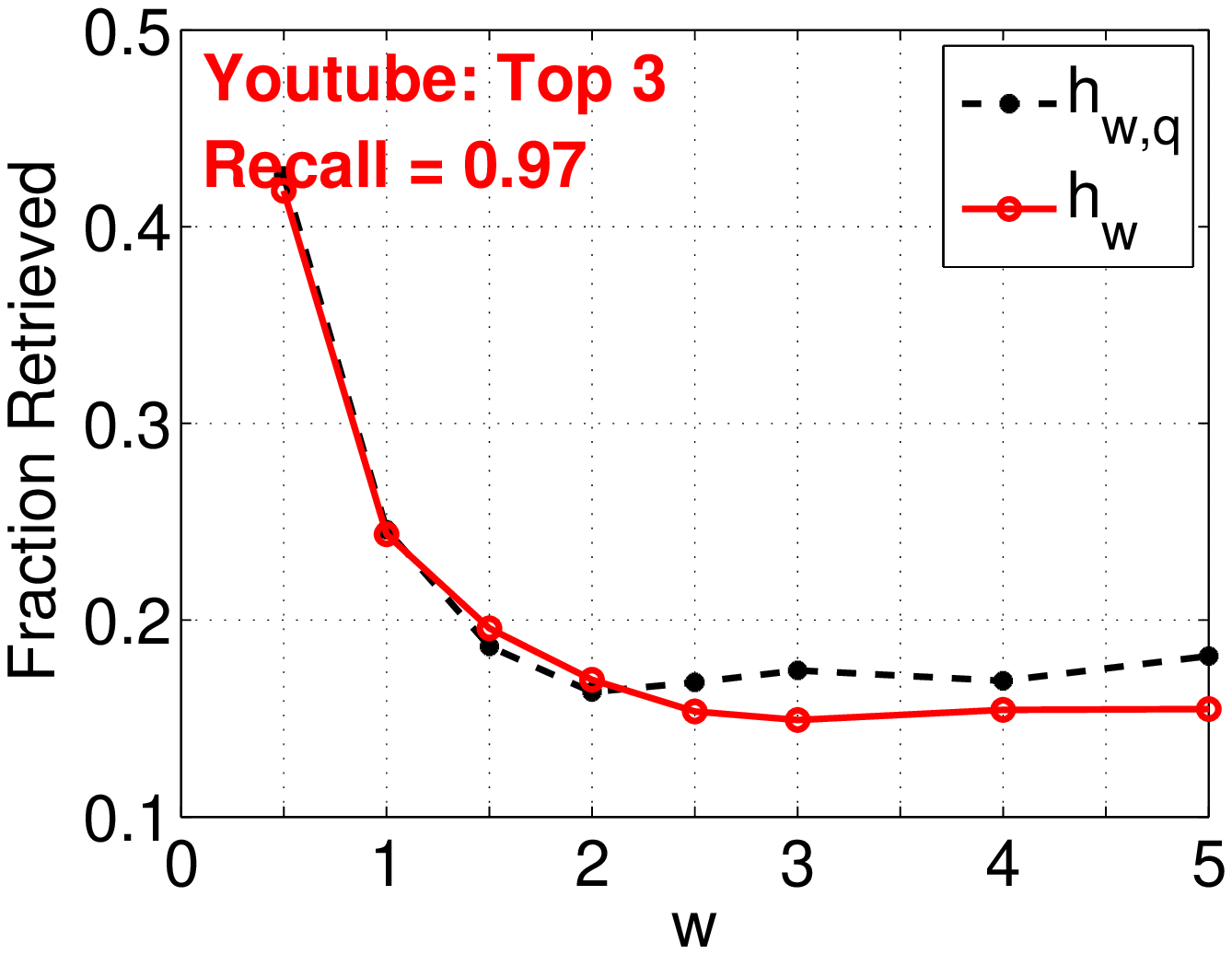}
}
\mbox{
\includegraphics[width = 2.7in]{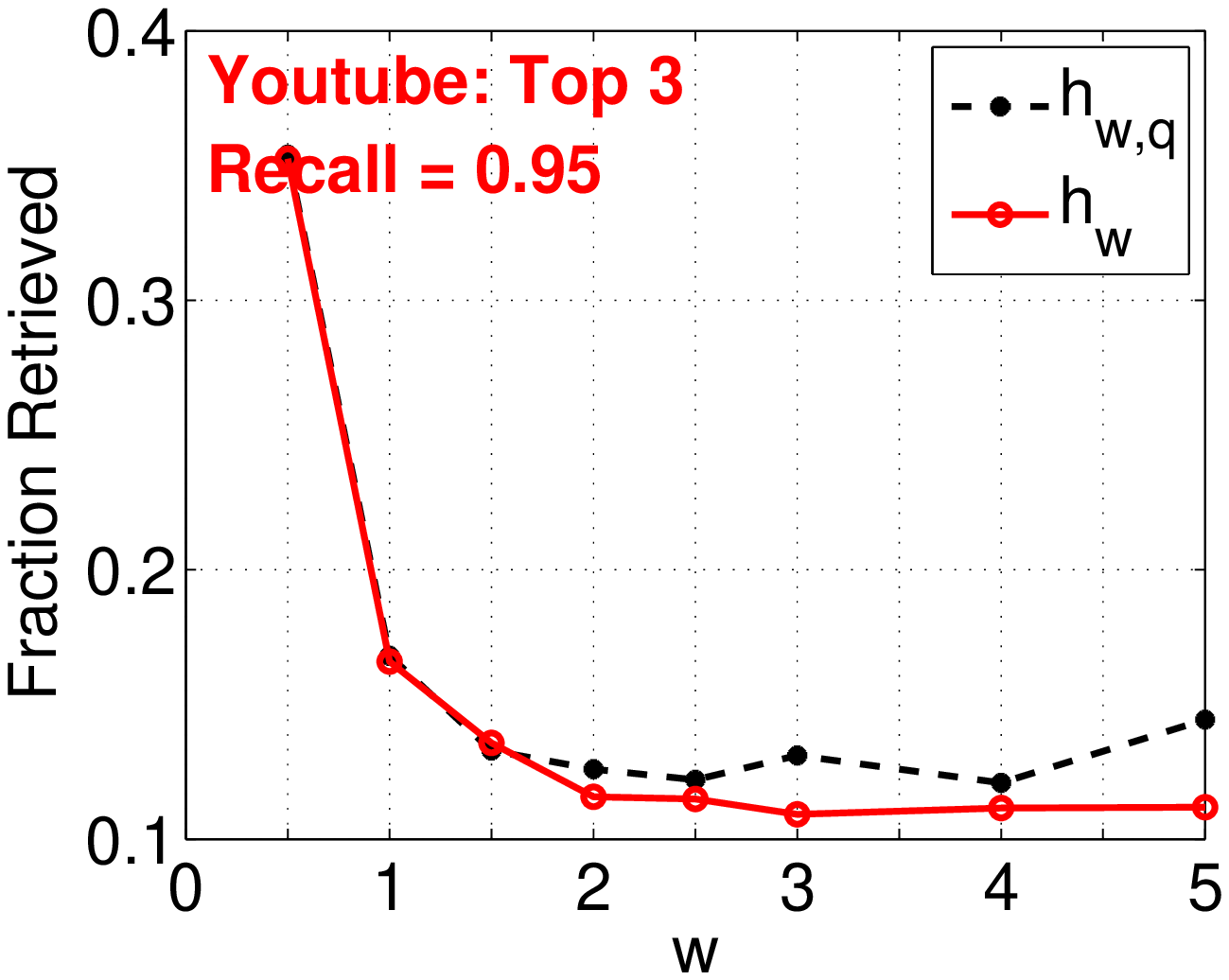}
\includegraphics[width = 2.7in]{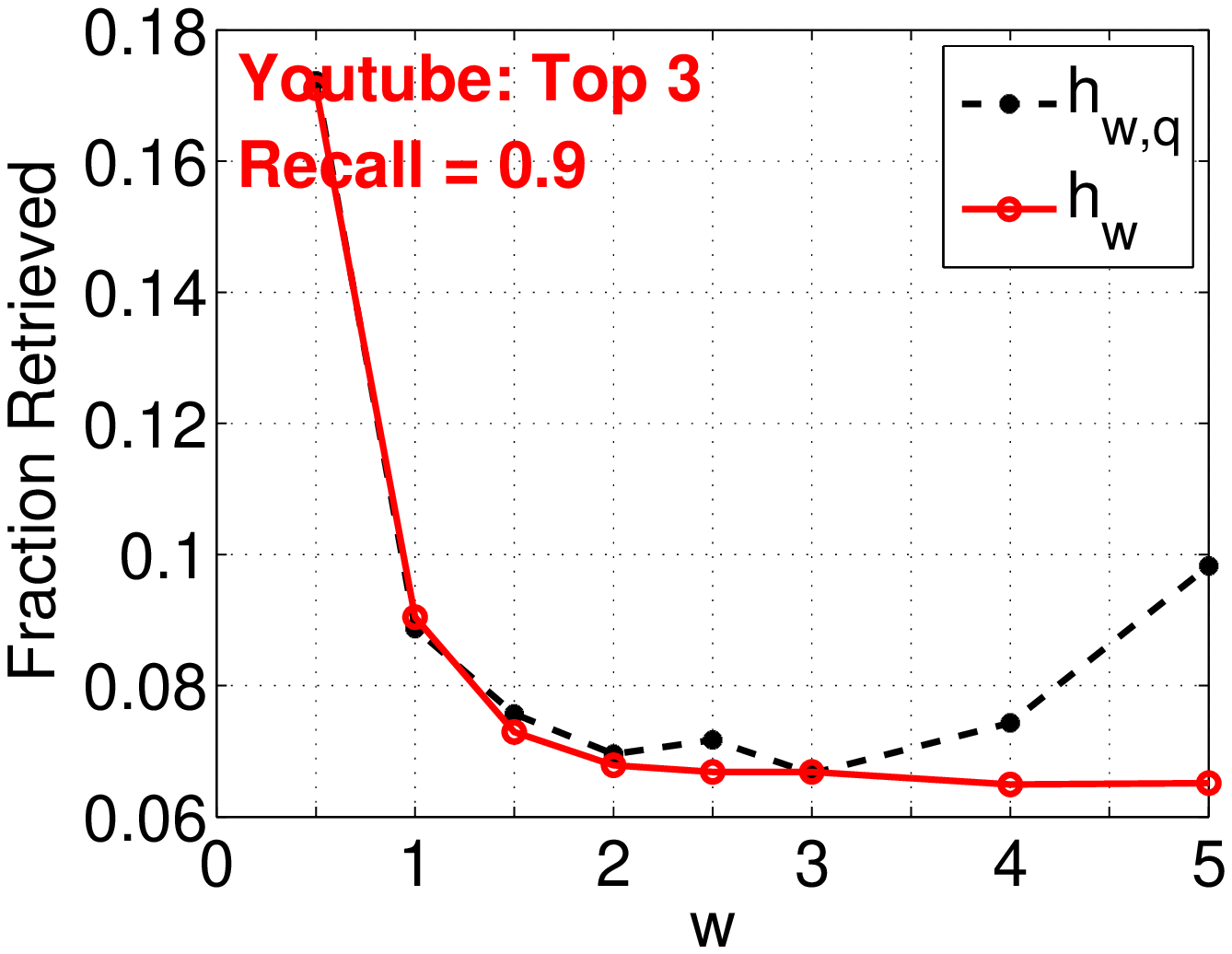}
}

\mbox{
\includegraphics[width = 2.7in]{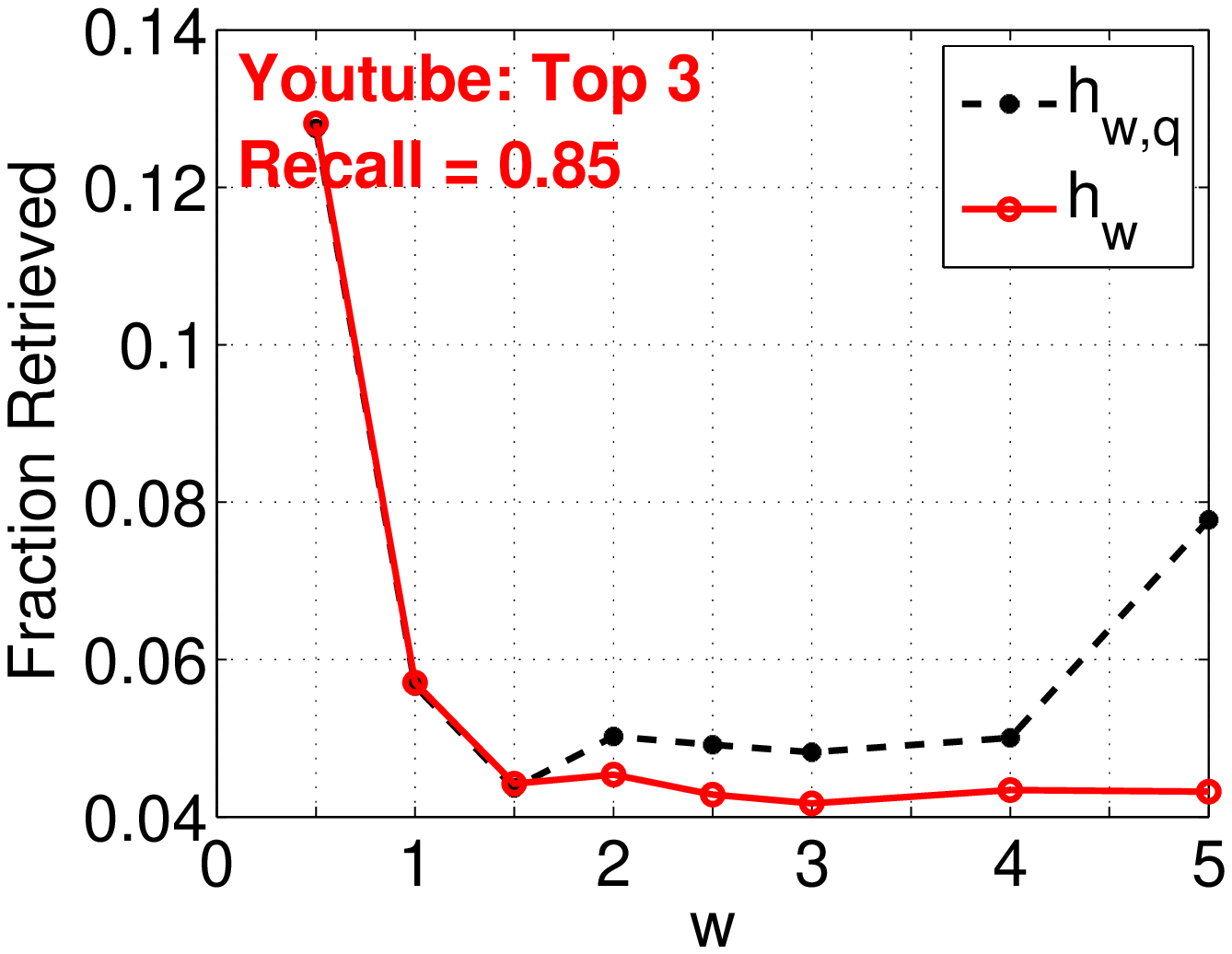}
\includegraphics[width = 2.7in]{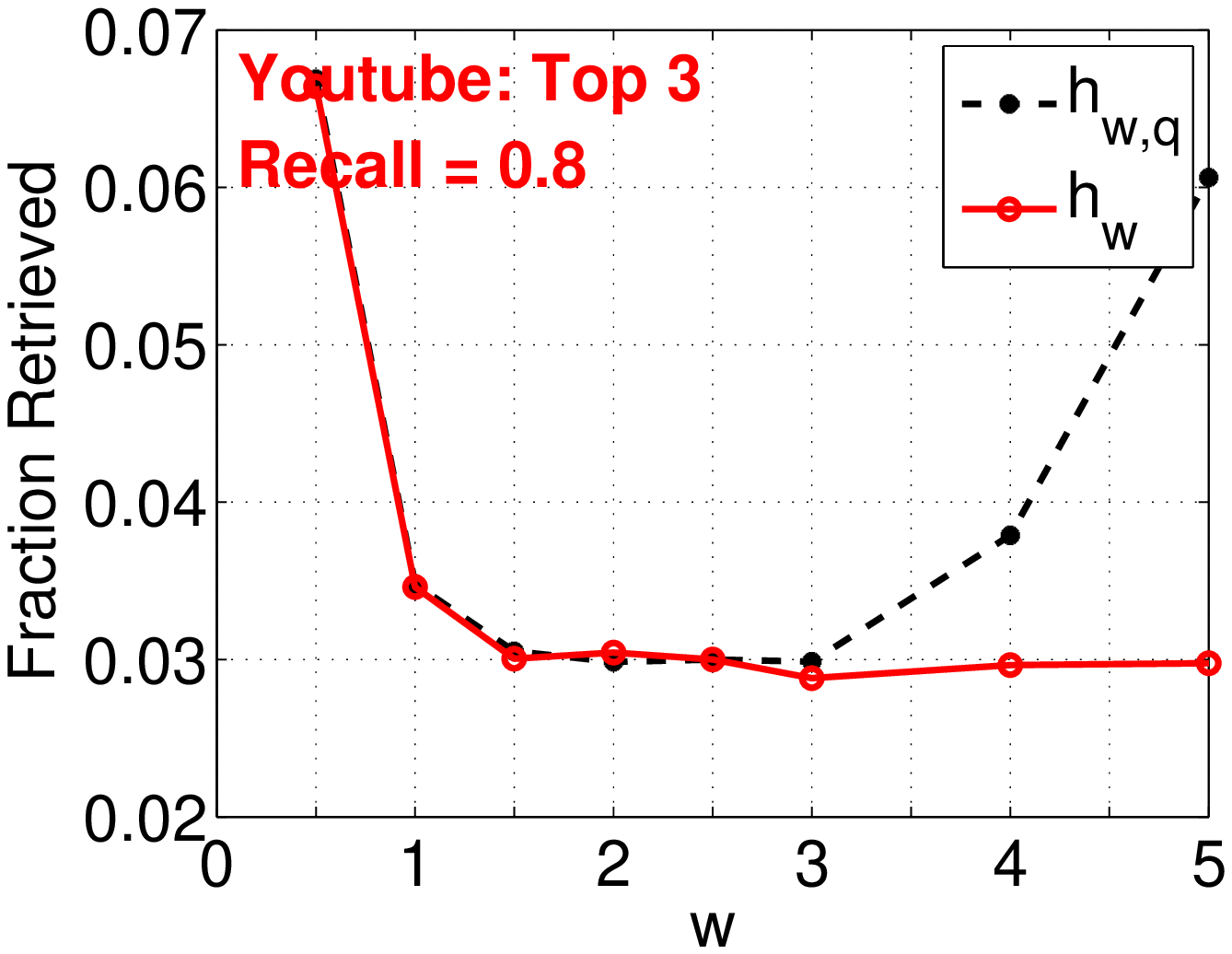}
}

\mbox{
\includegraphics[width = 2.7in]{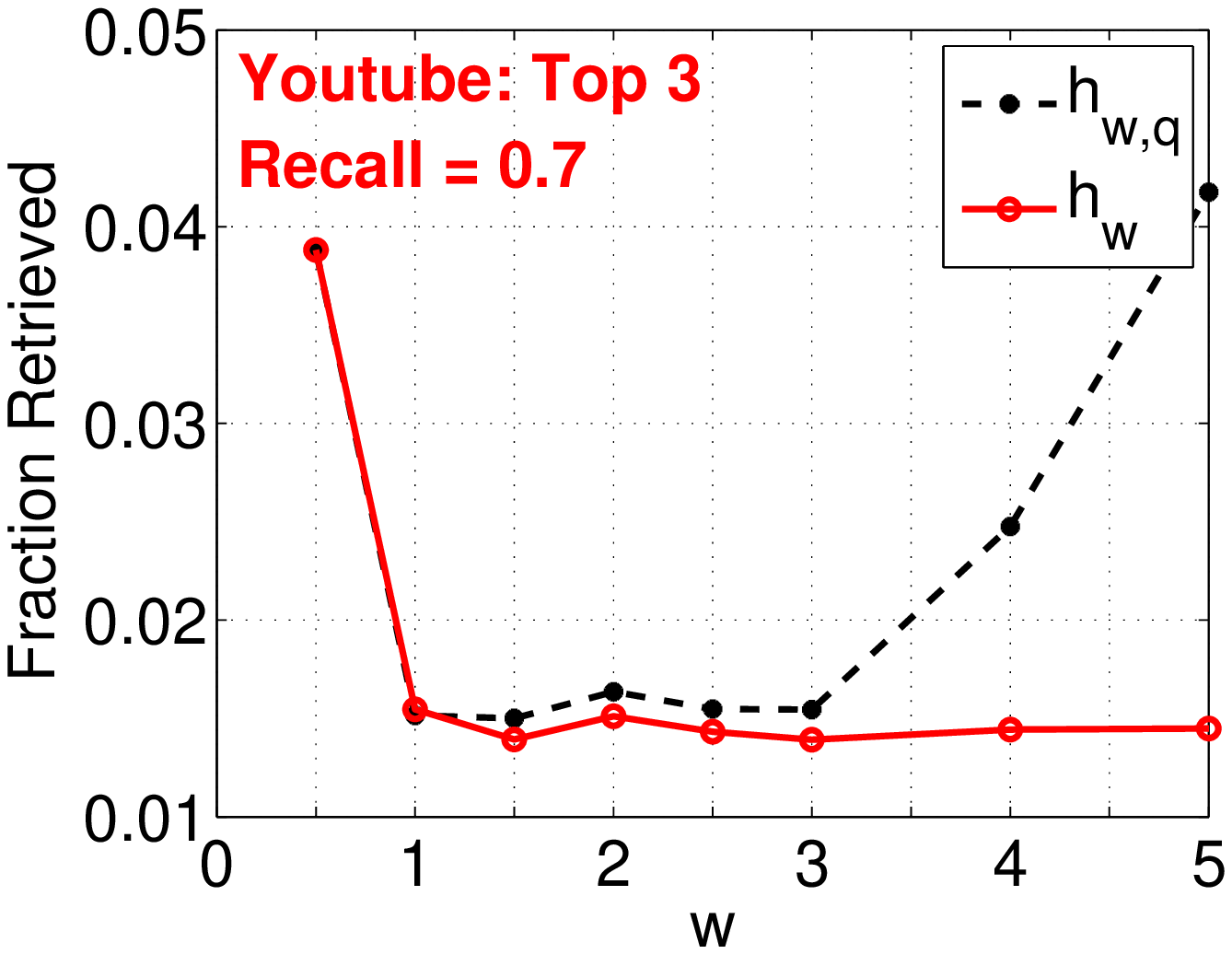}
\includegraphics[width = 2.7in]{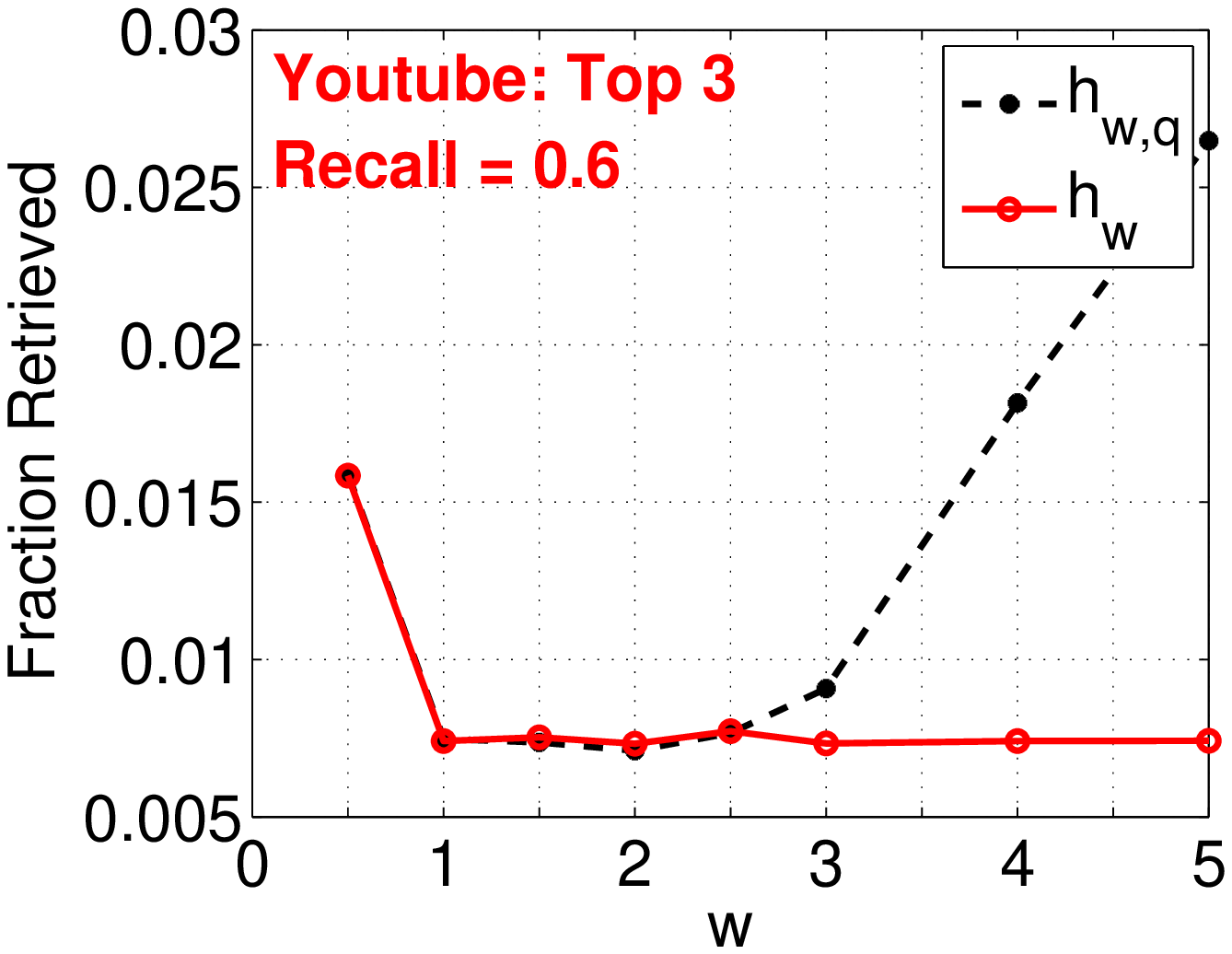}
}

\end{center}
\vspace{-.2in}
\caption{ \textbf{Youtube Top 3} . In each panel, we plot the optimal {\em fraction retrieved} at a target {\em recall} value (for top-3) with respect to $w$ for both coding schemes $h_w$ and $h_{w,q}$. }\label{fig_YoutubeRecallvsWT3}
\end{figure}

\begin{figure}
\begin{center}
\mbox{
\includegraphics[width = 2.7in]{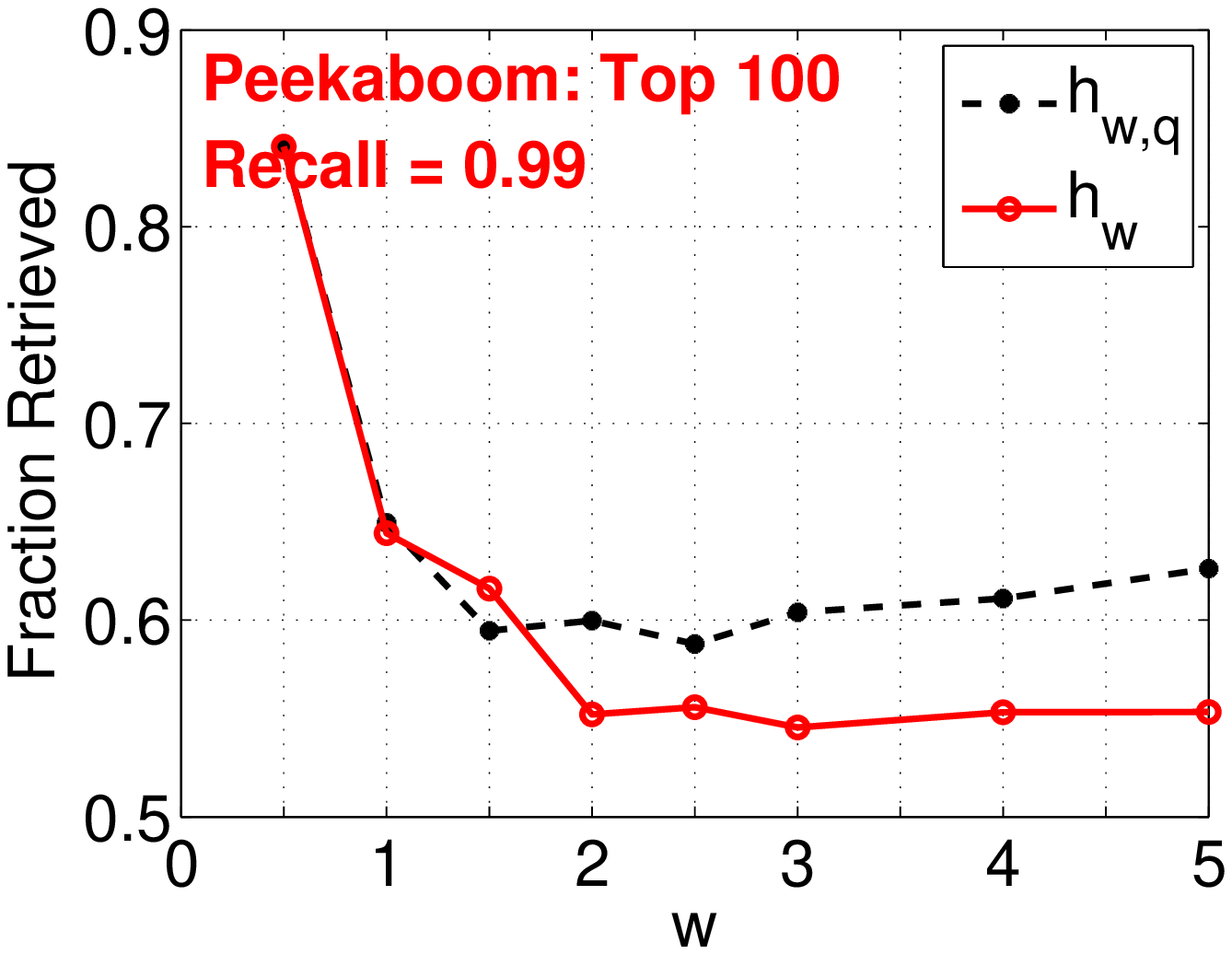}
\includegraphics[width = 2.7in]{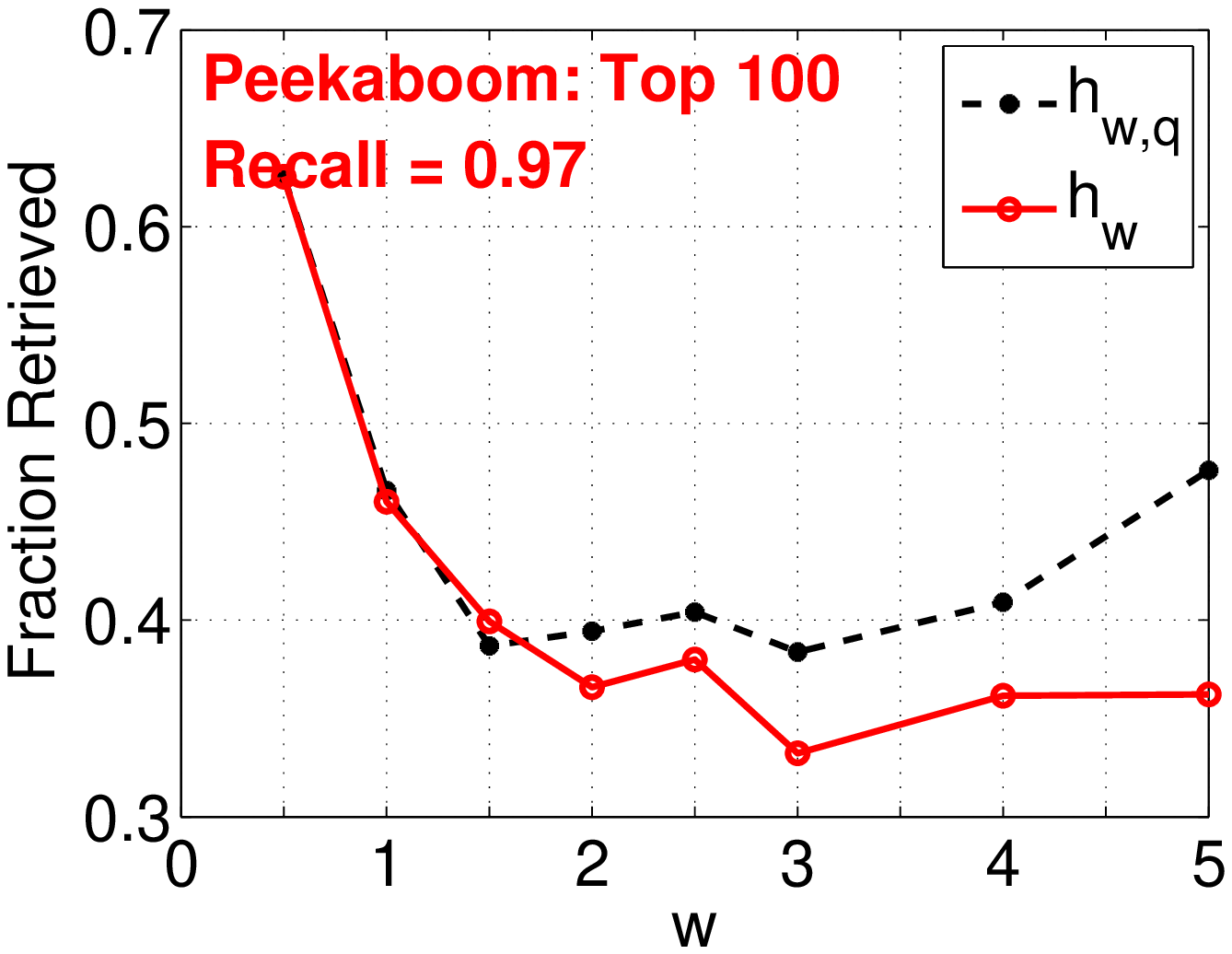}
}
\mbox{
\includegraphics[width = 2.7in]{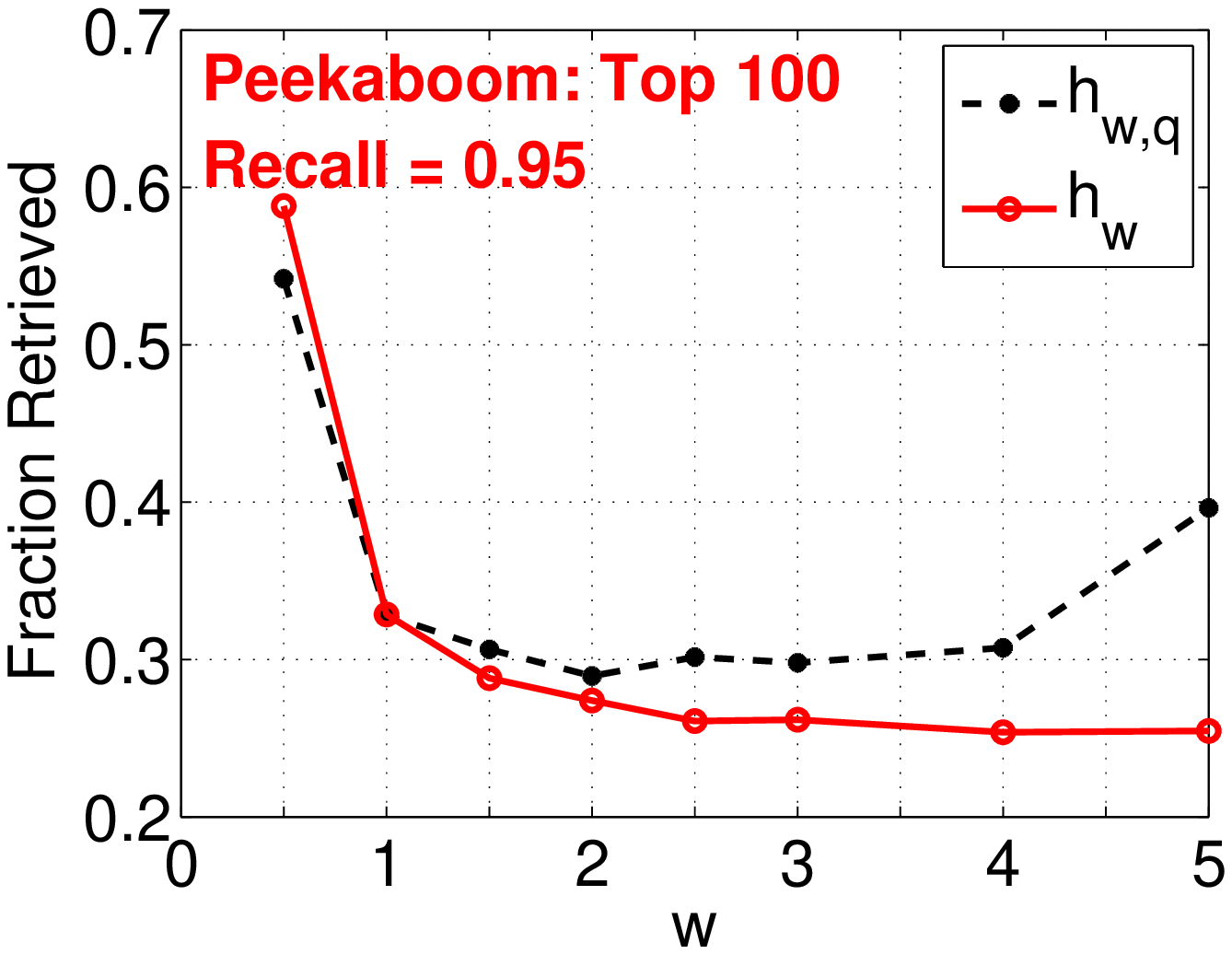}
\includegraphics[width = 2.7in]{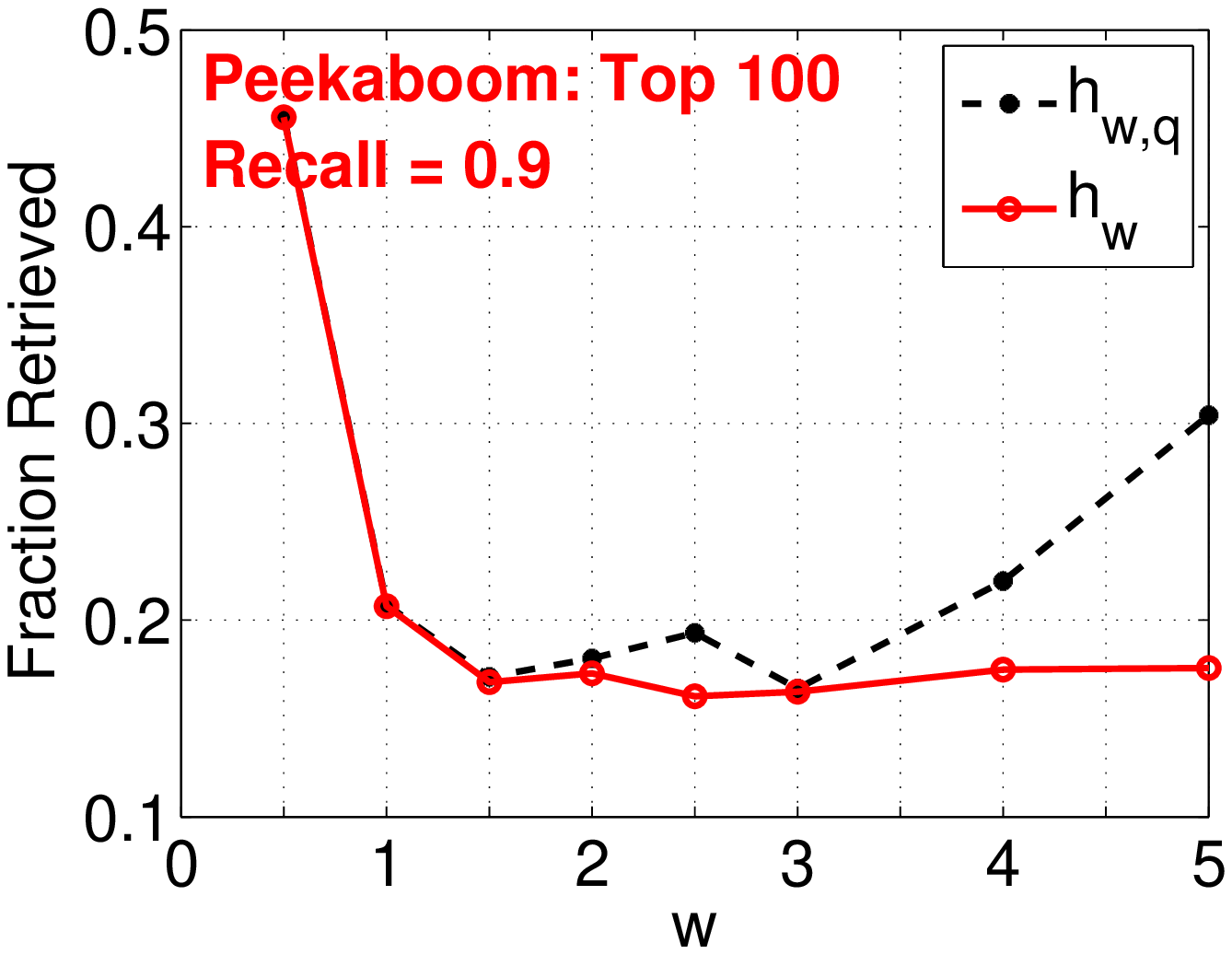}
}

\mbox{
\includegraphics[width = 2.7in]{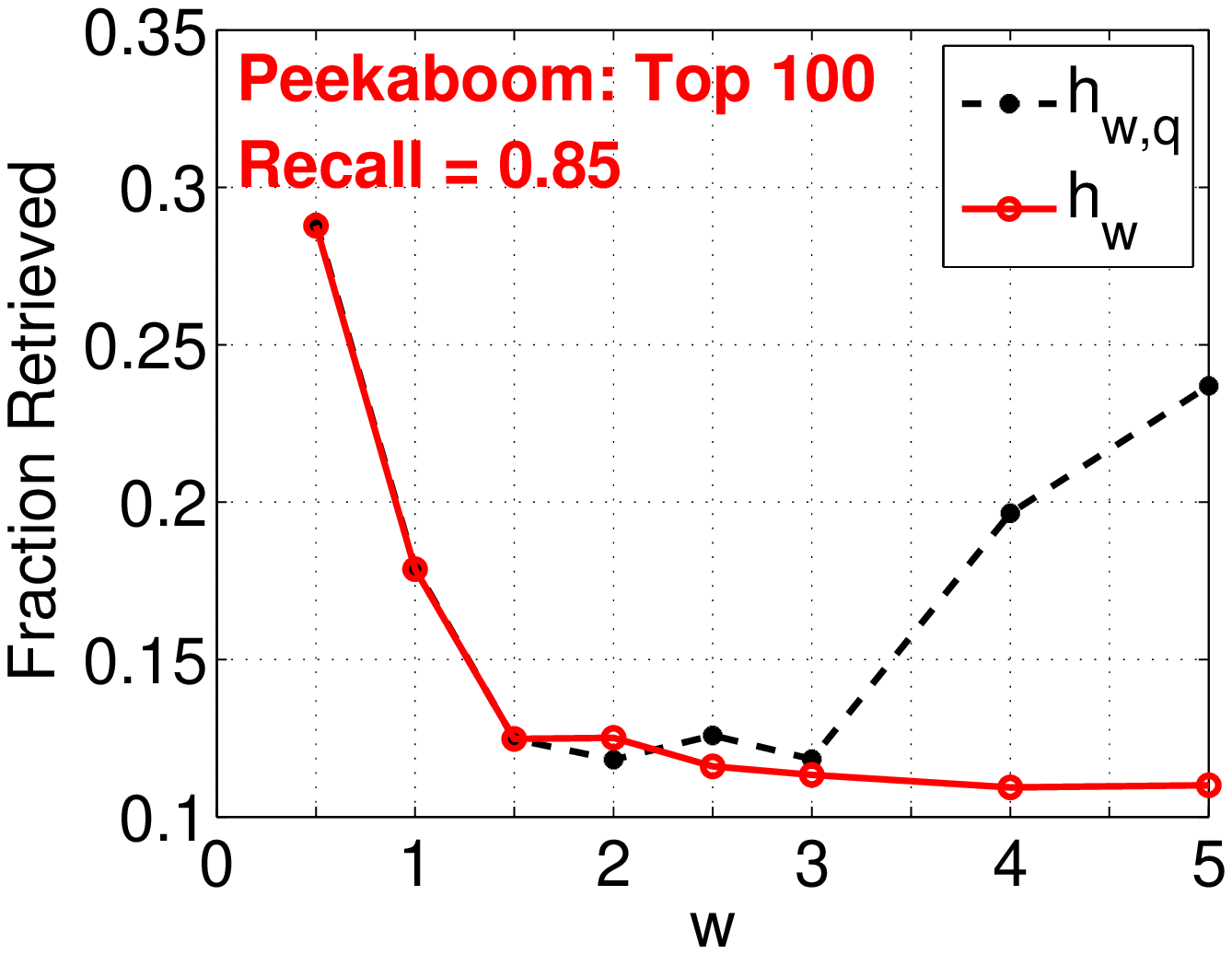}
\includegraphics[width = 2.7in]{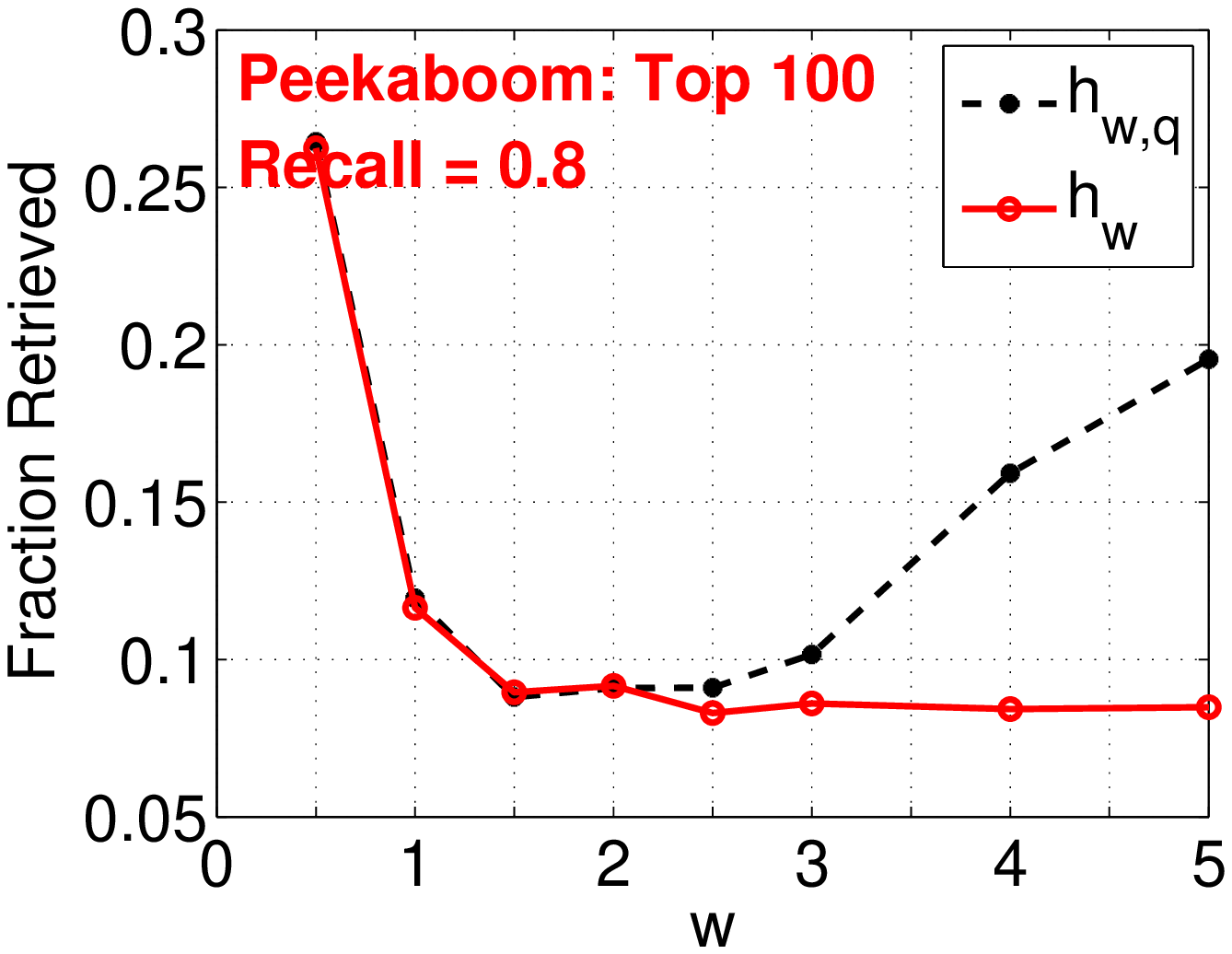}
}

\mbox{
\includegraphics[width = 2.7in]{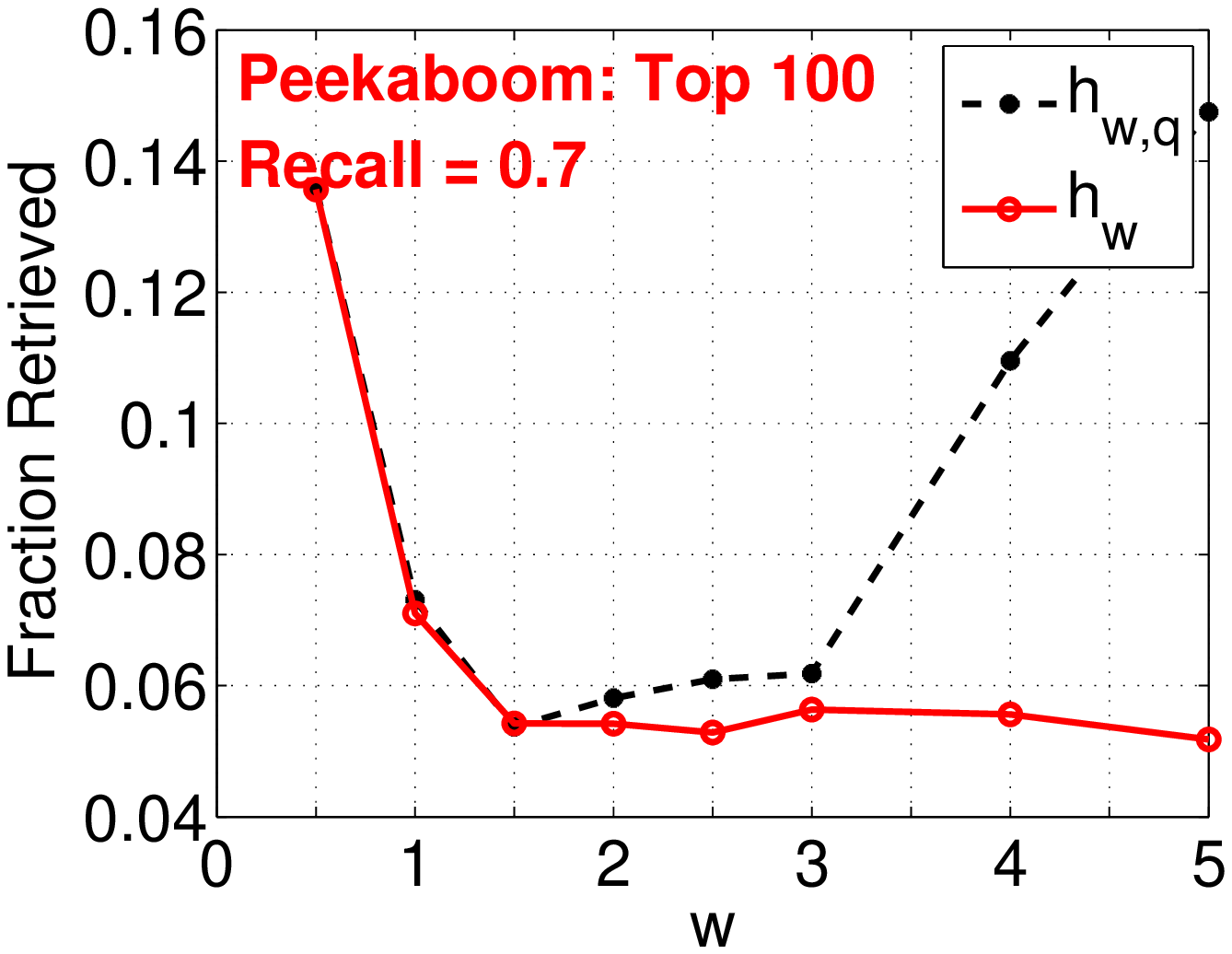}
\includegraphics[width = 2.7in]{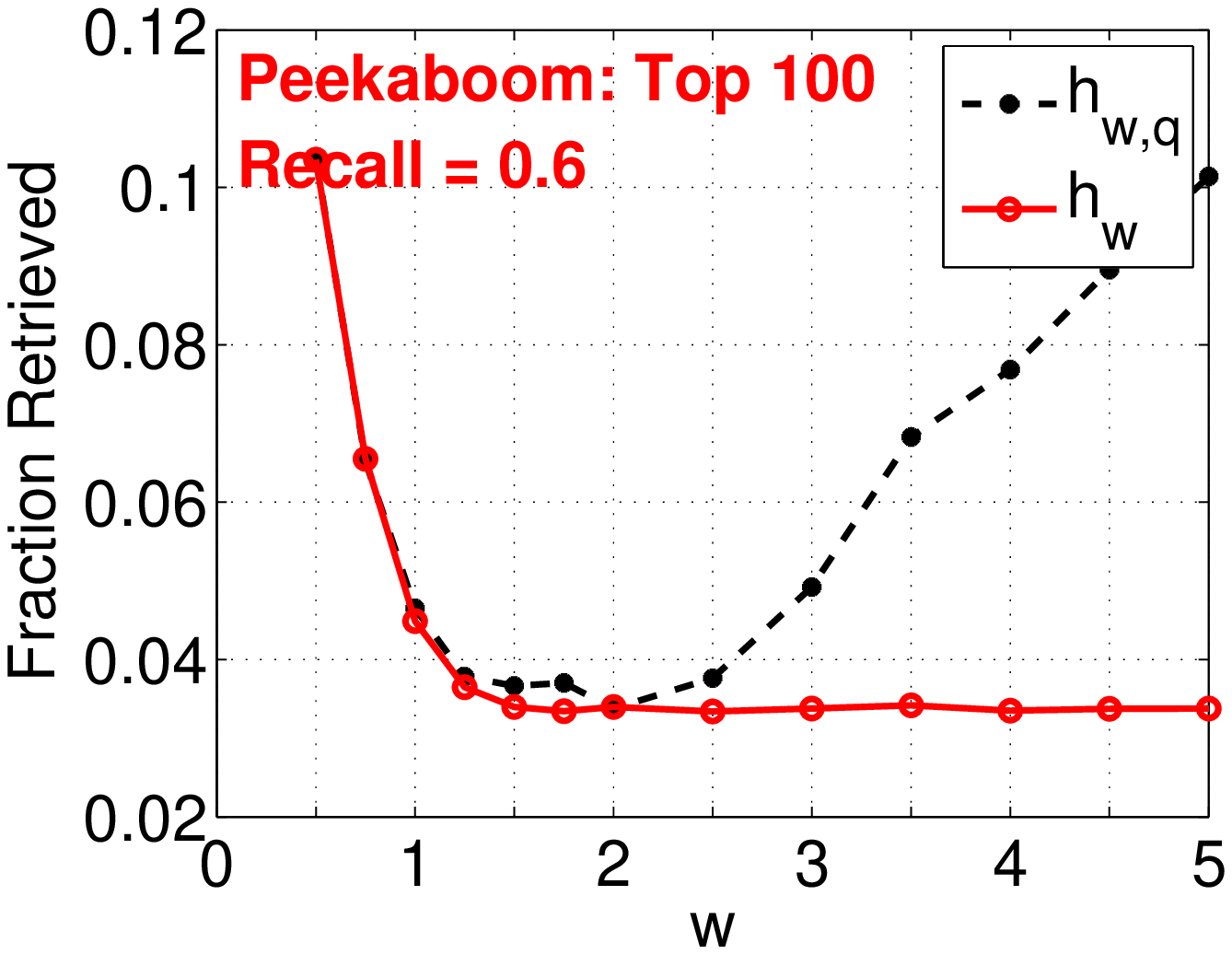}
}

\end{center}
\vspace{-.2in}
\caption{ \textbf{Peekaboom Top 100} . In each panel, we plot the optimal {\em fraction retrieved} at a target {\em recall} value (for top-100) with respect to $w$ for both coding schemes $h_w$ and $h_{w,q}$. }\label{fig_PeekaboomRecallvsWT100}
\end{figure}

\begin{figure}
\begin{center}
\mbox{
\includegraphics[width = 2.7in]{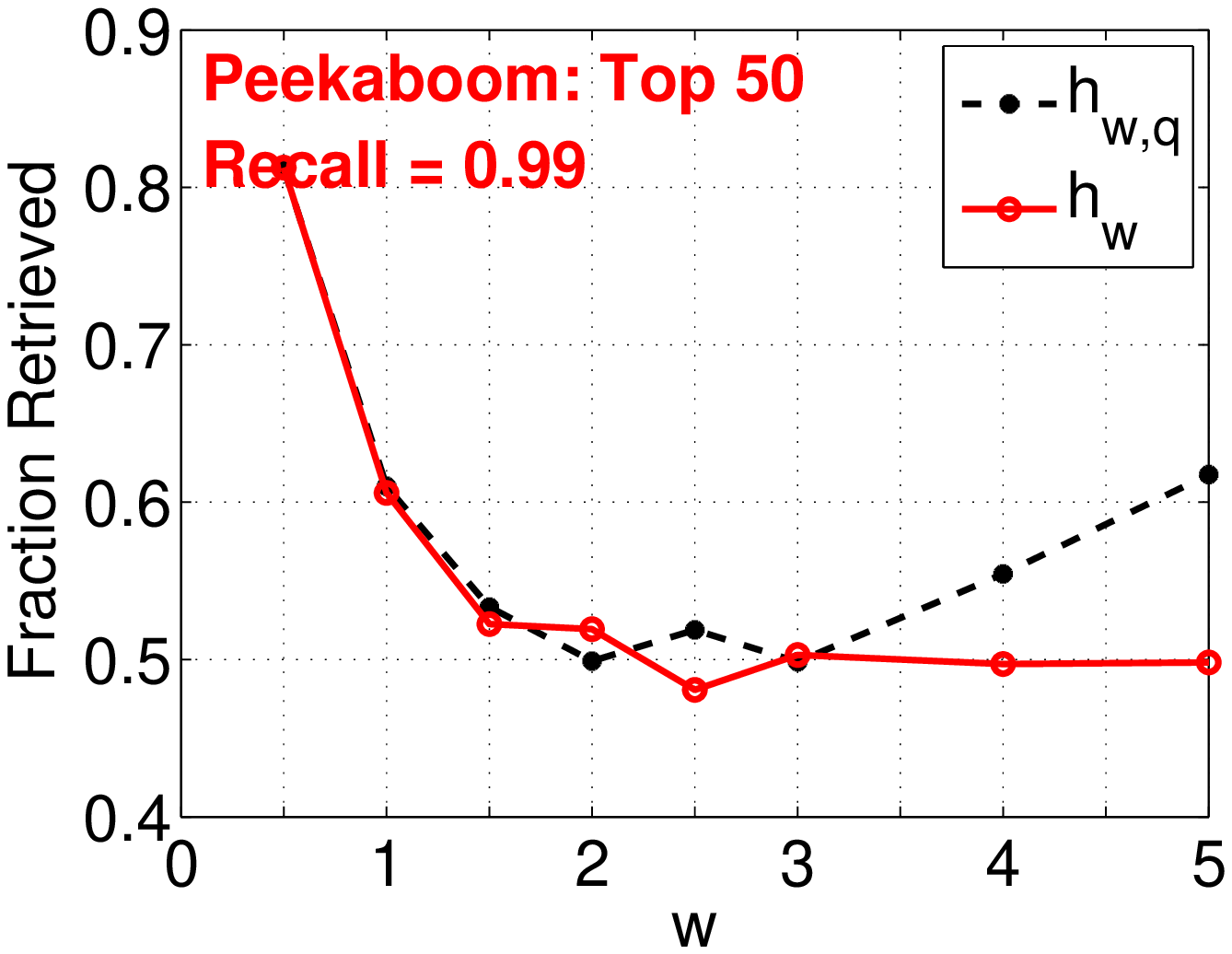}
\includegraphics[width = 2.7in]{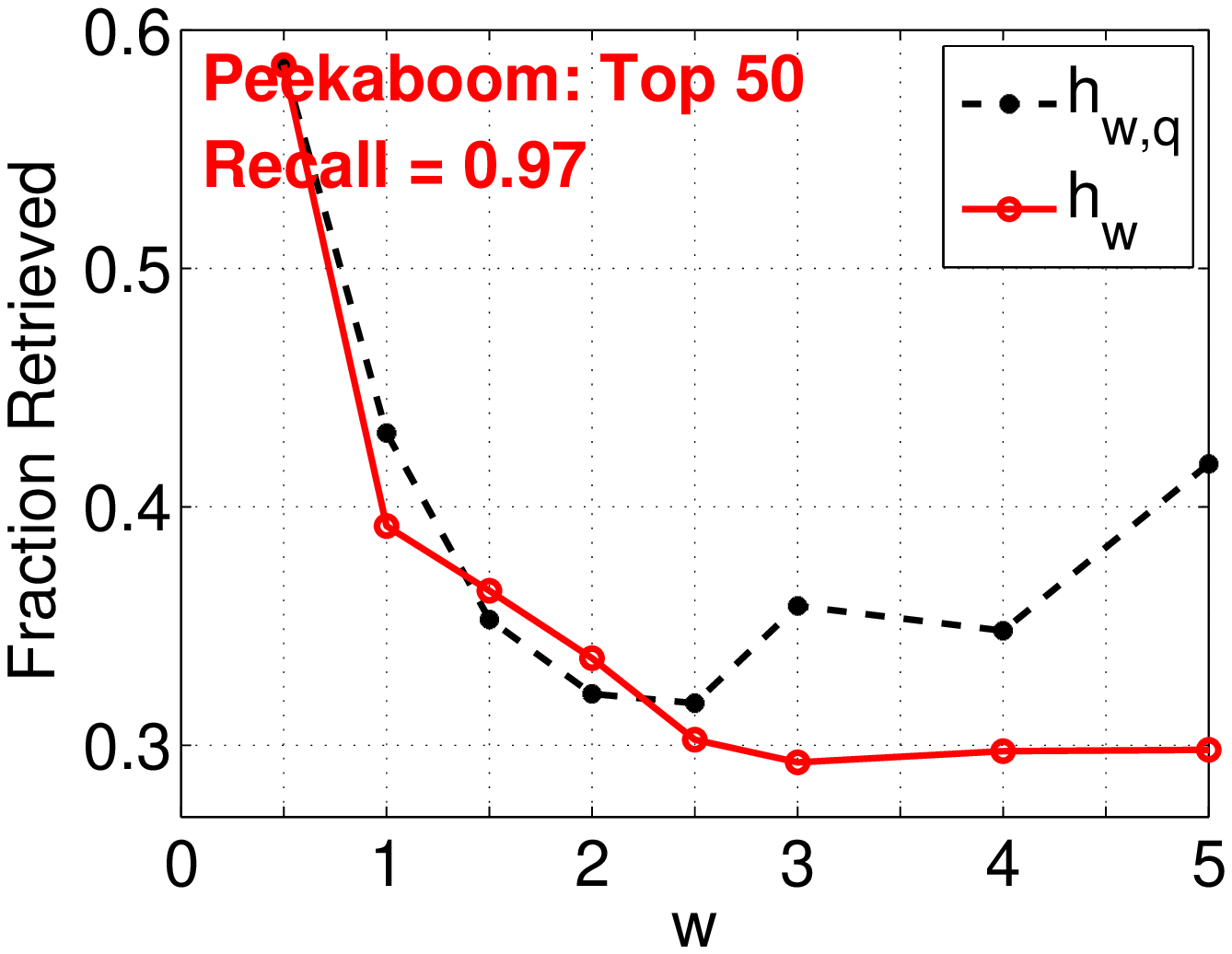}
}
\mbox{
\includegraphics[width = 2.7in]{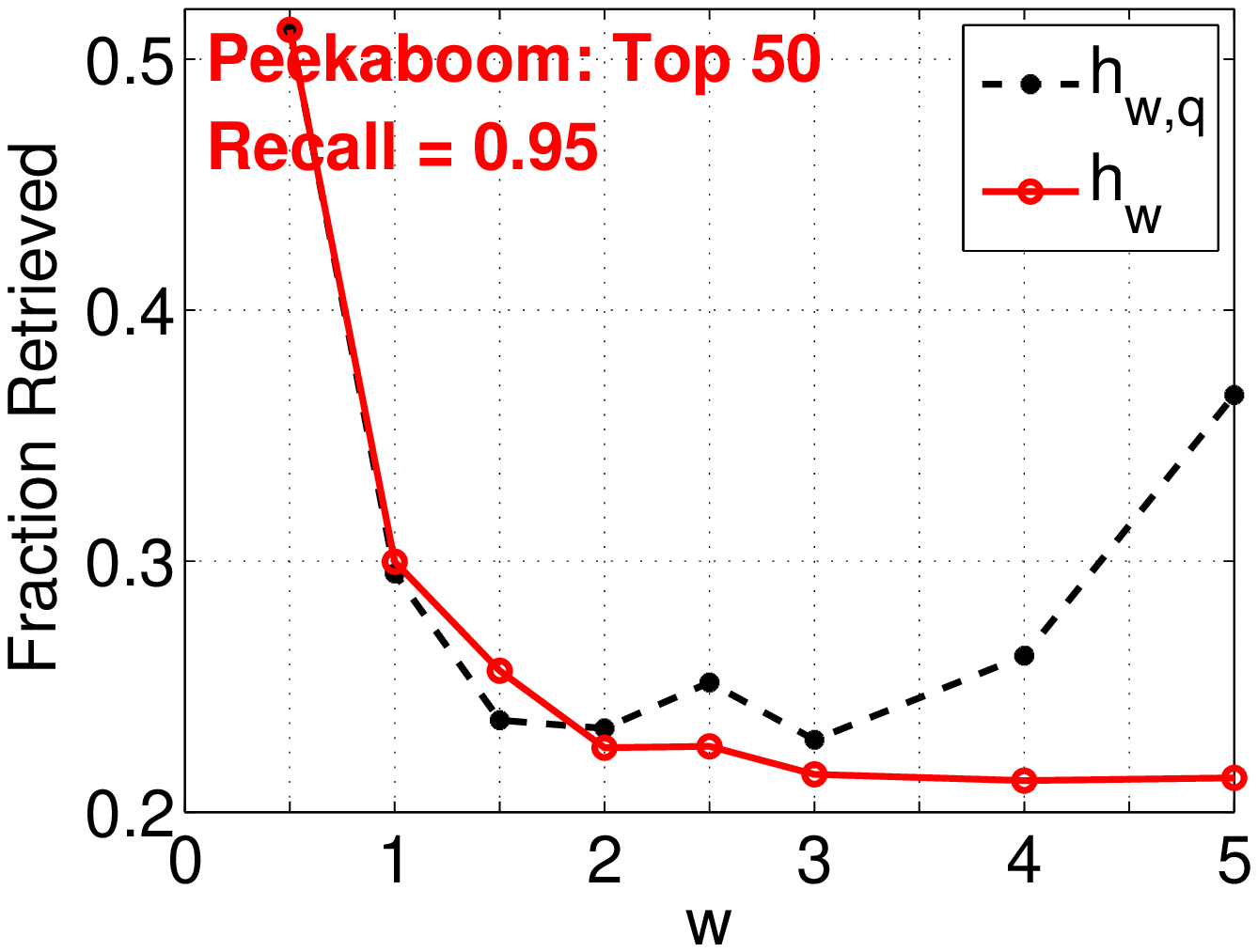}
\includegraphics[width = 2.7in]{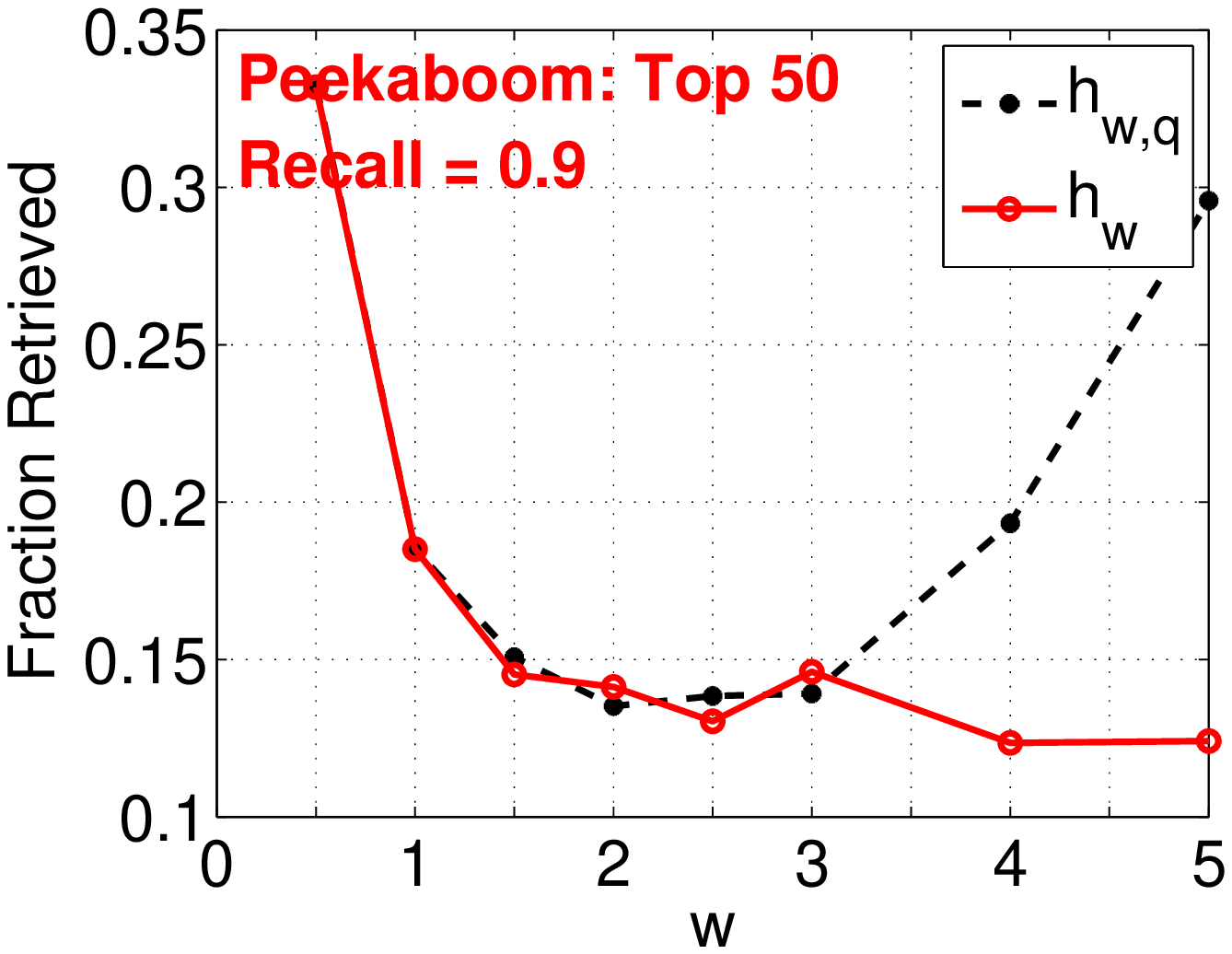}
}

\mbox{
\includegraphics[width = 2.7in]{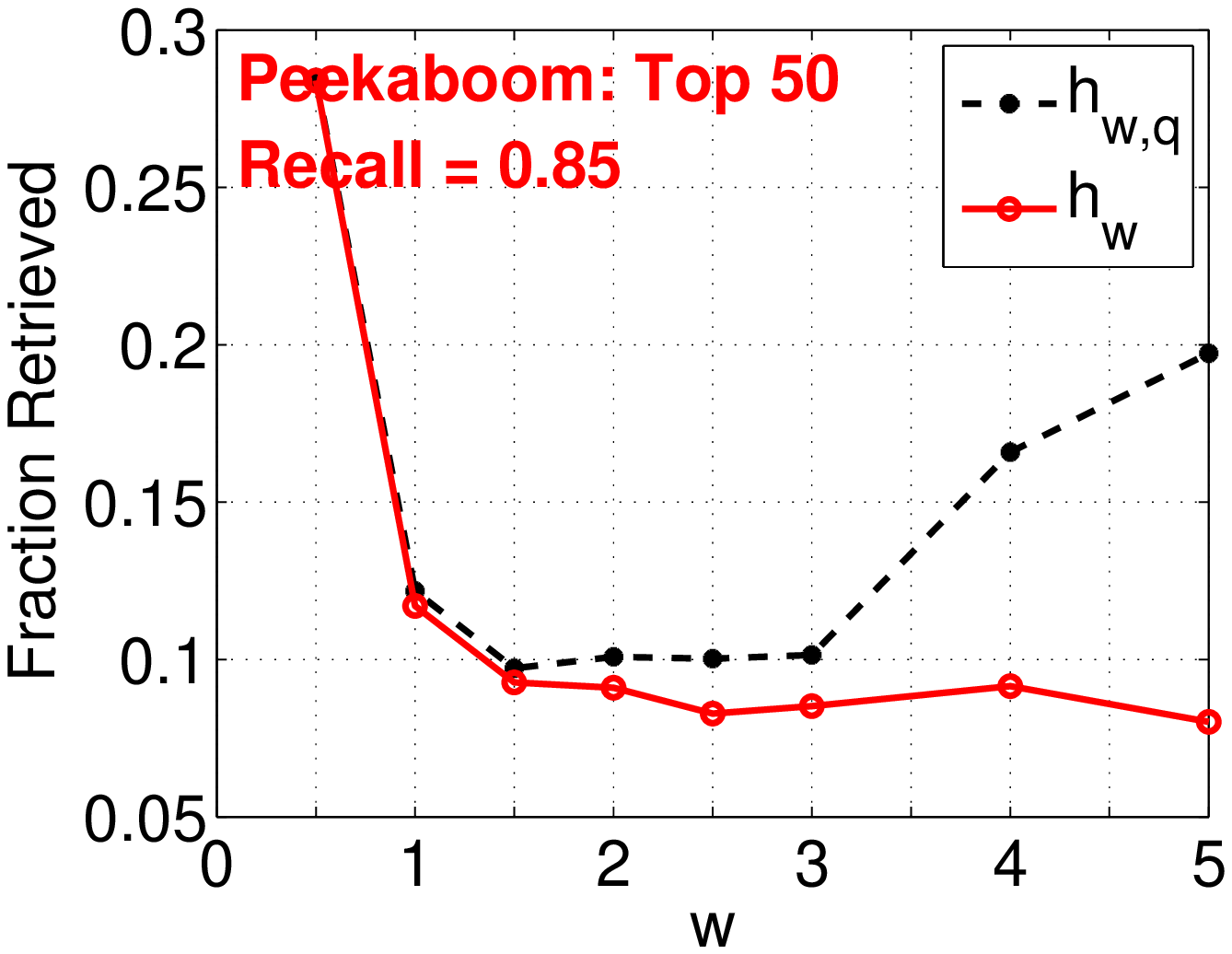}
\includegraphics[width = 2.7in]{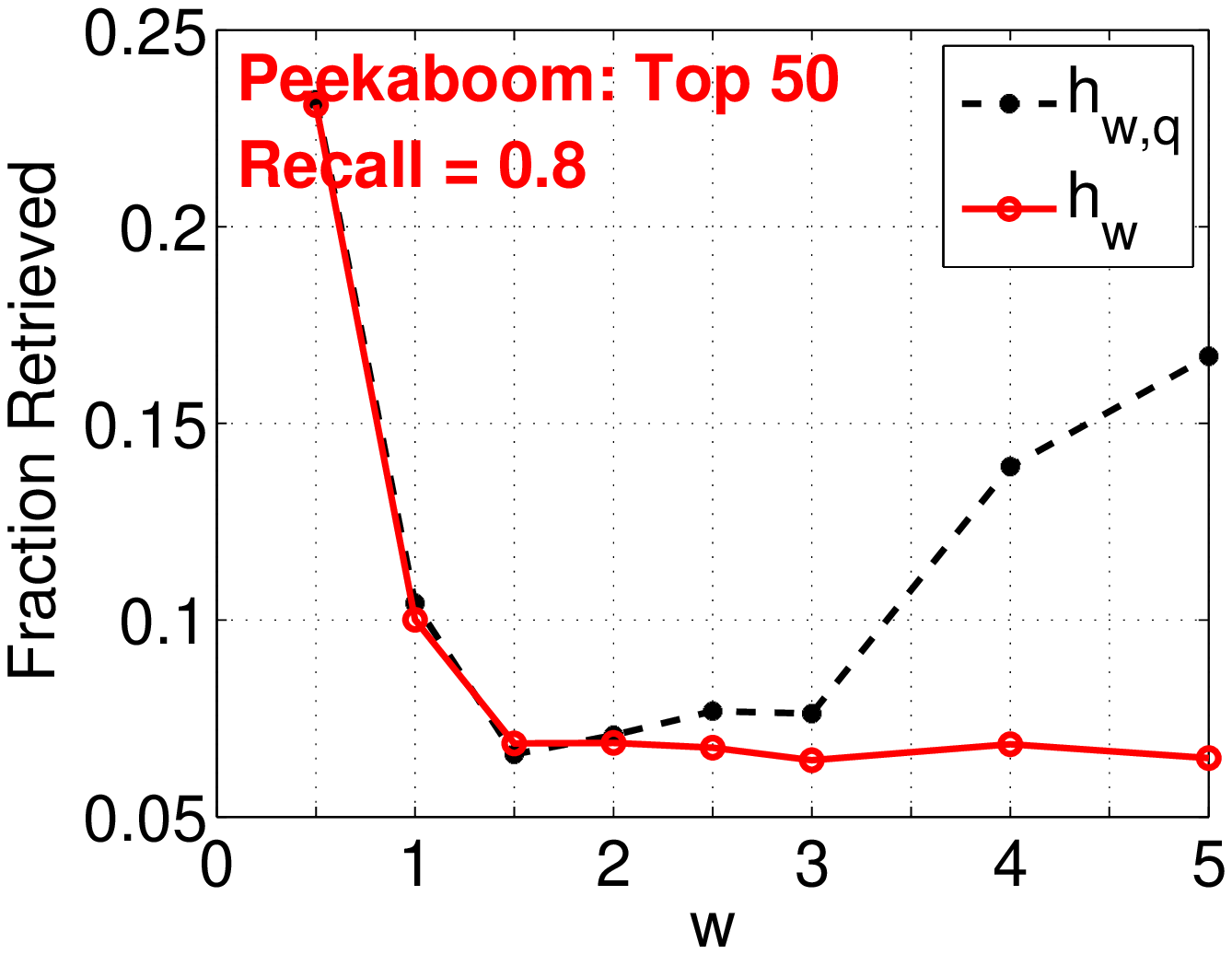}
}

\mbox{
\includegraphics[width = 2.7in]{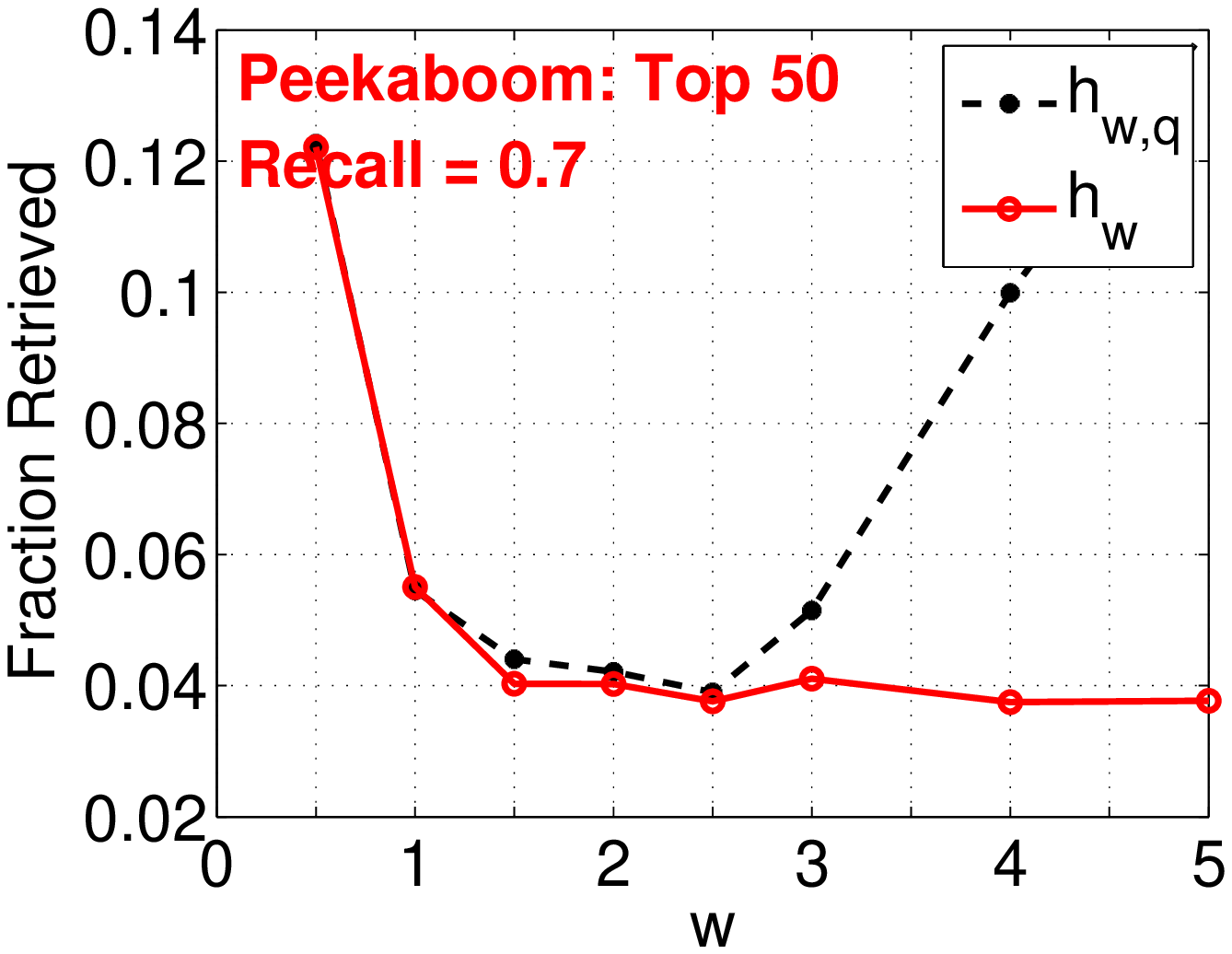}
\includegraphics[width = 2.7in]{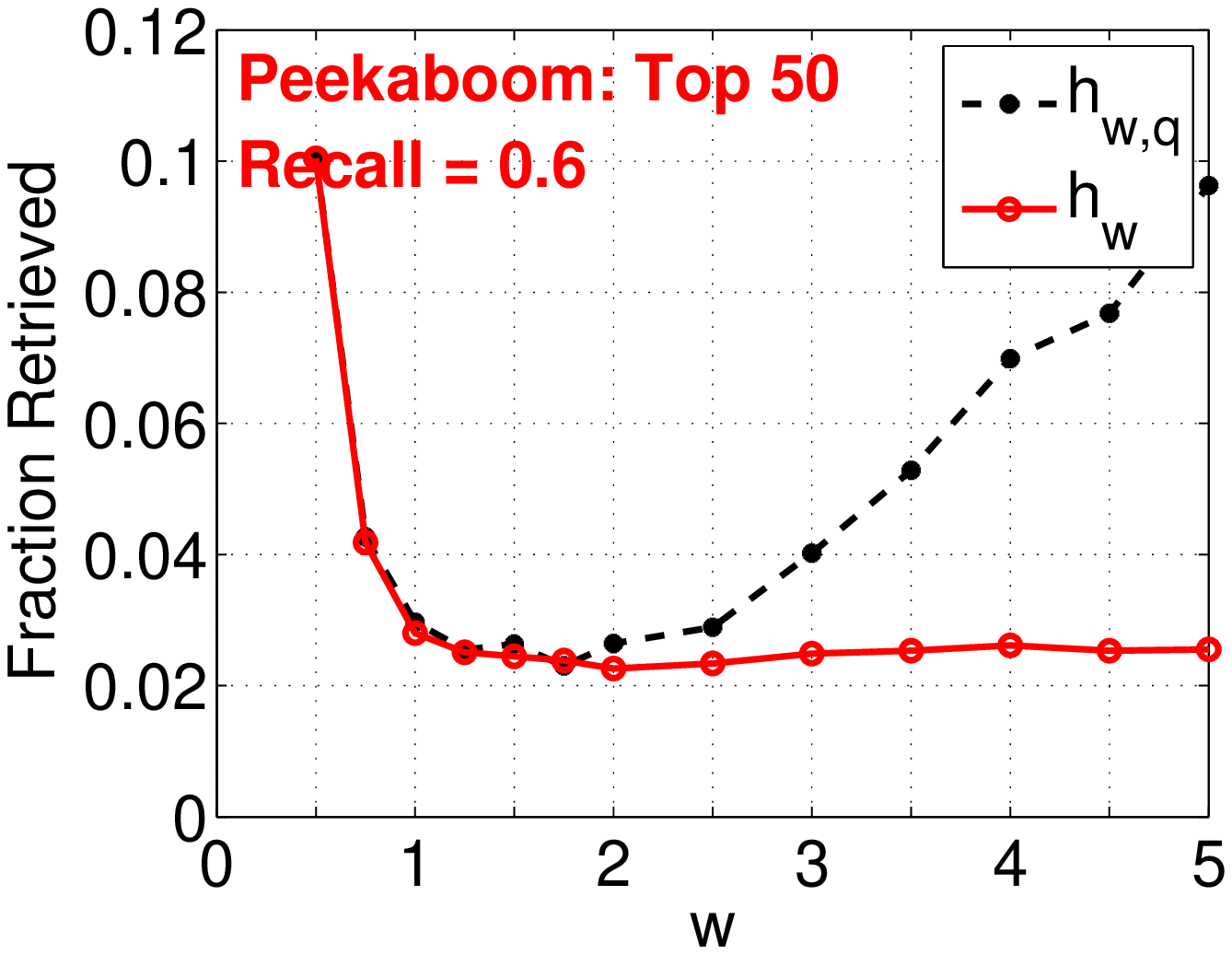}
}

\end{center}
\vspace{-.2in}
\caption{ \textbf{Peekaboom Top 50} . In each panel, we plot the optimal {\em fraction retrieved} at a target {\em recall} value (for top-50) with respect to $w$ for both coding schemes $h_w$ and $h_{w,q}$. }\label{fig_PeekaboomRecallvsWT50}
\end{figure}

\begin{figure}
\begin{center}
\mbox{
\includegraphics[width = 2.7in]{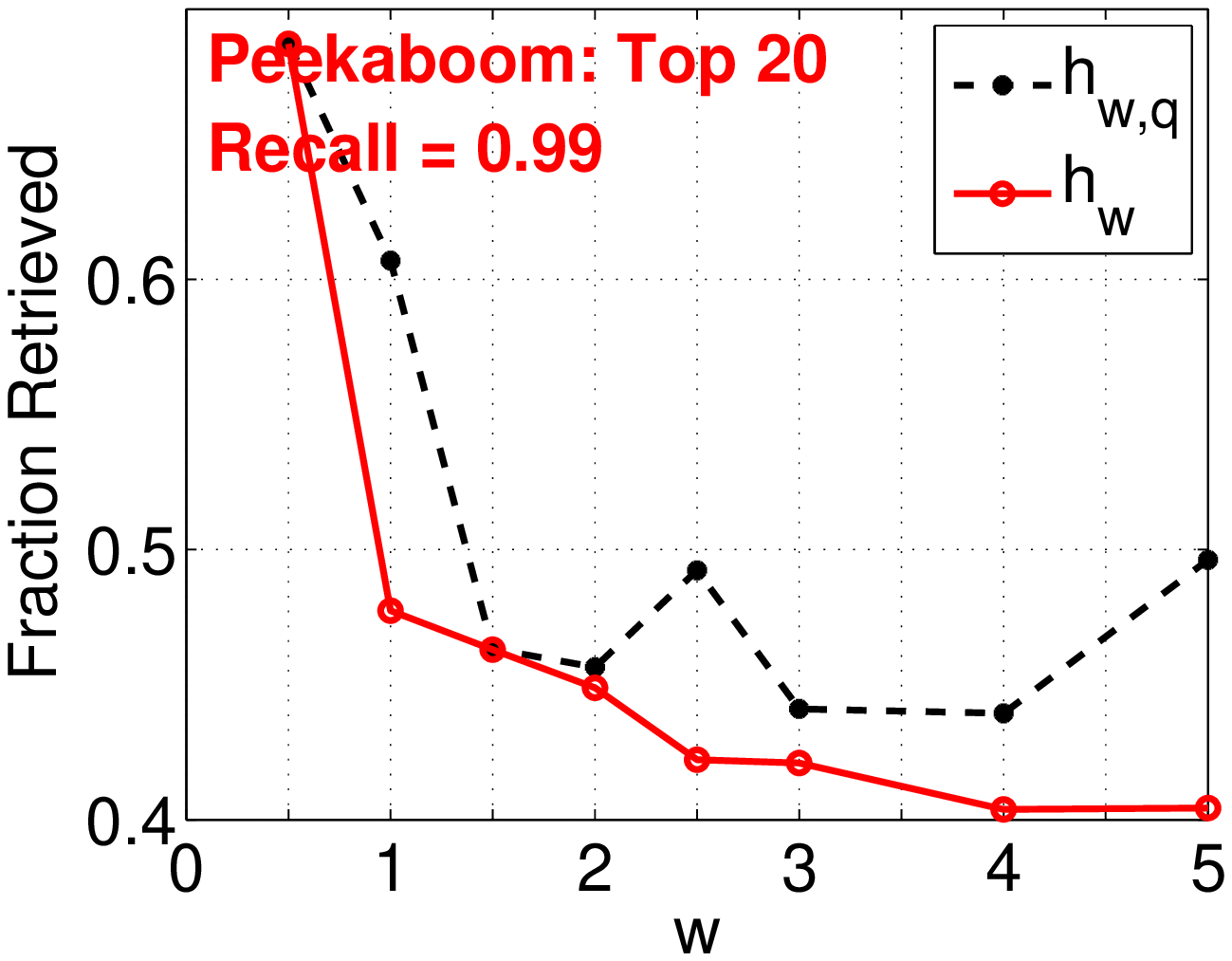}
\includegraphics[width = 2.7in]{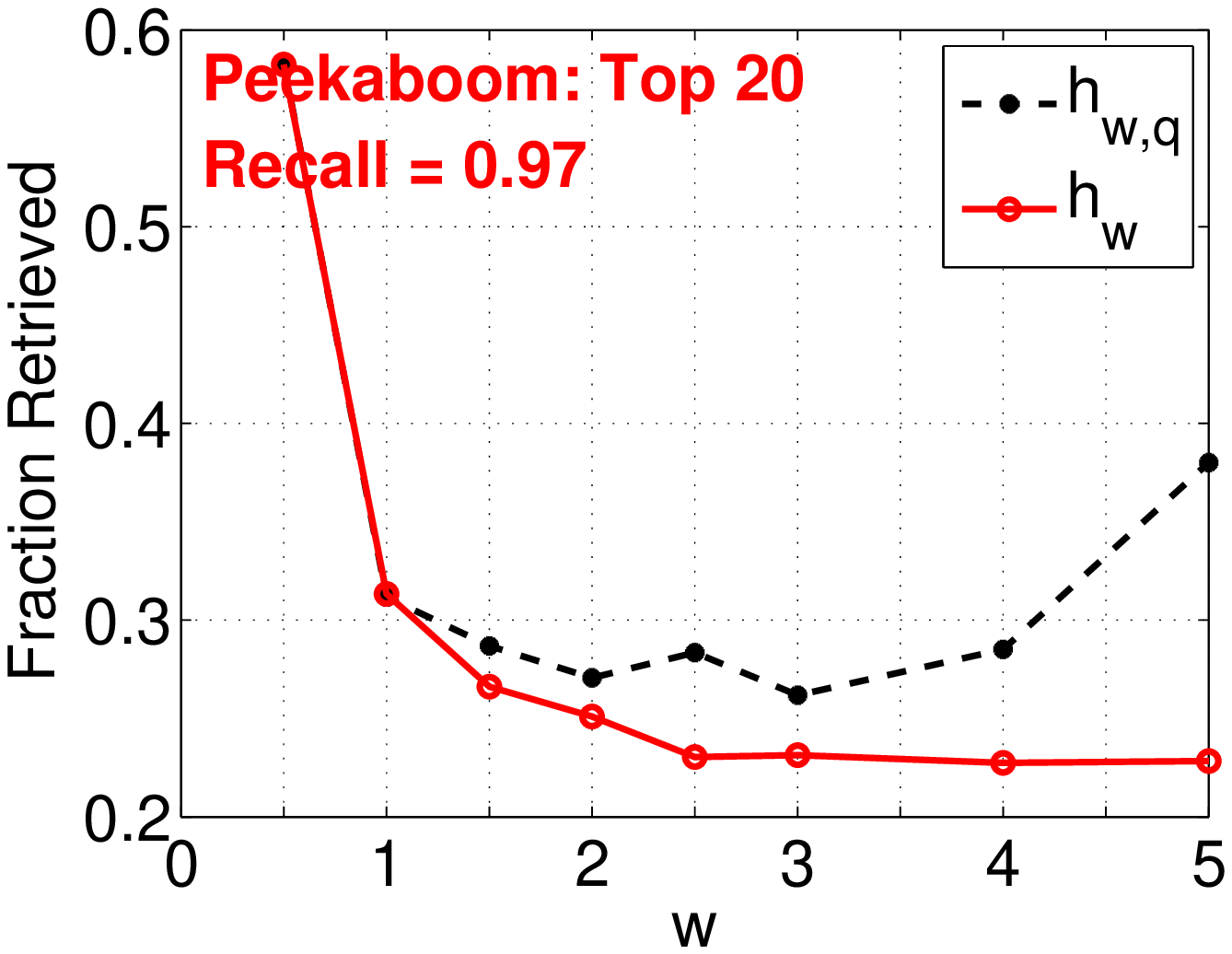}
}
\mbox{
\includegraphics[width = 2.7in]{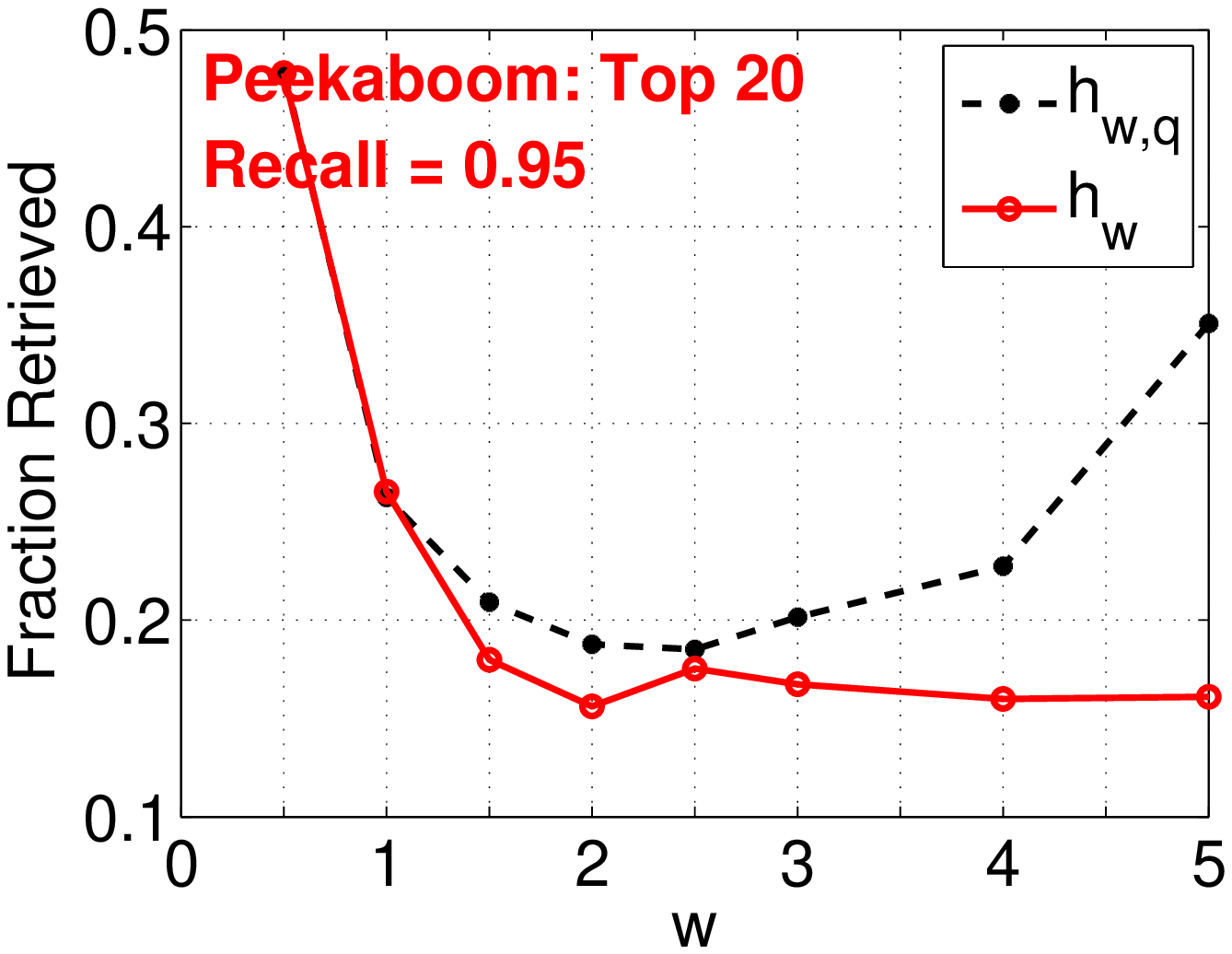}
\includegraphics[width = 2.7in]{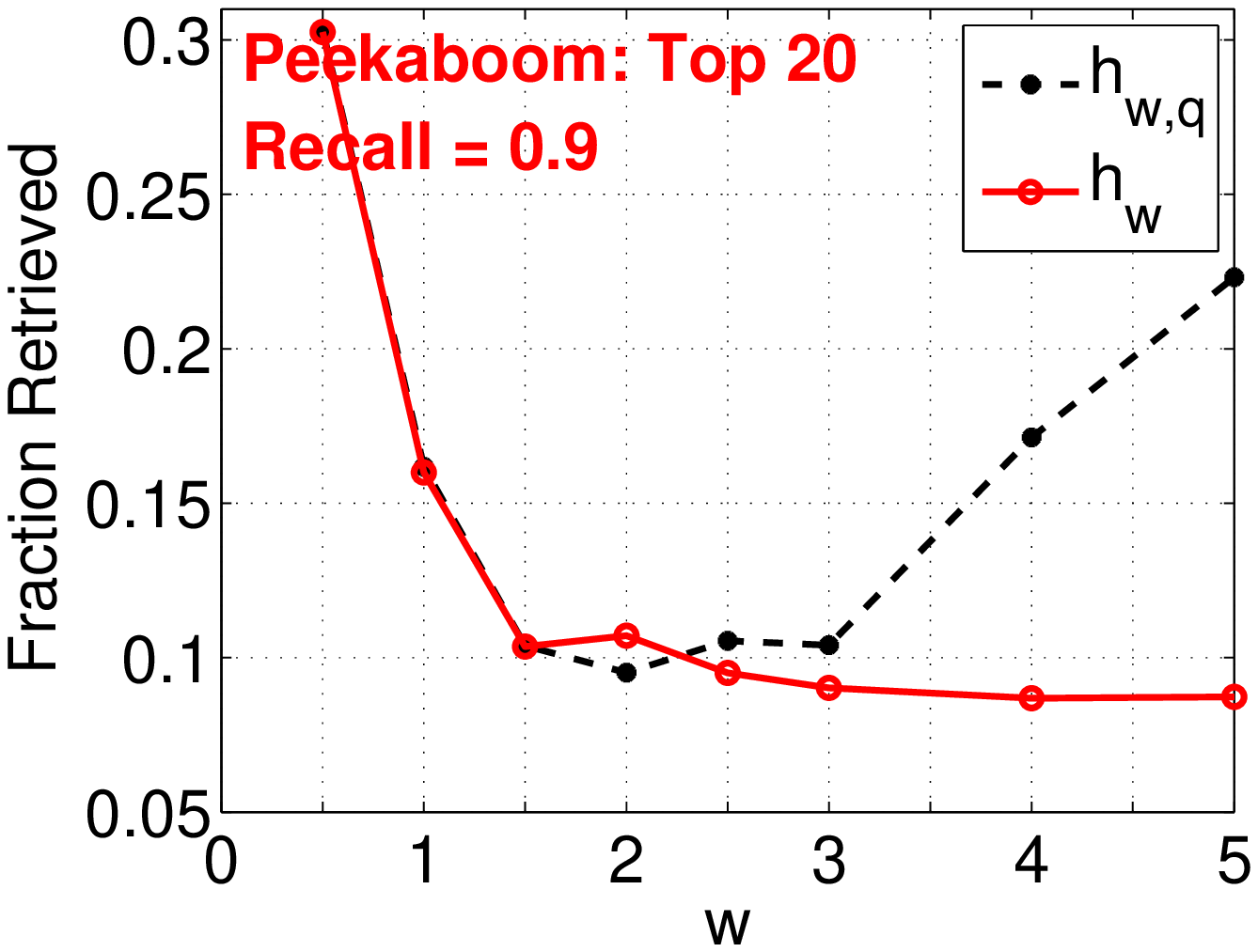}
}

\mbox{
\includegraphics[width = 2.7in]{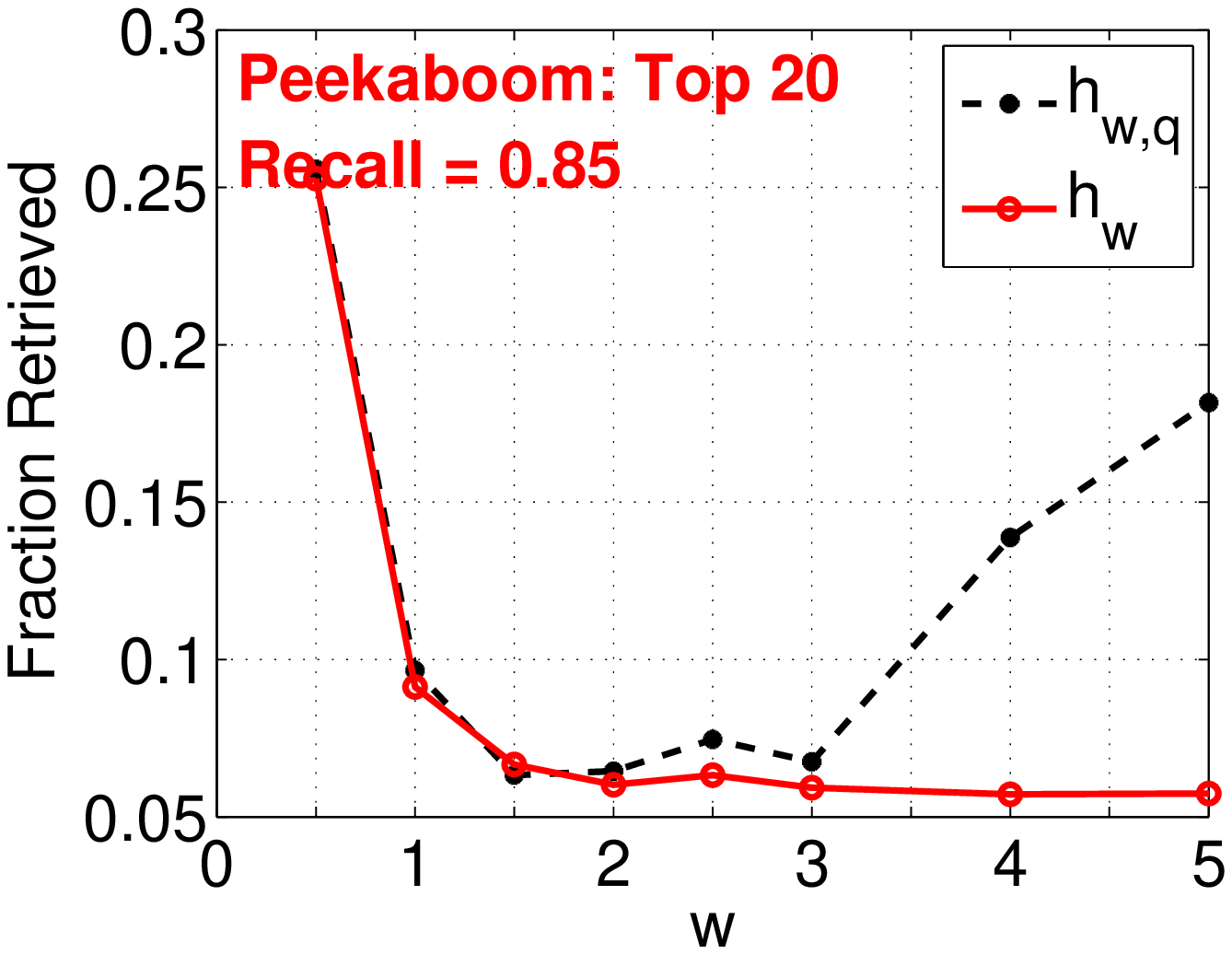}
\includegraphics[width = 2.7in]{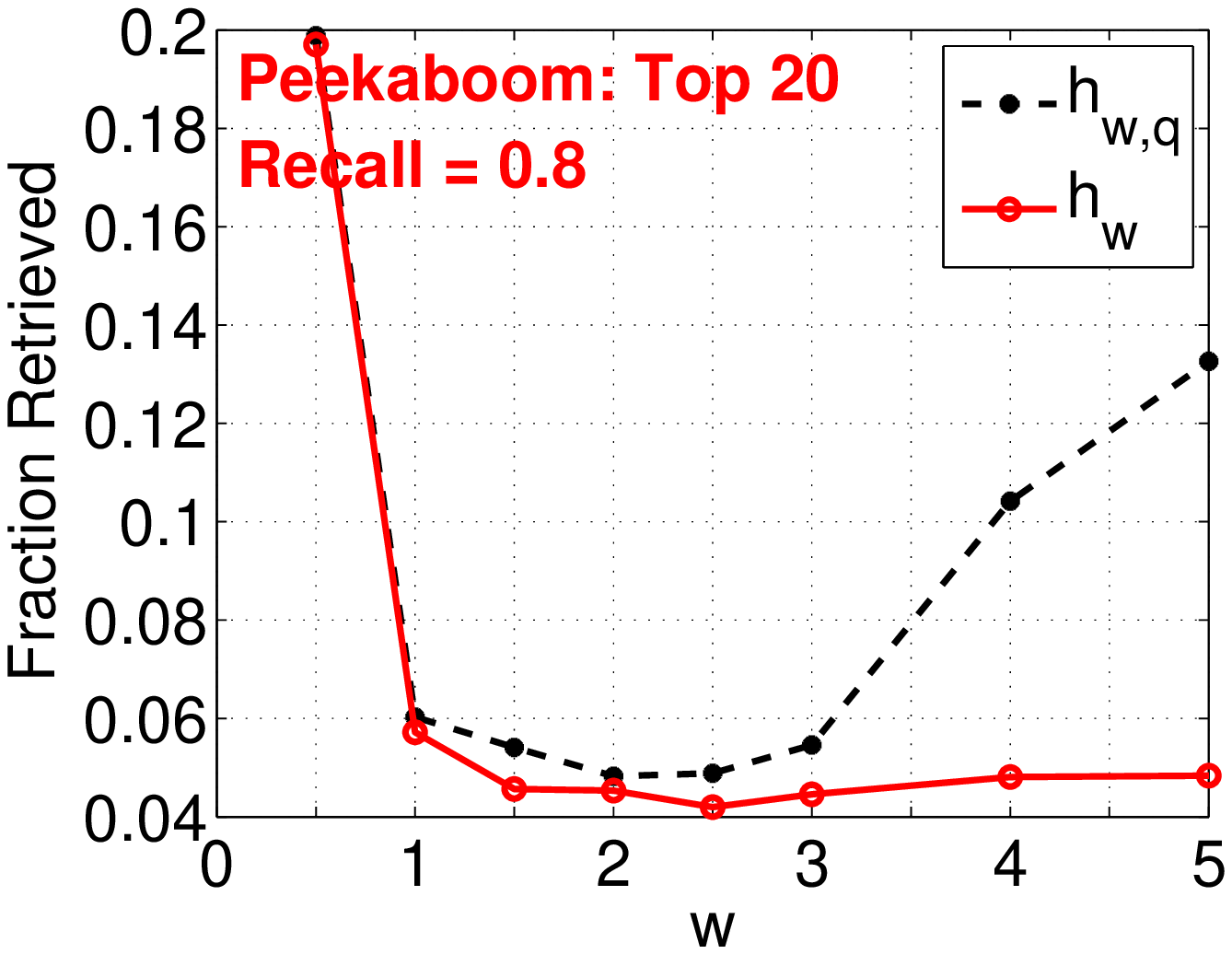}
}

\mbox{
\includegraphics[width = 2.7in]{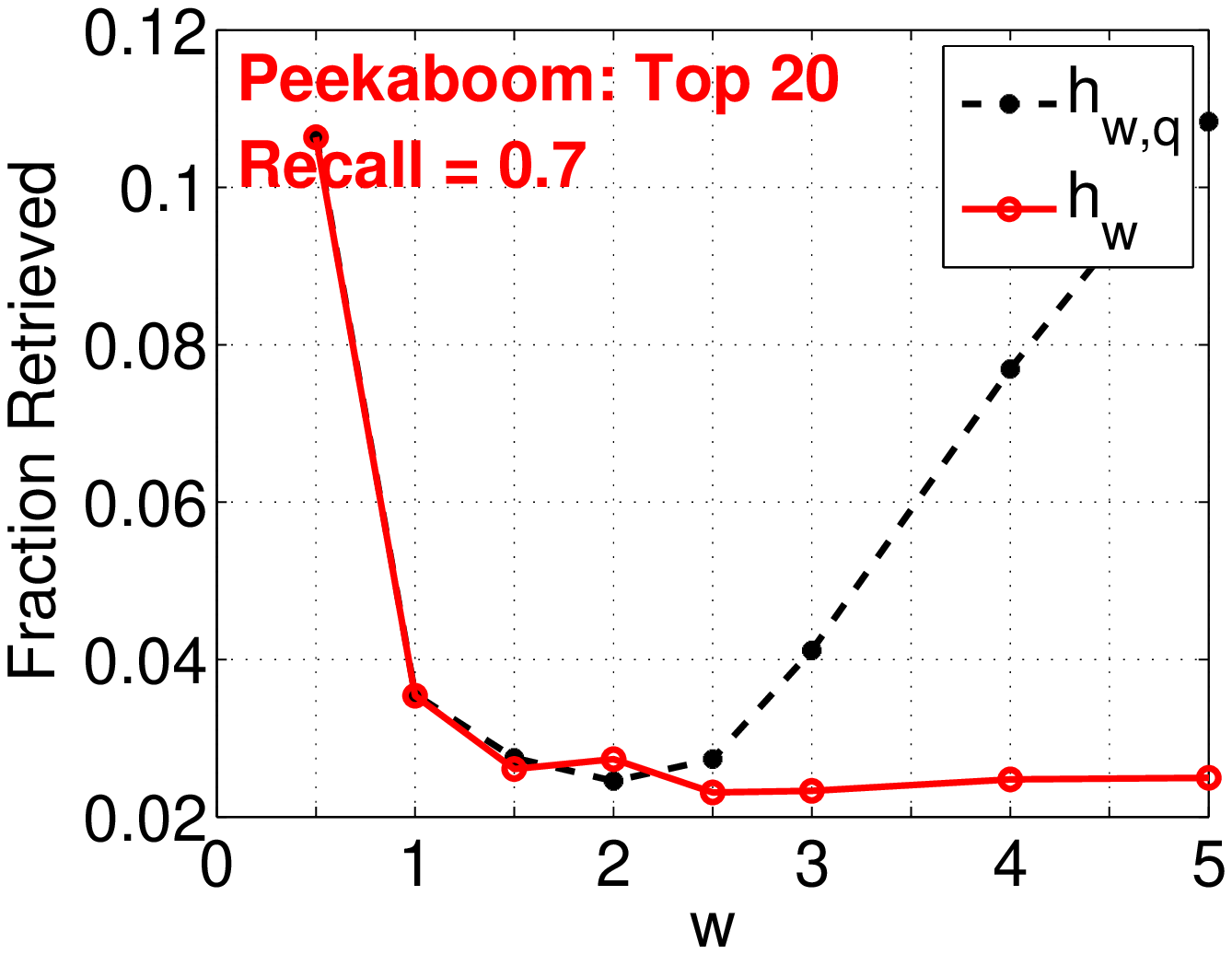}
\includegraphics[width = 2.7in]{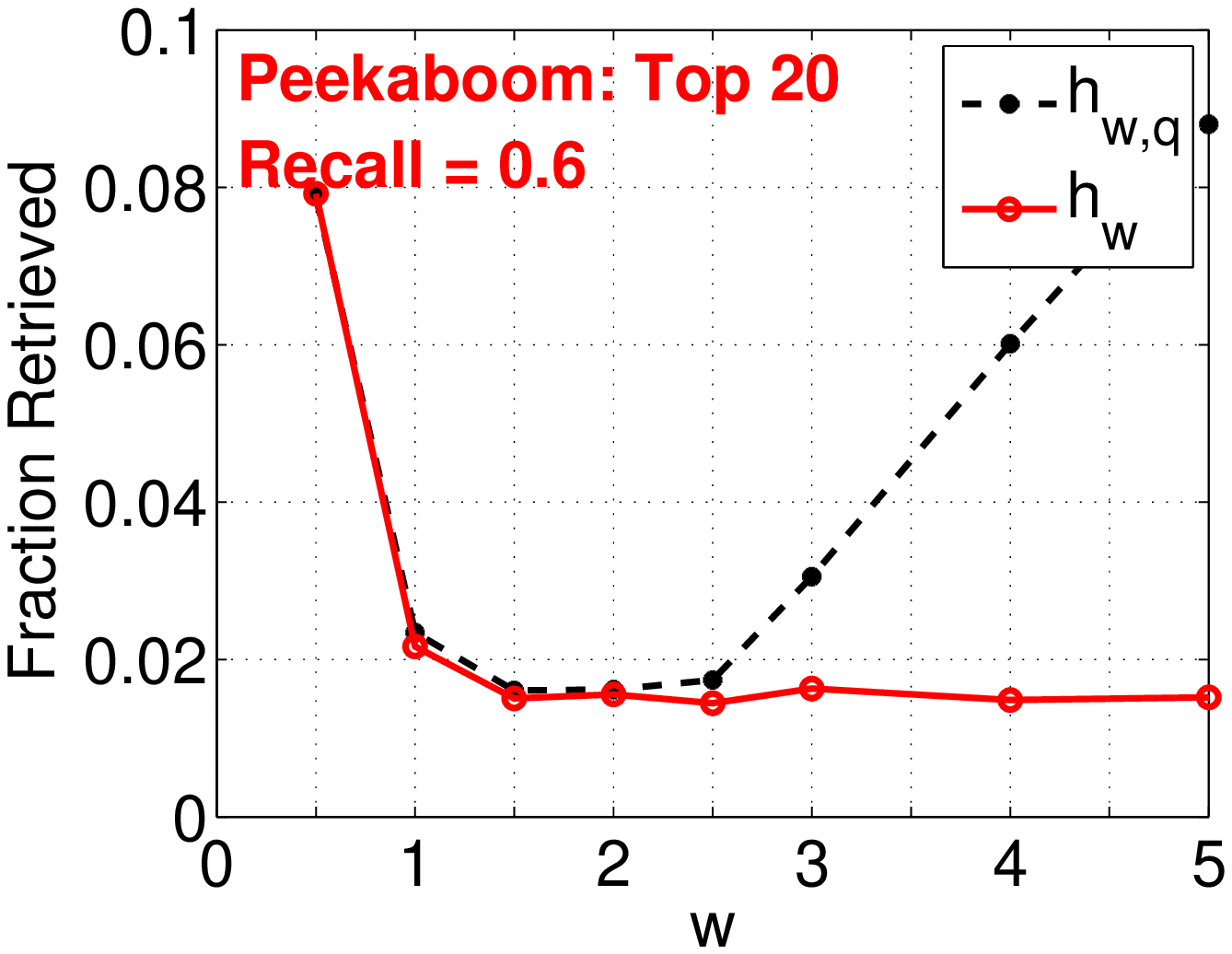}
}

\end{center}
\vspace{-.2in}
\caption{ \textbf{Peekaboom Top 20} . In each panel, we plot the optimal {\em fraction retrieved} at a target {\em recall} value (for top-20) with respect to $w$ for both coding schemes $h_w$ and $h_{w,q}$. }\label{fig_PeekaboomRecallvsWT20}
\end{figure}

\begin{figure}
\begin{center}
\mbox{
\includegraphics[width = 2.7in]{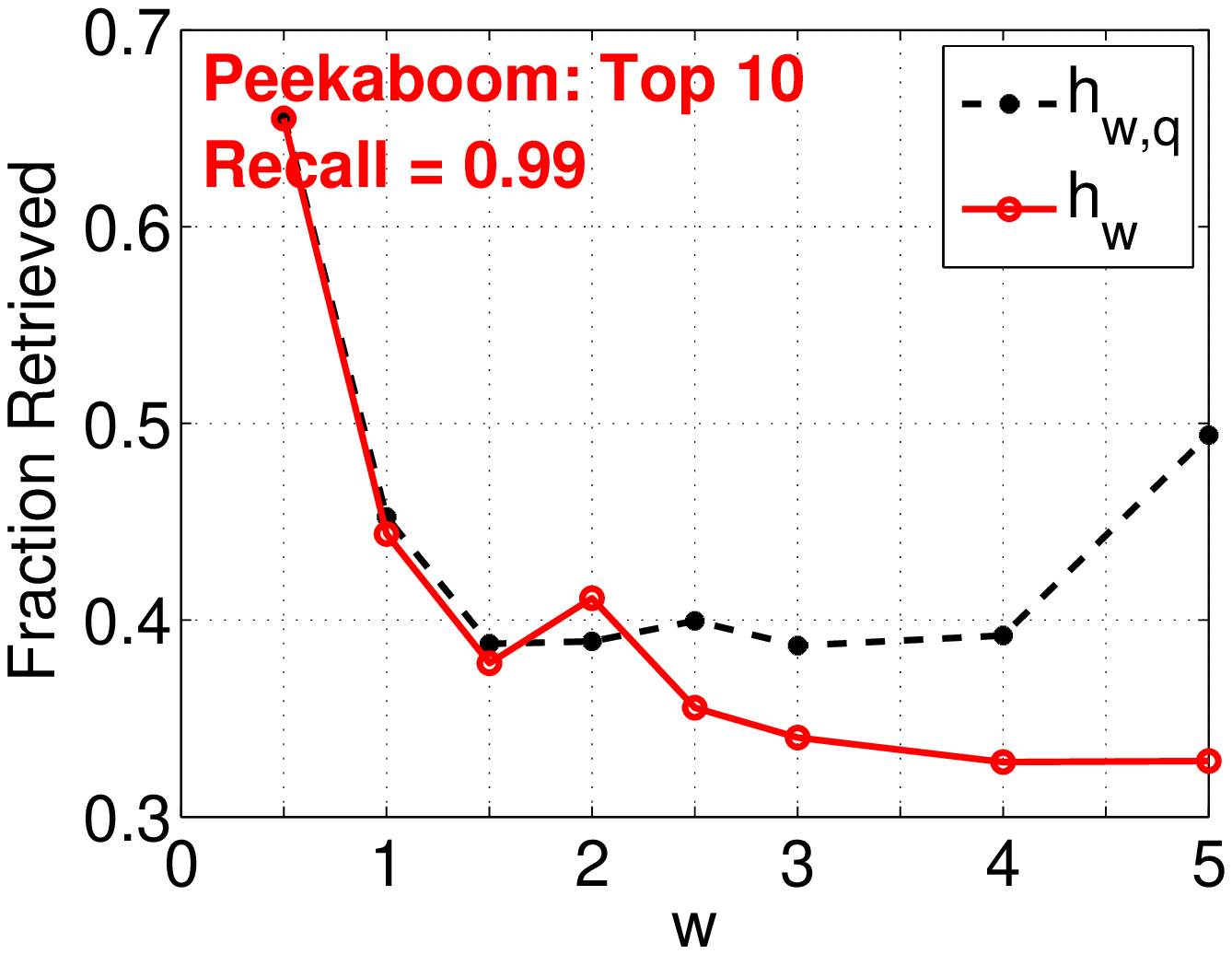}
\includegraphics[width = 2.7in]{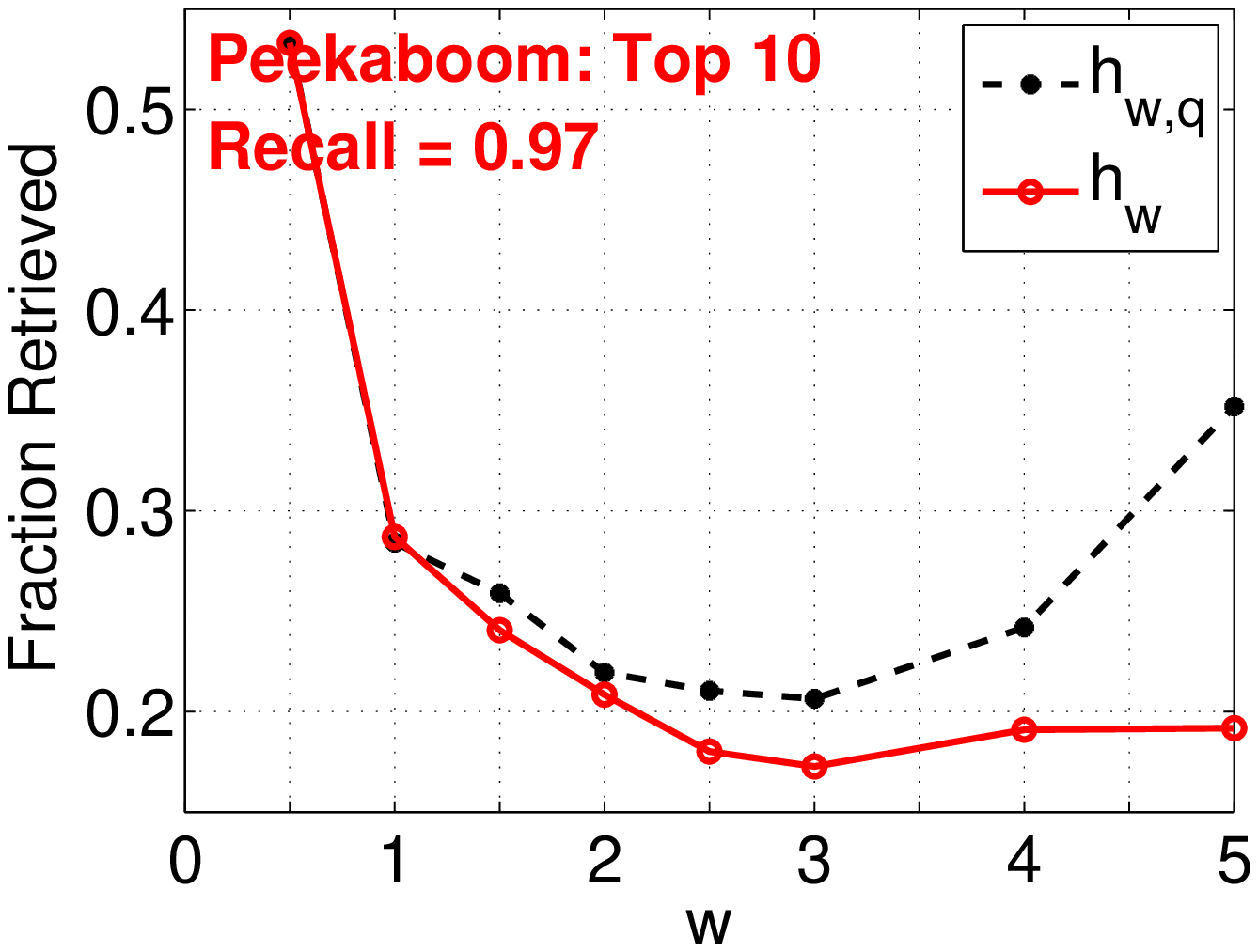}
}
\mbox{
\includegraphics[width = 2.7in]{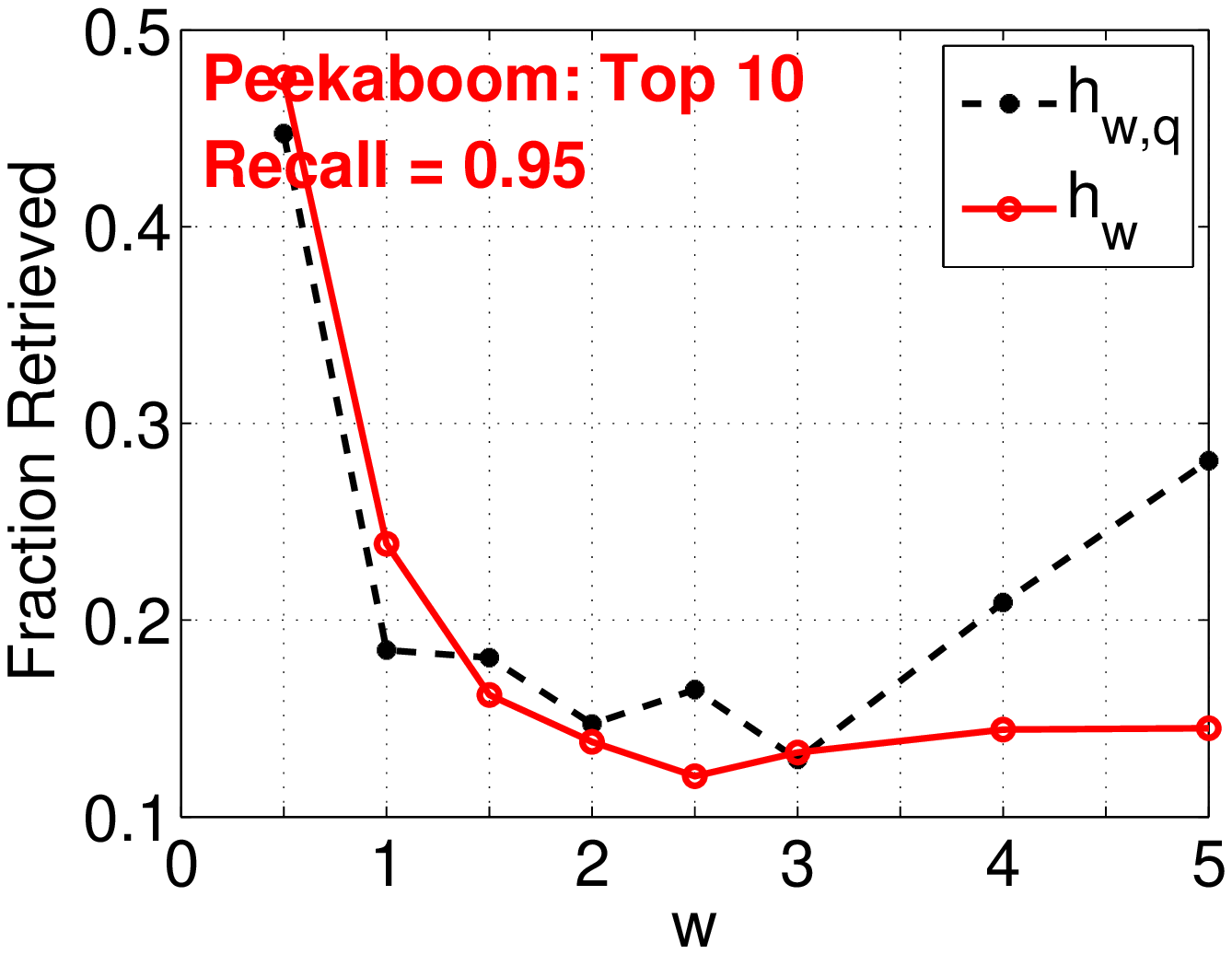}
\includegraphics[width = 2.7in]{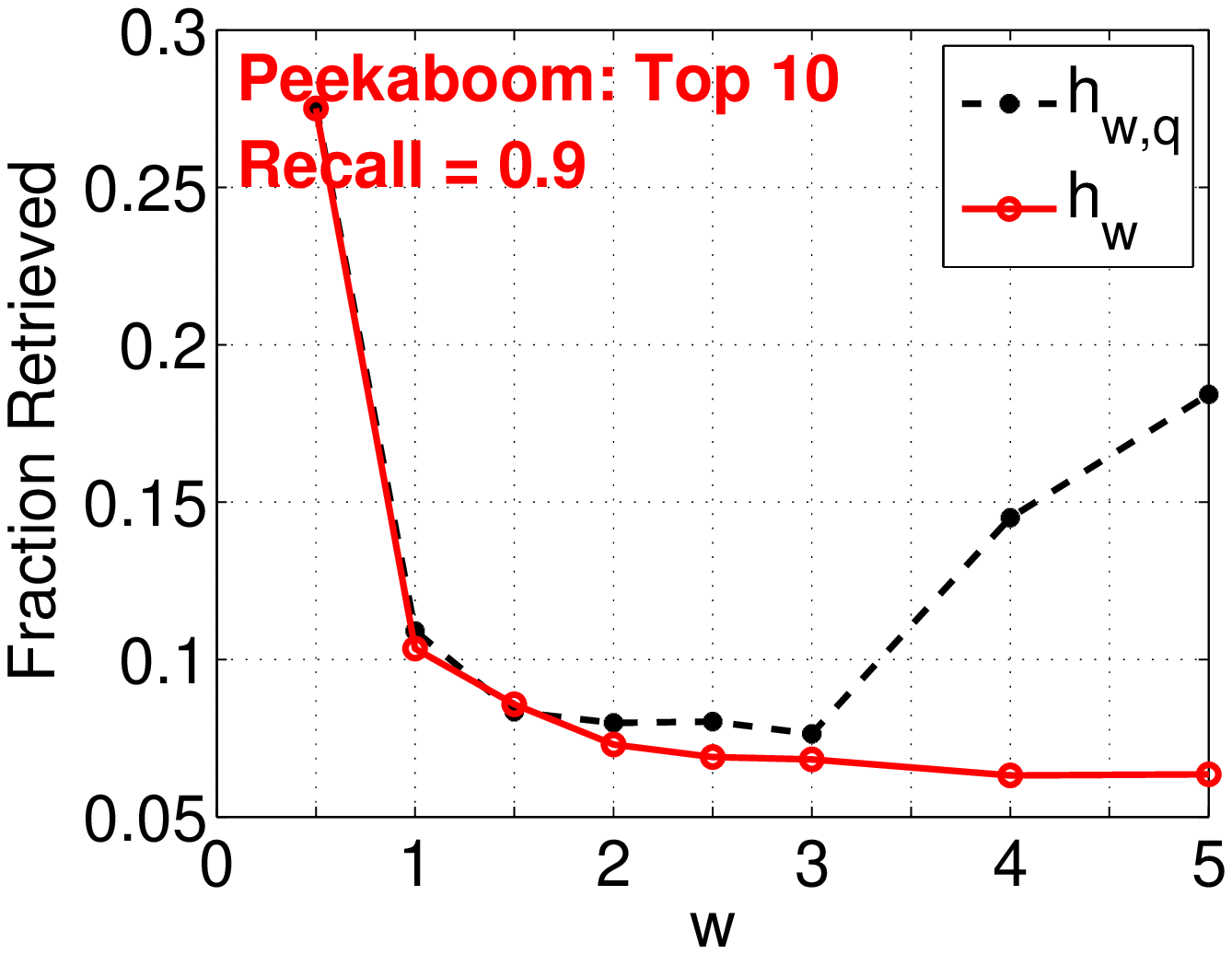}
}

\mbox{
\includegraphics[width = 2.7in]{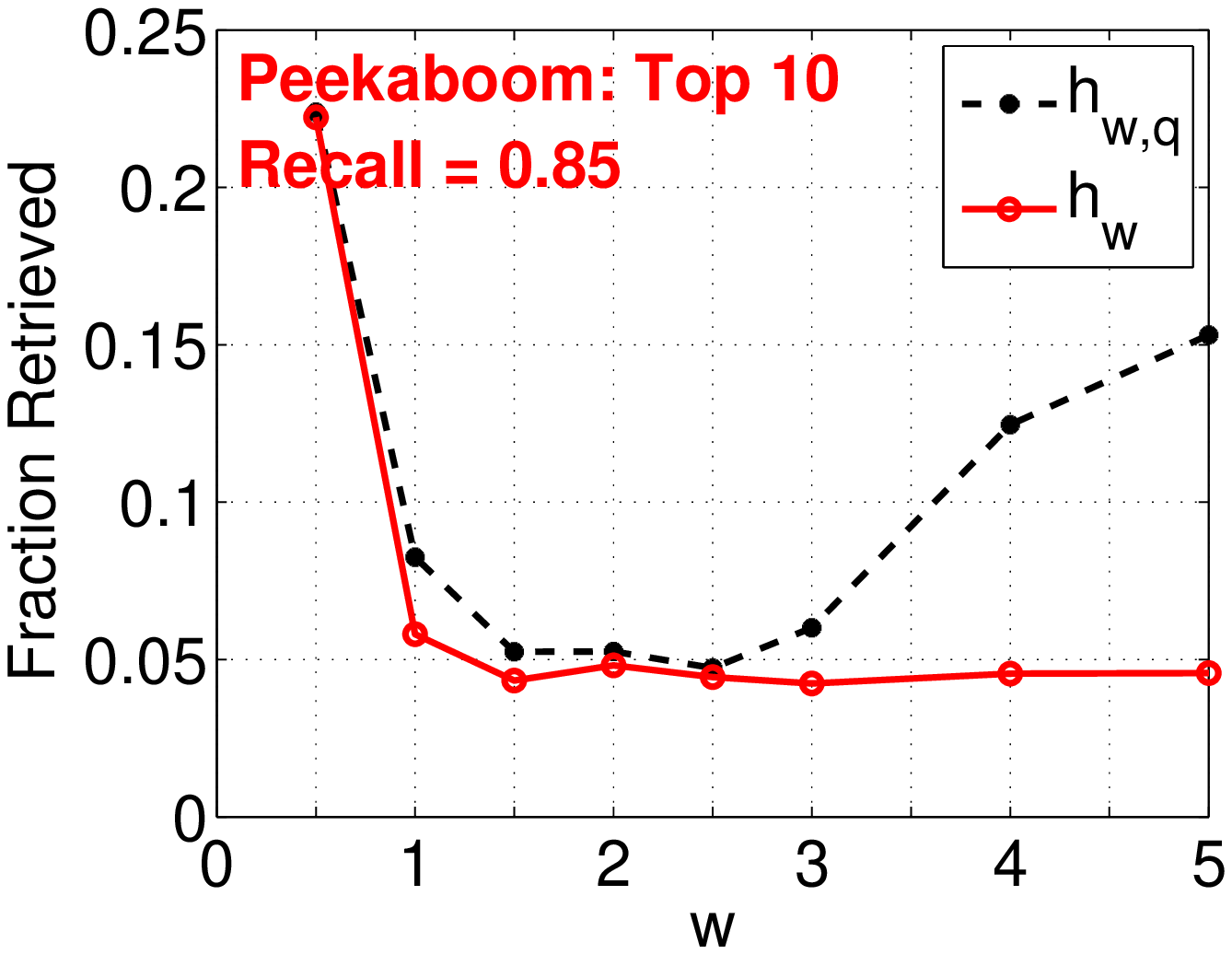}
\includegraphics[width = 2.7in]{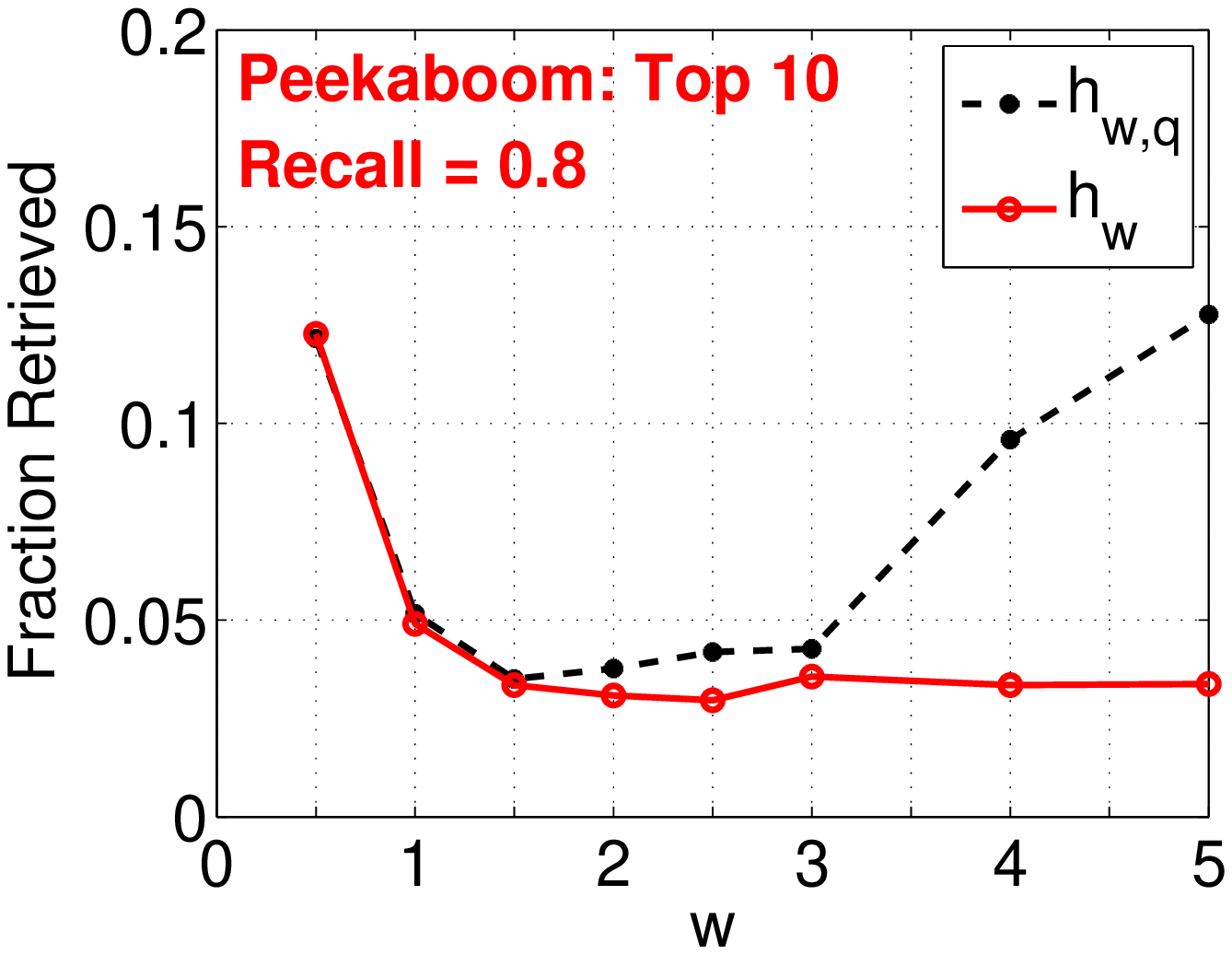}
}

\mbox{
\includegraphics[width = 2.7in]{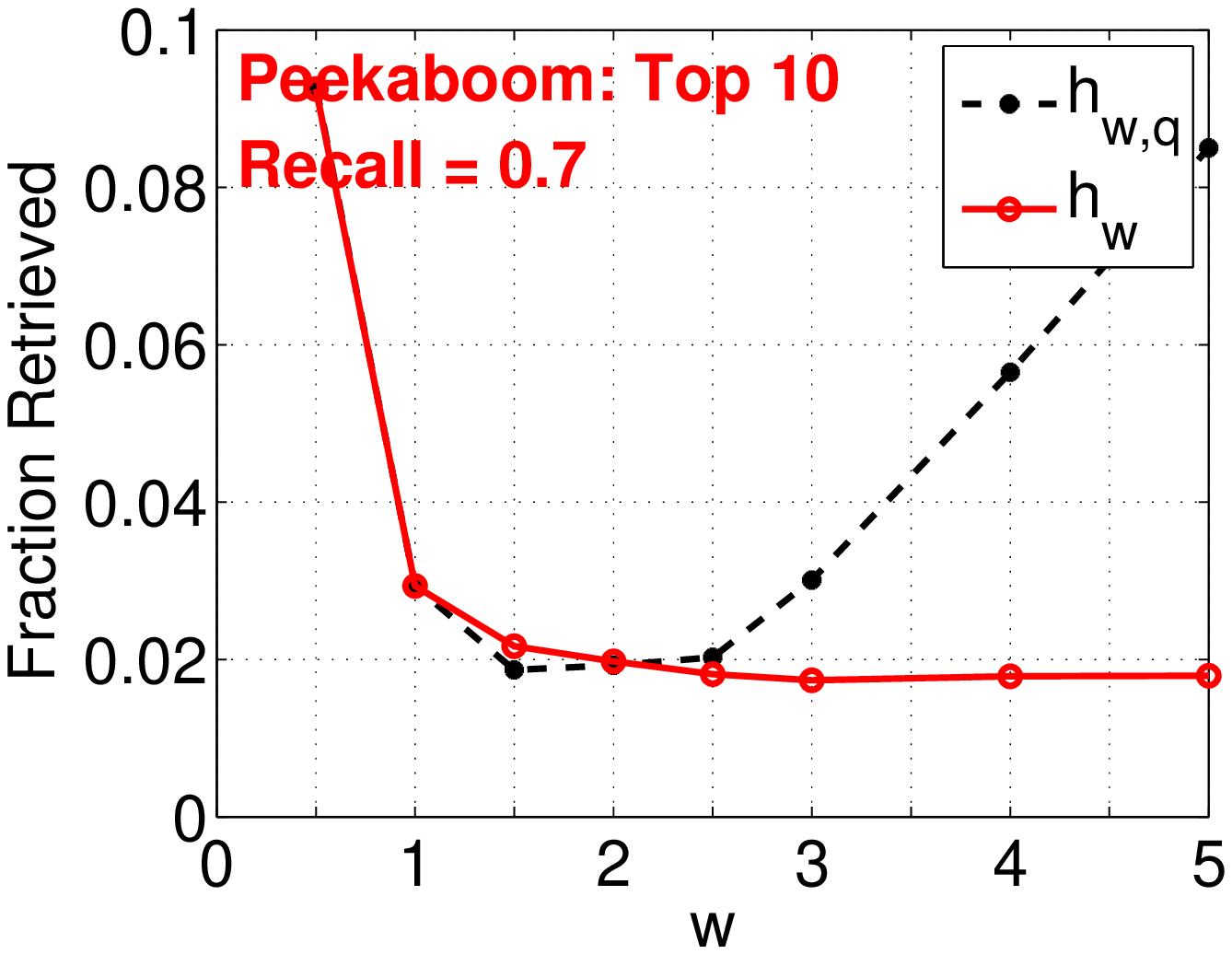}
\includegraphics[width = 2.7in]{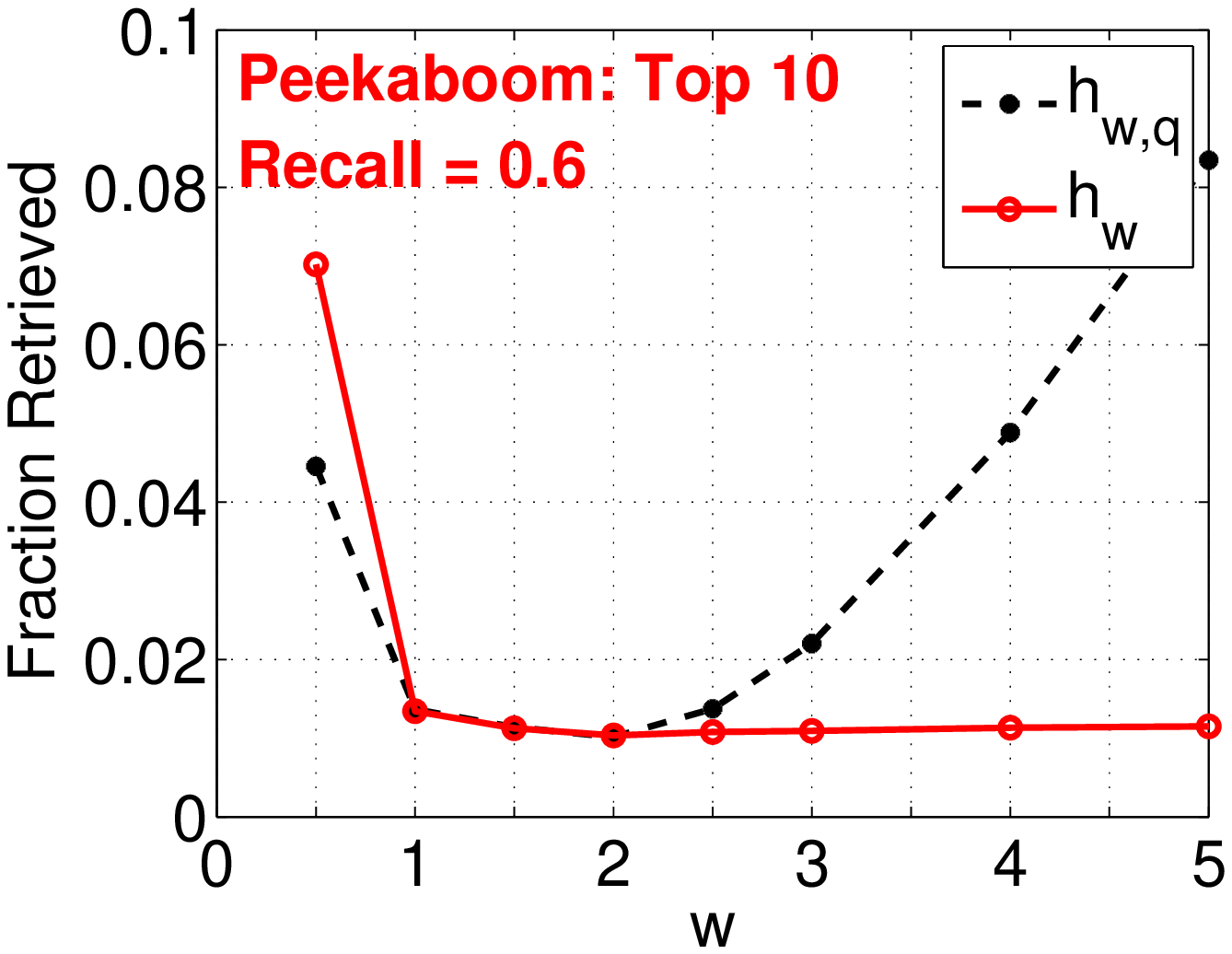}
}

\end{center}
\vspace{-.2in}
\caption{ \textbf{Peekaboom Top 10} . In each panel, we plot the optimal {\em fraction retrieved} at a target {\em recall} value (for top-10) with respect to $w$ for both coding schemes $h_w$ and $h_{w,q}$. }\label{fig_PeekaboomRecallvsWT10}
\end{figure}

\begin{figure}
\begin{center}
\mbox{
\includegraphics[width = 2.7in]{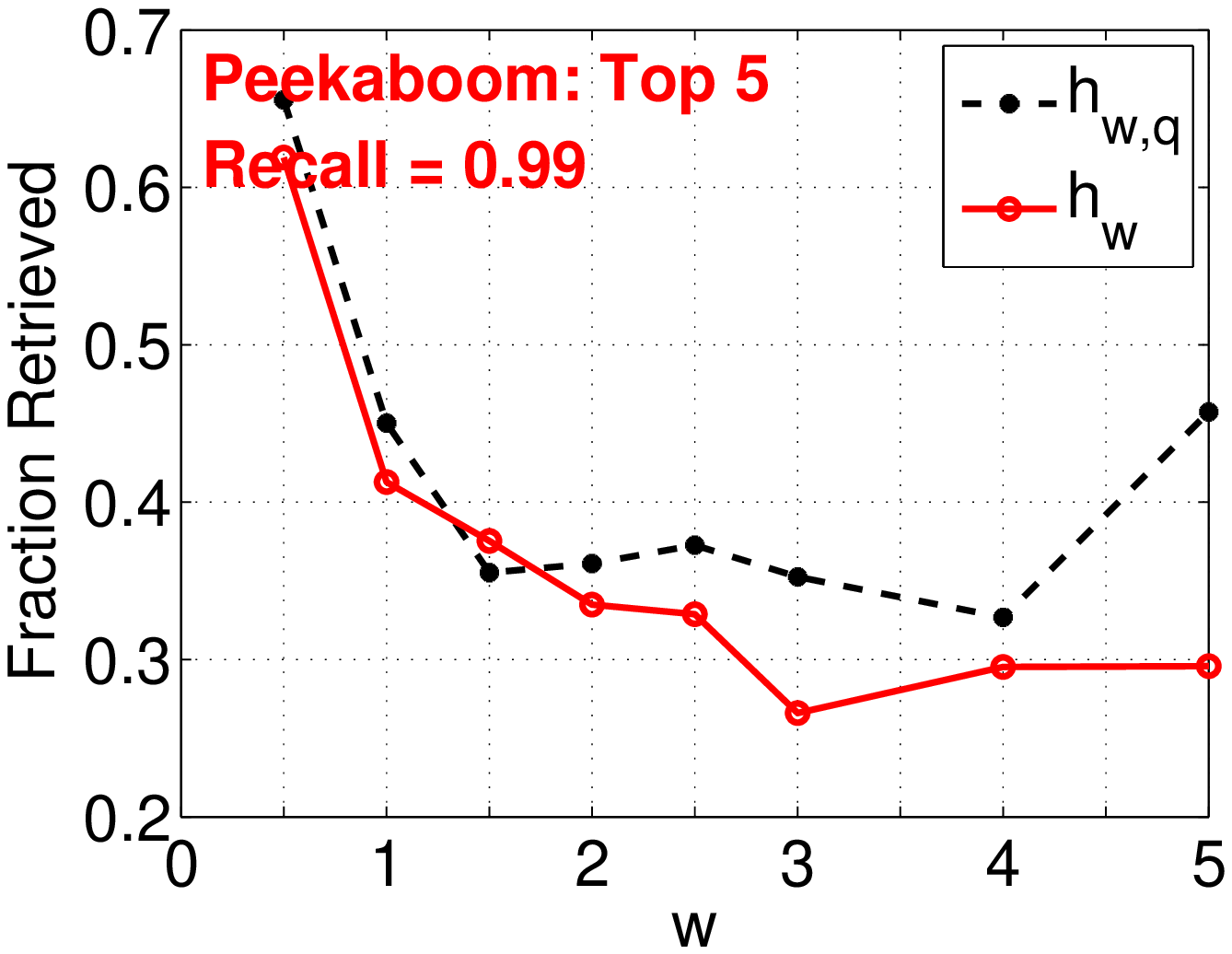}
\includegraphics[width = 2.7in]{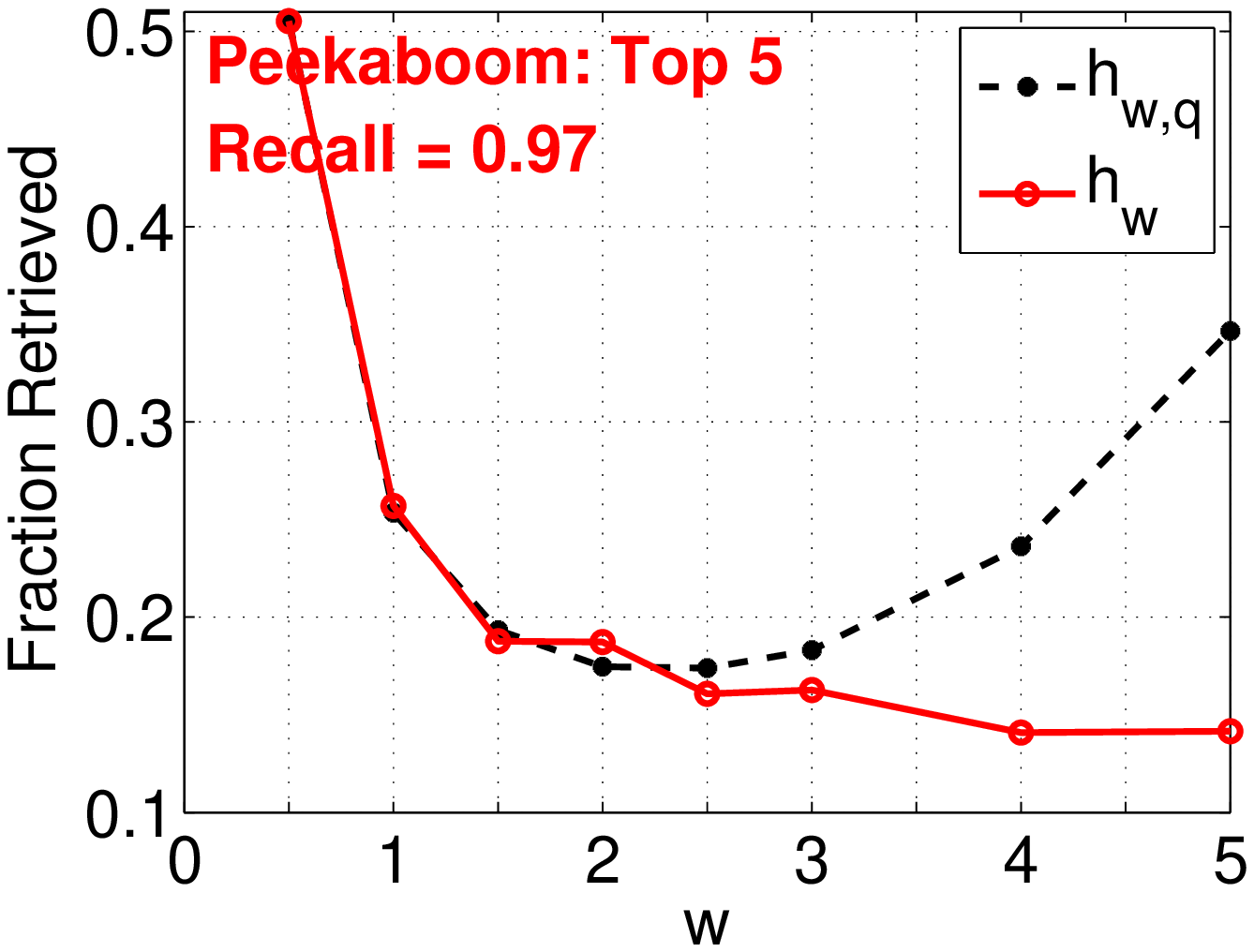}
}
\mbox{
\includegraphics[width = 2.7in]{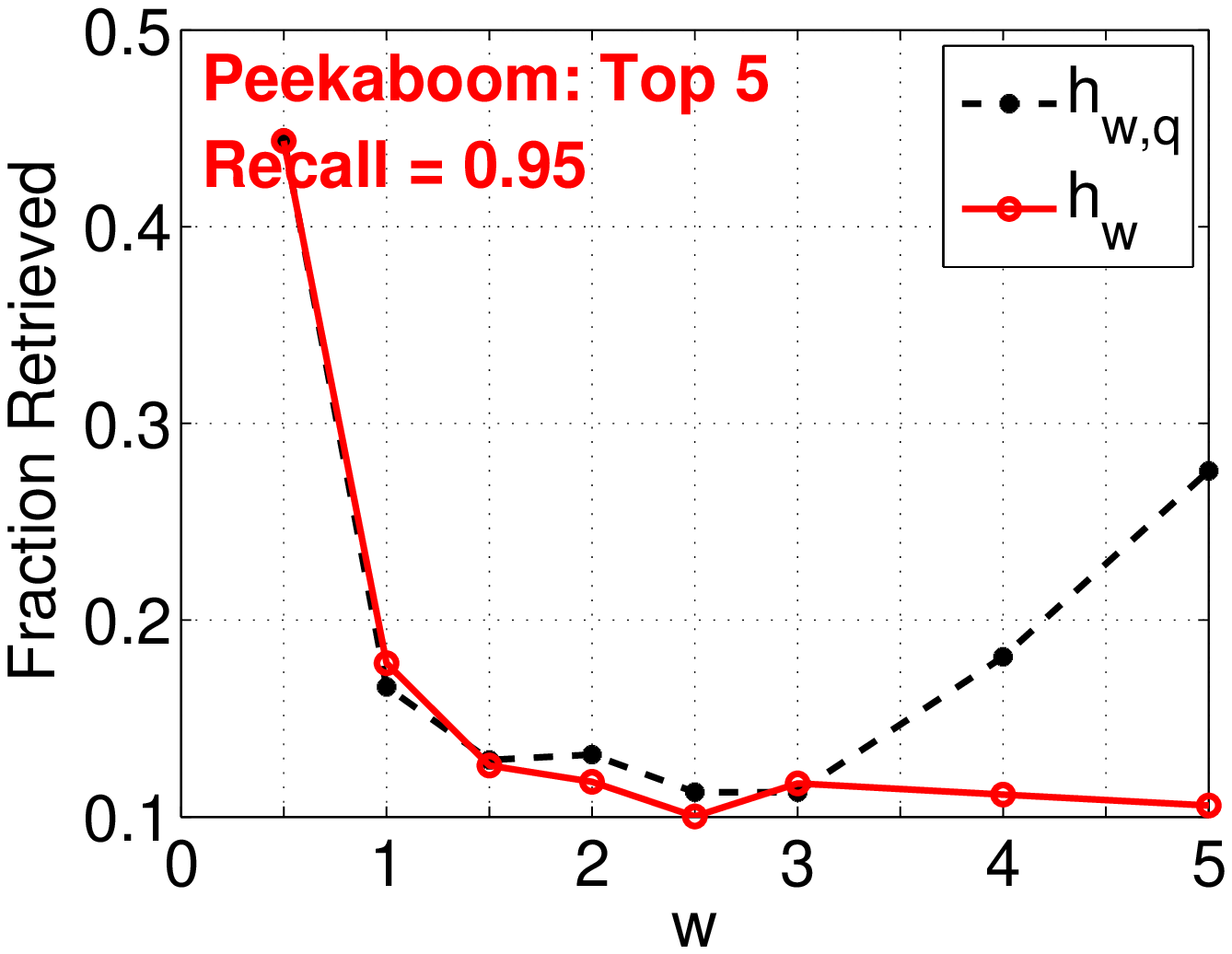}
\includegraphics[width = 2.7in]{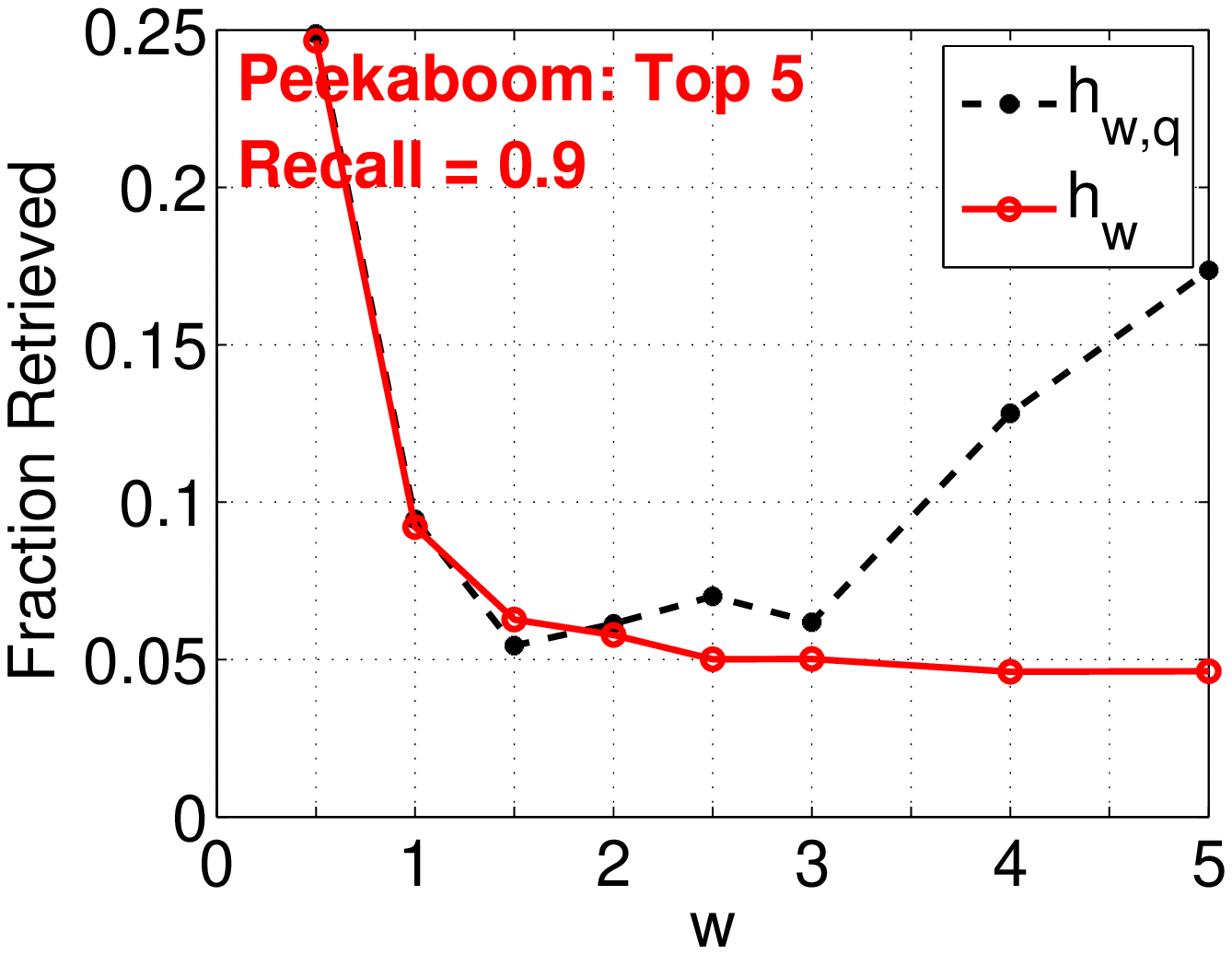}
}

\mbox{
\includegraphics[width = 2.7in]{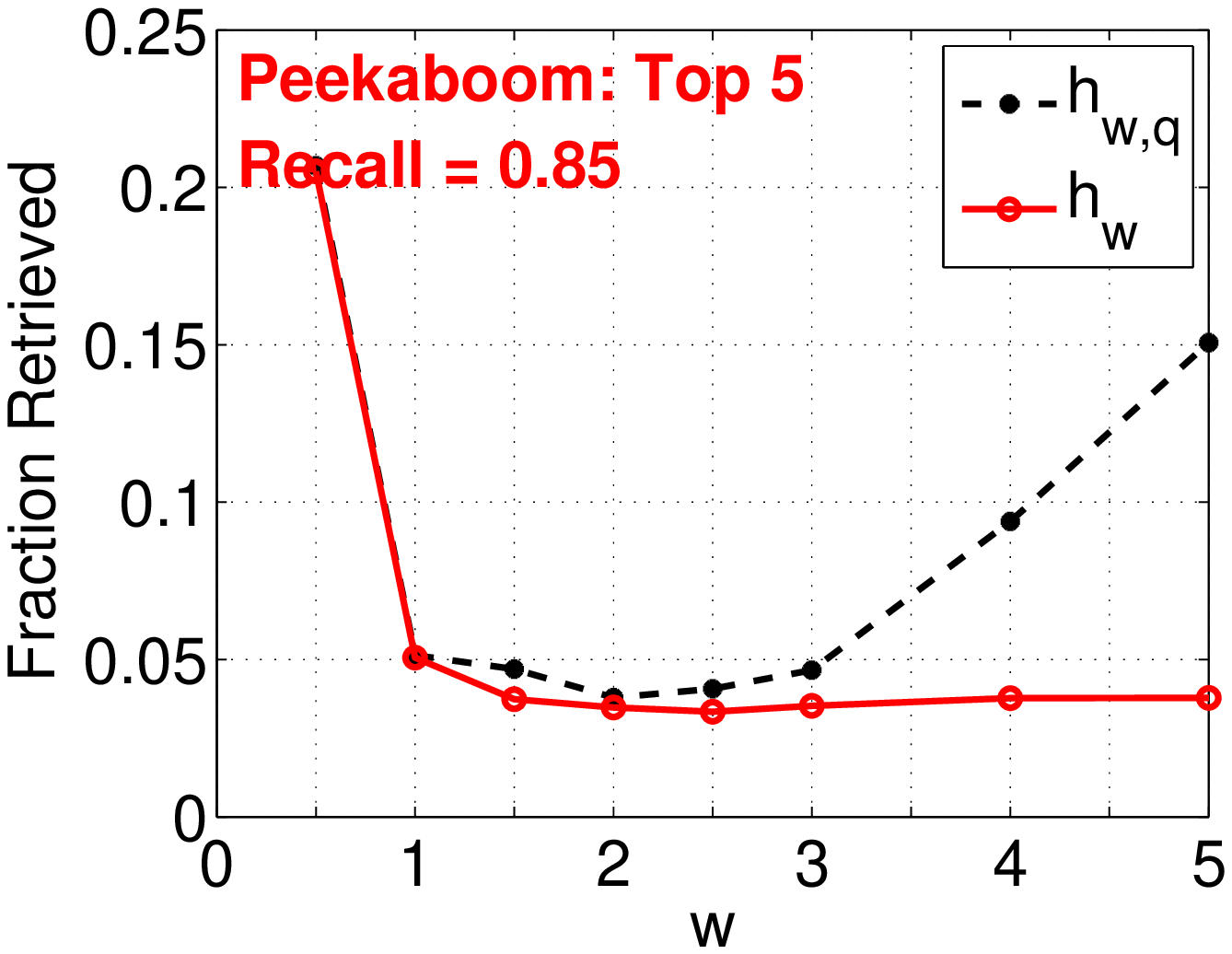}
\includegraphics[width = 2.7in]{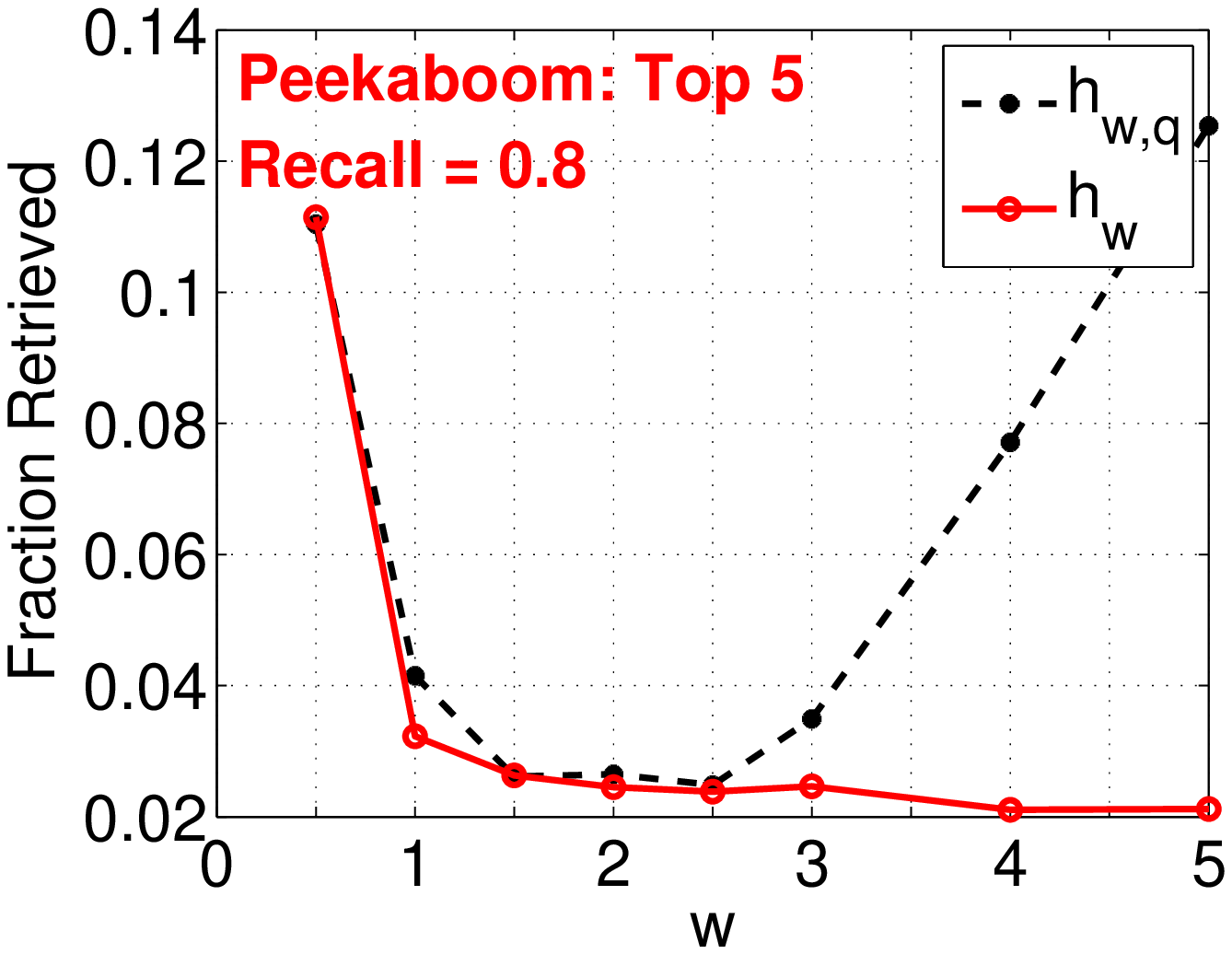}
}

\mbox{
\includegraphics[width = 2.7in]{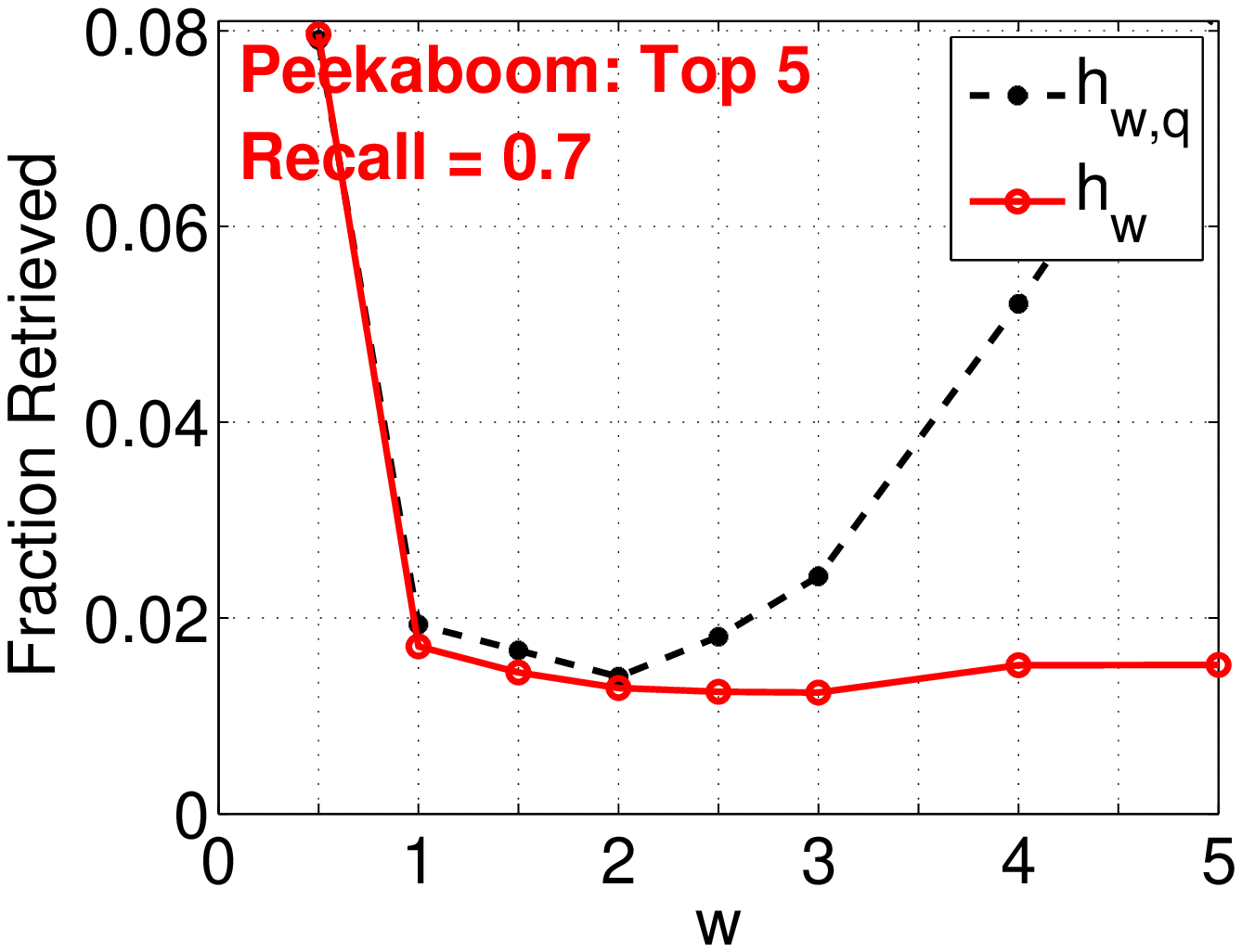}
\includegraphics[width = 2.7in]{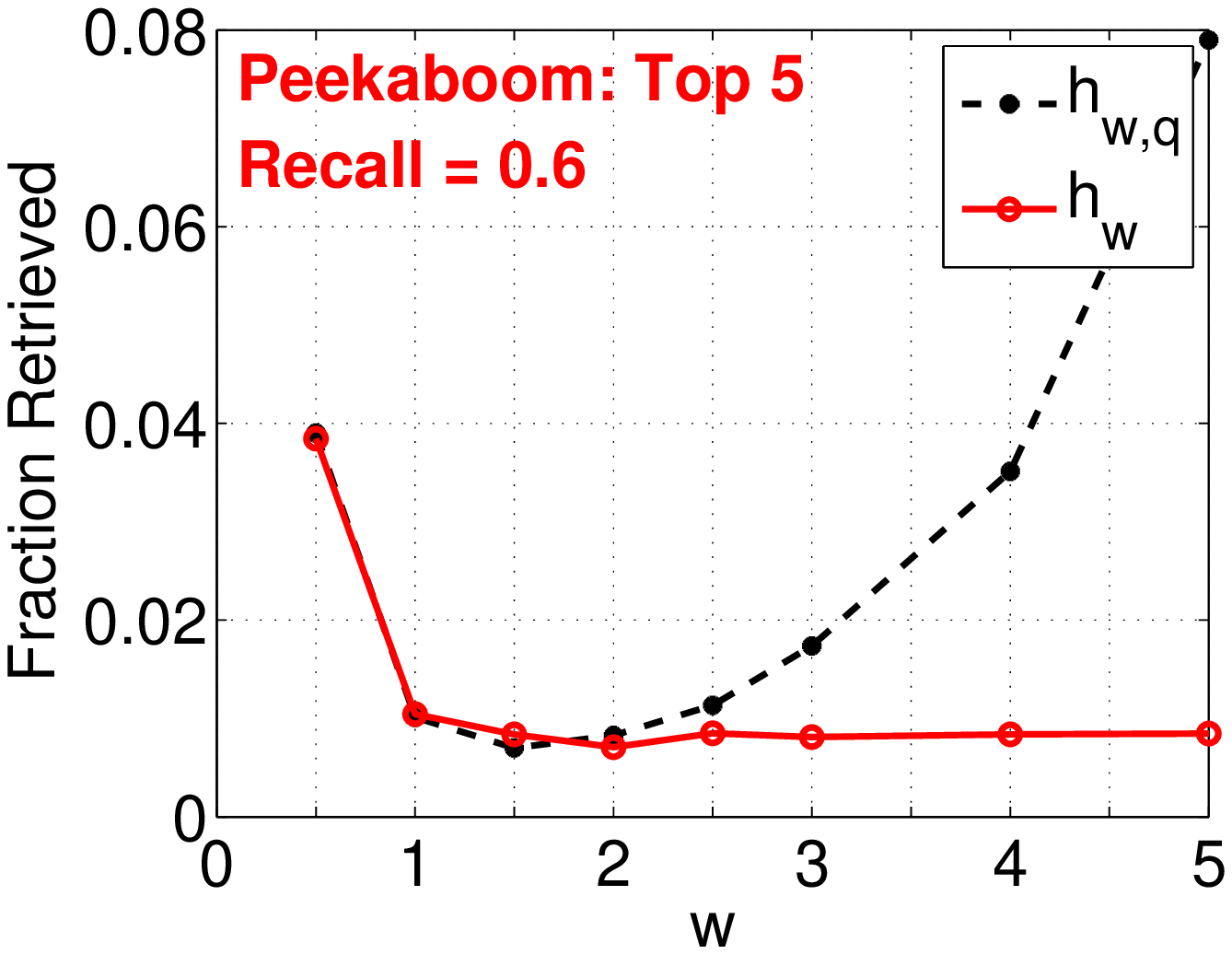}
}

\end{center}
\vspace{-.2in}
\caption{ \textbf{Peekaboom Top 5} . In each panel, we plot the optimal {\em fraction retrieved} at a target {\em recall} value (for top-5) with respect to $w$ for both coding schemes $h_w$ and $h_{w,q}$. }\label{fig_PeekaboomRecallvsWT5}
\end{figure}

\begin{figure}
\begin{center}
\mbox{
\includegraphics[width = 2.7in]{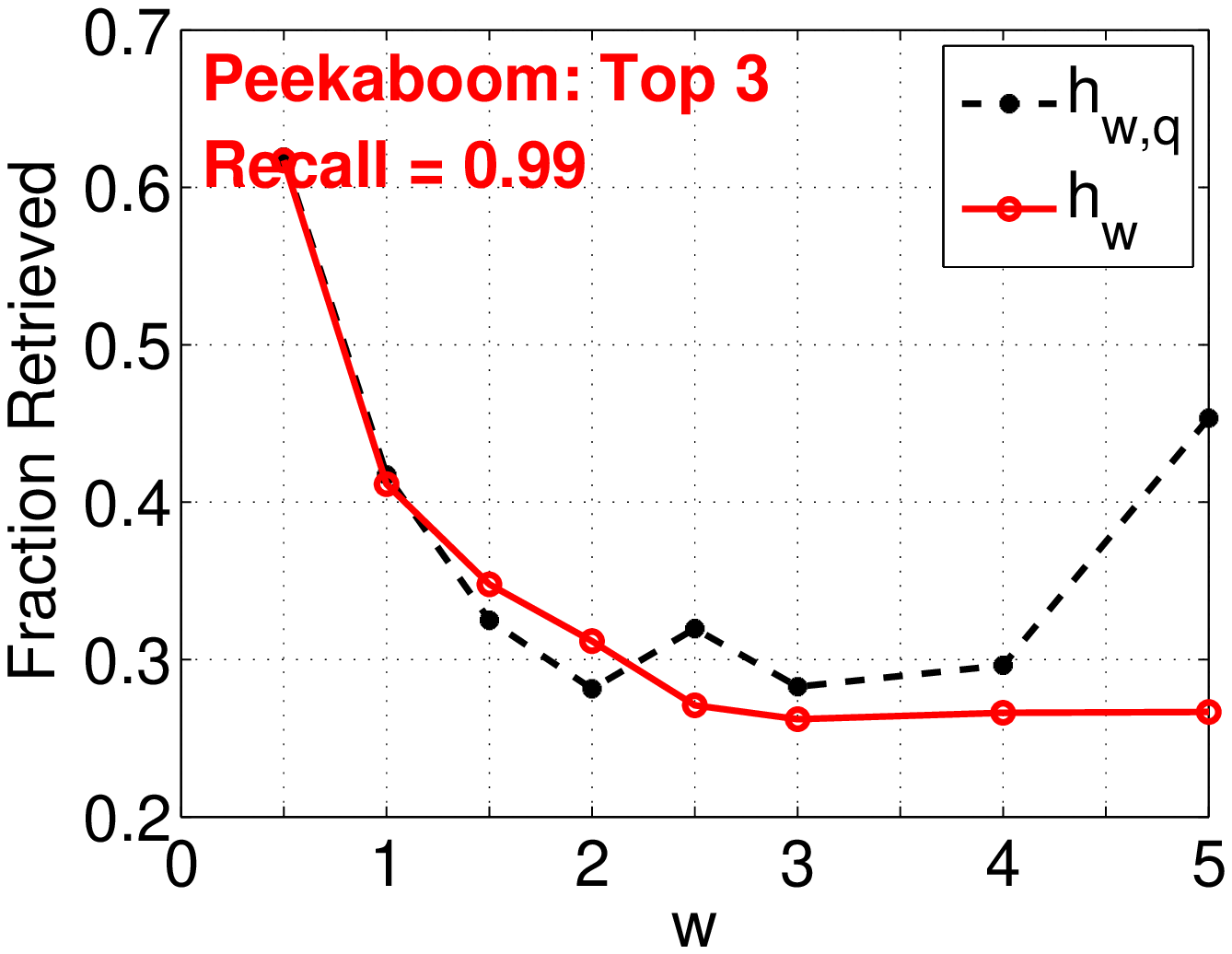}
\includegraphics[width = 2.7in]{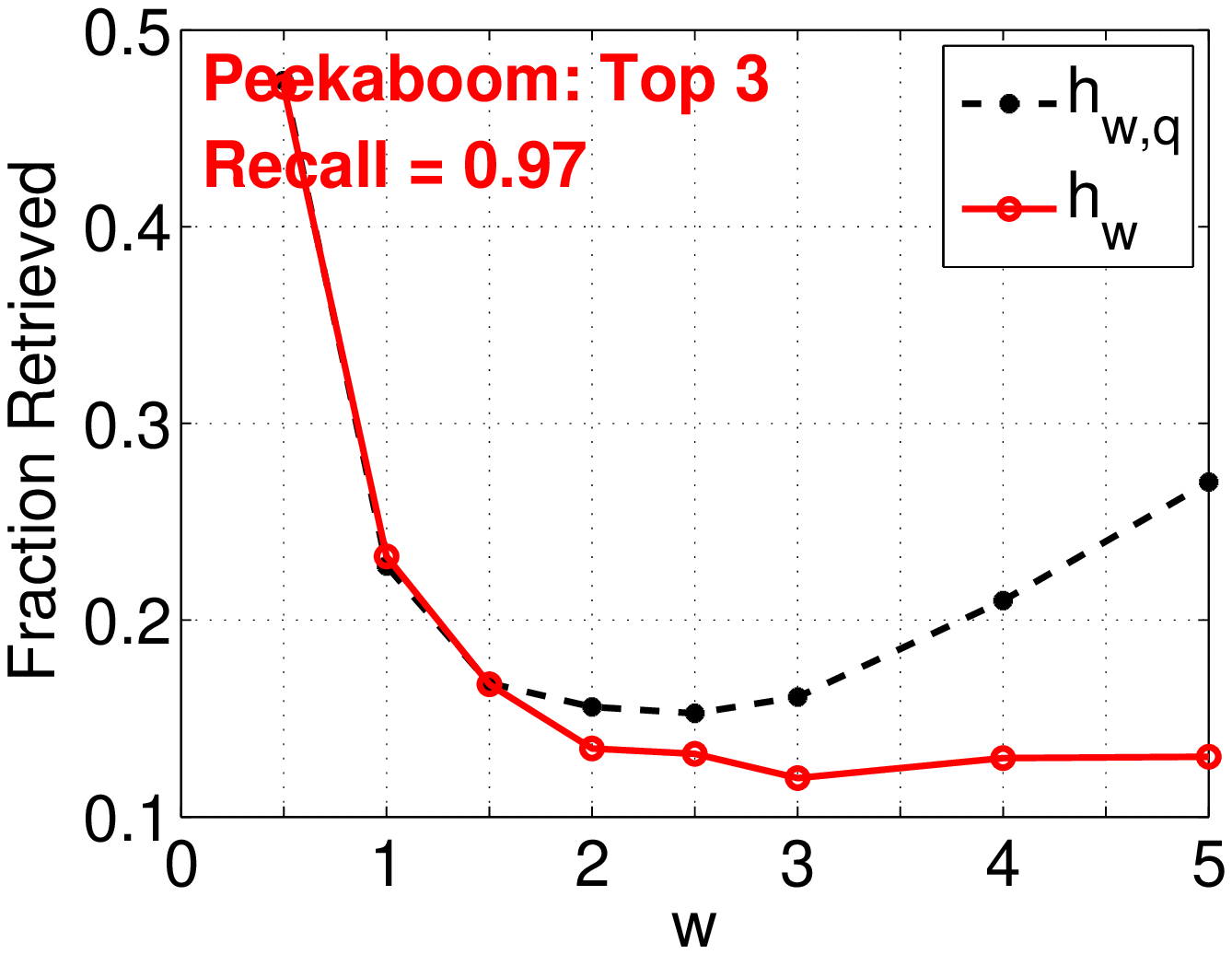}
}
\mbox{
\includegraphics[width = 2.7in]{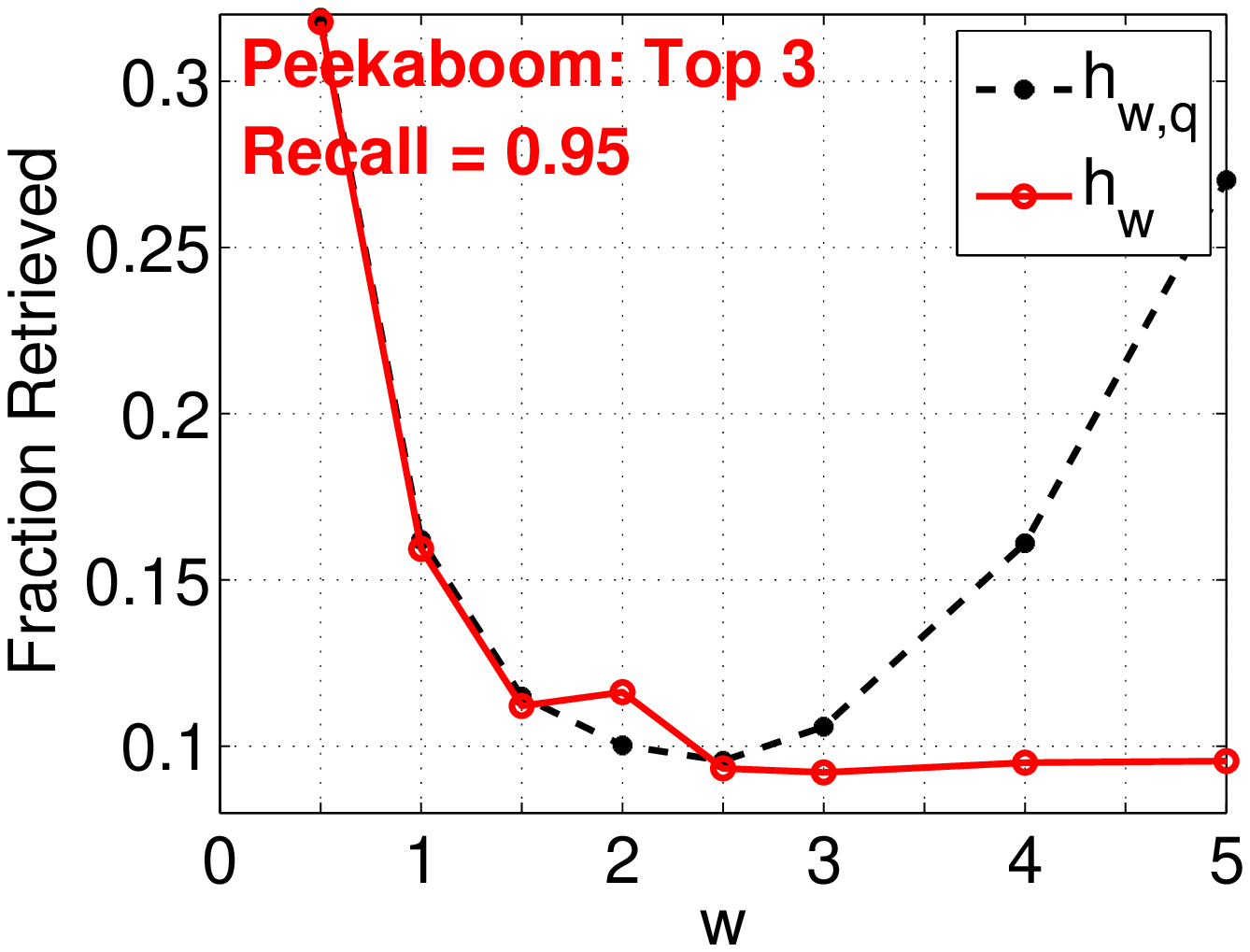}
\includegraphics[width = 2.7in]{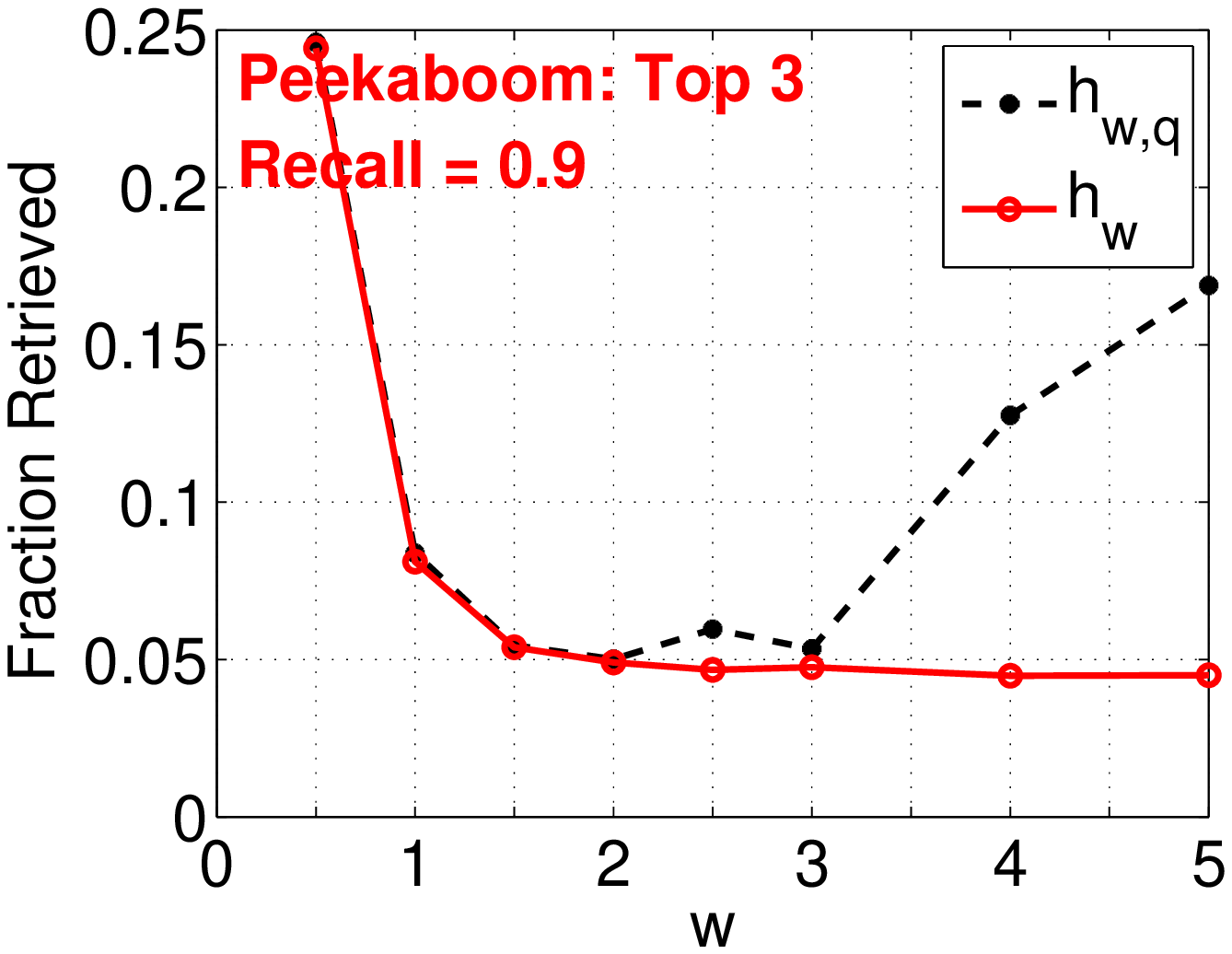}
}

\mbox{
\includegraphics[width = 2.7in]{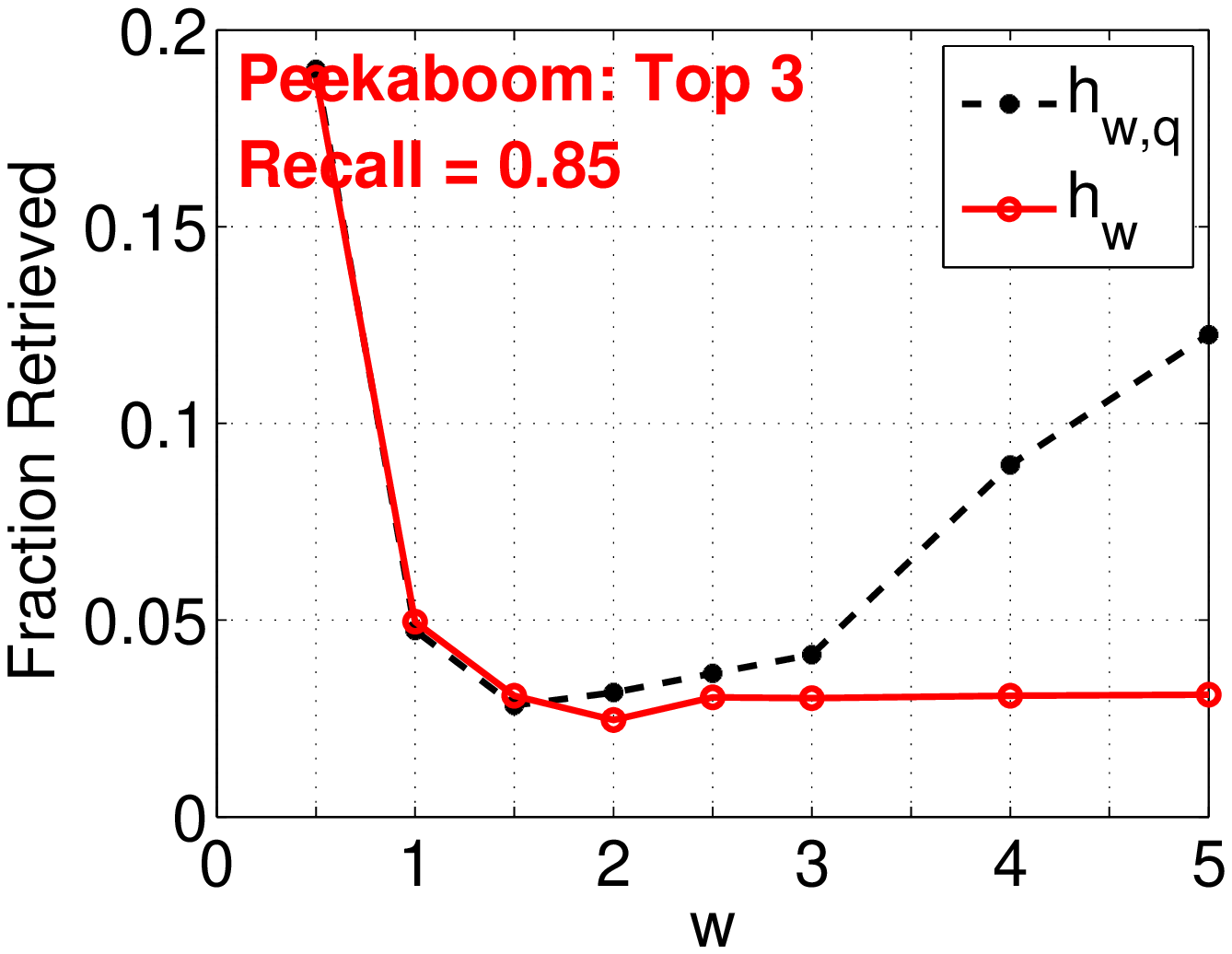}
\includegraphics[width = 2.7in]{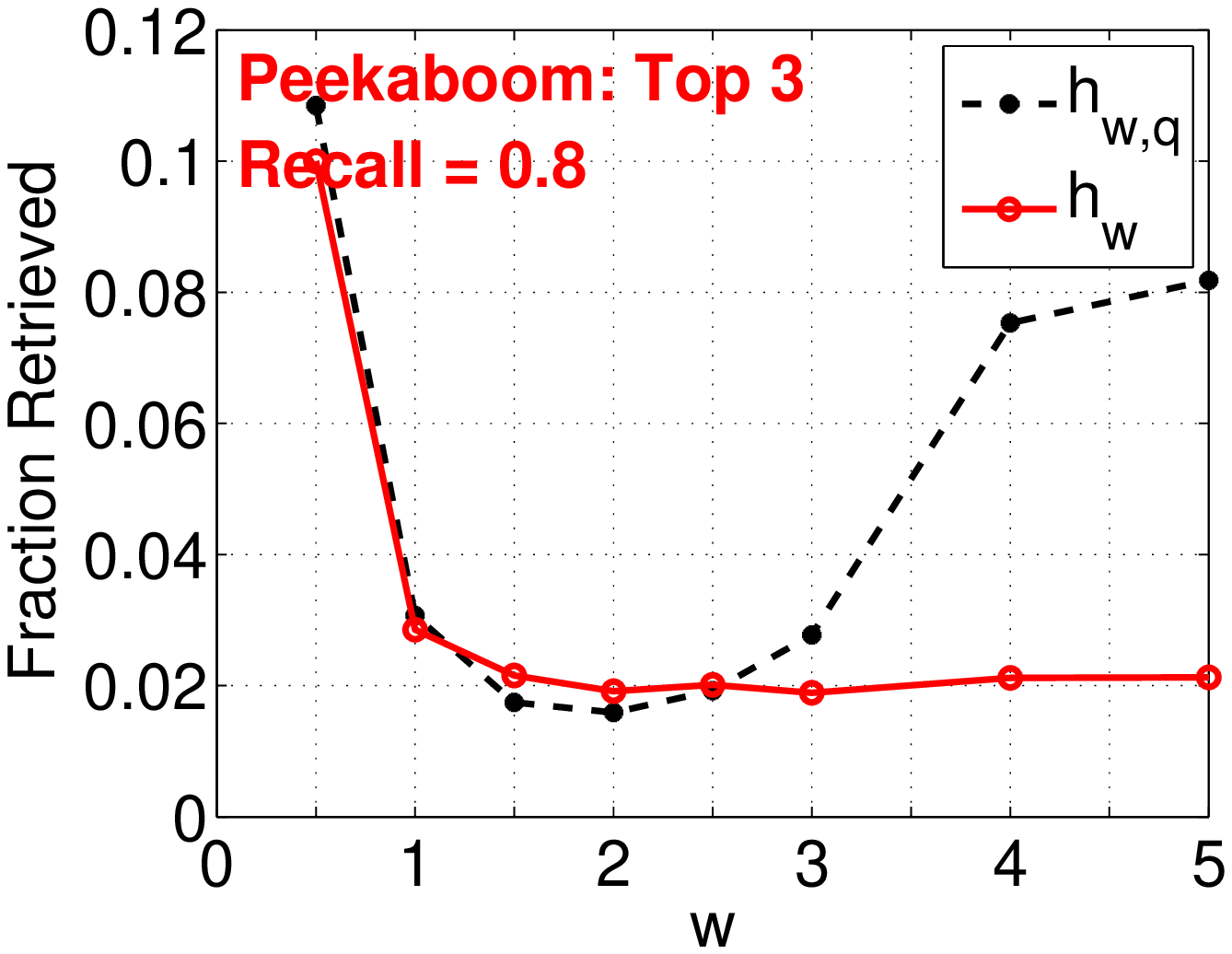}
}

\mbox{
\includegraphics[width = 2.7in]{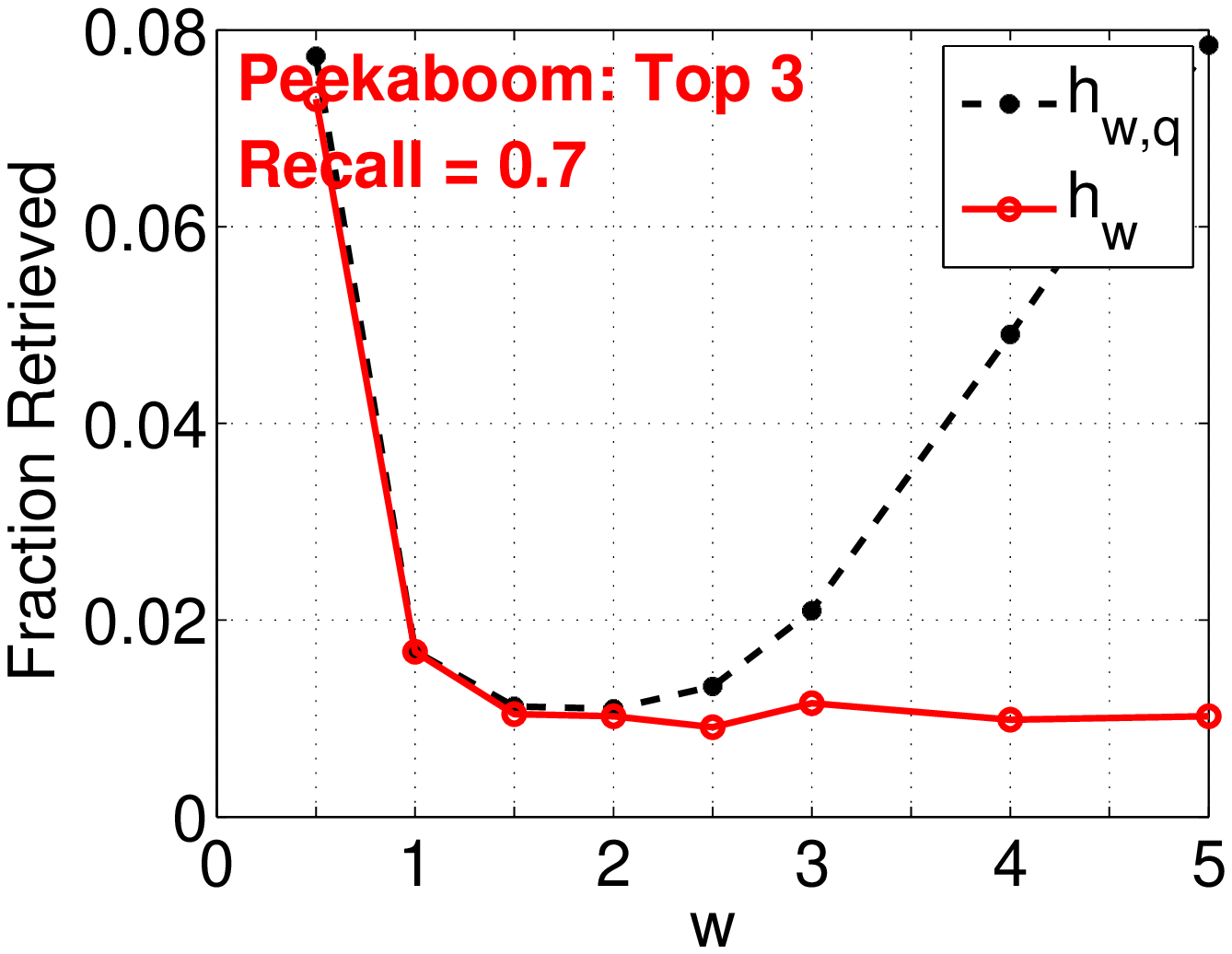}
\includegraphics[width = 2.7in]{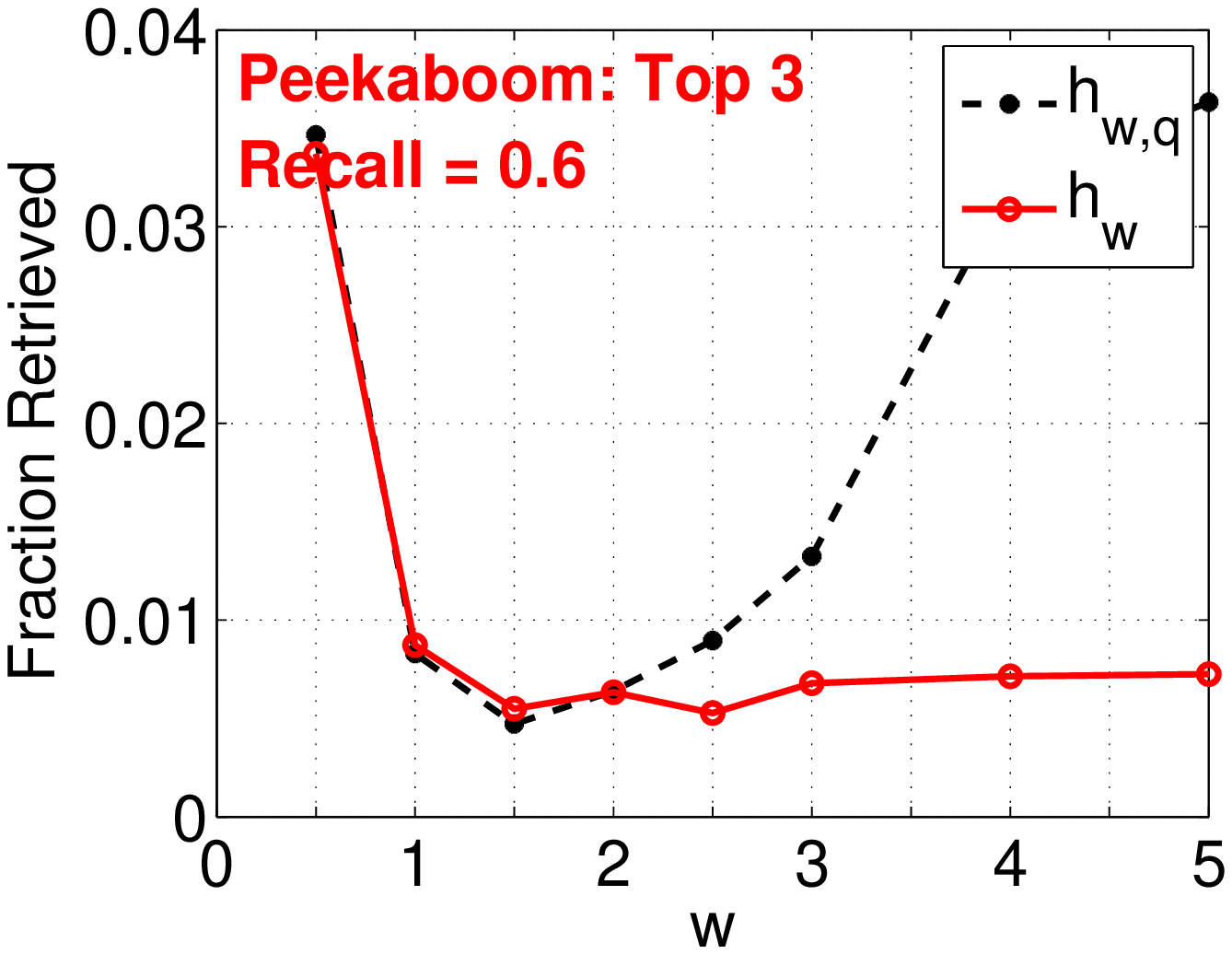}
}

\end{center}
\vspace{-.2in}
\caption{ \textbf{Peekaboom Top 3} . In each panel, we plot the optimal {\em fraction retrieved} at a target {\em recall} value (for top-3) with respect to $w$ for both coding schemes $h_w$ and $h_{w,q}$. }\label{fig_PeekaboomRecallvsWT3}
\end{figure}

Figures~\ref{fig_YoutubeRecallvsWT50} to~\ref{fig_YoutubeRecallvsWT3} present similar results on the {\em Youtube} dataset for $T=50, 20, 10, 5, 3$. We only include plots with relatively high recalls which are often more useful in practice. Figures~\ref{fig_PeekaboomRecallvsWT100} to~\ref{fig_PeekaboomRecallvsWT3} present the results on the {\em Peekaboom} dataset, which are essentially very similar to the results on the {\em Youtube} dataset. \\

These plots confirm the previous theoretical analysis: (i) it is essentially always better to use $h_w$ instead of $h_{w,q}$, i.e., the random offset is not needed; (ii) when using $h_w$ and the target recall is high (which essentially means when the target similarity is low), it is better to use a relatively large $w$ (e.g., $w=2\sim3$); (iii) when using $h_w$ and the target recall is low, it is better to use a smaller $w$ (e.g., $w=1.5$); (iv) when using $h_w$, the influence is $w$ is not that much as long as it is in a reasonable range, which is important in practice.


\clearpage\newpage

\section{Conclusion}

We have compared two quantization (coding) schemes for random projections in the context of sublinear time approximate near neighbor search. The recently proposed scheme based on uniform quantization~\cite{Report:RPCode2013} is simpler than the influential existing work~\cite{Proc:Datar_SCG04} (which used uniform quantization with a random offset). Our analysis confirms that, under the general theory of LSH, the new scheme~\cite{Report:RPCode2013} is simpler and more accurate than~\cite{Proc:Datar_SCG04}. In other words, the step of random offset in~\cite{Proc:Datar_SCG04} is not needed and may hurt the performance.

Our analysis provides the practical guidelines for using the proposed coding scheme to build hash tables. Our recommendation is to use a  bin width about $w=1.5$ when the target similarity is high and a bin width about $w=3$ when the target similarity is not that high. In addition, using the proposed coding scheme based on uniform quantization (without the random offset), the influence of $w$ is not very sensitive, which makes it very convenient in practical applications.

{

}

\end{document}